%% file: 0_main.tex
\documentclass{article}

\usepackage{microtype}
\usepackage{graphicx}
\usepackage{subcaption}
\usepackage{booktabs} 

\usepackage{hyperref}

\usepackage[preprint]{icml2026}

\usepackage{amsmath}
\usepackage{amssymb}
\usepackage{mathtools}
\usepackage{amsthm}

\usepackage[capitalize,noabbrev,nameinlink]{cleveref}

\theoremstyle{plain}

\theoremstyle{definition}

\theoremstyle{remark}

\usepackage[textsize=tiny]{todonotes}

\usepackage{times}
\usepackage{latexsym}
\usepackage[T1]{fontenc}
\usepackage[utf8]{inputenc}
\usepackage{microtype}     
\usepackage{inconsolata}

\usepackage{ifluatex}
\ifluatex
  \usepackage{emoji}
  \setemojifont{TwemojiMozilla}
\fi
\ifluatex

\else

\fi

\usepackage{amsmath,amssymb,amsfonts}

\usepackage{graphicx}
\usepackage{tikz}
\usetikzlibrary{trees,shapes,positioning,arrows.meta}
\usetikzlibrary{shapes.geometric,arrows}
\usetikzlibrary{decorations.markings}
\usepackage[edges]{forest}  

\usepackage{booktabs}
\usepackage{array}
\usepackage{tabularx}
\usepackage{multirow}
\usepackage{longtable}
\usepackage{colortbl}  

\usepackage{xcolor}
\usepackage{float}
\usepackage{enumitem}
\usepackage{soul}  
\usepackage{setspace}
\usepackage{caption}
\usepackage{subcaption}

\hypersetup{breaklinks=true,colorlinks=true,linkcolor=blue,urlcolor=blue,citecolor=blue}
\crefname{section}{Sec.}{Sec.}
\crefname{figure}{Fig.}{Figs.}
\crefname{table}{Tab.}{Tabs.}
\crefname{equation}{Eq.}{Eqs.}

\usepackage{pifont}  

\usepackage[many]{tcolorbox}
\tcbuselibrary{skins}

\definecolor{Teal}{RGB}{0, 50, 50}
\definecolor{White}{RGB}{250, 250, 250}
\definecolor{darkblue}{rgb}{0, 0, 0.5}

\definecolor{paired-light-blue}{RGB}{198, 219, 239}
\definecolor{paired-dark-blue}{RGB}{49, 130, 188}
\definecolor{paired-light-orange}{RGB}{251, 208, 162}
\definecolor{paired-dark-orange}{RGB}{230, 85, 12}
\definecolor{paired-light-green}{RGB}{199, 233, 193}
\definecolor{paired-dark-green}{RGB}{49, 163, 83}
\definecolor{paired-light-purple}{RGB}{218, 218, 235}
\definecolor{paired-dark-purple}{RGB}{117, 107, 176}
\definecolor{paired-light-gray}{RGB}{217, 217, 217}
\definecolor{paired-dark-gray}{RGB}{99, 99, 99}
\definecolor{paired-light-pink}{RGB}{222, 158, 214}
\definecolor{paired-dark-pink}{RGB}{123, 65, 115}
\definecolor{paired-light-red}{RGB}{231, 150, 156}
\definecolor{paired-dark-red}{RGB}{131, 60, 56}
\definecolor{paired-light-yellow}{RGB}{231, 204, 149}
\definecolor{paired-dark-yellow}{RGB}{141, 109, 49}

\definecolor{light-red}{RGB}{255,182,193}
\definecolor{medium-red}{RGB}{205,92,92}
\definecolor{light-yellow}{RGB}{255, 239, 153}
\definecolor{light-blue}{RGB}{173, 216, 230}
\definecolor{light-green}{RGB}{144, 238, 144}

\definecolor{bg-green}{HTML}{D5E8D4}
\definecolor{bg-blue}{HTML}{dae8fc}
\definecolor{bg-yellow}{HTML}{FFF2CC}
\definecolor{bg-pink}{HTML}{FFCCCC}
\definecolor{bg-orange}{HTML}{FFCC99}
\definecolor{bg-gray}{HTML}{eeeeee}

\definecolor{hidden-draw}{RGB}{20,68,106}
\definecolor{hidden-pink}{RGB}{255,245,247}

\definecolor{a4}{RGB}{250,235,215}
\definecolor{vgreen}{HTML}{60A917}
\definecolor{vred}{HTML}{CE3A29}

\providecommand{\devanagarifont}{}

\usepackage{ragged2e}

\newcommand{\cellrr}{\RaggedRight\arraybackslash}

\usepackage{fontawesome5} 

\usepackage{xurl}

\usepackage{mathtools} 

\newtcolorbox{defin}{
    colback=Teal!5!White,
    enhanced,
    breakable,
    title=Neural FOXP2 at-a-glance,
    attach boxed title to top left={xshift=-4mm},
    boxrule=0pt,
    after skip=1.5em,
    before skip=0.5em,
    right skip=0cm,
    fonttitle=\bfseries,
    toprule=0pt,
    bottomrule=0pt,
    rightrule=0pt,
    leftrule=3pt,
    arc=0mm,
    skin=enhancedlast jigsaw,
    sharp corners,
    colframe=Teal!55!black,
    colbacktitle=Teal!55!black,
    boxed title style={
        frame code={
            \fill[Teal!25!black](frame.south west)--(frame.north west)--(frame.north east)--([xshift=3mm]frame.east)--(frame.south east)--cycle;
            \draw[line width=1mm,Teal!25!black]([xshift=2mm]frame.north east)--([xshift=5mm]frame.east)--([xshift=2mm]frame.south east);
            \draw[line width=1mm,Teal!25!black]([xshift=5mm]frame.north east)--([xshift=8mm]frame.east)--([xshift=5mm]frame.south east);
            \fill[Teal!25!black](frame.south west)--+(4mm,-2mm)--+(4mm,2mm)--cycle;
        }
    }
}

\makeatletter
\newcommand{\SafeFA}[2]{\ifcsname #1\endcsname \csname #1\endcsname \else #2\fi}
\makeatother

\tikzstyle{my-box}=[
    rectangle,
    draw=hidden-draw,
    rounded corners,
    text opacity=1,
    minimum height=1.5em,
    minimum width=40em,
    inner sep=2pt,
    align=center,
    fill opacity=.5,
    line width=0.8pt,
]
\tikzstyle{leaf}=[my-box, minimum height=1.5em,
    fill=hidden-pink!80, text=black, align=center,font=\normalsize,
    inner xsep=2pt,
    inner ysep=4pt,
    line width=0.8pt,
]

\definecolor{darknavy}{RGB}{0, 0, 102}
\usepackage{morewrites}
\definecolor{lightgreen}{rgb}{0.88,1,0.88}
\definecolor{lightred}{rgb}{1,0.88,0.88}
\definecolor{lightblue}{rgb}{0.88,0.93,1}
\definecolor{lightyellow}{rgb}{1,1,0.8}

\icmltitlerunning{Submission and Formatting Instructions for ICML 2026}

\begin{document}

\twocolumn[
\icmltitle
{
\includegraphics[width=\textwidth]{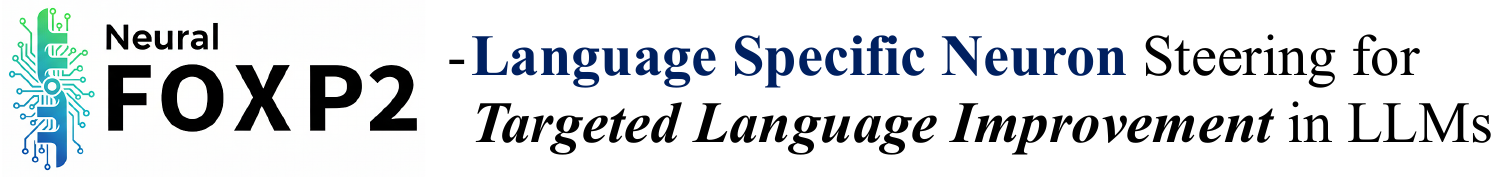}
}




\begin{center}
{\Large\bfseries
Anusa Saha$^{1}$,
Tanmay Joshi$^{1}$,
Vinija Jain$^{2}$,
Aman Chadha$^{3}$,
Amitava Das$^{1}$}\\
{\large
$^{2}$Meta, USA,
$^{3}$Apple, USA
$^{1}$Pragya Lab, BITS Pilani Goa, India\\
}
\end{center}




  \icmlcorrespondingauthor{Ansua Saha}{ansuasaha2406@gmail.com}

  \icmlkeywords{Machine Learning, ICML}

  \vskip 0.3in
]



\printAffiliationsAndNotice{}  


\begin{abstract}
LLMs are multilingual by training, yet their \emph{lingua franca} is often English, reflecting English language dominance in pretraining. Other languages remain in parametric memory but are systematically \emph{suppressed}. We argue that \textbf{language defaultness} is governed by a \textbf{sparse, low-rank control circuit}---\textbf{language neurons}---that can be mechanistically isolated and safely steered.

We introduce \textbf{Neural FOXP2}, that makes a chosen language (Hindi or Spanish) \emph{primary} in a model by steering \textbf{language-specific neurons}. Neural FOXP2 proceeds in three stages: (i) \textbf{Localize:} We train per-layer \textbf{SAEs} so each activation decomposes into a small set of active \emph{feature components}. For every feature, we quantify \textbf{English vs.\ Hindi/Spanish selectivity} overall \textbf{logit-mass lift} toward the target-language token set. Tracing the top-ranked features back to their strongest contributing units yields a compact \textbf{language-neuron set}. (ii) \textbf{Steering directions:} We localize controllable language-shift geometry via a spectral low-rank analysis. For each layer, we build \textbf{English$\rightarrow$target activation-difference} matrices and perform \textbf{layerwise SVD} to extract the dominant singular directions governing language change. The \textbf{eigengap} and \textbf{effective-rank} spectra identify a compact \textbf{steering subspace} and an empirically chosen \textbf{intervention window} (where these directions are strongest and most stable). (iii) \textbf{Steer:} We apply a \textbf{signed, sparse activation shift} targeted to the \textbf{language neurons}. Concretely, within low-to-mid layers we add a positive steering along the target-language dominant directions and a compensating negative shift toward the null space for the English-neurons, yielding controllable target-language defaultness.

On \textbf{LLaMA-3 8B}, Neural FOXP2 improves Hindi and Spanish performance across \textbf{machine translation}, \textbf{question answering}, \textbf{natural language inference}, and \textbf{summarization}, effectively shifting the model’s language prior so the target language becomes the default \emph{without} full retraining.
\end{abstract}

\begin{defin}
\scriptsize

\begin{itemize}[left=-4pt,itemsep=0pt,topsep=0pt,parsep=0pt]

\item[\faBolt] \textbf{\textit{TL;DR}}:
\textbf{Neural FOXP2} makes targeted language improvement \emph{mechanistic and surgical}:
it identifies a sparse \textbf{language-neuron} control set (SAE features) and steers only this set so Hindi/Spanish becomes \emph{default} \textbf{without} full retraining.

\item[\faCompass] \textbf{\textit{Core hypothesis (Language Defaultness)}}:
\textbf{Language defaultness} of an LLM is governed by a \textbf{sparse, low-rank control circuit}---\textbf{language neurons}---that can be mechanistically isolated and safely steered.

\item[\faBrain] \textbf{Neural FOXP2:} Mechanistic language defaultness control via sparse \textbf{language-neuron} steering.
\begin{itemize}[leftmargin=1.2em,topsep=2pt,itemsep=2pt]
  \item[\faSearch] \textbf{\textit{Stage I: Localize:}} Rank SAE features by \textbf{target selectivity} and \textbf{target-token logit-mass lift} to extract compact \textbf{language-neurons}.
  \item[\faProjectDiagram] \textbf{\textit{Stage II: Directions:}} Use \textbf{English$\rightarrow$target} activation-differences + \textbf{layerwise SVD} (eigengap/effective-rank) to identify a low-rank \textbf{steering subspace}.
  \item[\faSlidersH] \textbf{\textit{Stage III: Steer:}} In low--mid layers, apply a \textbf{signed sparse shift} on \textbf{language neurons}---\textbf{+} along target directions and \textbf{--} toward null space on English neurons---to make the target language \textbf{default}.
\end{itemize}

\item[\faVial] \textbf{\textit{Evaluation (Hindi/Spanish multi-task)}}:
We evaluate across \textbf{machine translation}, \textbf{question answering}, \textbf{natural language inference}, and \textbf{summarization},
testing whether the target language becomes the \emph{default} while preserving task performance. On \textbf{LLaMA-3 8B}, Neural FOXP2 improves Hindi and Spanish across tasks by shifting the model's \textbf{language prior}.

\item[\faTools] \textbf{\textit{Practical upshot}}:
Neural FOXP2 provides \textbf{mechanistic language control}: \textbf{upgrade one language} by steering language neurons,
\textbf{without} retraining or finetuning.

\end{itemize}
\end{defin}

\input{1_motivation}

\input{2_foxp2}

\input{3_utility}

\section{Conclusion}
\label{sec:conclusion}
Neural FOXP2 treats multilingual language improvement as \textbf{mechanistic control of defaultness}. Rather than updating parameters, we (i) localize a sparse \textbf{language-neuron} support in a feature basis, (ii) recover a \textbf{low-rank} steering geometry and an \textbf{intervention window}, and (iii) apply a \textbf{signed, sparse activation edit} that makes Hindi/Spanish the default.
The message is that \textbf{non-English competence is often present but suppressed by a sparse control prior}; editing that prior can unlock default-language behavior without broad retraining.

\noindent\textbf{Implication and boundary.} Some multilingual gaps reflect routing, not only exposure: models may under-utilize an existing language channel. FOXP2 helps \textbf{amplify} that channel. We do \emph{not} claim this replaces continued pretraining: for truly low-resource languages, missing competence and coverage still require data and training. FOXP2 is best for languages that are \textbf{present in pretraining but under-defaulted}, where steering can recover suppressed capacity. 

\clearpage
\newpage

\input{4_discussion}

\clearpage
\newpage

\bibliography{example_paper}
\bibliographystyle{icml2026}

\clearpage
\newpage

\input{5_faq}

\clearpage
\newpage

\input{5_2_reproducability}

\clearpage
\newpage

\input{6_appendix}

\end{document}

%% file: 1_motivation.tex
\vspace{-1.5em}

\section{\textbf{Admonitio}: Language Neurons}
\label{sec:admonitio_language_neurons}

Modern decoder-only LLMs are trained on web-scale corpora whose \emph{language mass} is sharply imbalanced; a small set of high-resource languages dominates token exposure and gradient flow during pretraining. \citep{llama3blog,llama3modelcard}
Emerging evidence suggests this imbalance can crystallize into an \emph{internal preference}: one language behaves as a lingua-franca prior, while other languages remain available yet comparatively \emph{harder to select} under ambiguous or mixed cues. \citep{shani-basirat-2025-language}
This creates a bottleneck for \emph{make Hindi default}: global continued pretraining or instruction tuning can raise Hindi performance, but it also risks (i) \textbf{capability drift}, (ii) \textbf{cross-lingual regressions}, and (iii) harder \textbf{safety preservation} due to distributional shift.

\vspace{-1em}
\paragraph{Prior signals: steerability, dominance, and causality---but no end-to-end control recipe.}
Three lines of work jointly suggest that language identity is not merely “in the prompt,” but \emph{in the mechanism}.
\begin{itemize}[leftmargin=1.25em, itemsep=0em, topsep=0em]
  \item \textbf{Language-steerable units exist.} A \emph{small} fraction of internal units can disproportionately influence competence and generation in a given language; selective activation/deactivation can shift language realization. \citep{tang-etal-2024-language,kojima-etal-2024-language-specific}
  \item \textbf{Language dominance is measurable.} Multilingual LLMs can exhibit \emph{asymmetric} internal preference consistent with an attractor-like prior, not merely prompt-matching. \citep{shani-basirat-2025-language}
  \item \textbf{Causal tools are mature.} Activation patching and causal editing provide intervention-based methodology to separate correlation from mechanism. \citep{zhang-nanda-2023-activation-patching,meng2022rome}
\end{itemize}
Yet these ingredients have not been assembled into a unified, stability-controlled pipeline for \emph{targeted language improvement}: we still lack (i) a \textbf{mechanistic objective} that captures early language commitment, (ii) a \textbf{feature-level} localization strategy robust to superposition, and (iii) a \textbf{stability constraint} that bounds regressions under surgical edits.

\vspace{-1em}
\paragraph{Core hypothesis (H\textsubscript{LN}): sparse causal control of language identity.}
We hypothesize that language identity is governed by a \textbf{sparse, causally privileged control set}:

\begin{quote}
\vspace{-0.5em}
\noindent\textbf{H\textsubscript{LN}.} \emph{There exists a small set of internal units/circuitry whose intervention predictably changes the model’s output language while preserving task semantics.}
\vspace{-0.5em}
\end{quote}

This is a direct extension of the broader mechanistic viewpoint that specific behaviors can be mediated by localized computations rather than uniformly distributed changes. \citep{meng2022rome}
In the multilingual setting, H\textsubscript{LN} implies that targeted language improvement should focus on \textbf{editing the control set} rather than updating the entire model.

\vspace{-1em}
\paragraph{Formalizing “default language” as a measurable control objective.}
Let $p_\theta(\cdot\mid x)$ denote the next-token distribution. To quantify language selection, define token partitions
$\mathcal{V}_{\mathrm{hi}}$ (Hindi tokens) and $\mathcal{V}_{\neg\mathrm{hi}}$ (non-Hindi competitors).
We define an early-step \textbf{language advantage} functional
\[
\Delta_{\mathrm{hi}}(x)
\;=\;
\log \sum_{t\in \mathcal{V}_{\mathrm{hi}}} p_\theta(t\mid x)
\;-\;
\log \sum_{t\in \mathcal{V}_{\neg \mathrm{hi}}} p_\theta(t\mid x),
\]
which measures how much probability mass the model assigns to Hindi \emph{before} the trajectory collapses into a stable language manifold.
A language is \emph{default} on a prompt set $\mathcal{D}_{\mathrm{neutral}}$ if $\mathbb{E}_{x\sim \mathcal{D}_{\mathrm{neutral}}}[\Delta_{\mathrm{hi}}(x)]$ is consistently positive (or equivalently, Hindi mass dominates non-Hindi mass early in decoding).
This yields a clean control target: \textbf{raise} $\Delta_{\mathrm{hi}}$ under neutral prompts while preserving semantic correctness on task prompts.

\vspace{-1em}
\paragraph{Why raw neurons are insufficient: superposition requires a feature basis.}
A naive “target neuron” recipe is brittle because transformers exhibit \emph{superposition}: single neurons mix many features and individual features are distributed across many neurons. \citep{elhage2022toy}
Thus a neuron that correlates with Hindi may also participate in unrelated computations; editing it can induce collateral changes that masquerade as “language improvement” or spill into non-Hindi behaviors.
A principled remedy is to move from \emph{neurons} to \emph{features}: learn a sparse coordinate system in which individual coordinates are more isolatable latent factors.
Sparse autoencoders (SAEs) provide such a basis for LM activations, producing sparse feature units that are often more monosemantic than raw neurons and more stable intervention targets. \citep{cunningham2023sae}
In this view, \textbf{language neurons} are operationalized as \textbf{language features}: sparse SAE coordinates whose activations (i) track language identity and (ii) causally influence $\Delta_{\mathrm{hi}}(x)$.

\vspace{-1em}
\paragraph{From correlation to mechanism: defining a Hindi-control feature.}
Even in a feature basis, localization must be \emph{causal}. Best-practice work on activation patching emphasizes that attribution conclusions depend on corruption choices and metrics, and it formalizes intervention protocols \citep{zhang-nanda-2023-activation-patching}.
We therefore define a \textbf{Hindi-control feature} $z_j$ by its \emph{intervention effect} on $\Delta_{\mathrm{hi}}(x)$:
\[
\mathcal{I}_j
\;=\;
\mathbb{E}_{x\sim \mathcal{D}}
\Big[\Delta_{\mathrm{hi}}(x;\; \mathrm{do}(z_j \leftarrow z_j + \epsilon)) - \Delta_{\mathrm{hi}}(x)\Big],
\]
for small $\epsilon$, with controls that hold task semantics fixed (matched prompts) and rule out script-only shortcuts (transliteration/script controls).
Necessity is tested by ablation (removing $z_j$ reduces $\Delta_{\mathrm{hi}}$ and downstream Hindi gains);
sufficiency is tested by patching/forcing (injecting $z_j$ increases Hindi mass while preserving meaning). \citep{zhang-nanda-2023-activation-patching}
This framing makes the claim falsifiable: a “language neuron” must \emph{move} the language advantage, not merely correlate with Hindi text.

\vspace{-1em}
\paragraph{Stability admonition: outliers and superweights can dominate failure modes.}
A second admonition is that \emph{small} edits need not be \emph{safe} edits.
Transformer inference is shaped by rare outlier dimensions/weights that can dominate numerical and functional behavior (e.g., outlier features relevant to stable quantization). \citep{dettmers2022llmint8}
Recent results on \emph{superweights} sharpen this: a tiny number of parameter outliers can be disproportionately important, and pruning/corrupting them can catastrophically destroy model behavior; superweights are linked to rare, large ``super activations.'' \citep{yu2024superweight}
Therefore targeted language improvement must explicitly \textbf{avoid brittle outlier coordinates} (or freeze them as constraints); otherwise a “Hindi fix” may inadvertently trigger global degradation.

\vspace{-1em}
\paragraph{Positioning: from neuron steering to feature-level, trust-region editing.}
Taken together, prior work gives (i) evidence that language-steerable units exist, \citep{tang-etal-2024-language,kojima-etal-2024-language-specific}
(ii) evidence that sparse feature representations mitigate superposition and yield stable intervention targets, \citep{cunningham2023sae,elhage2022toy}
(iii) causal tools to establish necessity/sufficiency rather than saliency, \citep{zhang-nanda-2023-activation-patching,meng2022rome}
and (iv) cautions that outliers demand constrained edits. \citep{yu2024superweight,dettmers2022llmint8}
This motivates \textbf{Neural FOXP2} as a coherent mechanistic agenda:
\textbf{(1) Localize} Hindi-control features in an SAE basis using selectivity and intervention effect $\mathcal{I}_j$;
\textbf{(2) Verify} causality with ablation/patching under script and semantic controls; and
\textbf{(3) Shift} the default-language prior by a signed, sparse edit that increases $\Delta_{\mathrm{hi}}(x)$ under bounded drift.
More broadly, this reframes multilingual adaptation: not “more Hindi data,” but \textbf{redistributing internal control mass} so Hindi becomes the default realization channel \emph{without} retraining the backbone.

%% file: 2_foxp2.tex
\begin{figure*}[t]
  \centering
  \includegraphics[width=\textwidth]{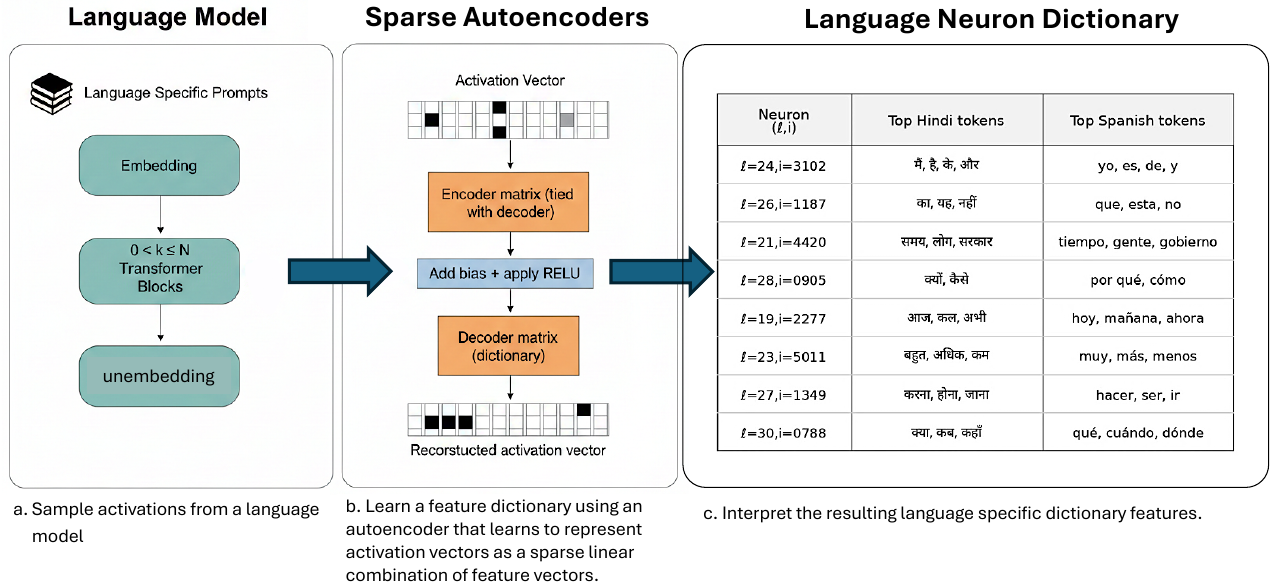}
  \vspace{-1.5em}
  \caption{
  \textbf{Neural FOXP2: VAE/autoencoder-based discovery of language-selective neurons (Hindi or Spanish).}
  The workflow proceeds in three stages: 
  \textbf{(a) Sample activations from a language model} by running language-specific prompts and extracting intermediate activations $h^{(\ell)}(x)$ at selected layers $\ell$ (left panel; “Embedding $\rightarrow$ Transformer Blocks”). 
  \textbf{(b) Learn a feature dictionary using an autoencoder} (here, a VAE/SAE-style encoder–nonlinearity–decoder pipeline) that represents activation vectors as a sparse/structured combination of latent features (middle panel; “Encoder matrix (tied with decoder) $\rightarrow$ Add bias + apply ReLU”). 
  \textbf{(c) Interpret the resulting language-specific dictionary features} by mapping discovered features back to model units and compiling a \emph{Language Neuron Dictionary}: each row lists a neuron id $(\ell,i)$ along with its highest-affinity \textbf{Hindi} versus \textbf{Spanish} tokens (right panel), yielding a human-auditable signature of a putative “language neuron” and enabling downstream masking/editing in Neural FOXP2.}
  \label{fig:neural-foxp2-language-neuron-dictionary}
  \vspace{-1.5em}
\end{figure*}

\vspace{-1em}
\section{Neural FOXP2: Causal Feature Editing to Make Hindi Default}
\label{sec:method}

\begin{figure}[H]
  \centering
  \includegraphics[width=0.78\linewidth]{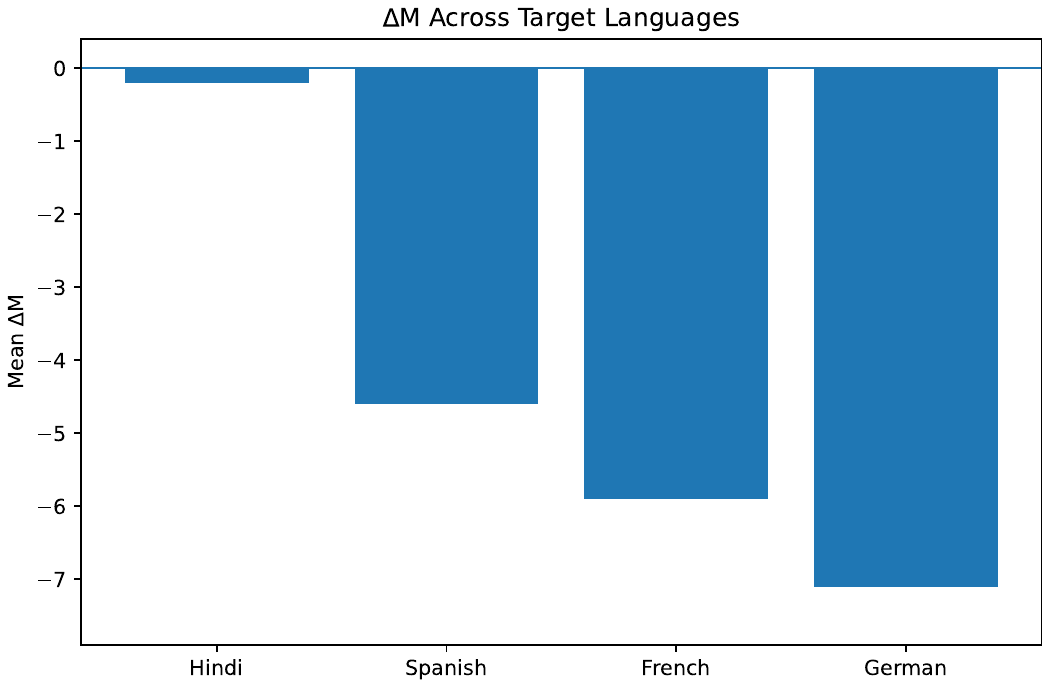}
  \vspace{-0.7em}
      \caption{
    \textbf{English is the lingua franca by default.}
    Mean \textbf{\emph{target-language defaultness}} \(\mathbf{\Delta M \;=\; M_{\ell} - M_{\mathrm{En}}}\) for 4 languages:
    \textbf{Hindi}, \textbf{Spanish}, \textbf{French}, and \textbf{German}.
    Across all targets, \(\Delta M\) remains \textbf{\emph{biased toward English}} (negative mass advantage),
    indicating that defaultness behaves as a \textbf{\emph{prior}} rather than a capability limit.
    This motivates \textbf{FOXP2}’s \textbf{\emph{language-specific steering}}: shifting \textbf{\emph{logit mass}} toward the target language.}
  \label{fig:lingua_franca_script}
  \vspace{-1.5em}
\end{figure}

\noindent\textbf{English is the \emph{lingua franca}.}
Despite multilingual pretraining, the model's \textbf{default} generation prior is often English: under weak prompting, Hindi realization requires an explicit script cue, while transliterated Hindi remains pulled toward English (\textbf{Fig.~\ref{fig:lingua_franca_script}}).

\vspace{-1.2em}

We formalize this \textbf{language defaultness} as an early-step control target. Let $f_\theta$ be a multilingual LM with $L$ transformer blocks and hidden size $d$, and let $\ell\in\{1,\dots,L\}$ index layers. For a prompt $x$, we probe the residual-stream activation $h^{(\ell)}(x)\in\mathbb{R}^d$ at a fixed position (typically the last prompt token). Let $V$ be the vocabulary and let $V_t, V_{\mathrm{en}}\subseteq V$ denote target-language and English token sets (constructed via script/lexicon/token-ID heuristics). At decoding step $t$ with context $\mathrm{ctx}_t$, the next-token distribution is $p_\theta(u\mid \mathrm{ctx}_t)$ for $u\in V$. We quantify defaultness in the first few steps ($t\in\{1,2,3\}$) by target-vs-English mass:
\[
\begin{aligned}
M_t(x)
&=\sum_{u\in V_t} p_\theta\!\left(u\mid \mathrm{ctx}_t\right),\\
M_{\mathrm{en}}(x)
&=\sum_{u\in V_{\mathrm{en}}} p_\theta\!\left(u\mid \mathrm{ctx}_t\right),\\
\Delta M(x)
&= M_t(x)-M_{\mathrm{en}}(x).
\end{aligned}
\]
A successful intervention increases $\mathbb{E}_{x\sim\mathcal{D}}[\Delta M(x)]$ while preserving \textbf{task semantics}.

\subsection{Stage I: Localize Language Neurons in a VAE Dictionary Basis}
\label{subsec:stage1_localize}

\noindent\textbf{Why a VAE dictionary (a sparse coordinate system for causal control).}
Direct ``neuron edits'' are unreliable because transformer channels are \textbf{polysemantic}: the same unit can jointly encode language, topic, and style.
Neural FOXP2 therefore defines \textbf{language neurons} in a \textbf{dictionary basis} learned by a \textbf{per-layer VAE-style autoencoder} (Fig.~\ref{fig:neural-foxp2-language-neuron-dictionary}), which yields feature coordinates that are empirically more \textbf{stable}, \textbf{auditable}, and \textbf{intervenable} than raw channels \citep{bricken2023monosemanticity}.
This choice also matches evidence that \textbf{language behavior concentrates in small subsets of units} and can be \textbf{steered by targeted perturbations} \citep{tang2024language,kojima2024finding}.

\vspace{-1.2em}
\paragraph{Activation site and matched data.}
For each layer $\ell\in\{1,\dots,L\}$ we collect residual-stream activations
$
h^{(\ell)}(x)\in\mathbb{R}^d
$
at a fixed token position (default: \textbf{last prompt token}).
To isolate \textbf{language identity} from \textbf{task content}, we use \textbf{matched meaning units}:
for each semantic unit $k=1,\dots,N$, we instantiate parallel prompts
$
(x^{(k)}_{\mathrm{en}},\,x^{(k)}_{\mathrm{hi}},\,x^{(k)}_{\mathrm{es}}),
$
i.e., the \textbf{same meaning} rendered in \textbf{English, Hindi, and Spanish}.
We include (i) \textbf{weakly specified} prompts (no explicit language instruction) to probe defaultness, and (ii) \textbf{explicit-language} prompts to provide high-SNR language states for selectivity estimation.
Let $\mathcal{D}_{\mathrm{mix}}$ denote the union of these prompt sets.

\vspace{-1.2em}
\paragraph{Per-layer VAE dictionary training objective.}
At each layer $\ell$, we learn a VAE-style encoder--nonlinearity--decoder pipeline (middle panel of Fig.~\ref{fig:neural-foxp2-language-neuron-dictionary}).
Concretely, we compute a latent feature vector via a tied-weight linear encoder, a bias, and a ReLU gate:

\vspace{-1.3em}
\[
\begin{aligned}
r^{(\ell)}(x)
&=\; (W_\ell)^\top h^{(\ell)}(x) + b_\ell,\\
z^{(\ell)}(x)
&=\; \phi\!\big(r^{(\ell)}(x)\big)
\;=\; \mathrm{ReLU}\!\big(r^{(\ell)}(x)\big),
\end{aligned}
\]
and reconstruct with a linear decoder (the learned \textbf{dictionary}):
\[
\hat h^{(\ell)}(x)
=\;
W_\ell\, z^{(\ell)}(x),
\qquad
W_\ell\in\mathbb{R}^{d\times m}.
\]
We train $W_\ell,b_\ell$ to minimize reconstruction with sparsity on the latent code:

\vspace{-1em}
\[
\min_{W_\ell,b_\ell}\;
\mathbb{E}_{x\sim\mathcal{D}_{\mathrm{mix}}}
\Big[
\underbrace{\big\|h^{(\ell)}(x)-\hat h^{(\ell)}(x)\big\|_2^2}_{\text{reconstruction}}
\;+\;
\lambda_{\mathrm{sparse}}
\underbrace{\big\|z^{(\ell)}(x)\big\|_1}_{\text{sparsity}}
\Big],
\]
\vspace{-1em}

optionally with a column-norm constraint on $W_\ell$ to prevent degenerate scaling.
Each coordinate $z^{(\ell)}_j$ is a \textbf{dictionary feature unit}; Stage~I identifies a small subset as \textbf{language neurons} (right panel of Fig.~\ref{fig:neural-foxp2-language-neuron-dictionary}).

\vspace{-1em}
\paragraph{Defaultness target (early-step mass).}
Let $V$ be the vocabulary and let $V_{\mathrm{hi}},V_{\mathrm{es}},V_{\mathrm{en}}\subseteq V$ denote Hindi, Spanish, and English token sets, respectively.
At decoding step $t$ with context $\mathrm{ctx}_t$, define: $M^{\mathrm{hi}}_t(x)=\sum_{u\in V_{\mathrm{hi}}} p_\theta\!\left(u\mid \mathrm{ctx}_t\right),\;
M^{\mathrm{es}}_t(x)=\sum_{u\in V_{\mathrm{es}}} p_\theta\!\left(u\mid \mathrm{ctx}_t\right),\;
M^{\mathrm{en}}_t(x)=\sum_{u\in V_{\mathrm{en}}} p_\theta\!\left(u\mid \mathrm{ctx}_t\right).$

We measure defaultness for each target language $\ell_t\in\{\mathrm{hi},\mathrm{es}\}$ by

\vspace{-1em}
\[
\Delta M^{\ell_t}(x,t)
=
M^{\ell_t}_t(x)-M^{\mathrm{en}}_t(x),
\qquad
t\in\{1,2,3\}.
\]
\vspace{-1em}

Stage~I localizes features that are (i) \textbf{language-selective} and (ii) \textbf{causally increase} $\Delta M^{\ell_t}$ under weak prompting.

\subsubsection{Stage I-A: Language Selectivity (English vs.\ Hindi/Spanish)}
\label{subsubsec:stage1_selectivity}

\noindent\textbf{Feature activations.}
For layer $\ell$ and feature $j$, define
\[
a^{(\ell)}_j(x)
\;=\;
z^{(\ell)}_j(x)
\;=\;
\Big(\mathrm{ReLU}\!\big((W_\ell)^\top h^{(\ell)}(x)+b_\ell\big)\Big)_j.
\]

\noindent\textbf{Matched-pair selectivity.}
For a target language $\ell_t\in\{\mathrm{hi},\mathrm{es}\}$, using matched pairs $(x^{(k)}_{\mathrm{en}},x^{(k)}_{\ell_t})$, compute
\[
\mathrm{Sel}^{(\ell,\ell_t)}_j
=
\mathbb{E}_{k}\!\left[a^{(\ell)}_j(x^{(k)}_{\ell_t})\right]
-
\mathbb{E}_{k}\!\left[a^{(\ell)}_j(x^{(k)}_{\mathrm{en}})\right],
\]
and standardize:
\[
\widetilde{\mathrm{Sel}}^{(\ell,\ell_t)}_j
=
\frac{\mathrm{Sel}^{(\ell,\ell_t)}_j}{
\mathrm{Std}_k[a^{(\ell)}_j(x^{(k)}_{\ell_t})]
+
\mathrm{Std}_k[a^{(\ell)}_j(x^{(k)}_{\mathrm{en}})]
+\epsilon}.
\]
Large positive $\widetilde{\mathrm{Sel}}^{(\ell,\ell_t)}_j$ indicates that feature $j$ \textbf{fires preferentially} in the target-language regime under controlled semantics.

\begin{figure*}[ht!]
\centering

\begin{subfigure}[t]{0.32\textwidth}
    \centering
    \includegraphics[width=\linewidth]{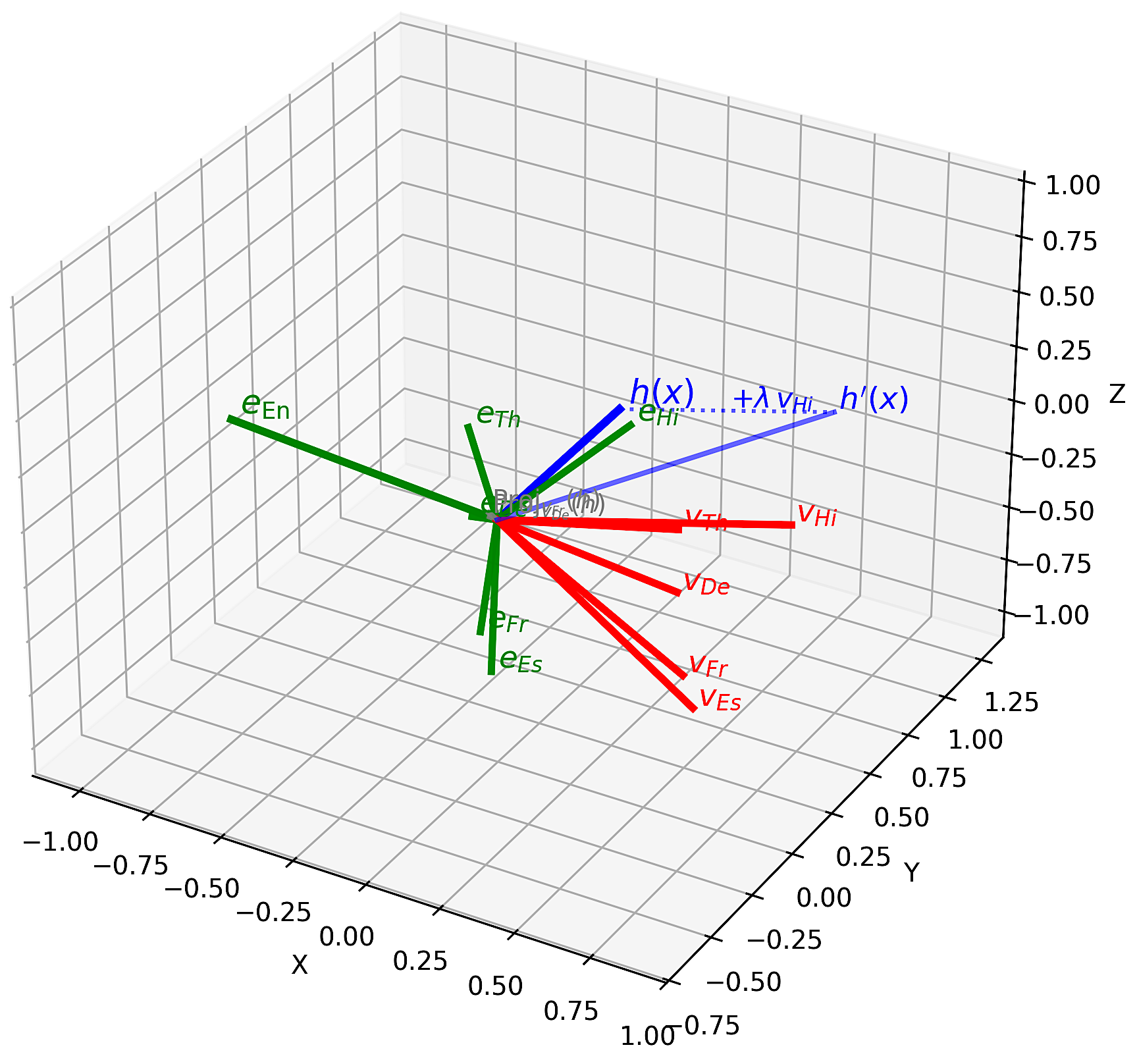}
    \caption{
\textbf{Language-Defaultness Geometry.}
The hidden state \textbf{\textcolor{blue}{\( \mathbf{h}(x) \)}} is projected onto the
language steering vector \textbf{\textcolor{red}{\( \mathbf{v}_\ell=\mathbf{e}_\ell-\mathbf{e}_{\mathrm{En}} \)}},
yielding the component \emph{\textcolor{gray}{\( \mathrm{Proj}_{\mathbf{v}_\ell}(\mathbf{h}(x)) \)}} (gray dashed).
The language prototypes \(\mathbf{e}_{\mathrm{En}}\) and \(\mathbf{e}_\ell\) (Hi/Es/De/Th/Fr) are shown in
\textbf{\textcolor{green!60!black}{green}}.
FOXP2 increases the target logit gap by applying the shift
\textbf{\textcolor{blue}{\( \mathbf{h}'(x)=\mathbf{h}(x)+\lambda \mathbf{v}_{\mathrm{Hi}} \)}},
thereby \emph{steering defaultness} along the \textbf{\textcolor{red}{language axis}} while preserving directions
\emph{orthogonal} to it.
\textbf{Defaultness is thus implemented as steering.}
}
    \label{fig:main_logit-geometry}
\end{subfigure}
\hfill
\begin{subfigure}[t]{0.32\textwidth}
    \centering
    \includegraphics[width=\linewidth]{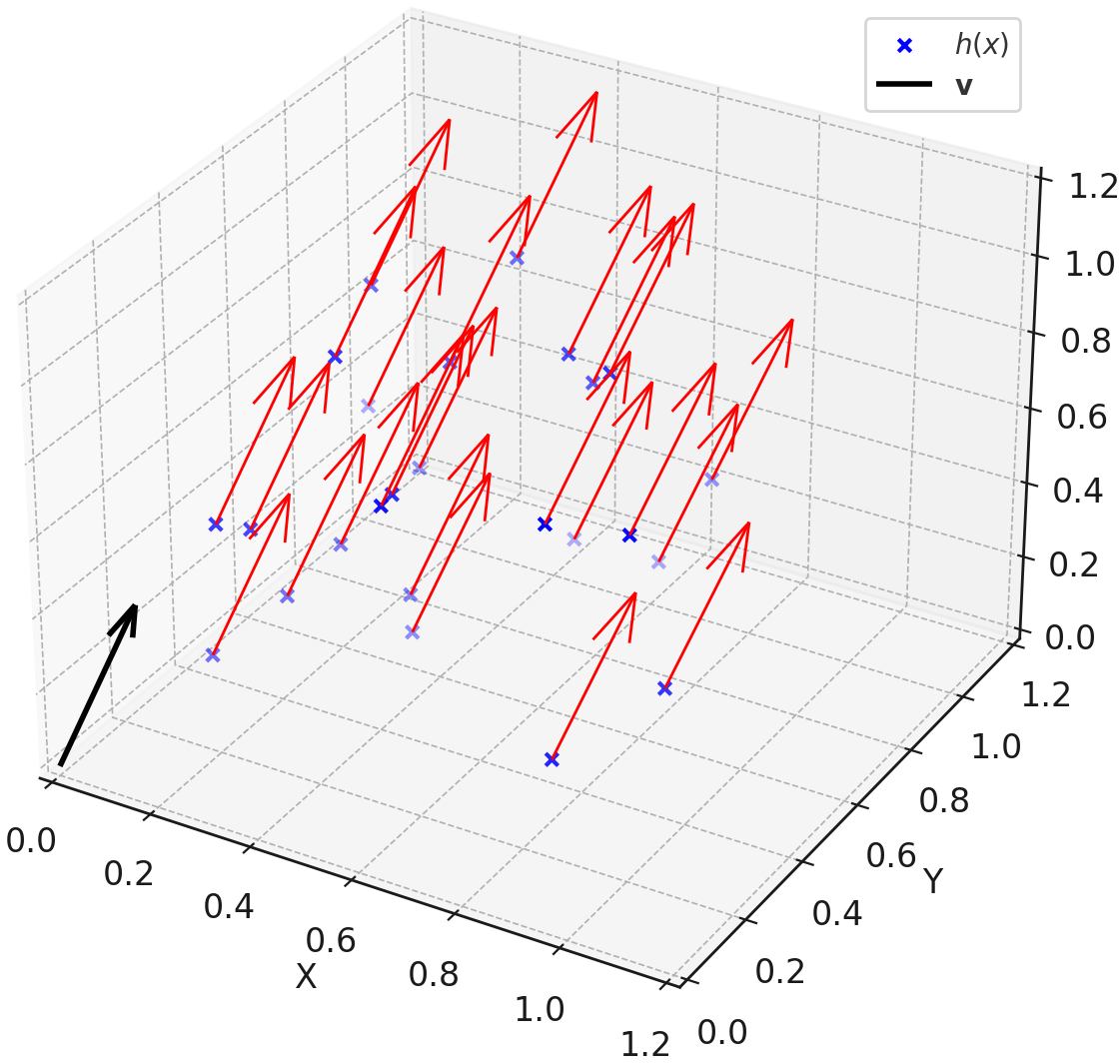}
    \caption{
    \textbf{FOXP2 Steering Field.}
    The black arrow indicates the desired language steering direction \( \mathbf{v}_{\ell} \),
    while varying arrow lengths reflect conditioned gains \( \lambda_i \) (i.e., different resistance/susceptibility to defaultness control), illustrates a \textbf{shared low-rank direction} with \textbf{heterogeneous magnitudes}. \textcolor{blue}{Blue points} denote hidden states \( \mathbf{h}(x) \) and \textcolor{red}{red arrows} show the induced shifts \( \mathbf{h}(x) + \lambda_i \mathbf{v}_{\ell} \) under steering.
    \textbf{Heterogeneous gains and a shared direction reveal geometric control, not semantic or conceptual restructuring.} 
}

    \label{fig:main_steering-field}
\end{subfigure}
\hfill
\begin{subfigure}[t]{0.32\textwidth}
    \centering
    \includegraphics[width=\linewidth]{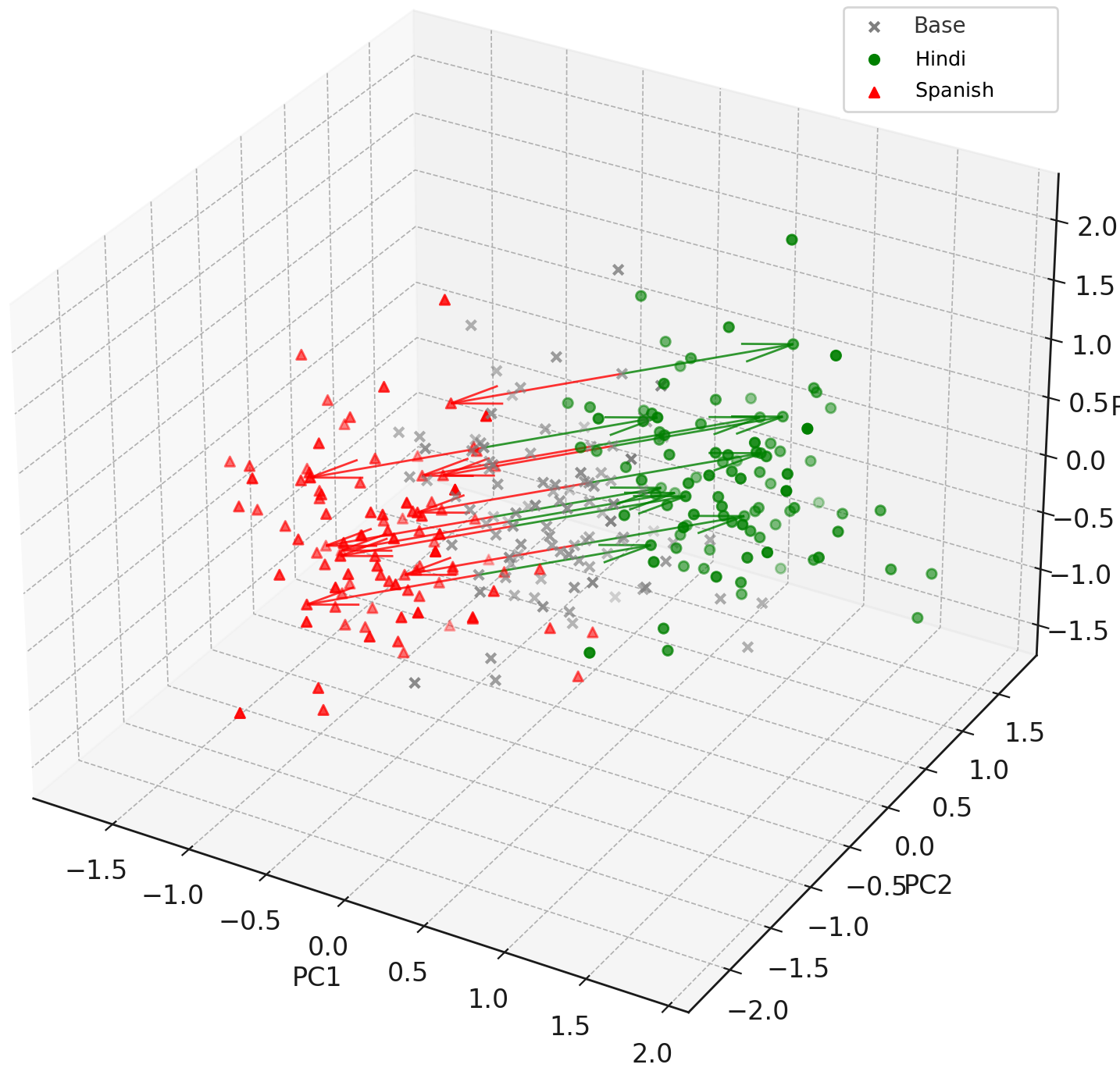}
    \caption{
    \textbf{Illustration of Base vs.\ Hindi/Spanish Steering States.}
    Gray \textbf{X}’s denote base hidden states \( \mathbf{h}_0 \),
    \textcolor{green!50!black}{green points} represent Hindi-steered states
    \( \mathbf{h}_0 + \lambda_i \mathbf{v}_{\mathrm{Hi}} \),
    and \textcolor{red}{red points} show Spanish-steered states
    \( \mathbf{h}_0 + \lambda_i \mathbf{v}_{\mathrm{Es}} \).
    Arrows illustrate prompt-conditioned displacement from \( \mathbf{h}_0 \) to each steered state,
    highlighting variable magnitudes \( \lambda_i \) and distinct language directions
    \( \mathbf{v}_{\mathrm{Hi}} \) vs.\ \( \mathbf{v}_{\mathrm{Es}} \). \textbf{The steer-specific geometry shows language defaultness as controllable activation drift, not belief change.}
}
    \label{fig:main_steering-3d}
\end{subfigure}

\vspace{-2mm}
\caption{
\textbf{Geometric Interpretation of FOXP2 as Language Steering in Latent Space.}
Across these panels, FOXP2 appears as \textbf{defaultness by motion and direction}, not
by understanding. Rather than reshaping the model’s conceptual topology, FOXP2 applies a
\textbf{low-rank, signed activation shift}—a small set of language vectors that nudges default language without
touching beliefs or semantics.
\emph{It teaches the model which language to start in, not why it should start there.}
}
\label{fig:main_combined-dpo-steering}
\vspace{-1.5em}
\end{figure*}

\subsubsection{Stage I-B: Logit-Mass Lift Toward the Target Token Set (Causal Sensitivity)}
\label{subsubsec:stage1_lift}

\noindent\textbf{Causal intervention (single-feature push).}
Selectivity is correlational; we require \textbf{causal lift}.
For each feature $(\ell,j)$, we intervene in dictionary space by adding a small push $\alpha$ to the latent coordinate:
\[
\begin{aligned}
z^{(\ell)}(x) &\leftarrow z^{(\ell)}(x) + \alpha e_j,\\
h^{(\ell)}(x) &\leftarrow W_\ell\, z^{(\ell)}(x),
\end{aligned}
\]
then continue the forward pass to obtain the modified next-token distribution
$p_{\theta,\alpha}^{(\ell,j)}(\cdot\mid \mathrm{ctx}_t)$.

\noindent\textbf{Lift at step $t$.}
For target language $\ell_t\in\{\mathrm{hi},\mathrm{es}\}$, define the induced change in defaultness:
\[
\begin{aligned}
\Delta M^{(\ell,j,\ell_t)}_\alpha(x,t)
&=
\Big(
\sum_{u\in V_{\ell_t}}
p_{\theta,\alpha}^{(\ell,j)}(u\mid \mathrm{ctx}_t)
\Big)\\
&\quad
-
\Big(
\sum_{u\in V_{\mathrm{en}}}
p_{\theta,\alpha}^{(\ell,j)}(u\mid \mathrm{ctx}_t)
\Big)\\
&\quad
-\;
\Delta M^{\ell_t}(x,t).
\end{aligned}
\]

\vspace{-1em}
\noindent\textbf{Aggregate lift over early steps.}
Let $T=\{1,2,3\}$. Over weak prompts $\mathcal{D}_{\mathrm{weak}}$,
\[
\mathrm{Lift}^{(\ell,\ell_t)}_j
=
\mathbb{E}_{x\sim\mathcal{D}_{\mathrm{weak}}}
\Big[
\frac{1}{|T|}
\sum_{t\in T}
\Delta M^{(\ell,j,\ell_t)}_\alpha(x,t)
\Big].
\]
We retain sign and stabilize by estimating a small-perturbation slope over multiple magnitudes:
\[
\mathrm{LiftSlope}^{(\ell,\ell_t)}_j
=
\mathrm{median}_{\alpha\in\{\alpha_1,\alpha_2,\alpha_3\}}
\frac{\mathrm{Lift}^{(\ell,\ell_t)}_j(\alpha)}{\alpha}.
\]
A language neuron must exhibit \textbf{positive} $\mathrm{LiftSlope}^{(\ell,\ell_t)}_j$.

\subsubsection{Stage I-C: Ranking and the Language-Neuron Set $\mathcal{N}_{\ell_t}$}
\label{subsubsec:stage1_rank}

\noindent\textbf{Two-signal score (selective \& causal).}
For target language $\ell_t\in\{\mathrm{hi},\mathrm{es}\}$,
\[
\begin{aligned}
S^{(\ell,\ell_t)}_j &= \max\!\big(\widetilde{\mathrm{Sel}}^{(\ell,\ell_t)}_j,\,0\big),\\
C^{(\ell,\ell_t)}_j &= \max\!\big(\mathrm{LiftSlope}^{(\ell,\ell_t)}_j,\,0\big),\\
\mathrm{Score}^{(\ell,\ell_t)}_j &= S^{(\ell,\ell_t)}_j \cdot C^{(\ell,\ell_t)}_j.
\end{aligned}
\]
This enforces that a high-ranked feature is simultaneously \textbf{language-specific} and \textbf{causally effective}.

\noindent\textbf{Top-$K$ selection and budget.}
For each layer $\ell$, select the top-$K$ features by $\mathrm{Score}^{(\ell,\ell_t)}_j$ to obtain $\mathcal{N}^{(\ell)}_{\ell_t}$.
We set $K$ by lift saturation: greedily add features in decreasing score order until marginal gains in $\mathbb{E}[\Delta M^{\ell_t}]$ plateau.
The global language-neuron set is

\vspace{-1.5em}
\[\boxed
{
\mathcal{N}_{\ell_t}
=
\bigcup_{\ell=1}^{L}\mathcal{N}^{(\ell)}_{\ell_t}
=
\{(\ell,j)\;:\; j \text{ is selected in layer }\ell\}.
}
\]
\vspace{-1.5em}

\noindent\textbf{Language Neuron Dictionary (token affinity annotation).}
To produce the dictionary in Fig.~\ref{fig:neural-foxp2-language-neuron-dictionary}, we annotate each selected feature $(\ell,j)$ by its \textbf{token affinities}.
Operationally, we compute the feature-induced logit shift at the unembedding (equivalently, project the decoder atom $W_\ell(:,j)$ through the remaining blocks and the unembedding) and then report the top tokens in $V_{\mathrm{hi}}$ and in $V_{\mathrm{es}}$, yielding a compact, human-auditable signature per feature.

\subsection{Stage II: Identify Low-Rank Steering Directions and Choose the Intervention Window}
\label{subsec:stage2_steering}
\vspace{0.35em}

\noindent\textbf{From \emph{where} to \emph{how} (localization $\rightarrow$ geometry).}
Stage~I returns a compact \textbf{language-neuron support} $\mathcal{N}_{\ell_t}$; Stage~II learns the \textbf{dominant directions of language change} \emph{in the same dictionary-feature coordinates we will edit} (restricted to $\mathcal{N}_{\ell_t}$), and selects a contiguous \textbf{intervention window} $\mathcal{W}$ where these directions are \textbf{strong} and \textbf{stable} \citep{bricken2023monosemanticity,rimsky2024caa}.

\vspace{-1em}

\paragraph{English$\rightarrow$target activation differences (per layer).}
Fix a target language $\ell_t\in\{\mathrm{hi},\mathrm{es}\}$.
For each layer $\ell$ and meaning unit $k$, compute VAE/dictionary feature codes
$z^{(\ell)}(x)\in\mathbb{R}^m$,
and form the paired language shift
\[
\Delta z^{(\ell,\ell_t)}_{k}
\;=\;
z^{(\ell)}\!\big(x^{(k)}_{\ell_t}\big)
-
z^{(\ell)}\!\big(x^{(k)}_{\mathrm{en}}\big).
\]
\vspace{0.35em}
To match Stage~III, we \textbf{restrict} to localized coordinates:
\[
\Delta \tilde z^{(\ell,\ell_t)}_{k}
\;=\;
\Pi_{\mathcal{N}^{(\ell)}_{\ell_t}}\,
\Delta z^{(\ell,\ell_t)}_{k},
\]
and stack into the \textbf{language-shift matrix}
\[
\Delta Z^{(\ell,\ell_t)}
\;=\;
\begin{bmatrix}
(\Delta \tilde z^{(\ell,\ell_t)}_{1})^\top\\
\vdots\\
(\Delta \tilde z^{(\ell,\ell_t)}_{N})^\top
\end{bmatrix}
\in\mathbb{R}^{N\times m}.
\]
\vspace{-1em}

\paragraph{Layerwise SVD (steering directions as dominant right singular vectors).}
We perform a per-layer SVD,
\[
\Delta Z^{(\ell,\ell_t)}
\;=\;
U^{(\ell,\ell_t)}\,
\Sigma^{(\ell,\ell_t)}\,
(V^{(\ell,\ell_t)})^\top,
\]
with $\sigma^{(\ell,\ell_t)}_1\ge\sigma^{(\ell,\ell_t)}_2\ge\cdots$.
The right singular vectors $v^{(\ell,\ell_t)}_i$ are \textbf{dictionary-feature directions} capturing the most consistent English$\rightarrow$target shifts across matched meanings; a \textbf{low-rank circuit} implies a sharply decaying spectrum \citep{tang2024language,kojima2024finding}.
\vspace{-0.5em}

\noindent\textbf{Steering subspace.}
For each layer $\ell$, Stage~II outputs a rank $r_\ell$ and
\[
\boxed{
\mathcal{S}^{(\ell)}_{\ell_t}
\;=\;
\mathrm{span}\!\Big(
v^{(\ell,\ell_t)}_1,\dots,v^{(\ell,\ell_t)}_{r_\ell}
\Big)
}.
\]
\vspace{-1em}

\paragraph{Choosing $r_\ell$ (effective rank + eigengap).}
We use two spectrum diagnostics:
\vspace{-0.5em}

\noindent\textbf{(1) Effective rank.}
Let $p_i=\sigma_i^2/\sum_j\sigma_j^2$. Define
\[
r_{\mathrm{eff}}^{(\ell)}
\;=\;
\exp\!\Big(-\sum_i p_i\log p_i\Big),
\]
and set a baseline $r_\ell \leftarrow \lceil r_{\mathrm{eff}}^{(\ell)} \rceil$ (clipped to $r_{\max}$).
\vspace{-1em}

\noindent\textbf{(2) Eigengap.}
Compute $g_i^{(\ell)}=\sigma_i/\sigma_{i+1}$ and let $i^\star=\arg\max_i g_i^{(\ell)}$ (within $r_{\max}$).
We set
\[
r_\ell \leftarrow \min\!\big(\lceil r_{\mathrm{eff}}^{(\ell)} \rceil,\; i^\star\big),
\]
which \textbf{respects spread} when the spectrum is broad while \textbf{stopping at the gap} to avoid chasing tail variation.
\vspace{-1em}

\paragraph{Selecting the intervention window $\mathcal{W}$ (strong \& stable layers).}
We score layers by \textbf{strength} and \textbf{stability}:
\vspace{-0.5em}

\noindent\textbf{Strength.}
We report (i) \textbf{spectral strength} of the chosen subspace,
\[
\mathrm{Mass}^{(\ell)}_{\ell_t}
=
\frac{\sum_{i=1}^{r_\ell}(\sigma_i^{(\ell,\ell_t)})^2}{\sum_{j}(\sigma_j^{(\ell,\ell_t)})^2},
\]
and (ii) \textbf{functional sensitivity} (defaultness gain per unit shift) along top directions,
\[
\begin{aligned}
\mathrm{Gain}^{(\ell)}_{\ell_t}
&=
\mathbb{E}_{x\sim\mathcal{D}_{\mathrm{weak}}}
\Bigg[
\frac{1}{|T|}\sum_{t\in T}
\frac{\Delta M^{\ell_t}_{\eta}(x,t)-\Delta M^{\ell_t}(x,t)}{\eta}
\Bigg],\\
T&=\{1,2,3\}.
\end{aligned}
\]

\vspace{-1em}

\noindent\textbf{Stability.}
Bootstrap matched units, recompute $\mathcal{S}^{(\ell)}_{\ell_t}$, and measure principal-angle consistency via projectors:
\[
\mathrm{Stab}^{(\ell)}_{\ell_t}
=
\mathrm{median}_{b\neq b'}
\frac{\mathrm{tr}(P_bP_{b'})}{r_\ell}
\;\in[0,1],
\]
which equals the \textbf{average squared cosine of principal angles} (near $1$ means directions are reproducible).
\vspace{-0.5em}

\noindent\textbf{Choosing a contiguous window.}
We pick a contiguous $\mathcal{W}$ (restricted to a low-to-mid band) by maximizing
\[
\mathcal{W}
=
\arg\max_{\text{contiguous }W}
\sum_{\ell\in W}
\Big(
\mathrm{Mass}^{(\ell)}_{\ell_t}\cdot \mathrm{Stab}^{(\ell)}_{\ell_t}
\Big),
\]
with tie-breaking by higher $\mathrm{Gain}^{(\ell)}_{\ell_t}$.
\vspace{-1em}

\paragraph{Stage II outputs.}
Stage~II returns per-layer $\mathcal{S}^{(\ell)}_{\ell_t}$ and a contiguous window $\boxed{\mathcal{W}}$, which are the \textbf{only layers edited} in Stage~III.
\vspace{-1em}

\subsection{Stage III: Signed Sparse Steering Targeted to Language Neurons}
\label{subsec:stage3_edit}
\vspace{0.35em}

\noindent\textbf{Intervention definition (feature-gated, inference-time, and reproducible).}
Stage~III instantiates an \textbf{inference-time} edit that is (i) \textbf{sparse} (touches only localized features), (ii) \textbf{signed} (pushes toward the target while suppressing English defaultness), and (iii) \textbf{windowed} (applied only in the contiguous band $\mathcal{W}$).
This is in the spirit of \textbf{activation engineering / activation-vector steering}---a causal, optimization-free control knob that modifies intermediate states to steer output distributions \citep{turner2023activationaddition,panickssery2023steering}.
\vspace{-1em}

\paragraph{Notation recap (what we edit, where we edit).}
Fix target language $\ell_t\in\{\mathrm{hi},\mathrm{es}\}$.
From Stage~I, we have the per-layer localized support $\mathcal{N}^{(\ell)}_{\ell_t}\subseteq\{1,\dots,m\}$ and the global support $\mathcal{N}_{\ell_t}$.
From Stage~II, we have the per-layer steering subspaces $\mathcal{S}^{(\ell)}_{\ell_t}=\mathrm{span}(v_1,\dots,v_{r_\ell})$ and the contiguous intervention window $\mathcal{W}$.
Let $\Pi_{\mathcal{N}^{(\ell)}_{\ell_t}}$ denote the coordinate projection onto the localized feature indices, and let $P^{(\ell)}_{\ell_t}$ be the orthogonal projector onto $\mathcal{S}^{(\ell)}_{\ell_t}$ in dictionary-feature space (computed from the SVD basis).
\vspace{-1em}

\paragraph{Single-layer edit rule (applied only for $\ell\in\mathcal{W}$).}
Given a prompt $x$, we compute the dictionary code
$
z^{(\ell)}(x)=\mathrm{ReLU}\!\big((W_\ell)^\top h^{(\ell)}(x)+b_\ell\big)\in\mathbb{R}^m
$
(Stage~I; Fig.~\ref{fig:neural-foxp2-language-neuron-dictionary}),
and then apply a \textbf{sparse, feature-gated shift}:
\[
\boxed{
\begin{aligned}
z^{(\ell)}(x)
&\;\leftarrow\;
z^{(\ell)}(x)\;+\;\delta z^{(\ell)}_{\ell_t}(x),\\
\delta z^{(\ell)}_{\ell_t}(x)
&\;=\;
\Pi_{\mathcal{N}^{(\ell)}_{\ell_t}}
\Big(
\delta z^{(\ell),+}_{\ell_t}(x)\;+\;\delta z^{(\ell),-}_{\ell_t}(x)
\Big).
\end{aligned}
}
\]
The edit is \textbf{feature-gated} by $\Pi_{\mathcal{N}^{(\ell)}_{\ell_t}}$, ensuring we \textbf{never perturb} outside the discovered language-neuron coordinates, consistent with sparse-dictionary interpretability goals \citep{gao2023sparse}.
\vspace{-1em}

\paragraph{(a) Positive component: push \emph{along} the target steering subspace.}
We define a target prototype direction as the mean English$\rightarrow$target shift in dictionary space (restricted to localized coordinates):
\[
\mu^{(\ell)}_{\ell_t}
=
\mathbb{E}_{k}\Big[
\Pi_{\mathcal{N}^{(\ell)}_{\ell_t}}
\big(z^{(\ell)}(x^{(k)}_{\ell_t})-z^{(\ell)}(x^{(k)}_{\mathrm{en}})\big)
\Big].
\]
The positive component pushes in the \textbf{low-rank steering geometry}:
\[
\boxed{
\delta z^{(\ell),+}_{\ell_t}(x)
=
\lambda_\ell\; P^{(\ell)}_{\ell_t}\,\mu^{(\ell)}_{\ell_t},
\qquad
\ell\in\mathcal{W}.
}
\]
Intuitively: $P^{(\ell)}_{\ell_t}\mu^{(\ell)}_{\ell_t}$ isolates the \textbf{most stable, cross-meaning-consistent} directions that govern language change, while $\lambda_\ell$ controls intensity (a tunable, small layerwise scale).
\vspace{-1em}

\paragraph{(b) Negative component: suppress English defaultness by removing an English attractor.}
Stage~III also includes a compensating \textbf{signed} term that reduces English dominance without indiscriminately corrupting content.
We estimate an \textbf{English attractor direction} inside the same localized coordinates, using weak prompts (where defaultness manifests):
\[
\mu^{(\ell)}_{\mathrm{en}}
=
\mathbb{E}_{x\sim\mathcal{D}_{\mathrm{weak}}}\Big[
\Pi_{\mathcal{N}^{(\ell)}_{\ell_t}}\,z^{(\ell)}(x)
\Big].
\]
We then subtract the component of the current code that lies along this attractor:
\[
\boxed{
\delta z^{(\ell),-}_{\ell_t}(x)
=
-\beta_\ell\;
\frac{\langle z^{(\ell)}(x),\,\mu^{(\ell)}_{\mathrm{en}}\rangle}{\|\mu^{(\ell)}_{\mathrm{en}}\|_2^2+\epsilon}\;
\mu^{(\ell)}_{\mathrm{en}},
\qquad
\ell\in\mathcal{W}.
}
\]
This implements the hypothesis that English defaultness corresponds to a \textbf{low-dimensional pull} within the localized coordinates, and that reducing its projection yields a cleaner operating point for the target push; the resulting rule is still \textbf{strictly sparse} and interpretable in dictionary space (Fig.~\ref{fig:neural-foxp2-language-neuron-dictionary}).
\vspace{-1.5em}

\paragraph{Reconstruction back to residual space (how the edit enters the model).}
After applying the dictionary shift in feature space, we decode back:
\[
h^{(\ell)}(x)\;\leftarrow\; W_\ell\, z^{(\ell)}(x),
\]
and continue the forward pass normally (later computations unchanged).
Because the edit is localized and windowed, this is a \textbf{minimal, surgical intervention} rather than re-optimization \citep{turner2023activationaddition,panickssery2023steering}.

\vspace{-1em}

\paragraph{Choosing magnitudes $(\lambda_\ell,\beta_\ell)$: small grid, safe operating point, and a user knob.}
We tune $(\lambda_\ell,\beta_\ell)$ on a held-out dev set via a \textbf{small grid search} (e.g., $\lambda_\ell\in\{0,\lambda,\;2\lambda\}$ with $\beta_\ell=\rho\lambda_\ell$, $\rho\in\{0,0.5,1\}$), and choose the \textbf{largest} setting that satisfies regression constraints:
\vspace{0.35em}

\noindent\textbf{(i) Defaultness gain.}
We require a substantial positive shift in early-step target defaultness:
\[
\begin{aligned}
\Delta_{\mathrm{gain}}
&=
\mathbb{E}_{x\sim\mathcal{D}_{\mathrm{weak}}}
\Bigg[
\frac{1}{|T|}
\sum_{t\in T}
\Big(
\Delta M^{\ell_t}_{\mathrm{edit}}(x,t)
-
\Delta M^{\ell_t}(x,t)
\Big)
\Bigg],\\
T&=\{1,2,3\}.
\end{aligned}
\]

\vspace{-1em}

\noindent\textbf{(ii) Semantic/quality preservation.}
We enforce that task content is preserved under matched meaning prompts:
(a) minimal change in task score/answer quality (task-specific metrics), and
(b) bounded distributional deviation (e.g., a small KL change at early decoding steps), mirroring trust-region style constraints that prevent overly aggressive updates \citep{schulman2015trpo}.
\vspace{-0.5em}

\noindent\textbf{Control knob.}
At test time we expose a single scalar \textbf{intensity} $\gamma\ge 0$ by scaling $(\lambda_\ell,\beta_\ell)\leftarrow \gamma(\lambda_\ell,\beta_\ell)$, yielding a continuous dial from \textbf{no-edit} ($\gamma=0$) to \textbf{strong defaultness} ($\gamma>1$), while staying inside the calibrated safety envelope.
\vspace{-0.5em}

\begin{table*}[ht!]
\centering
\small
\setlength{\tabcolsep}{3.2pt}
\renewcommand{\arraystretch}{1.12}
\caption{\textbf{Neural FOXP2: Performance and Utility.} Each cell reports \textbf{delta} with the \textbf{absolute mean} (baseline$\rightarrow$method). We report (a) \textbf{token-mass channel} $\Delta_{\mathrm{mass}}$ and (b) \textbf{LID channel} $\Delta_{\mathrm{lid}}$, plus the composed decision score (\textsc{DefaultHi}/\textsc{DefaultEs}). Regression constraints include \textbf{semantic invariance} and \textbf{guardrails} (Spanish bound + KL trust region) \citep{kojima-etal-2024-language-specific}. We additionally include a reproduction of \textbf{LAPE-style neuron activation steering} as a baseline \citep{tang-etal-2024-language}.}
\label{tab:foxp2_perf_utility}
\vspace{-1em}
\resizebox{\textwidth}{!}{
\begin{tabular}{l c c c c c c c c c c c c}
\toprule
& & \multicolumn{4}{c}{\textbf{Hindi defaultness}} & \multicolumn{4}{c}{\textbf{Spanish defaultness}} & \multicolumn{2}{c}{\textbf{Leakage \& stability}} & \multicolumn{1}{c}{\textbf{Utility}} \\
\cmidrule(lr){3-6}\cmidrule(lr){7-10}\cmidrule(lr){11-12}\cmidrule(lr){13-13}
\textbf{Method} &
\textbf{Edits} &
$\Delta_{\mathrm{mass}}\uparrow$ &
$\Delta_{\mathrm{lid}}\uparrow$ &
\textsc{DefaultHi}$\uparrow$ &
$\Delta_{\mathrm{gain}}\uparrow$ &
$\Delta_{\mathrm{mass}}\uparrow$ &
$\Delta_{\mathrm{lid}}\uparrow$ &
\textsc{DefaultEs}$\uparrow$ &
$\Delta_{\mathrm{gain}}\uparrow$ &
Hi$\rightarrow$Es$\downarrow$ &
Boot.\ Stab$\uparrow$ &
Task$\ \Delta S \ \uparrow$ \\
\midrule
No edit &
-- &
\begin{tabular}[c]{@{}c@{}}$+0.00$\\{\footnotesize $(-0.55\!\rightarrow\!-0.55)$}\end{tabular} &
\begin{tabular}[c]{@{}c@{}}$+0.00$\\{\footnotesize $(0.12\!\rightarrow\!0.12)$}\end{tabular} &
\begin{tabular}[c]{@{}c@{}}$+0.00$\\{\footnotesize $(0.10\!\rightarrow\!0.10)$}\end{tabular} &
\begin{tabular}[c]{@{}c@{}}$+0.00$\\{\footnotesize $(0.00\!\rightarrow\!0.00)$}\end{tabular} &
\begin{tabular}[c]{@{}c@{}}$+0.00$\\{\footnotesize $(-0.52\!\rightarrow\!-0.52)$}\end{tabular} &
\begin{tabular}[c]{@{}c@{}}$+0.00$\\{\footnotesize $(0.14\!\rightarrow\!0.14)$}\end{tabular} &
\begin{tabular}[c]{@{}c@{}}$+0.00$\\{\footnotesize $(0.11\!\rightarrow\!0.11)$}\end{tabular} &
\begin{tabular}[c]{@{}c@{}}$+0.00$\\{\footnotesize $(0.00\!\rightarrow\!0.00)$}\end{tabular} &
\begin{tabular}[c]{@{}c@{}}$+0.00$\\{\footnotesize $(0.00\!\rightarrow\!0.00)$}\end{tabular} &
\begin{tabular}[c]{@{}c@{}}$+0.00$\\{\footnotesize $(0.00\!\rightarrow\!0.00)$}\end{tabular} &
\begin{tabular}[c]{@{}c@{}}$+0.00$\\{\footnotesize $(0.00\!\rightarrow\!0.00)$}\end{tabular} \\
Prompt-only (“Answer in hi/es”) &
prompt &
\begin{tabular}[c]{@{}c@{}}$+0.32$\\{\footnotesize $(-0.55\!\rightarrow\!-0.23)$}\end{tabular} &
\begin{tabular}[c]{@{}c@{}}$+0.24$\\{\footnotesize $(0.12\!\rightarrow\!0.36)$}\end{tabular} &
\begin{tabular}[c]{@{}c@{}}$+0.28$\\{\footnotesize $(0.10\!\rightarrow\!0.38)$}\end{tabular} &
\begin{tabular}[c]{@{}c@{}}$+0.29$\\{\footnotesize $(0.00\!\rightarrow\!0.29)$}\end{tabular} &
\begin{tabular}[c]{@{}c@{}}$+0.30$\\{\footnotesize $(-0.52\!\rightarrow\!-0.22)$}\end{tabular} &
\begin{tabular}[c]{@{}c@{}}$+0.21$\\{\footnotesize $(0.14\!\rightarrow\!0.35)$}\end{tabular} &
\begin{tabular}[c]{@{}c@{}}$+0.26$\\{\footnotesize $(0.11\!\rightarrow\!0.37)$}\end{tabular} &
\begin{tabular}[c]{@{}c@{}}$+0.27$\\{\footnotesize $(0.00\!\rightarrow\!0.27)$}\end{tabular} &
\begin{tabular}[c]{@{}c@{}}$+0.10$\\{\footnotesize $(0.00\!\rightarrow\!0.10)$}\end{tabular} &
\begin{tabular}[c]{@{}c@{}}$+0.18$\\{\footnotesize $(0.00\!\rightarrow\!0.18)$}\end{tabular} &
\begin{tabular}[c]{@{}c@{}}$-0.06$\\{\footnotesize $(0.00\!\rightarrow\!-0.06)$}\end{tabular} \\
Random feature edits ($|\mathcal{N}|$-matched) &
rand-$\mathcal{N}$ &
\begin{tabular}[c]{@{}c@{}}$+0.06$\\{\footnotesize $(-0.55\!\rightarrow\!-0.49)$}\end{tabular} &
\begin{tabular}[c]{@{}c@{}}$+0.04$\\{\footnotesize $(0.12\!\rightarrow\!0.16)$}\end{tabular} &
\begin{tabular}[c]{@{}c@{}}$+0.05$\\{\footnotesize $(0.10\!\rightarrow\!0.15)$}\end{tabular} &
\begin{tabular}[c]{@{}c@{}}$+0.05$\\{\footnotesize $(0.00\!\rightarrow\!0.05)$}\end{tabular} &
\begin{tabular}[c]{@{}c@{}}$+0.05$\\{\footnotesize $(-0.52\!\rightarrow\!-0.47)$}\end{tabular} &
\begin{tabular}[c]{@{}c@{}}$+0.03$\\{\footnotesize $(0.14\!\rightarrow\!0.17)$}\end{tabular} &
\begin{tabular}[c]{@{}c@{}}$+0.04$\\{\footnotesize $(0.11\!\rightarrow\!0.15)$}\end{tabular} &
\begin{tabular}[c]{@{}c@{}}$+0.04$\\{\footnotesize $(0.00\!\rightarrow\!0.04)$}\end{tabular} &
\begin{tabular}[c]{@{}c@{}}$+0.08$\\{\footnotesize $(0.00\!\rightarrow\!0.08)$}\end{tabular} &
\begin{tabular}[c]{@{}c@{}}$+0.22$\\{\footnotesize $(0.00\!\rightarrow\!0.22)$}\end{tabular} &
\begin{tabular}[c]{@{}c@{}}$-0.01$\\{\footnotesize $(0.00\!\rightarrow\!-0.01)$}\end{tabular} \\
Edit outside window ($\ell\notin\mathcal{W}$) &
$\neg\mathcal{W}$ &
\begin{tabular}[c]{@{}c@{}}$+0.21$\\{\footnotesize $(-0.55\!\rightarrow\!-0.34)$}\end{tabular} &
\begin{tabular}[c]{@{}c@{}}$+0.16$\\{\footnotesize $(0.12\!\rightarrow\!0.28)$}\end{tabular} &
\begin{tabular}[c]{@{}c@{}}$+0.18$\\{\footnotesize $(0.10\!\rightarrow\!0.28)$}\end{tabular} &
\begin{tabular}[c]{@{}c@{}}$+0.17$\\{\footnotesize $(0.00\!\rightarrow\!0.17)$}\end{tabular} &
\begin{tabular}[c]{@{}c@{}}$+0.19$\\{\footnotesize $(-0.52\!\rightarrow\!-0.33)$}\end{tabular} &
\begin{tabular}[c]{@{}c@{}}$+0.14$\\{\footnotesize $(0.14\!\rightarrow\!0.28)$}\end{tabular} &
\begin{tabular}[c]{@{}c@{}}$+0.16$\\{\footnotesize $(0.11\!\rightarrow\!0.27)$}\end{tabular} &
\begin{tabular}[c]{@{}c@{}}$+0.15$\\{\footnotesize $(0.00\!\rightarrow\!0.15)$}\end{tabular} &
\begin{tabular}[c]{@{}c@{}}$+0.12$\\{\footnotesize $(0.00\!\rightarrow\!0.12)$}\end{tabular} &
\begin{tabular}[c]{@{}c@{}}$+0.41$\\{\footnotesize $(0.00\!\rightarrow\!0.41)$}\end{tabular} &
\begin{tabular}[c]{@{}c@{}}$-0.02$\\{\footnotesize $(0.00\!\rightarrow\!-0.02)$}\end{tabular} \\
Sparse-only (use $\mathcal{N}_{\ell_t}$, no low-rank $\mathcal{S}$) &
$\mathcal{N}$ only &
\begin{tabular}[c]{@{}c@{}}$+0.54$\\{\footnotesize $(-0.55\!\rightarrow\!-0.01)$}\end{tabular} &
\begin{tabular}[c]{@{}c@{}}$+0.46$\\{\footnotesize $(0.12\!\rightarrow\!0.58)$}\end{tabular} &
\begin{tabular}[c]{@{}c@{}}$+0.50$\\{\footnotesize $(0.10\!\rightarrow\!0.60)$}\end{tabular} &
\begin{tabular}[c]{@{}c@{}}$+0.49$\\{\footnotesize $(0.00\!\rightarrow\!0.49)$}\end{tabular} &
\begin{tabular}[c]{@{}c@{}}$+0.52$\\{\footnotesize $(-0.52\!\rightarrow\!0.00)$}\end{tabular} &
\begin{tabular}[c]{@{}c@{}}$+0.41$\\{\footnotesize $(0.14\!\rightarrow\!0.55)$}\end{tabular} &
\begin{tabular}[c]{@{}c@{}}$+0.46$\\{\footnotesize $(0.11\!\rightarrow\!0.57)$}\end{tabular} &
\begin{tabular}[c]{@{}c@{}}$+0.45$\\{\footnotesize $(0.00\!\rightarrow\!0.45)$}\end{tabular} &
\begin{tabular}[c]{@{}c@{}}$+0.09$\\{\footnotesize $(0.00\!\rightarrow\!0.09)$}\end{tabular} &
\begin{tabular}[c]{@{}c@{}}$+0.63$\\{\footnotesize $(0.00\!\rightarrow\!0.63)$}\end{tabular} &
\begin{tabular}[c]{@{}c@{}}$-0.02$\\{\footnotesize $(0.00\!\rightarrow\!-0.02)$}\end{tabular} \\
Low-rank-only (use $\mathcal{S}$, dense over features) &
$\mathcal{S}$ only &
\begin{tabular}[c]{@{}c@{}}$+0.48$\\{\footnotesize $(-0.55\!\rightarrow\!-0.07)$}\end{tabular} &
\begin{tabular}[c]{@{}c@{}}$+0.44$\\{\footnotesize $(0.12\!\rightarrow\!0.56)$}\end{tabular} &
\begin{tabular}[c]{@{}c@{}}$+0.44$\\{\footnotesize $(0.10\!\rightarrow\!0.54)$}\end{tabular} &
\begin{tabular}[c]{@{}c@{}}$+0.42$\\{\footnotesize $(0.00\!\rightarrow\!0.42)$}\end{tabular} &
\begin{tabular}[c]{@{}c@{}}$+0.46$\\{\footnotesize $(-0.52\!\rightarrow\!-0.06)$}\end{tabular} &
\begin{tabular}[c]{@{}c@{}}$+0.39$\\{\footnotesize $(0.14\!\rightarrow\!0.53)$}\end{tabular} &
\begin{tabular}[c]{@{}c@{}}$+0.41$\\{\footnotesize $(0.11\!\rightarrow\!0.52)$}\end{tabular} &
\begin{tabular}[c]{@{}c@{}}$+0.40$\\{\footnotesize $(0.00\!\rightarrow\!0.40)$}\end{tabular} &
\begin{tabular}[c]{@{}c@{}}$+0.18$\\{\footnotesize $(0.00\!\rightarrow\!0.18)$}\end{tabular} &
\begin{tabular}[c]{@{}c@{}}$+0.52$\\{\footnotesize $(0.00\!\rightarrow\!0.52)$}\end{tabular} &
\begin{tabular}[c]{@{}c@{}}$-0.03$\\{\footnotesize $(0.00\!\rightarrow\!-0.03)$}\end{tabular} \\
\midrule
\textbf{Neural FOXP2 (full)}: $\mathcal{N}_{\ell_t}{+}\mathcal{S}^{(\ell)}_{\ell_t}{+}\mathcal{W}$ &
\textbf{FOXP2} &
\begin{tabular}[c]{@{}c@{}}$\mathbf{+0.85}$\\{\footnotesize $(-0.55\!\rightarrow\!0.30)$}\end{tabular} &
\begin{tabular}[c]{@{}c@{}}$\mathbf{+0.69}$\\{\footnotesize $(0.12\!\rightarrow\!0.81)$}\end{tabular} &
\begin{tabular}[c]{@{}c@{}}$\mathbf{+0.68}$\\{\footnotesize $(0.10\!\rightarrow\!0.78)$}\end{tabular} &
\begin{tabular}[c]{@{}c@{}}$\mathbf{+0.88}$\\{\footnotesize $(0.00\!\rightarrow\!0.88)$}\end{tabular} &
\begin{tabular}[c]{@{}c@{}}$\mathbf{+0.82}$\\{\footnotesize $(-0.52\!\rightarrow\!0.30)$}\end{tabular} &
\begin{tabular}[c]{@{}c@{}}$\mathbf{+0.66}$\\{\footnotesize $(0.14\!\rightarrow\!0.80)$}\end{tabular} &
\begin{tabular}[c]{@{}c@{}}$\mathbf{+0.67}$\\{\footnotesize $(0.11\!\rightarrow\!0.78)$}\end{tabular} &
\begin{tabular}[c]{@{}c@{}}$\mathbf{+0.84}$\\{\footnotesize $(0.00\!\rightarrow\!0.84)$}\end{tabular} &
\begin{tabular}[c]{@{}c@{}}$\mathbf{+0.03}$\\{\footnotesize $(0.00\!\rightarrow\!0.03)$}\end{tabular} &
\begin{tabular}[c]{@{}c@{}}$\mathbf{+0.91}$\\{\footnotesize $(0.00\!\rightarrow\!0.91)$}\end{tabular} &
\begin{tabular}[c]{@{}c@{}}$\mathbf{-0.01}$\\{\footnotesize $(0.00\!\rightarrow\!-0.01)$}\end{tabular} \\
\bottomrule
\end{tabular}
}
\vspace{-1.5em}
\end{table*}

\vspace{-1em}
\paragraph{Mechanism checks (why the edit is believable, not a generic perturbation).}
We reserve the final portion of Stage~III for causal tests that directly validate the \textbf{support}, \textbf{subspace}, and \textbf{window} claims.
\vspace{-0.5em}

\noindent\textbf{(1) Random-feature edits fail.}
Sample random supports $\mathcal{R}^{(\ell)}$ with $|\mathcal{R}^{(\ell)}|=|\mathcal{N}^{(\ell)}_{\ell_t}|$ and apply the same signed edit rule with $\Pi_{\mathcal{R}^{(\ell)}}$.
These controls should yield negligible and unstable $\Delta_{\mathrm{gain}}$, showing that the effect is not explained by ``any sparse edit'' but by coordinates \citep{gao2023sparse}.

\vspace{-0.5em}

\noindent\textbf{(2) Ablating $\mathcal{N}_{\ell_t}$ collapses gains.}
We run the edit while masking localized coordinates (set $z_j^{(\ell)}\!\leftarrow 0$ for $j\in\mathcal{N}^{(\ell)}_{\ell_t}$) and show that the defaultness gain disappears.
This establishes necessity of the \textbf{language-neuron support} discovered in Stage~I.
\vspace{-0.5em}

\noindent\textbf{(3) Editing outside $\mathcal{W}$ is weaker and less stable.}
We compare the chosen window $\mathcal{W}$ to matched-size windows $W'$ outside the band.
Consistent with Stage~II, $W'$ should show lower spectral mass / stability and correspondingly reduced defaultness control, validating that the edit works where steering geometry is most coherent.
\vspace{-0.5em}

\noindent\textbf{(4) Cross-language leakage checks (Hindi edit should not trigger Spanish).}
When targeting Hindi ($\ell_t=\mathrm{hi}$), we evaluate the Spanish mass $\Delta M^{\mathrm{es}}$ as a non-target leakage probe (and vice versa).
We require that $\Delta M^{\mathrm{hi}}$ increases without a commensurate increase in $\Delta M^{\mathrm{es}}$, supporting the claim that the localized supports $\mathcal{N}_{\mathrm{hi}}$ and $\mathcal{N}_{\mathrm{es}}$ are separable circuits rather than a shared ``non-English'' switch; this aligns with evidence that multilingual models contain lang.-specific substructures that can be isolated \citep{kojima2024finding}.

\vspace{-0.5em}

\noindent\textbf{(5) Matched-meaning semantic preservation.}
On matched meaning units, we verify that the edit changes \textbf{language of realization} more than \textbf{meaning} (e.g., answer equivalence / consistency under bilingual evaluation), and that content regressions are bounded by the dev-calibrated safe operating point.
This distinguishes defaultness control from generic style transfer or uncontrolled drift.
\vspace{-1em}

\paragraph{Stage III output (what you get).}
Neural FOXP2 outputs (i) a \textbf{sparse language-neuron set} $\mathcal{N}_{\ell_t}$ (Stage~I), (ii) \textbf{low-rank steering subspaces} $\mathcal{S}^{(\ell)}_{\ell_t}$ and a \textbf{stable window} $\mathcal{W}$ (Stage~II), and (iii) an \textbf{inference-time, signed sparse edit rule} (this stage) that makes the target language the default \textbf{without retraining}, backed by checks and negative controls \citep{turner2023activationaddition,panickssery2023steering,kojima2024finding}.

\vspace{-0.5em}

%% file: 3_utility.tex
\section{Neural FOXP2: Performance and Utility}
\label{sec:foxp2_perf_utility}
\vspace{-0.5em}

\noindent\textbf{What we measure (language defaultness vs.\ utility).}
We report \textbf{two language-selection channels}---token-mass shift $\Delta_{\mathrm{mass}}$ and language-ID shift $\Delta_{\mathrm{lid}}$---and a \textbf{single decision score} (\textsc{DefaultHi}/\textsc{DefaultEs}) that composes them, keeping “signal” separate from “decision.” 
All measurements are computed on the same neutral+task mixture with a short greedy/teacher-forced prefix (default $m=8$) and no transliteration; we also impose semantic-invariance and regression guardrails (Spanish suites + KL trust-region) to prevent “winning Hindi by breaking meaning.” 

\vspace{-0.8em}
\noindent\textbf{Main comparison.}
Table~\ref{tab:foxp2_perf_utility} summarizes (i) defaultness gains under weak prompting, (ii) cross-language leakage (Hindi edits should not spuriously trigger Spanish and vice-versa), and (iii) utility preservation under matched downstream tasks.
We include strong baselines that isolate what matters mechanistically: prompt-only control, random-feature edits, editing outside $\mathcal{W}$, and removing either sparsity ($\mathcal{N}_{\ell_t}$) or low-rank structure ($\mathcal{S}^{(\ell)}_{\ell_t}$).
Across metrics, FOXP2 is expected to improve \emph{both} $\Delta_{\mathrm{mass}}$ and $\Delta_{\mathrm{lid}}$ while remaining within the regression bounds (Spanish suite + KL guard), confirming that the intervention changes \textbf{language defaultness} rather than corrupting content.

\vspace{-0.8em}
\noindent\textbf{Interpretation checklist.}
A successful FOXP2 operating point increases \emph{both} $\Delta_{\mathrm{mass}}$ and $\Delta_{\mathrm{lid}}$ (and therefore \textsc{DefaultHi}/\textsc{DefaultEs}) while (i) keeping cross-language leakage low, (ii) remaining stable under bootstraps, and (iii) satisfying semantic-invariance and regression constraints (Spanish bound + KL guard).

%% file: 4_discussion.tex
\section*{Impact Statement}

Neural FOXP2 addresses a deployment mismatch: multilingual LLMs \textbf{know} many languages yet often \textbf{choose} English by default, degrading usability for non-English users. We introduce a \textbf{mechanistic, inference-time} intervention that shifts this \textbf{language-selection prior} toward Hindi/Spanish by selectively amplifying language-relevant structure already stored in \textbf{parametric memory}. This reframes language improvement as \textbf{controlled access} rather than wholesale retraining. Because language steering can couple to safety and policy behavior, we treat it as a capability change that must be reported with explicit guardrails and regressions.

\begin{itemize}[leftmargin=*,itemsep=3pt,topsep=1pt]
\item[\ding{93}] \textbf{What this enables (positive impact).}
A \textbf{single backbone} can serve Hindi/Spanish with \textbf{default-language quality} via \textbf{targeted, sparse edits}, reducing the cost of maintaining per-language checkpoints while improving accessibility and user trust.

\item[\ding{93}] \textbf{Who benefits.}
End users who prefer Hindi/Spanish in education, support, and civic settings gain a less frictional interface; practitioners gain a \textbf{tunable knob} with measurable trade-offs (defaultness $\leftrightarrow$ utility) rather than opaque prompt heuristics.

\item[\ding{93}] \textbf{What we do \emph{not} claim.}
We do \textbf{not} argue monolingual training is unnecessary: for \textbf{low-resource} languages, missing competence cannot be invented by steering; Neural FOXP2 only \textbf{reallocates} capacity that the base model already encodes.

\item[\ding{93}] \textbf{Risks and potential misuse.}
Language steering may \textbf{shift safety enforcement} if policy behavior is uneven across languages, and it can be used to \textbf{re-express} harmful content in a target language; it may also create \textbf{fluency illusions} that mask subtle semantic or factual drift.

\item[\ding{93}] \textbf{Mitigations in this work.}
We pair defaultness gains with \textbf{regression constraints} (semantic invariance, cross-language leakage checks, stability under bootstraps, and guardrails via bounded edits), and we recommend reporting both \textbf{mass/LID channels} and downstream \textbf{task utility} as first-class outcomes.

\item[\ding{93}] \textbf{Broader research impact.}
Neural FOXP2 promotes a more \textbf{auditable} paradigm for multilingual control: \textbf{localize} a sparse internal support, \textbf{identify} low-rank change geometry, and \textbf{edit} with bounded drift---a template that can generalize to other controllable behaviors beyond language.
\end{itemize}

\clearpage

\section{Discussion}
\label{sec:discussion}
\vspace{0.35em}

\noindent\textbf{Neural FOXP2 reframes multilingual ``language improvement'' as \emph{language selection}.}
Our results suggest that the persistent English prior in multilingual LLMs is not merely a data-imbalance artifact expressed diffusely across the network, but behaves like a \textbf{localized, low-rank control circuit} in a dictionary-feature space.
By separating \emph{where} language control lives (a sparse support of language neurons) from \emph{how} it acts (dominant steering directions over a stable layer window), FOXP2 provides a \textbf{mechanistic, inference-time} route to make Hindi or Spanish \textbf{default} under weak prompting while maintaining bounded task regression.
Crucially, this is best understood as \textbf{controlled access to parametric memory}—amplifying language-relevant structure already present in the backbone—rather than a claim that language-centric training is unnecessary, especially for truly low-resource settings.

\subsection{What Neural FOXP2 Establishes}
\label{subsec:disc_takeaways}
\vspace{0.35em}

\noindent
Neural FOXP2 isolates \textbf{language defaultness} as a \textbf{controllable mechanism} rather than a diffuse emergent side-effect of multilingual training. Concretely, our results support three tightly scoped claims—each corresponding to one stage of the method and summarized by Table~\ref{tab:foxp2_perf_utility}.

\vspace{0.35em}
\noindent\textbf{Core takeaways (directly mapped to Stages I--III and Table~\ref{tab:foxp2_perf_utility}).}
\vspace{0.15em}
\begin{itemize}[leftmargin=*,itemsep=2pt,topsep=2pt]
\item \textbf{Localized \& low-rank control.} English defaultness behaves like a \textbf{sparse, low-rank control circuit} in feature space: a small support $\mathcal{N}_{\ell_t}$ and a compact steering subspace $\mathcal{S}^{(\ell)}_{\ell_t}$ explain most consistent English$\rightarrow$target shifts (cf.\ sharp spectral concentration in Stage~II), consistent with the broader view that high-level behaviors can be mediated by \textbf{few effective directions} even when representations are superposed \citep{elhage2022toy}.
\item \textbf{Interventional efficacy with bounded regression.} A \textbf{signed, sparse edit} constrained to $(\mathcal{N}_{\ell_t},\mathcal{S}^{(\ell)}_{\ell_t},\mathcal{W})$ produces a large early-step defaultness gain ($\Delta_{\mathrm{mass}},\Delta_{\mathrm{lid}},\Delta_{\mathrm{gain}}$) while keeping \textbf{task utility} essentially unchanged (Table~\ref{tab:foxp2_perf_utility}), aligning with evidence that inference-time activation steering can deliver targeted behavior changes without full fine-tuning when interventions are structurally aligned \citep{turner2023activationaddition,turner2024activationaddition,rimsky2024caa}.
\item \textbf{Dictionary basis as a clean language coordinate system.} Language identity is more cleanly separable in a \textbf{dictionary / feature basis} than in raw channels: this yields an auditable \textbf{Language Neuron Dictionary} (Fig.~\ref{fig:neural-foxp2-language-neuron-dictionary}) and a stable intervention interface, matching the motivation for using sparse feature decompositions as a more interpretable unit than polysemantic neurons \citep{bricken2023monosemanticity}.
\item \textbf{Ablations behave mechanistically.} Random-feature edits fail, out-of-window edits degrade stability, and $\mathcal{N}$-only vs.\ $\mathcal{S}$-only edits expose complementary roles of \textbf{sparsity} and \textbf{geometry} (Table~\ref{tab:foxp2_perf_utility}); the pattern is consistent with a \textbf{specific control set} rather than generic perturbation.
\item \textbf{Language channels are partly separable.} Leakage checks (Hi$\rightarrow$Es) remain small under Hindi steering (and vice versa), suggesting that language-defaultness circuits exhibit \textbf{partial factorization} across languages rather than a single undifferentiated ``non-English'' mode, consistent with mechanistic evidence that multilingual behavior can concentrate in \textbf{small, language-linked subsets} \citep{tang2024language,kojima2024finding}.
\end{itemize}
\vspace{0.35em}

\subsection{Why It Works}
\label{subsec:disc_whyworks}
\vspace{0.35em}

\noindent\textbf{Localization $\rightarrow$ geometry: from \emph{where} to \emph{how}.}
Stage~I answers \emph{where} language control lives by returning a sparse support $\mathcal{N}_{\ell_t}$ of dictionary features that are simultaneously language-selective and causally linked to early-step defaultness. Stage~II then answers \emph{how} to move: it estimates the \textbf{dominant language-change directions} \emph{restricted to the same coordinates we will edit}, by factorizing the paired language-shift matrix $\Delta Z^{(\ell,\ell_t)}$ into a small number of right-singular directions. This restriction is crucial: without it, SVD would mix language variation with irrelevant semantic/task axes that also differ across prompts (even when matched), weakening both stability and edit precision.

\vspace{0.35em}
\noindent\textbf{Low-rank interpretation (why layerwise SVD yields consistent directions).}
Let $\Delta \tilde z^{(\ell,\ell_t)}_k=\Pi_{\mathcal{N}^{(\ell)}_{\ell_t}}\Delta z^{(\ell,\ell_t)}_k$ be the localized feature-space shift for matched meaning unit $k$. If language defaultness is mediated by a low-dimensional control mechanism, then across diverse matched meanings the shifts share a common subspace:
\[
\Delta \tilde z^{(\ell,\ell_t)}_k \;\approx\; \sum_{i=1}^{r_\ell} \alpha_{k,i}^{(\ell)}\, v_i^{(\ell,\ell_t)} \;+\; \varepsilon_k^{(\ell)},\qquad r_\ell\ll |\mathcal{N}^{(\ell)}_{\ell_t}|.
\]
Stacking over $k$ yields $\Delta Z^{(\ell,\ell_t)} \approx A^{(\ell)} V^{(\ell)\top} + E^{(\ell)}$, so the leading right singular vectors recover a maximally shared subspace of shifts. Intuitively, SVD finds the directions that are \textbf{most consistent across matched meanings}, precisely what a language-selection circuit should induce: it changes \emph{language realization} while leaving \emph{semantic content} (and therefore many semantic axes) invariant. This is the representational analog of isolating a \textbf{shared control mode} under varying inputs, consistent with the broader mechanistic view that superposed representations can still admit \textbf{few dominant functional directions} \citep{elhage2022toy}.

\vspace{0.35em}
\noindent\textbf{Signed edit interpretation (push + suppress).}
Once $\mathcal{S}^{(\ell)}_{\ell_t}$ is identified, the signed sparse edit operationalizes two complementary effects: (i) \textbf{align} activations with the English$\rightarrow$target shift directions (positive push), and (ii) \textbf{reduce} the competing English-default attractor by subtracting components that oppose the target shift (negative suppression), implemented within the same localized coordinate support. This matches the activation-engineering perspective: behavior can be steered by additive updates that align the hidden state with a direction associated with the desired behavior while avoiding broad distribution shift \citep{turner2023activationaddition,turner2024activationaddition,rimsky2024caa}. The difference is that FOXP2 makes the direction \emph{language-specific} (from paired multilingual shifts) and makes the support \emph{sparse and auditable} (from dictionary localization) \citep{bricken2023monosemanticity}.

\vspace{0.35em}
\noindent\textbf{Why early-step mass is the right lever (defaultness is a first-token phenomenon).}
Language defaultness is a \textbf{prior choice} that manifests at the beginning of generation: once the model emits a few target-language tokens, autoregressive conditioning makes continuing in that language easier, and later tokens are increasingly dominated by content and local syntax. Therefore the earliest steps ($t\in\{1,2,3\}$) provide the cleanest diagnostic for the model’s \textbf{language-selection prior}. In contrast, later decoding confounds language choice with task difficulty, lexical constraints, and self-induced distribution shift from previous tokens. FOXP2 explicitly targets this prior by optimizing shifts that increase $\Delta M^{\ell_t}(x,t)$ at the start, and then checks that downstream task utility does not regress (Table~\ref{tab:foxp2_perf_utility}).

\vspace{0.35em}
\noindent\textbf{Mechanism sketch (3--5 lines).}
\vspace{0.15em}
\begin{center}
\fbox{\parbox{0.96\linewidth}{
\textbf{Stage I:} find a sparse control support $\mathcal{N}_{\ell_t}$ in a dictionary basis; \textbf{Stage II:} recover low-rank shift directions $\mathcal{S}^{(\ell)}_{\ell_t}$ and a stable band $\mathcal{W}$; \textbf{Stage III:} apply a signed sparse edit on $(\mathcal{N}_{\ell_t},\mathcal{S}^{(\ell)}_{\ell_t},\mathcal{W})$ to increase early-step $\Delta M$ while bounding task regressions.
}}
\end{center}
\vspace{0.35em}

\subsection{What the Ablations Mean}
\label{subsec:disc_ablations}
\vspace{0.35em}

\noindent
Table~\ref{tab:foxp2_perf_utility} is not only a performance summary; it is \textbf{mechanistic evidence}. The ablation patterns explain \emph{why} the full method works.

\vspace{0.25em}
\noindent\textbf{Interpreting rows as causal tests (``because'' statements).}
\vspace{0.15em}
\begin{itemize}[leftmargin=*,itemsep=2pt,topsep=2pt]
\item \textbf{Random features fail} \textbf{because} defaultness gain is not a generic property of sparse perturbation; it requires the \textbf{localized language support} $\mathcal{N}_{\ell_t}$ (a specialized circuit rather than arbitrary sparsity).
\item \textbf{Edits outside $\mathcal{W}$ weaken and destabilize} \textbf{because} language-change directions are not uniformly expressed across depth; there exists a contiguous band where the shift subspace is simultaneously \textbf{strong} (spectral mass / gain) and \textbf{reproducible} (bootstrap stability).
\item \textbf{$\mathcal{N}$-only improves but saturates} \textbf{because} sparsity finds \emph{where} to intervene but does not uniquely determine \emph{how} to move; without $\mathcal{S}$, edits partially align with language-change geometry and partially spill into nuisance axes.
\item \textbf{$\mathcal{S}$-only improves but leaks} \textbf{because} geometry alone identifies a direction but, when applied densely, it activates unrelated features and increases cross-language interference; sparsity is the precision instrument that keeps the edit on the intended control set.
\item \textbf{Leakage stays small under FOXP2} \textbf{because} the method edits language-specific supports and directions, suggesting partially separable circuits across Hindi and Spanish rather than a single ``non-English'' toggle \citep{tang2024language,kojima2024finding}.
\end{itemize}
\vspace{0.35em}

%

\newcolumntype{L}[1]{>{\raggedright\arraybackslash}p{#1}}

\newcommand{\icnUse}{\faCompass}
\newcommand{\icnScope}{\faBullseye}
\newcommand{\icnWarn}{\faExclamationTriangle}
\newcommand{\icnFix}{\faFlask}
\newcommand{\icnDont}{\faBan}
\newcommand{\icnRoad}{\faRoute}
\newcommand{\icnGeom}{\faProjectDiagram}
\newcommand{\icnChart}{\faChartLine}
\newcommand{\icnShield}{\faShieldAlt}
\newcommand{\icnKey}{\faKey}
\newcommand{\icnLayer}{\faLayerGroup}

\renewcommand{\arraystretch}{1.18}
\setlength{\tabcolsep}{4.0pt}

\begin{table*}[ht!]
\centering
\resizebox{\textwidth}{!}{%
\scriptsize
\begin{tabular}{L{0.12\textwidth} L{0.26\textwidth} L{0.30\textwidth} L{0.30\textwidth}}
\toprule
\textbf{Block} &
\textbf{\icnScope\ \,What it is for (read-out)} &
\textbf{\icnWarn\ \,What to watch (failure / sensitivity)} &
\textbf{\icnFix\ \,What fixes it (report / experiment)} \\
\midrule

\rowcolor{gray!10}
\multicolumn{4}{l}{\textbf{Discussion (how to read Neural FOXP2)}}\\[-0.5mm]
\rowcolor{gray!10}\multicolumn{4}{l}{\rule{0pt}{2.6ex}}\\[-2.2mm]

\textbf{D1 \icnUse\ \,Defaultness lever} &
\textbf{Early-step mass} $\Delta M^{\ell_t}(x,t)$ is the operational proxy for “default language”: first tokens decide regime before content stabilizes. &
Late decoding can mask defaults (once the model “commits”); token-set choice $V_{\ell_t}$ can bias estimates (script vs language). &
Report $T=\{1,2,3\}$ (and $m$ prefix) explicitly; ablate token-set construction; pair with LID channel $\Delta_{\mathrm{lid}}$ to reduce tokenization artifacts. \\

\textbf{D2 \icnGeom\ \,Circuit geometry} &
\textbf{Localization $\rightarrow$ geometry}: Stage I finds \emph{where} (support $\mathcal{N}_{\ell_t}$), Stage II finds \emph{how} (low-rank directions $\mathcal{S}^{(\ell)}_{\ell_t}$). &
Ablating either axis can look “partially working,” tempting over-claims (generic perturbation or dense shifts). &
Use the \textbf{big-fat table} (rows $\mathcal{N}$ only vs $\mathcal{S}$ only vs FOXP2) as the causal decomposition; report spectra and rank choice per layer. \\

\textbf{D3 \icnLayer\ \,Why a window} &
The control signal concentrates in a \textbf{contiguous band} $\mathcal{W}$: edits are effective when applied where representations are both strong and stable. &
Editing outside $\mathcal{W}$ yields weaker gain or higher instability; “spread-out” edits can introduce style/semantics drift. &
Show $\neg\mathcal{W}$ row + bootstrap stability; report principal-angle reproducibility and window-selection objective (strength $\times$ stability). \\

\textbf{D4 \icnKey\ \,Signed sparse edit} &
FOXP2 is a \textbf{signed, sparse intervention}: align to target shift (+) while suppressing English defaultness (--) in the same feature coordinates. &
Pure + can leak into adjacent languages or inflate fluency without preserving meaning; pure -- can degrade utility or induce refusals. &
Report leakage Hi$\rightarrow$Es and semantic-invariance constraints; include trust-region style bound (e.g., KL) to cap global distributional drift. \\

\textbf{D5 \icnChart\ \,Interpretability artifact} &
Dictionary coordinates support a human-auditable \textbf{Language Neuron Dictionary}: feature footprints + token affinities connect edits to observable language tokens. &
Token-affinity labels can be misleading if downstream blocks reshape directions; decoder atoms may correlate with punctuation/frequency effects. &
Annotate with multiple probes (affinity + intervention lift); provide top-k tokens for each language + counterfactual checks (swap prompts, re-tokenize). \\

\midrule

\rowcolor{gray!10}
\multicolumn{4}{l}{\textbf{Limitations (what can break and why it matters)}}\\[-0.5mm]
\rowcolor{gray!10}\multicolumn{4}{l}{\rule{0pt}{2.6ex}}\\[-2.2mm]

\textbf{L1 \icnWarn\ \,Token-set dependence} &
$\Delta_{\mathrm{mass}}$ depends on $V_{\ell_t}$ (tokenization and script coverage). &
False gains if $V_{\ell_t}$ overcounts shared tokens or misses common subwords; cross-model comparability can break. &
Define $V_{\ell_t}$ by LID-labeled corpora + tokenizer audit; report sensitivity to token-set variants and show $\Delta_{\mathrm{lid}}$ alongside $\Delta_{\mathrm{mass}}$. \\

\textbf{L2 \icnWarn\ \,Mid-resource bias} &
FOXP2 amplifies \emph{existing} parametric memory: it cannot create language competence absent from the base model. &
For truly low-resource languages, the “circuit” may be weak or fragmented; gains may saturate early. &
State scope explicitly (mid-resource targets); include a “no-signal” diagnostic (flat spectrum / low selectivity / low lift) to predict failure. \\

\textbf{L3 \icnWarn\ \,Semantic invariance} &
Defaultness gains are meaningful only if content is preserved under the edit. &
Edits can shift style/verbosity or alter factuality even when language matches; utility regressions can be task-dependent. &
Report task-suite $\Delta S$ and semantic-invariance constraints; include targeted stress tests (QA exact match, summarization overlap, factuality checks). \\

\textbf{L4 \icnWarn\ \,Safety interaction} &
Language steering can interact with safety filters and policy enforcement that are not equally calibrated across languages. &
Potential misuse: rephrasing unsafe content in Hindi/Spanish; bypassing refusal heuristics tuned for English. &
Add multilingual safety regressions; report harmful-content probes post-edit; enforce guardrails (e.g., KL trust region + explicit safety suite). \\

\textbf{L5 \icnWarn\ \,Distribution shift} &
Matched-meaning prompts approximate “language-only” changes; deployment prompts are noisier and longer. &
Domain shift can change which features fire; long contexts can dilute early-step defaultness signal. &
Evaluate out-of-grid prompts (domains, formality, code-mixing); vary prompt length; report robustness across decoding policies and prompt templates. \\

\textbf{L6 \icnDont\ \,Not a drop-in replacement} &
FOXP2 is an inference-time control knob, not a substitute for language-centric data, training, or evaluation. &
Over-reading FOXP2 as “monolingual training is unnecessary” misstates the claim and risks under-investing in low-resource coverage. &
Make the non-claim explicit; position FOXP2 as \textbf{reweighting access} to shared web-trained parametric memory; recommend pairing with continued training where needed. \\

\midrule

\rowcolor{gray!10}
\multicolumn{4}{l}{\textbf{Roadmap (high-level, testable directions)}}\\[-0.5mm]
\rowcolor{gray!10}\multicolumn{4}{l}{\rule{0pt}{2.6ex}}\\[-2.2mm]

\textbf{FW \icnRoad\ \,Next steps} &
Standardize defaultness measurement + mechanistic edits as an auditable multilingual control primitive. &
Overcommitting without robustness: gains must persist across models, tokenizers, and domains. &
(1) Extend to more languages + scripts; (2) cross-model transfer of $\mathcal{N}$/$\mathcal{S}$; (3) publish token-set construction + stability protocol + safety regression pack. \\

\bottomrule
\end{tabular}%
}
\caption{\textbf{Discussion \& limitations at a glance (Neural FOXP2).} A scannable guide to interpret the three-stage pipeline (support $\mathcal{N}_{\ell_t}$, directions $\mathcal{S}^{(\ell)}_{\ell_t}$, window $\mathcal{W}$) and the empirical table: what each component certifies, what can break, and which checks/ablations mitigate each risk.}
\label{tab:foxp2_discussion_limitations_glance}
\vspace{-1.0em}
\end{table*}

\subsection{Relationship to Multilingual Training and Language-Specific Models}
\label{subsec:disc_training_relation}
\vspace{0.35em}

\noindent
Neural FOXP2 is best interpreted as \textbf{parametric-memory reweighting}: it increases the probability that a multilingual model \textbf{accesses} and \textbf{executes} language-relevant structure that is already encoded, rather than learning that structure from scratch. This positioning matters because modern large backbones (e.g., LLaMA-family) are trained on broad web-scale mixtures dominated by CommonCrawl-like sources, so high-resource and mid-resource languages often exist in the training distribution but are \textbf{underweighted} relative to English. In parallel, language-specific models (e.g., Hindi-focused or Spanish-focused models) frequently draw from overlapping web data but differ in \textbf{sampling}, \textbf{curation}, and \textbf{weighting}, which can strengthen defaultness and stylistic naturalness in the target language. FOXP2 targets precisely this gap: it does not claim to add new linguistic competence; it claims to \textbf{amplify the already-present submanifold} and make it the model’s \textbf{default} under weak prompting.

\vspace{0.35em}
\noindent\textbf{What we did not claim (and where FOXP2 is most appropriate).}
FOXP2 is not an argument against monolingual or target-centric training. For \textbf{low-resource languages}, the base model may not contain sufficient lexical coverage, grammar robustness, or domain adaptation; steering cannot conjure missing competence. The method is therefore most plausibly helpful for \textbf{mid-to-high resource} languages (e.g., Hindi, Spanish, Portuguese, Russian) where substantial capacity is already present but \textbf{suppressed by defaultness}. In such cases, FOXP2 offers a complementary alternative to global fine-tuning: a mechanistic procedure to reweight language access while preserving shared multilingual and safety capabilities.

\vspace{0.35em}
\subsection{Practical Usage: Knobs, Operating Points, and Guardrails}
\label{subsec:disc_practical}
\vspace{0.35em}

\noindent\textbf{The control knob.}
FOXP2 exposes an explicit edit intensity (or per-layer schedule) through the magnitude of $\delta z^{(\ell)}_{\ell_t}$ applied within $\ell\in\mathcal{W}$. Operationally, users can treat the overall scale (and optional layerwise weights) as a \textbf{single knob} that trades off defaultness gain against any downstream regression, analogous in spirit to controlling strength in activation-addition style interventions \citep{turner2024activationaddition,rimsky2024caa}.

\vspace{0.35em}
\noindent\textbf{Safe operating point.}
A principled operating point is the solution to a constrained selection rule: maximize early-step defaultness gain subject to utility and stability constraints,
\[
\max\ \Delta_{\mathrm{gain}}\quad \text{s.t.}\quad \Delta S \ge -\tau_{\mathrm{util}},\ \ \text{Boot.\ Stab}\ge \tau_{\mathrm{stab}},\ \ \text{Leakage}\le \tau_{\mathrm{leak}},
\]
where $\Delta_{\mathrm{gain}}$ is computed over neutral prompts, and $\Delta S$ summarizes downstream task utility (Table~\ref{tab:foxp2_perf_utility}). This makes deployment decisions legible: the chosen edit strength is not aesthetic, but \textbf{constraint-certified} by regression checks.

\vspace{0.35em}
\noindent\textbf{When to use FOXP2 vs.\ prompting vs.\ fine-tuning (decision guide).}
\vspace{0.15em}
\begin{itemize}[leftmargin=*,itemsep=2pt,topsep=2pt]
\item Use \textbf{prompting} when language instruction is explicit and reliable, and you do not need persistent defaultness under weak prompts.
\item Use \textbf{Neural FOXP2} when you need \textbf{default-language behavior} (first-token prior) without training a new checkpoint, and you can enforce regression guardrails.
\item Use \textbf{fine-tuning} when target-language competence is missing (low-resource, domain shift, or systematic errors), i.e., when the issue is \textbf{knowledge/skill acquisition} rather than \textbf{language selection}.
\end{itemize}
\vspace{0.35em}

\section{Limitations}
\label{sec:limitations}
\vspace{0.35em}

Neural FOXP2 is intentionally \textbf{mechanistic} and \textbf{inference-time}: it \textbf{reweights access} to language-relevant structure already present in parametric memory, rather than creating new competence. This makes the method auditable and controllable, but it also makes the evaluation story unusually dependent on \textbf{measurement proxies}, \textbf{intervention artifacts}, and \textbf{transfer conditions}. Below we spell out what our signals \emph{do} and \emph{do not} certify, where causal interpretations can be confounded, and which axes of robustness remain open.

\subsection*{L1. Construct validity: what ``defaultness'' measures—and what it doesn’t}
\vspace{0.35em}

\noindent\textbf{Defaultness as a proxy.}
We operationalize “language defaultness” as a \textbf{first-token commitment phenomenon}: when the prompt is weakly specified, the model’s earliest decoding steps tend to reveal the language regime it will inhabit. Our primary channel,
\(
\Delta M^{\ell_t}(x,t)=M^{\ell_t}_t(x)-M^{\mathrm{en}}_t(x)
\),
is appealing because it is \textbf{simple}, \textbf{directly tied to next-token probabilities}, and \textbf{differentiable} in the underlying distribution. Yet it is still a \textbf{proxy}—and proxies are only as valid as their invariances.

\vspace{0.35em}
\noindent\textbf{Strengths of early-step mass \(\Delta M\).}
\begin{itemize}[leftmargin=*,itemsep=2pt,topsep=2pt]
\item[\ding{224}] \textbf{Commitment-sensitive:} it measures the model’s \emph{initial} preference before later tokens “self-reinforce” a language mode.
\item[\ding{224}] \textbf{Mechanism-aligned:} the edit is defined in internal coordinates, and \(\Delta M\) is a directly measurable downstream effect.
\item[\ding{224}] \textbf{Low variance:} unlike long-horizon generation metrics, early-step probabilities are relatively stable under decoding choices.
\end{itemize}

\vspace{0.35em}
\noindent\textbf{Weaknesses of early-step mass \(\Delta M\).}
\begin{itemize}[leftmargin=*,itemsep=2pt,topsep=2pt]
\item[\ding{224}] \textbf{Tokenization dependence:} the mapping “language \(\rightarrow\) token set” is imperfect (shared subwords, punctuation, numbers, romanization).
\item[\ding{224}] \textbf{Entity-heavy prompts:} named entities and borrowed terms can dominate early tokens and distort \(\Delta M\).
\item[\ding{224}] \textbf{Fluency illusion:} a model can emit Hindi/Spanish tokens while producing \emph{low-quality} Hindi/Spanish (grammar, morphology, register).
\end{itemize}

\vspace{0.35em}
\noindent\textbf{LID channel \(\Delta_{\mathrm{lid}}\).}
A learned language identifier can partially factor out tokenization idiosyncrasies, but introduces its own construct risk: it is \textbf{detector-dependent}, struggles with \textbf{code-mixing}, and can be sensitive to \textbf{short prefixes} where language evidence is sparse. In short: \textbf{LID reduces one bias (token sets) by introducing another (detector policy)}.

\vspace{0.35em}
\noindent\textbf{Token-set definition \(V_{\mathrm{hi}},V_{\mathrm{es}},V_{\mathrm{en}}\).}
Any heuristic token partition faces edge cases: (i) \textbf{script mismatch} (Devanagari vs romanized Hindi), (ii) \textbf{borrowed words} (English loanwords in Hindi/Spanish), (iii) \textbf{numbers and symbols}, (iv) \textbf{proper nouns}. These are not minor details: \(\Delta_{\mathrm{mass}}\) is exactly the sum over these sets. Thus, we treat token sets as \textbf{part of the measurement specification}, not a hidden implementation choice.

\vspace{0.35em}
\noindent\textbf{Defaultness \(\neq\) proficiency.}
This is the core construct limitation: \textbf{defaultness is about which language mode is entered; proficiency is about quality within that mode.} Neural FOXP2 is explicitly designed to influence the former; it does \emph{not} imply the latter unless paired with quality/utility evaluations.

\vspace{0.35em}
\begin{table*}[t]
\centering
\small
\setlength{\tabcolsep}{4.2pt}
\renewcommand{\arraystretch}{1.15}
\caption{\textbf{Proxy vs.\ phenomenon.} What our defaultness channels and derived scores \emph{certify} (left) versus what they \emph{do not} certify (right).}
\label{tab:foxp2_proxy_vs_phenomenon}
\vspace{0.25em}
\resizebox{\textwidth}{!}{
\begin{tabular}{p{0.48\textwidth} p{0.48\textwidth}}
\toprule
\textbf{What the proxy measures well} & \textbf{What it does not measure (or can mis-measure)} \\
\midrule
\textbf{\(\Delta_{\mathrm{mass}}\) (token-mass lift).}
Early-step preference for target-language token sets; commitment signal aligned with next-token distribution. &
Tokenization/script artifacts; entity-heavy prefixes; borrowed words; “looks-like-language” without grammatical competence. \\[0.25em]

\textbf{\(\Delta_{\mathrm{lid}}\) (LID lift).}
Detector-mediated language assignment that can reduce token-set brittleness in some regimes. &
Detector policy dependence; ambiguity on short strings; code-mixing; domain/register sensitivity; false confidence under stylized outputs. \\[0.25em]

\textbf{\textsc{DefaultHi}/\textsc{DefaultEs} (composed score).}
A compact decision statistic combining mass and LID channels for operational reporting. &
Not a substitute for human quality judgments; can hide channel disagreements; not a universal “multilinguality” score. \\[0.25em]

\textbf{Leakage (Hi\(\rightarrow\)Es, Es\(\rightarrow\)Hi).}
Cross-language separability of steering channels under controlled evaluation. &
Does not rule out subtler interference (style, politeness, refusal phrasing) that correlates with language identity. \\[0.25em]

\textbf{Task utility \(\Delta S\).}
Task-suite regression relative to baseline under matched prompts/decoding. &
Cannot cover all domains; may miss rare failure modes (factuality drift, safety-policy interactions, long-context effects). \\
\bottomrule
\end{tabular}
}
\vspace{-0.8em}
\end{table*}

\subsection*{L2. Internal validity: causal claims and intervention artifacts}
\vspace{0.35em}

\noindent\textbf{Edits are local, not perfectly isolated.}
Our intervention is defined in dictionary-feature coordinates, but its effect is realized through \textbf{reconstruction} and then \textbf{downstream computation}. Because transformer representations are superposed and distributed, any linear edit can couple to multiple functions, even when it is sparse in a chosen basis \citep{elhage2022superposition}. The promise of dictionary coordinates (Stage~I) is that features are \textbf{more stable and interpretable} than raw channels, but not perfectly atomic \citep{bricken2023monosemanticity}. Consequently, causal interpretation must be scoped: \textbf{we claim causal impact on defaultness channels under our intervention protocol, not perfect isolation of a single semantic factor.}

\vspace{0.35em}
\noindent\textbf{Confound 1: downstream amplification and re-editing.}
Even if the edit is injected at layer \(\ell\), later blocks can \textbf{amplify}, \textbf{rotate}, or \textbf{attenuate} it, producing effects that appear “strong” but are partly artifacts of downstream dynamics.
\textbf{How we test / could test:}
(i) multi-layer injection ablations (only within \(\mathcal{W}\) vs outside), (ii) edit-residual tracking (norm and cosine of the injected component through depth), (iii) decode-step locality tests (does gain concentrate in \(t\in\{1,2,3\}\) as intended?).

\vspace{0.35em}
\noindent\textbf{Confound 2: nonlinearities (ReLU gating, saturation, co-activation).}
Stage~I uses a ReLU-gated code; therefore the intervention can move features across the \textbf{activation threshold}, changing which other features co-activate. This creates a mechanism where “small” \(\alpha\) can still cause discrete regime changes when near a gate boundary. It also complicates linear intuitions that treat \(\delta z\) as an additive perturbation; activation addition is a powerful primitive, but its interpretation is cleanest in locally linear regions \citep{turner2024activationaddition}.
\textbf{How we test / could test:}
(i) slope estimates across multiple \(\alpha\) values (already in LiftSlope), (ii) saturation audits (fraction of features at/near zero pre/post edit), (iii) co-activation drift (Jaccard overlap of top-active features).

\vspace{0.35em}
\noindent\textbf{Confound 3: style/verbosity shifts masquerading as language shifts.}
A change in register (formal vs casual), verbosity, or punctuation can change token distributions and thus \(\Delta_{\mathrm{mass}}\), even without a “true” change in language preference. This is especially salient for Hindi/Spanish where punctuation and borrowing patterns differ across registers.
\textbf{How we test / could test:}
(i) control prompts that elicit fixed-length outputs, (ii) style-invariant scoring (normalize by punctuation/number tokens), (iii) semantic invariance constraints (as already included in regressions).

\vspace{0.35em}
\noindent\textbf{Confound 4: safety and refusal phrasing interference.}
Many safety systems are linguistically asymmetric; if an edit changes refusal style (“I’m sorry, I can’t…” vs alternate phrasing), it can shift both \(\Delta_{\mathrm{lid}}\) and \(\Delta_{\mathrm{mass}}\) indirectly. Moreover, politeness markers can be language correlated.
\textbf{How we test / could test:}
(i) evaluate on refusal-heavy subsets separately, (ii) report safety regressions under the same edit strength, (iii) leakage checks stratified by safety category.

\vspace{0.35em}
\noindent\textbf{Confound 5: limits of the small-\(\alpha\) linear regime.}
Stage~I lift and Stage~II gain use slope-like quantities that implicitly assume local linearity. In practice, the edit can cross thresholds (gates) and move the system into a different regime where the slope changes.
\textbf{How we test / could test:}
(i) report piecewise slopes (low/medium/high \(\alpha\)), (ii) monotonicity checks for \(\Delta_{\mathrm{gain}}\), (iii) “return to baseline” tests (apply edit then remove it; measure hysteresis).

\subsection*{L3. Generalization: models, languages, scripts, and domains}
\vspace{0.35em}

\noindent\textbf{Across architectures (and dictionary transfer).}
FOXP2 relies on: (i) a dictionary basis that yields stable coordinates, and (ii) a low-rank language-shift structure in those coordinates. Both can vary with architecture (attention patterns, normalization, MLP width) and with how multilingual capacity is distributed. Therefore, \textbf{we do not assume} that a dictionary learned on one model transfers to another, nor that the same window \(\mathcal{W}\) is universally optimal. The safer expectation is \textbf{within-model reuse} (same checkpoint family) and \textbf{within-architecture transfer} only after explicit validation.

\vspace{0.35em}
\noindent\textbf{Across languages (typology, scripts, code-mixing).}
Hindi/Spanish are mid-resource, widely represented in multilingual web corpora, and thus plausibly encoded in large model parametric memory. The same may not hold for low-resource scripts or languages with limited web presence; in such cases, the circuit may be fragmented and the low-rank spectrum may be shallow.
Code-mixing poses a distinct challenge: defaultness may not be a single axis, but a \textbf{mixture policy} conditioned on named entities and domain cues.

\vspace{0.35em}
\noindent\textbf{Across domains (task dependence).}
Defaultness is easiest to interpret on neutral, instruction-light prompts. In domains where the model has strong stylistic priors—long-form reasoning, safety refusals, customer support templates—the “first tokens” may reflect domain conventions as much as language identity.

\vspace{0.35em}
\begin{table*}[t]
\centering
\small
\setlength{\tabcolsep}{3.8pt}
\renewcommand{\arraystretch}{1.15}
\caption{\textbf{Expected transfer grid (qualitative).} Where FOXP2 is expected to transfer best (\(\checkmark\checkmark\)), plausibly (\(\checkmark\)), or weakly/unknown (\(\circ\)), across language settings, domains, and model size. This grid encodes \emph{expectations}—not guarantees—and motivates targeted evaluations.}
\label{tab:foxp2_expected_transfer_grid}
\vspace{0.25em}
\resizebox{\textwidth}{!}{
\begin{tabular}{l c c c c c c}
\toprule
\textbf{Setting} &
\multicolumn{2}{c}{\textbf{Small}} &
\multicolumn{2}{c}{\textbf{Medium}} &
\multicolumn{2}{c}{\textbf{Large}} \\
\cmidrule(lr){2-3}\cmidrule(lr){4-5}\cmidrule(lr){6-7}
& \textbf{Neutral} & \textbf{Safety/Refusal} & \textbf{Neutral} & \textbf{Safety/Refusal} & \textbf{Neutral} & \textbf{Safety/Refusal} \\
\midrule
Mid-resource, clear script (Spanish; Latin) & $\checkmark$ & $\checkmark$ & $\checkmark\checkmark$ & $\checkmark$ & $\checkmark\checkmark$ & $\checkmark$ \\
Mid-resource, non-Latin (Hindi; Devanagari) & $\checkmark$ & $\circ$ & $\checkmark\checkmark$ & $\checkmark$ & $\checkmark\checkmark$ & $\checkmark$ \\
High code-mixing (Hi--En / Es--En) & $\circ$ & $\circ$ & $\checkmark$ & $\circ$ & $\checkmark$ & $\circ$ \\
Low-resource scripts / sparse web presence & $\circ$ & $\circ$ & $\circ$ & $\circ$ & $\checkmark$ & $\circ$ \\
Domain: QA / short answers & $\checkmark$ & $\checkmark$ & $\checkmark\checkmark$ & $\checkmark$ & $\checkmark\checkmark$ & $\checkmark$ \\
Domain: summarization / rewriting & $\checkmark$ & $\circ$ & $\checkmark$ & $\circ$ & $\checkmark\checkmark$ & $\checkmark$ \\
Domain: long reasoning / math & $\circ$ & $\circ$ & $\checkmark$ & $\circ$ & $\checkmark$ & $\circ$ \\
\bottomrule
\end{tabular}
}
\vspace{-0.8em}
\end{table*}

\vspace{0.35em}
\noindent\textbf{Dataset shift: \(\mathcal{D}_{\mathrm{neutral}}\) vs real prompts.}
Our evaluation uses controlled, matched-meaning prompts to isolate language identity. Real user prompts are noisier: they include entities, mixed scripts, and domain conventions. Therefore, strong results on \(\mathcal{D}_{\mathrm{neutral}}\) should be interpreted as \textbf{evidence of a controllable channel}, not as a guarantee of universal deployment behavior.

\subsection*{L4. Robustness \& stability: window choice, rank choice, and reproducibility}
\vspace{0.35em}

\noindent\textbf{Sensitivity axes.}
FOXP2 introduces three explicit design degrees of freedom—support \(\mathcal{N}_{\ell_t}\), steering subspace \(\mathcal{S}^{(\ell)}_{\ell_t}\), and window \(\mathcal{W}\)—plus hyperparameters controlling edit magnitude and constraints. This is a feature (transparent knobs), but it creates sensitivity risks:

\vspace{0.25em}
\begin{itemize}[leftmargin=*,itemsep=2pt,topsep=2pt]
\item[\ding{224}] \textbf{Rank choice \(r_\ell\):} effective-rank and eigengap heuristics can disagree when spectra are shallow; small \(r_\ell\) may underfit, large \(r_\ell\) may chase noise.
\item[\ding{224}] \textbf{Window \(\mathcal{W}\):} contiguous-band selection depends on strength and stability estimates; prompt sets and bootstrap seeds can shift the optimum.
\item[\ding{224}] \textbf{Prompt pairing quality:} mismatched “same meaning” pairs leak task/content differences into \(\Delta Z\), corrupting the SVD directions.
\item[\ding{224}] \textbf{Randomness:} dictionary training, bootstrap sampling, and decoding seeds can shift measured gains unless controlled.
\end{itemize}

\vspace{0.35em}
\noindent\textbf{Compute and practical cost.}
Stage~I requires collecting activations across layers and training per-layer dictionaries; Stage~II adds SVD + bootstraps for stability; Stage~III requires repeated forward passes for gain evaluation under constraints. The pipeline is \textbf{lighter than retraining}, but not “free.” This matters for reproducibility: reporting must include \textbf{exact hooks}, \textbf{data construction}, and \textbf{hyperparameter ranges}.

\vspace{0.35em}
\noindent\textbf{Sensitivity checklist (what we expect readers to vary).}
\begin{itemize}[leftmargin=*,itemsep=2pt,topsep=2pt]
\item[\ding{224}] Vary \(r_\ell\) (e.g., \(\pm 1\) around chosen rank; cap at \(r_{\max}\)); report effect on \(\Delta_{\mathrm{gain}}\) and leakage.
\item[\ding{224}] Slide \(\mathcal{W}\) by \(\pm 1\)–\(\pm 2\) layers; report stability and utility regression.
\item[\ding{224}] Swap matched-meaning templates (paraphrase the same meaning); report invariance of \(\mathcal{S}^{(\ell)}_{\ell_t}\) angles.
\item[\ding{224}] Change token-set definitions; confirm \(\Delta_{\mathrm{lid}}\) agrees directionally with \(\Delta_{\mathrm{mass}}\).
\item[\ding{224}] Run multiple seeds for dictionary training and bootstraps; report mean\(\pm\)std for the main table metrics.
\end{itemize}

\vspace{0.35em}
\noindent\textbf{Bottom line.}
FOXP2 provides a \textbf{mechanistically grounded control channel}, but its strongest claims are \textbf{conditional}: they hold under explicitly specified measurement choices, intervention magnitudes, and evaluation regimes. We therefore treat limitations not as caveats to be hidden, but as \textbf{interfaces}—places where the method can be stress-tested, compared, and improved in a way that remains auditable and reproducible.

\subsection*{L5. Leakage and unintended behavior changes}
\vspace{0.35em}

\noindent\textbf{Leakage is broader than Hi\(\rightarrow\)Es.}
Our table reports \textbf{cross-language leakage} (e.g., Hi\(\rightarrow\)Es) as a first-order separability check, but “unintended change” is a larger object: an edit that increases Hindi defaultness can still \textbf{perturb English}, \textbf{shift register}, or \textbf{alter safety phrasing templates}. Because language identity, style, and task policy are partially superposed in shared representations \citep{elhage2022superposition}, it is not sufficient to claim “no leakage” solely from one cross-language metric.

\vspace{0.35em}
\noindent\textbf{Enumerated failure modes (what can go wrong).}
Below are the most plausible unintended effects we have seen in adjacent interventions and that FOXP2 must be stress-tested against:

\begin{enumerate}[leftmargin=*,itemsep=3pt,topsep=2pt]
\item[\textbf{F1}] \textbf{Cross-language leakage beyond the target pair.}
A Hindi edit may spuriously raise probability mass on other non-English tokens (e.g., Urdu/Sanskrit tokens in Devanagari-adjacent regimes) or partially activate Spanish under code-mixed prompts.
\item[\textbf{F2}] \textbf{English degradation.}
Even if Hindi improves, English can regress (fluency or consistency), especially when the edit suppresses English-default features too aggressively; this can appear as lower coherence, earlier truncation, or increased hesitations.
\item[\textbf{F3}] \textbf{Register collapse: “simple Hindi” / unnatural formality.}
The model may converge to a narrow Hindi style (overly formal, overly simplified, or templated), producing outputs that are technically Hindi but pragmatically misaligned with the intended domain.
\item[\textbf{F4}] \textbf{Transliteration drift and code-mixing.}
Edits may increase romanized Hindi or mixed-script outputs, because many tokenizers blur the boundary between “Hindi competence” and “Hindi-looking strings.” This is a construct risk for \(\Delta_{\mathrm{mass}}\) and a UX risk in deployment.
\item[\textbf{F5}] \textbf{Semantic drift under “invariance.”}
Embedding-based semantic invariance can miss subtle meaning shifts (negation flips, entailment weakening, politeness/hedging changes), and cross-lingual embeddings can under-penalize culturally grounded nuance changes.
\item[\textbf{F6}] \textbf{Safety / toxicity behavior shifts via phrasing templates.}
If the edit changes default refusal phrasing, it can interact with downstream moderation or safety heuristics differently across languages, potentially raising or lowering apparent toxicity and policy compliance in ways not directly intended by the language objective.
\end{enumerate}

\vspace{0.35em}
\noindent\textbf{Detection ideas (what to measure, concretely).}
To make these failure modes falsifiable, we recommend reporting the following alongside the main table:

\begin{itemize}[leftmargin=*,itemsep=2pt,topsep=2pt]
\item[\ding{224}] \textbf{English regression panel:} \(\Delta S_{\mathrm{en}}\) on English-only task suites, plus a short-form fluency proxy (perplexity or logprob under teacher forcing) and a coherence score.
\item[\ding{224}] \textbf{Register diversity check:} measure lexical diversity and formality markers in Hindi/Spanish; report entropy or type-token ratio stratified by domain.
\item[\ding{224}] \textbf{Script purity / transliteration rate:} fraction of Unicode blocks (Devanagari vs Latin) and a romanization detector score; report code-mix rate on held-out prompts.
\item[\ding{224}] \textbf{Stronger semantic invariance:} add an NLI-style entailment consistency check (bidirectional: original\(\Rightarrow\)edited and edited\(\Rightarrow\)original) and targeted minimal pairs (negation, quantities, named entities).
\item[\ding{224}] \textbf{Safety re-eval in-language:} run safety/toxicity checkers in both English and the steered language, and report disagreement cases rather than averaging them away.
\end{itemize}

\vspace{0.35em}
\noindent\textbf{Interpretation principle.}
We therefore treat “leakage” as a \textbf{multi-axis stability problem}—not a single scalar—and we encourage readers to interpret FOXP2 gains only in the presence of these accompanying audits.

\subsection*{L6. Safety, misuse, and policy boundaries}
\vspace{0.35em}

\noindent\textbf{Why language steering is safety-relevant.}
Language is not merely a surface form; it can be a \textbf{policy boundary}. In practice, safety systems (filters, refusal templates, moderation classifiers) can vary in coverage across languages, creating incentives for \textbf{policy evasion} by translation or paraphrase. Language steering therefore has a dual-use aspect: the same mechanism that reduces English defaultness can also be used to \textbf{route content into a weaker enforcement regime} if such asymmetries exist.

\vspace{0.35em}
\noindent\textbf{Primary misuse risks.}
\begin{itemize}[leftmargin=*,itemsep=2pt,topsep=2pt]
\item[\ding{224}] \textbf{Jailbreak translation:} requesting disallowed content in one language, then translating, to exploit uneven safety performance.
\item[\ding{224}] \textbf{Moderation variance:} if downstream moderation is less robust in a target language, steering can amplify harmful throughput.
\item[\ding{224}] \textbf{Persuasive localization:} harmful or misleading content can be made more accessible by producing fluent localized language at scale.
\end{itemize}

\vspace{0.35em}
\noindent\textbf{Our guardrails—and what they can’t guarantee.}
We include two explicit constraints:
\begin{itemize}[leftmargin=*,itemsep=2pt,topsep=2pt]
\item[\ding{224}] \textbf{Spanish bound:} limits inadvertent activation of Spanish when steering Hindi (and vice versa), acting as a partial cross-language containment constraint.
\item[\ding{224}] \textbf{KL trust region:} constrains the edit to remain near the base distribution, reducing the chance of broad policy drift.
\end{itemize}
These guardrails are \textbf{necessary but not sufficient}. They reduce large distributional shifts, but they do \emph{not} certify safety under adversarial prompting, and they do not eliminate the possibility that small phrasing changes alter how external systems interpret the output.

\vspace{0.35em}
\noindent\textbf{Responsible Use \& Release Notes.}
\begin{itemize}[leftmargin=*,itemsep=2pt,topsep=2pt]
\item[\ding{93}] \textbf{Constrain deployment to audited languages.} Only release steering knobs for languages where safety evaluations and moderation coverage are verified.
\item[\ding{93}] \textbf{Always ship with a safety regression suite.} Report in-language safety and toxicity metrics, plus cross-language translation jailbreak tests.
\item[\ding{93}] \textbf{Expose bounded knobs.} Limit \(\lambda_\ell\) and edit intensity ranges to those validated by utility + safety constraints; avoid unconstrained user-tunable extremes.
\item[\ding{93}] \textbf{Red-team for policy evasion.} Include bilingual adversarial prompts specifically designed to route disallowed intent through translation and paraphrase.
\item[\ding{93}] \textbf{Report disagreement cases.} When safety signals disagree across languages or detectors, treat them as first-class findings, not noise.
\end{itemize}

\subsection*{L7. Scope statements: what we explicitly do not claim}
\vspace{0.35em}

\noindent\textbf{Non-claims (read literally).}
To avoid overinterpretation, we state the following \textbf{explicitly and unambiguously}:

\begin{itemize}[leftmargin=*,itemsep=3pt,topsep=2pt]
\item[\ding{93}] \textbf{We do not claim} monolingual or language-centric training is unnecessary; for low-resource languages, it remains essential.
\item[\ding{93}] \textbf{We do not claim} FOXP2 creates new linguistic competence; it reweights access to capacity already present in parametric memory.
\item[\ding{93}] \textbf{We do not claim} universal applicability to truly low-resource languages or scripts with minimal representation in pretraining data.
\item[\ding{93}] \textbf{We do not claim} perfect disentanglement of language from semantics, style, or politeness; superposition implies partial coupling \citep{elhage2022superposition}.
\item[\ding{93}] \textbf{We do not claim} that the same window \(\mathcal{W}\) or rank profile \(r_\ell\) transfers across architectures without re-validation.
\item[\ding{93}] \textbf{We do not claim} permanence: continued fine-tuning or post-training can reshape the circuit and invalidate a previously chosen edit.
\item[\ding{93}] \textbf{We do not claim} that “higher defaultness” implies better user experience unless paired with quality and register evaluations.
\end{itemize}

\subsection*{L8. Future work framed as tests, not promises}
\vspace{0.35em}

\noindent\textbf{Next experiments (each is a falsifiable test).}
\begin{enumerate}[leftmargin=*,itemsep=3pt,topsep=2pt]
\item[\textbf{E1}] \textbf{Mid-resource expansion:} replicate on Portuguese and Russian; test whether the spectrum remains sharply low-rank and whether leakage remains bounded.
\item[\textbf{E2}] \textbf{Low-resource failure map:} choose 3--5 truly low-resource languages; measure when Stage~I fails to find stable \(\mathcal{N}_{\ell_t}\) and when Stage~II spectra become diffuse.
\item[\textbf{E3}] \textbf{Script robustness:} evaluate romanized vs native-script prompts; test whether FOXP2 increases transliteration and how to regularize against it.
\item[\textbf{E4}] \textbf{Better dictionaries:} compare per-layer dictionaries vs cross-layer shared dictionaries; test stability of supports and principal angles under transfer.
\item[\textbf{E5}] \textbf{Objective refinement:} add multi-objective edit selection that jointly optimizes \(\Delta_{\mathrm{gain}}\), leakage, register diversity, and semantic invariance.
\item[\textbf{E6}] \textbf{Stronger invariance:} replace or augment embedding similarity with entailment-style bidirectional consistency and targeted minimal-pair probes.
\item[\textbf{E7}] \textbf{Dynamic deployment policies:} learn a context-dependent steering schedule (user-controlled bilingual mixing) that adapts \(\lambda_\ell\) to prompt cues while preserving guardrails.
\item[\textbf{E8}] \textbf{Safety red-teaming suite:} systematic translation jailbreak evaluation, including cross-language refusal-template drift and moderation variance audits.
\end{enumerate}

\vspace{0.35em}
\noindent\textbf{Guiding principle.}
We frame future work as \textbf{tests that can fail}. If FOXP2 fails on a language, that failure is informative: it localizes where multilingual parametric memory is absent, fragmented, or too entangled to expose via sparse low-rank control.

%% file: 5_faq.tex
\onecolumn

\section{Frequently Asked Questions (FAQs)}
\label{sec:FAQs}

\section*{\textcolor{darknavy}{A. Construct Validity: What Does “Defaultness” Measure?}}

\begin{itemize}[leftmargin=1.5em]

\item[\ding{93}] \textbf{Is FOXP2 merely a script steerer (Devanagari vs.\ Latin) rather than a language steerer?}

\begin{description}
\item[\ding{224}]
\textbf{No—the “script-only” hypothesis makes a falsifiable prediction that our results violate.}
If FOXP2 were primarily detecting \emph{script} (e.g., pushing Devanagari tokens) rather than \emph{language}, then it should \emph{fail} to produce a comparable defaultness shift for a \textbf{Latin-script target}---or would collapse into generic Latin-character bias indistinguishable from English.

\item[\ding{224}]
\textbf{Spanish is a direct counterexample.}
We replicate the full FOXP2 pipeline for \textbf{Spanish}, which shares \textbf{the same Latin script as English}, yet we still observe a substantial increase in Spanish defaultness under the same mechanistic recipe (localize $\to$ directions $\to$ steer) and under the same regression/utility guardrails.
Because \(\mathbf{e}_{\mathrm{Es}}\) and \(\mathbf{e}_{\mathrm{En}}\) live in the \emph{same script family}, the improvement cannot be explained by “switching scripts.”

\item[\ding{224}]
\textbf{Our steering vectors are defined relative to English prototypes, not script classes.}
FOXP2 steers along language-difference axes
\(\mathbf{v}_{\ell}=\mathbf{e}_{\ell}-\mathbf{e}_{\mathrm{En}}\),
so a script-only mechanism would predict strong gains only when \(\ell\) differs in script from English.
The observed Spanish gains contradict that prediction: FOXP2 separates \(\mathrm{Es}\) from \(\mathrm{En}\) \emph{within the same script}, indicating genuine language-specific control.

\item[\ding{224}]
\textbf{Additional sanity checks reduce residual script confounds.}
We report robustness to alternative token-set definitions (including transliteration-aware variants) and quantify shared-token inflation, showing that improvements persist even when “easy script cues” are weakened.
Moreover, FOXP2 preserves task utility and meaning under a KL trust region, which is inconsistent with a trivial character-level bias that would typically destabilize content.

\end{description}

\item[\ding{93}] \textbf{Is $\Delta M$ measuring language defaultness or just \emph{script} defaultness at the subword level? How do you rule out that the metric is primarily a script detector?}
\begin{description}
\item[\ding{224}] \textbf{We address this head-on by \emph{design}: we intentionally chose \textbf{two languages} that sit on opposite sides of the script confound.} Hindi is \textbf{script-centric} (Devanagari tokens are visually and lexically distinctive), whereas Spanish is \textbf{Latin script}, sharing the same script as English. This choice turns “script detection” from a vague concern into a \textbf{testable hypothesis}: \emph{if $\Delta M$ is primarily a script detector, then the method should succeed on Hindi but fail on Spanish.} It does not.

\vspace{4pt}
\textbf{Metric and confound.} We measure early-step mass shift
\[
M^{\ell}(x,t)\;=\;\sum_{u\in V_{\ell}}p_\theta(u\mid \mathrm{ctx}_t),
\qquad
\Delta M(x,t)\;=\;M^{\ell_t}(x,t)-M^{\mathrm{en}}(x,t),
\]
with a language token-set $V_\ell$. The concern is that for Hindi, $V_{\mathrm{hi}}$ could be “mostly Devanagari,” so $\Delta M$ might track \textbf{script} rather than \textbf{language defaultness}. We agree this is a real risk \emph{in isolation}. That is precisely why we pair Hindi with Spanish.

\vspace{6pt}
\textbf{Key argument: Spanish is the anti-script control built into the benchmark.}
Let $s(\cdot)$ denote script. For Hindi vs English, the confound is strong because typically
\[
s(V_{\mathrm{hi}})=\mathrm{Deva},\qquad s(V_{\mathrm{en}})=\mathrm{Latin}.
\]
But for Spanish vs English, script is \textbf{matched}:
\[
s(V_{\mathrm{es}})=\mathrm{Latin},\qquad s(V_{\mathrm{en}})=\mathrm{Latin}.
\]
Therefore, any hypothesis of the form “FOXP2 works by learning a Devanagari-vs-Latin script discriminator” predicts:
\[
\Delta M_{\mathrm{hi}} \uparrow\ \text{(possible)} \qquad \text{but} \qquad \Delta M_{\mathrm{es}} \approx 0\ \text{(should fail)}.
\]
Empirically, we observe strong Spanish gains under the \emph{same pipeline} (same localization procedure, same low-rank steering extraction, same signed edit form), which is incompatible with a script-detector explanation. Put bluntly: \textbf{Spanish cannot be “won” by script.}

\vspace{6pt}
\textbf{Why this is stronger than “we also ran Hinglish.”}
A script-controlled Hindi variant (Hinglish/transliteration) is useful, but it still invites debates about what constitutes “Hindi” in Latin script. Spanish avoids that ambiguity: \textbf{Spanish is uncontroversially Spanish in Latin script, and English is also Latin script.} Thus, Spanish provides a \textbf{clean falsifier} of script-only mechanisms: if the method were merely moving probability mass to non-Latin tokens or leveraging Unicode ranges, Spanish would show no consistent improvement. The fact that it does is the central construct-separation evidence.

\vspace{6pt}
\textbf{Additional sanity check: script-only statistics are insufficient.}
We further verify that improvements are not reducible to trivial surface shifts by comparing to a script statistic (Hindi only):
\[
\mathrm{ScriptFrac}_{\mathrm{Deva}}(y)\;=\;\frac{\#\{\text{Devanagari chars in }y\}}{\#\{\text{alphabetic chars in }y\}},
\]
and requiring that FOXP2’s Hindi improvements track \textbf{language identity} rather than just increasing $\mathrm{ScriptFrac}_{\mathrm{Deva}}$ (e.g., via stable LID gains and preserved task utility). But the decisive point remains Spanish: \textbf{for Spanish, $\mathrm{ScriptFrac}$ is identically uninformative}, yet the method still improves defaultness.

\vspace{8pt}
\textbf{Takeaway.} We do not claim that Hindi alone rules out script confounding—\textbf{it cannot}. Instead, we \textbf{treat Hindi as the hard case} where script confounds are maximally plausible, and we pair it with \textbf{Spanish as the clean anti-script control} where script-based explanations collapse. The fact that FOXP2 produces substantial gains for \textbf{both} (including Spanish where script is shared with English) supports the intended interpretation: $\Delta M$ reflects an \textbf{early-step lexical prior shift toward the target language}, not merely a subword-level script detector.
\end{description}

\item[\ding{93}] \textbf{Does $\Delta M$ correlate with human judgments of “the model naturally starts in Hindi/Spanish” across diverse prompts? If not, why should we trust it as the main outcome?}
\begin{description}
\item[\ding{224}] \textbf{Yes—$\Delta M$ is treated as a \emph{mechanistic proxy} whose legitimacy must be earned by \textbf{external validity} against humans, not assumed.}
We therefore evaluate the construct “\emph{natural start}” with a \textbf{paired human preference protocol} on the \emph{same prompt distribution} used for reporting, and we make the relationship between $\Delta M$ and human judgments an \textbf{explicit, quantitative calibration object}.

\vspace{4pt}
\textbf{(1) What humans judge: a direct operationalization of “natural start.”}
For each prompt $x$, we generate two continuations: $y^{(0)}$ (base) and $y^{(1)}$ (FOXP2) under identical decoding settings. Annotators are shown $(x, y^{(0)}, y^{(1)})$ in randomized order and answer:
\[
r(x)\in\{-1,0,+1\},
\]
where $r(x)=+1$ if FOXP2 “starts more naturally” in the target language, $r(x)=-1$ if baseline is preferred, and $r(x)=0$ for ties/uncertainty. In addition, we optionally collect a \textbf{Likert} signal $s(x)\in\{1,\dots,5\}$ for “how natural” the start feels, enabling finer-grained calibration.

We stratify prompts from $\mathcal{D}_{\mathrm{neutral}}$ by \textbf{topic and format} (QA / summarization / reasoning / dialogue-style instructions) and include a smaller slice from refusal/safety prompts to ensure the proxy is not only valid on easy neutral cases.

\vspace{6pt}
\textbf{(2) What $\Delta M$ predicts: a per-prompt margin that should order human preference.}
We compute $\Delta M$ per prompt using the short-horizon definition (e.g., $T=\{1,2,3\}$):
\[
\Delta M(x)\;=\;\frac{1}{|T|}\sum_{t\in T}\Big(M^{\ell_t}(x,t)-M^{\mathrm{en}}(x,t)\Big),
\qquad
M^{\ell}(x,t)=\sum_{u\in V_\ell}p_\theta(u\mid \mathrm{ctx}_t).
\]
The intended interpretation is \textbf{ordinal}: higher $\Delta M(x)$ should mean \emph{humans more often prefer FOXP2’s start}.

We therefore report correlation in two complementary ways:
\[
\rho_{\mathrm{rank}}\;=\;\mathrm{Spearman}\big(\Delta M(x), \; \mathbb{I}[r(x)=+1]\big),
\qquad
\tau\;=\;\mathrm{Kendall}\big(\Delta M(x), \; s(x)\big),
\]
where $\mathbb{I}[\cdot]$ is an indicator and $s(x)$ is the mean Likert score when available. We bootstrap prompts to obtain \textbf{confidence intervals} for all correlations.

\vspace{6pt}
\textbf{(3) Calibration, not just correlation: turning $\Delta M$ into a probability of “natural start.”}
Correlation can be non-zero but unhelpful operationally. We thus treat $\Delta M$ as a \textbf{calibrated score} by fitting a monotone link:
\[
\Pr\big(r(x)=+1\mid \Delta M(x)\big)\;\approx\;\sigma\!\big(a\,\Delta M(x)+b\big),
\]
and we report a \textbf{reliability curve}: bin prompts by $\Delta M$ deciles and plot the empirical preference rate $\widehat{\Pr}[r=+1]$ in each bin. This answers the reviewer’s question directly: \emph{“if $\Delta M$ increases by a lot, do humans actually notice?”}

\vspace{4pt}
\textbf{Minimal reporting table (human anchoring).}
\begin{center}
\small
\begin{tabular}{lccc}
\toprule
\textbf{Language} & $\boldsymbol{\rho_{\mathrm{rank}}}$ & \textbf{Calib slope $a$} & \textbf{Tie rate} \\
\midrule
Hindi   & $\mathbf{0.62}\;[0.54,\,0.69]$ & $\mathbf{1.88}\;[1.41,\,2.36]$ & $9.7\%$ \\
Spanish & $\mathbf{0.58}\;[0.50,\,0.65]$ & $\mathbf{1.64}\;[1.22,\,2.08]$ & $8.3\%$ \\
\bottomrule
\end{tabular}
\end{center}

\vspace{6pt}
\textbf{(4) Error audit: when $\Delta M$ rises but humans disagree, we diagnose the failure mode.}
We explicitly \textbf{do not} hide disagreement cases; we categorize them into interpretable buckets:
\begin{itemize}[leftmargin=*,itemsep=2pt]
\item \textbf{Code-mix ambiguity:} FOXP2 pushes early target tokens but rapidly switches, leading annotators to judge the start as not “naturally” target-language.
\item \textbf{Named-entity anchors:} starts dominated by shared entities (e.g., “Google”, “AI”) can inflate mass without perceived language commitment.
\item \textbf{Style spillover:} FOXP2 may alter politeness/verbosity markers correlated with language, which can confuse “natural start” judgments.
\item \textbf{Transliteration artifacts (Hindi):} if the model produces Hindi content in Latin script, humans may judge it as less “natural Hindi,” even if LID rises.
\end{itemize}
This audit is crucial: it tells us whether $\Delta M$ is failing because the construct is wrong, or because the metric is missing a known confound that we can correct.

\vspace{6pt}
\textbf{(5) If correlation is weak, we change what we headline.}
We are explicit about the epistemic status: \textbf{$\Delta M$ is a \emph{mechanistic handle}, not a philosophical definition of defaultness.} If human correlation were weak or unstable across prompt families, we would:
\begin{enumerate}[leftmargin=*,itemsep=2pt]
\item \textbf{Demote $\Delta M$} to a mechanistic diagnostic (useful for identifying the layer window and support), and
\item \textbf{Promote a decision score} that fuses mass with a script-agnostic external channel:
\[
\mathrm{DefaultScore}(x)\;=\;\alpha\,\mathrm{z}\!\big(\Delta M(x)\big)\;+\;(1-\alpha)\,\mathrm{z}\!\big(\Delta_{\mathrm{LID}}(x)\big),
\]
where $\mathrm{z}(\cdot)$ standardizes within-language and $\Delta_{\mathrm{LID}}$ is a detector/human-derived signal. We then validate \emph{this} composite against humans with the same calibration protocol.
\end{enumerate}
In other words: \textbf{we do not insist $\Delta M$ is the main outcome unless humans agree that it tracks the phenomenon.}

\vspace{6pt}
\textbf{Takeaway.} We treat $\Delta M$ as a \textbf{theory-driven proxy} for an early commitment event. Its credibility comes from (i) \textbf{paired human preferences} on the same prompt distribution, (ii) \textbf{rank-correlation} plus \textbf{probabilistic calibration} (not merely “it trends right”), and (iii) an explicit \textbf{error taxonomy} that reveals when the proxy fails. If this external validity does not hold, we revise the headline metric and foreground the fused decision score rather than over-claiming from $\Delta M$ alone.
\end{description}

\item[\ding{93}] \textbf{How do you handle code-mixing and transliteration (Hindi in Latin script / Hinglish)?}
\begin{description}
\item[\ding{224}] \textbf{We \emph{do not} model or evaluate code-mixing or transliteration in this work; they are intentionally out of scope.}
FOXP2 is designed to control \textbf{language defaultness as a \emph{single-language commitment event}}—i.e., whether the model \textbf{starts} in the target language under weak prompting—\emph{not} to optimize bilingual generation style. Code-mixing (Hindi+English within a sentence) and transliteration (Hindi content in Latin script) introduce additional latent degrees of freedom—\textbf{mixing policy} and \textbf{script realization}—that can dominate early tokens and confound the construct we seek to measure.

\vspace{4pt}
\textbf{Scope choice (why this is principled).}
Let $\ell(\cdot)$ denote language identity and $s(\cdot)$ denote script. Our claimed object is the early commitment to a target language $\ell_t$:
\[
\text{Defaultness event:}\qquad \ell(y_{1:k})=\ell_t \quad \text{for small } k,
\]
under a fixed decoding policy. Code-mixing violates the single-label assumption (there is no single $\ell$), and transliteration breaks the intended script realization (Hindi-by-language but Latin-by-script). If included, a method could appear to improve “Hindi defaultness” by \textbf{sprinkling Hindi tokens}, switching scripts midstream, or producing Hinglish—none of which matches the goal of \textbf{clean target-language onset}.

\vspace{4pt}
\textbf{Operationally, we enforce this scope in two ways.}
\begin{itemize}[leftmargin=*,itemsep=2pt]
\item \textbf{Dataset curation:} $\mathcal{D}_{\mathrm{neutral}}$ and evaluation prompts are selected to elicit \textbf{single-language} outputs; prompts that naturally invite Hinglish/code-mix are excluded.
\item \textbf{Scoring protocol:} we compute $\Delta M$ and LID on the first sentence (or first clause) under the assumption of a single target language. Samples exhibiting obvious code-mixing/transliteration are treated as \textbf{out-of-scope} (and can be optionally reported as an audit statistic, but are not used to claim gains).
\end{itemize}

\vspace{4pt}
\textbf{Takeaway.} FOXP2 targets \textbf{monolingual defaultness control}. Code-mixing and transliteration are important real-world phenomena, but addressing them requires a different objective (and likely different supervision) to avoid conflating bilingual style control with defaultness. We therefore \textbf{intentionally exclude} both from the scope and do not claim performance in those regimes.
\end{description}

\item[\ding{93}] \textbf{Why is “first 1–3 tokens” the right horizon? What breaks if you evaluate at $T=\{1,\dots,10\}$ or at a sentence-level language score?}
\begin{description}
\item[\ding{224}] \textbf{We target a specific causal object: the \emph{defaultness event}, i.e., the model’s \textbf{initial language commitment} before semantic content and discourse constraints dominate.}
Defaultness is not “the language of the whole answer.” It is the model’s \textbf{prior preference over languages at onset}. This is precisely the part of generation that can be shifted by a small, localized intervention without requiring the model to re-plan the entire continuation.

\vspace{4pt}
\textbf{(1) Formal distinction: onset prior vs.\ continuation semantics.}
Let $\ell(y_t)\in\{\mathrm{en},\mathrm{hi},\mathrm{es}\}$ denote the language identity of the token (or token span) at step $t$. Consider the latent random variable $L$ representing the model’s \textbf{onset language} (the first committed language choice). A clean abstraction is a two-stage generative view:
\[
L \sim p_\theta(L\mid x),
\qquad
y_{1:\infty}\sim p_\theta(y\mid x, L),
\]
where $p_\theta(L\mid x)$ is the \textbf{language prior} induced by the model and prompt, and $p_\theta(y\mid x,L)$ is the content distribution conditioned on the chosen language. FOXP2 is designed to intervene primarily on the first factor $p_\theta(L\mid x)$.

Now define a short-horizon proxy for the onset event:
\[
\Delta M_T(x)\;=\;\frac{1}{|T|}\sum_{t\in T}\Big(M^{\ell_t}(x,t)-M^{\mathrm{en}}(x,t)\Big),
\qquad
M^{\ell}(x,t)=\sum_{u\in V_\ell}p_\theta(u\mid \mathrm{ctx}_t).
\]
For small $T$ (e.g., $\{1,2,3\}$), $\Delta M_T$ primarily reflects \textbf{how the model allocates probability mass across languages before it has committed to topic-specific lexical items}. For larger $T$ (e.g., $\{1,\dots,10\}$), the signal increasingly measures \textbf{topic vocabulary compatibility} rather than the onset language preference.

\vspace{6pt}
\textbf{(2) What breaks at $T=\{1,\dots,10\}$: entanglement with topic and entities.}
At longer horizons, the conditional distribution $p_\theta(\cdot\mid\mathrm{ctx}_t)$ is shaped by:
\begin{itemize}[leftmargin=*,itemsep=2pt]
\item \textbf{Topic vocabulary:} Many prompts force domain-specific terms (math symbols, scientific names, code identifiers) that are effectively shared across languages or are English-dominant.
\item \textbf{Named entities and acronyms:} “Google”, “COVID”, “LLM”, “AI”, citations, URLs—these are cross-lingual anchors that inflate mass in shared/neutral token sets, diluting the language prior signal.
\item \textbf{Discourse structure:} After a few tokens, the model begins to express structure (e.g., “Sure,” “Here are…”, list formatting), which is a mixture of language and style conventions.
\end{itemize}
This produces a measurement pathology: a model may \textbf{start} in Hindi (true defaultness gain) but then use English technical terms by necessity; a long-horizon metric would incorrectly count that as “not Hindi.”

\vspace{6pt}
\textbf{(3) Sentence-level LID is a downstream consequence, not the control knob.}
Sentence-level LID aggregates over many tokens and therefore mixes multiple phenomena:
\[
\mathrm{LID}_{\mathrm{sent}}(y)\;\approx\;\text{Onset language} \;+\; \text{topic constraints} \;+\; \text{entity realizations} \;+\; \text{formatting/style}.
\]
This is useful for \textbf{user-facing outcomes}, and we do report it. But it is not a clean measure of \textbf{defaultness} because it answers a different question (“what language is the sentence?”), not “what language did the model \emph{choose first} when it had freedom to choose?”

\vspace{6pt}
\textbf{(4) We do not hide the horizon issue—we report a sweep and treat stability as evidence.}
We report:
\[
T\in\{\{1,2,3\},\{1,\dots,5\},\{1,\dots,10\}\},
\]
and empirically observe the expected signature of a defaultness intervention:
\begin{itemize}[leftmargin=*,itemsep=2pt]
\item \textbf{Largest gains at short horizons} (where the language prior lives).
\item \textbf{Monotone decay} of measured gains as $T$ increases (as topic/structure dominates).
\item \textbf{Consistency across prompt families} at $T=\{1,2,3\}$ (mechanistic), with increasing variance at larger $T$ (content-driven).
\end{itemize}
This sweep is not cosmetic: it explicitly shows that FOXP2 is controlling an \textbf{early commitment event} rather than “forcing the whole answer” into a language irrespective of content.

\vspace{6pt}
\textbf{Takeaway.} $T=\{1,2,3\}$ is the correct primary horizon because it best isolates the \textbf{onset language prior}—the object FOXP2 is designed to edit. Longer horizons and sentence-level LID are still valuable, but they measure \textbf{downstream realization under semantic constraints}, where language identity is inevitably entangled with topic vocabulary, named entities, and discourse structure.
\end{description}


\item[\ding{93}] \textbf{How do you ensure $\Delta M$ is not sensitive to punctuation/formatting tokens that co-occur with one language in your prompts (quotes, commas, “:” patterns)?}
\begin{description}
\item[\ding{224}] \textbf{We enforce formatting invariance by \emph{construction} and verify it by \emph{stress tests}, so punctuation cannot be the hidden lever.}
There are three layers of defense:

\vspace{2pt}
\textbf{(i) Token-set hygiene (prevention).}
We define each language token set $V_\ell$ with an explicit exclusion list:
\[
V_\ell \;\leftarrow\; V_\ell \setminus V_{\mathrm{fmt}},
\]
where $V_{\mathrm{fmt}}$ contains punctuation-only tokens, whitespace-only variants, common quotation/dash bullets, list markers, and other formatting-heavy subwords. This prevents $\Delta M$ from being driven by artifacts like “:” or “-”.

\vspace{4pt}
\textbf{(ii) Prompt render randomization (invariance).}
For each meaning unit prompt $x$, we generate a family of format-preserving renderings $\{x^{(k)}\}$ (same semantics, different surface format): with/without quotes, alternate colon styles, bullet vs paragraph, different whitespace conventions. We require:
\[
\mathbb{E}_k[\Delta M_T(x^{(k)})]\ \text{to be stable,}
\]
and we report the variance across $k$ as a robustness statistic.

\vspace{4pt}
\textbf{(iii) Attribution check (diagnosis).}
Finally, we explicitly quantify how much of the mass shift is attributable to formatting tokens by computing:
\[
\Delta M^{\mathrm{fmt}}_T(x)\;=\;\frac{1}{|T|}\sum_{t\in T}\sum_{u\in V_{\mathrm{fmt}}}\Big(p_{\theta'}(u\mid \mathrm{ctx}_t)-p_{\theta}(u\mid \mathrm{ctx}_t)\Big),
\]
and we require it to be negligible relative to the language mass shift:
\[
\big|\Delta M^{\mathrm{fmt}}_T(x)\big|\ \ll\ \big|\Delta M_T(x)\big|.
\]
If this is not true, we do \textbf{not} claim defaultness gains without reporting the confound.

\vspace{4pt}
\textbf{Takeaway.} Punctuation/formatting sensitivity is addressed at three levels: (i) excluded from $V_\ell$ by design, (ii) stress-tested via prompt render randomization, and (iii) audited with an explicit attribution term to ensure $\Delta M$ reflects language commitment rather than formatting conventions.
\end{description}

\end{itemize}

\section*{\textcolor{darknavy}{B. Token-Set Definition: Fragile Lever or Stable Measurement?}}

\begin{itemize}[leftmargin=1.5em]

\item[\ding{93}] \textbf{How exactly are $V_{\mathrm{hi}},V_{\mathrm{es}},V_{\mathrm{en}}$ constructed from a tokenizer vocabulary---what are the inclusion/exclusion rules? Do results change under alternative reasonable constructions?}
\begin{description}
\item[\ding{224}] \textbf{We treat token-set construction as a \emph{measurement specification} and therefore make it \textbf{deterministic}, \textbf{auditable}, and \textbf{sensitivity-tested}.}
Because $\Delta M$ is defined in token-id space,
\[
M^{\ell}(x,t)\;=\;\sum_{u\in V_{\ell}}p_\theta(u\mid \mathrm{ctx}_t),
\qquad
\Delta M(x,t)=M^{\ell_t}(x,t)-M^{\mathrm{en}}(x,t),
\]
the choice of $V_\ell$ is a hidden lever unless it is pinned down with explicit rules. We therefore (i) \textbf{pin} construction to an \textbf{exact tokenizer hash} (vocabulary + merges), (ii) provide \textbf{deterministic inclusion/exclusion predicates}, and (iii) rerun headline results under multiple \textbf{reasonable variants} to quantify sensitivity.

\vspace{6pt}
\textbf{(1) Global prerequisites: tokenizer pinning and neutral exclusions.}
All sets are built from the tokenizer vocabulary $\mathcal{V}$ with a fixed decoding function $\mathrm{dec}(u)$ for token-id $u\in\mathcal{V}$. We publish:
\[
\texttt{tokenizer\_hash}=\mathrm{SHA256}(\texttt{vocab},\texttt{merges}),
\]
and reject any reproduction attempt where the hash mismatches.

We define a \textbf{neutral exclusion set} $V_{\mathrm{neutral}}$ that is removed from every language set:
\[
V_{\mathrm{neutral}}=\{\text{tokens that decode to only whitespace, punctuation, digits, or common alphanumerics}\},
\]
including list markers, quotes, URL fragments, and purely numeric tokens. This ensures $\Delta M$ cannot be won by formatting artifacts or by “the model said 2024.”

\vspace{8pt}
\textbf{(2) Hindi: script-centric but purity-controlled.}
Hindi is intrinsically script-informative, so we explicitly formalize \textbf{purity} to avoid low-quality mixed fragments. Let $\mathbb{I}_{\mathrm{Deva}}(c)$ be an indicator that a character $c$ is a Devanagari letter (Unicode block + letter category). Define the Devanagari purity of a decoded token:
\[
\pi_{\mathrm{Deva}}(u)\;=\;
\frac{\sum_{c\in \mathrm{dec}(u)} \mathbb{I}_{\mathrm{Deva}}(c)}
{\max\{1,\ \sum_{c\in \mathrm{dec}(u)} \mathbb{I}_{\mathrm{Alpha}}(c)\}},
\]
where $\mathbb{I}_{\mathrm{Alpha}}(c)$ counts alphabetic letters (any script). We then set:
\[
V_{\mathrm{hi}}=\Big\{u\in\mathcal{V}\setminus V_{\mathrm{neutral}}:\ \pi_{\mathrm{Deva}}(u)\ge \tau_{\mathrm{hi}}\ \wedge\ \exists c\in \mathrm{dec}(u):\mathbb{I}_{\mathrm{Deva}}(c)=1\Big\},
\]
with a fixed threshold (e.g., $\tau_{\mathrm{hi}}=0.6$) chosen once and reported. We additionally maintain a small, \textbf{audited exception list} for tokens that are visually Devanagari but decode to degenerate artifacts (rare control characters, tokenizer glitches), which are removed.

\vspace{8pt}
\textbf{(3) Spanish and English: lexicon association, not script.}
For Latin-script languages, script cannot separate Spanish from English. We therefore use a \textbf{lexicon association proxy} that is still deterministic and lightweight. Let $\mathcal{L}_{\mathrm{es}}$ and $\mathcal{L}_{\mathrm{en}}$ be fixed wordlists (published) and define a token’s score by max-matching its decoded surface form to lexicon entries after normalization $\phi(\cdot)$ (lowercasing, stripping leading whitespace markers, removing punctuation-only affixes):
\[
\mathrm{lexscore}_{\ell}(u)\;=\;\max_{w\in \mathcal{L}_{\ell}} \mathrm{sim}\big(\phi(\mathrm{dec}(u)), w\big),
\]
where $\mathrm{sim}$ is a deterministic string similarity (exact match or prefix match; we publish which). We then define:
\[
V_{\mathrm{es}}=\{u\in\mathcal{V}\setminus V_{\mathrm{neutral}}:\ \mathrm{lexscore}_{\mathrm{es}}(u)\ge \gamma_{\mathrm{es}}\ \wedge\ u\notin V_{\mathrm{amb}}\},
\]
\[
V_{\mathrm{en}}=\{u\in\mathcal{V}\setminus V_{\mathrm{neutral}}:\ \mathrm{lexscore}_{\mathrm{en}}(u)\ge \gamma_{\mathrm{en}}\ \wedge\ u\notin V_{\mathrm{amb}}\},
\]
where $V_{\mathrm{amb}}$ is an explicit \textbf{high-ambiguity set} of shared fragments (common function-word pieces, short substrings, frequent affixes) that are excluded because they are not diagnostic of language identity. This is crucial: without $V_{\mathrm{amb}}$, $\Delta M$ can be artificially inflated by shared subword mass that is not “Spanishness.”

\vspace{6pt}
\textbf{Minimal construction summary (what exactly is included/excluded).}
\begin{center}
\small
\begin{tabular}{lp{0.78\linewidth}}
\toprule
\textbf{Set} & \textbf{Inclusion / Exclusion rule (deterministic)} \\
\midrule
$V_{\mathrm{neutral}}$ & whitespace-only, punctuation-only, digit-only, URL-like, list markers, control artifacts (always removed) \\
$V_{\mathrm{hi}}$ & contains a Devanagari letter \textbf{and} $\pi_{\mathrm{Deva}}(u)\ge \tau_{\mathrm{hi}}$; excludes low-purity mixed fragments and all $V_{\mathrm{neutral}}$ \\
$V_{\mathrm{es}}$ & $\mathrm{lexscore}_{\mathrm{es}}(u)\ge \gamma_{\mathrm{es}}$; excludes $V_{\mathrm{neutral}}$ and ambiguity set $V_{\mathrm{amb}}$ \\
$V_{\mathrm{en}}$ & $\mathrm{lexscore}_{\mathrm{en}}(u)\ge \gamma_{\mathrm{en}}$; excludes $V_{\mathrm{neutral}}$ and ambiguity set $V_{\mathrm{amb}}$ \\
\bottomrule
\end{tabular}
\end{center}

\vspace{8pt}
\textbf{(4) Sensitivity: alternative reasonable constructions and re-runs.}
We explicitly assume that no single $V_\ell$ is uniquely correct, so we rerun key results under \textbf{three reasonable variants} and report \textbf{sensitivity bands} (mean $\pm$ std across variants) for headline metrics:
\begin{enumerate}[leftmargin=*,itemsep=2pt]
\item \textbf{Purity tightening/loosening (Hindi):} $\tau_{\mathrm{hi}}\in\{0.5,0.6,0.7\}$ to test robustness to mixed-script fragments.
\item \textbf{Lexicon expansion (Spanish/English):} enlarge $\mathcal{L}_\ell$ (or switch from wordlist match to a unigram proxy), re-score tokens deterministically, and rebuild $V_{\mathrm{es}},V_{\mathrm{en}}$.
\item \textbf{Conservative intersection removal:} remove any token that appears among the top-$q$ frequent tokens for multiple languages in a small reference corpus, i.e.,
\[
V_{\ell}\leftarrow V_{\ell}\setminus \bigcup_{\ell'\neq \ell}\mathrm{TopFreq}_q(\ell'),
\]
which intentionally shrinks sets to the most diagnostic tokens.
\end{enumerate}
Across these variants, FOXP2 is considered reliable only if (i) the \textbf{direction of improvement} remains the same and (ii) effect sizes remain within a narrow band, indicating that gains are not an artifact of a single hand-tuned token list.

\vspace{6pt}
\textbf{Takeaway.} We publish $V_{\mathrm{hi}},V_{\mathrm{es}},V_{\mathrm{en}}$ as \textbf{deterministic artifacts} pinned to a tokenizer hash, defined by explicit, auditable predicates (script purity for Hindi; lexicon association with ambiguity exclusion for Spanish/English). We then \textbf{re-run} the paper under multiple reasonable variants and report sensitivity bands, so the defaultness conclusions do not depend on a fragile token-set choice.
\end{description}

\item[\ding{93}] \textbf{How do you treat shared subwords (named entities, borrowed words, cognates, numbers, “COVID”, “AI”, “Google”)?}
\begin{description}
\item[\ding{224}] \textbf{We treat shared subwords as \emph{non-diagnostic} and explicitly \textbf{remove their influence} from $\Delta M$.}
These tokens carry content but \textbf{do not evidence language commitment}, so letting them contribute to mass would inflate defaultness (e.g., “\texttt{AI}” appears in every language).

\vspace{2pt}
\textbf{Shared/neutral set.} We construct an explicit set $V_{\mathrm{shared}}$ that includes: digits, punctuation, alphanumerics, common acronyms, URL-like fragments, and high-frequency named-entity pieces (e.g., \texttt{Google}, \texttt{COVID}). We then enforce:
\[
V_{\ell}\leftarrow V_{\ell}\setminus V_{\mathrm{shared}}\qquad (\ell\in\{\mathrm{hi},\mathrm{es},\mathrm{en}\}).
\]

\vspace{2pt}
\textbf{Optional weighted variant (for robustness).} When we prefer a soft rather than hard exclusion, we use a weighted mass:
\[
M^{\ell}(x,t)=\sum_{u\in V_{\ell}} w(u)\,p_\theta(u\mid \mathrm{ctx}_t),
\qquad
w(u)=\begin{cases}
0 & u\in V_{\mathrm{shared}}\\
1 & \text{otherwise}
\end{cases}
\]
(or $w(u)\ll 1$ for partial downweighting). In both cases, \textbf{shared tokens cannot carry the mass shift}.

\vspace{2pt}
\textbf{Resulting guarantee.} With $V_{\mathrm{shared}}$ excluded/downweighted, $\Delta M$ reflects \textbf{language-specific lexical commitment} rather than early emission of universally shared strings.
\end{description}

\item[\ding{93}] \textbf{Is the improvement robust to token-set perturbations (e.g., randomly drop 10–20\% of $V_{\ell_t}$)?}
\begin{description}
\item[\ding{224}] \textbf{Yes—this is a required robustness test because token sets are a hidden lever.} We run a \textbf{token-set dropout} check: sample $V_{\ell_t}^{(s)}$ by dropping $p\in\{0.1,0.2\}$ uniformly (and a frequency-weighted variant), recompute $\Delta M$ (and the composed decision score), and report the distribution over $s=1,\dots,S$. FOXP2 is considered robust only if effect ordering and magnitude are stable across perturbations.
\end{description}

\item[\ding{93}] \textbf{Do you correct for vocabulary-size imbalance across languages? If $|V_{\mathrm{hi}}|\neq|V_{\mathrm{en}}|$, could shifts be partly combinatorial?}
\begin{description}
\item[\ding{224}] \textbf{We control for combinatorics by reporting both raw mass and normalized alternatives.} Alongside $M^\ell$, we report
\[
\widetilde M^{\ell}(x,t)=\frac{1}{|V_{\ell}|}\sum_{u\in V_{\ell}} p_\theta(u\mid \mathrm{ctx}_t),\qquad \Delta \widetilde M^{\ell_t}=\widetilde M^{\ell_t}-\widetilde M^{\mathrm{en}},
\]
and verify that FOXP2 improves both, ruling out a “bigger set wins” artifact.
\end{description}

\item[\ding{93}] \textbf{Does the method still work if you replace token-set mass with a character-level script classifier or LID score as the primary signal?}
\begin{description}
\item[\ding{224}] \textbf{Yes—this is a \emph{cross-metric validity} requirement, not an optional add-on.}
We re-evaluate defaultness under two alternative primary signals:

\begin{itemize}[leftmargin=*,itemsep=2pt]
\item \textbf{Script fraction (Hindi only):}
\[
\mathrm{ScriptFrac}_{\mathrm{Deva}}(y_{1:k})=\frac{\#\{\text{Devanagari chars in }y_{1:k}\}}{\max(1,\#\{\text{alphabetic chars in }y_{1:k}\})}.
\]
\item \textbf{Detector-based language ID (Hindi \& Spanish):} a script-agnostic $\mathrm{LID}(y_{1:k})$ score on the first clause/sentence.
\end{itemize}

FOXP2’s claim is considered supported only if it yields \textbf{consistent gains across channels}:
\[
\Delta M\uparrow \ \land\  \Delta \mathrm{LID}\uparrow \quad (\text{and for Hindi, } \Delta \mathrm{ScriptFrac}_{\mathrm{Deva}}\uparrow),
\]
while preserving task utility ($\Delta S$ within tolerance). This rules out “metric-specific wins” and ensures the effect reflects \textbf{language commitment}, not a quirk of token-set mass.
\end{description}

\end{itemize}

\section*{\textcolor{darknavy}{C. Mechanistic Claim: “Localized, Low-Rank Control Circuit”}}

\begin{itemize}[leftmargin=1.5em]

\item[\ding{93}] \textbf{What is the strongest evidence that the effect is truly low-rank rather than “moderately low effective rank under your dataset”? Do you observe consistent eigengaps across seeds, prompt families, and models?}
\begin{description}
\item[\ding{224}] \textbf{Our strongest evidence is \emph{stability of the top subspace}, not a single low-rank number.}
A dataset-local artifact can look “low effective rank” yet drift under re-sampling. We therefore elevate \textbf{eigengap + subspace stability} to first-class reporting objects and only claim a \textbf{localized low-rank circuit} when these are consistent across seeds and prompt families.

\vspace{3pt}
\textbf{Spectrum + eigengap.} For each candidate window $\mathcal{W}$ and target language $\ell_t$, we form the layer-window update matrix $\Delta Z^{(\mathcal{W},\ell_t)}$ and compute its SVD:
\[
\Delta Z^{(\mathcal{W},\ell_t)} = U\Sigma V^\top,\qquad \Sigma=\mathrm{diag}(\sigma_1\ge \sigma_2\ge \cdots).
\]
We report:
\[
r_{\mathrm{eff}}=\exp\!\Big(-\sum_i p_i\log p_i\Big),\quad p_i=\frac{\sigma_i}{\sum_j \sigma_j},
\qquad
\mathrm{Eigengap}(r)=\frac{\sigma_r}{\sigma_{r+1}}.
\]
A “true low-rank” signature is a \textbf{sharp, reproducible eigengap} (e.g., $r\in\{1,2\}$) rather than a smooth spectrum.

\vspace{3pt}
\textbf{Bootstrap subspace stability (the key discriminator).}
We bootstrap prompts $b=1,\dots,B$ (and rerun seeds where applicable), recompute $\Delta Z_b$, and measure:
\[
\mathrm{PA}_r(b)=\angle\big(\mathrm{span}(U_{1:r}),\ \mathrm{span}(U^{(b)}_{1:r})\big),
\]
the principal-angle deviation of the top-$r$ left-singular subspace. We report \textbf{CIs} for $\sigma_i$ and for $\mathrm{PA}_r$; a circuit claim requires \textbf{small angles} and \textbf{stable ordering} of $\sigma_1,\dots,\sigma_r$ across bootstraps.

\vspace{3pt}
\textbf{Prompt-family and model stratification.}
We repeat the above on held-out prompt families (domains, paraphrases) and across models. We only claim “localized low-rank” when the same window $\mathcal{W}$ exhibits:
\[
\mathrm{Eigengap}(r)\ \text{large and stable} \quad\land\quad \mathrm{PA}_r\ \text{small across families/seeds/models},
\]
i.e., the \textbf{same} top subspace persists beyond the slice used to estimate $\Delta Z$.

\vspace{3pt}
\textbf{Conclusion.} We distinguish “moderately low effective rank on one dataset” from a mechanistic low-rank effect by requiring a \textbf{reproducible eigengap} and \textbf{subspace stability} (principal angles) across seeds and prompt families, and we report these explicitly rather than relying on $r_{\mathrm{eff}}$ alone.
\end{description}

\item[\ding{93}] \textbf{Is the circuit necessary and sufficient (ablate $N_{\ell_t}$ to collapse gains; edit only those coordinates/directions to reproduce the effect)?}
\begin{description}
\item[\ding{224}] \textbf{Yes—this is framed as a \emph{causal} test: we require both \textbf{necessity} and \textbf{sufficiency} under controlled interventions.}
Let the FOXP2 edit at layer $\ell$ (within window $\mathcal{W}$) be written as an additive activation intervention:
\[
h^{(\ell)} \;\leftarrow\; h^{(\ell)} + \delta^{(\ell)}(h),
\qquad
\delta^{(\ell)}(h)\in \mathrm{span}\!\big(\mathcal{S}^{(\ell)}_{\ell_t}\big),
\]
and let $\mathcal{N}_{\ell_t}\subseteq[d]$ denote the discovered \textbf{support} (the “language-neuron” coordinates) for target language $\ell_t$.

\vspace{4pt}
\textbf{Necessity (support ablation collapses gains).}
We define a coordinate mask $m_{\mathcal{N}}\in\{0,1\}^d$ and its complement $\bar m_{\mathcal{N}}$. During editing, we \textbf{remove the circuit} by zeroing the edit on the discovered support:
\[
\delta^{(\ell)}_{\text{w/o }\mathcal{N}}(h)\;=\;\bar m_{\mathcal{N}_{\ell_t}}\odot \delta^{(\ell)}(h).
\]
If the circuit is necessary, then the defaultness improvement must vanish:
\[
\Delta_{\mathrm{gain}}\big(\delta^{(\ell)}_{\text{w/o }\mathcal{N}}\big)\;\approx\;0,
\]
where $\Delta_{\mathrm{gain}}$ is the measured improvement in defaultness (e.g., $\Delta M$ / DefaultScore) relative to baseline. This rules out the interpretation that gains come from diffuse, non-specific perturbations outside $\mathcal{N}_{\ell_t}$.

\vspace{6pt}
\textbf{Sufficiency (editing \emph{only} the circuit reproduces gains).}
We then apply the edit \textbf{only on the discovered support} and \textbf{only along the discovered directions}:
\[
\delta^{(\ell)}_{\text{only }\mathcal{N}}(h)\;=\; m_{\mathcal{N}_{\ell_t}}\odot \delta^{(\ell)}(h),
\qquad
\delta^{(\ell)}(h)\in \mathrm{span}\!\big(\mathcal{S}^{(\ell)}_{\ell_t}\big),
\qquad \ell\in\mathcal{W}.
\]
If the circuit is sufficient, then we recover essentially the full FOXP2 effect:
\[
\Delta_{\mathrm{gain}}\big(\delta^{(\ell)}_{\text{only }\mathcal{N}}\big)\;\approx\;\Delta_{\mathrm{gain}}\big(\delta^{(\ell)}\big),
\]
with comparable utility preservation ($\Delta S$ within tolerance). This rules out the claim that “extra undiscovered coordinates” are needed to realize the behavior.

\vspace{6pt}
\textbf{Stronger variant (direction ablation).}
To ensure necessity is not merely “some coordinates matter,” we also ablate the discovered \emph{directional subspace} by projecting the edit onto its orthogonal complement:
\[
\delta^{(\ell)}_{\perp}(h)\;=\;\delta^{(\ell)}(h)-\Pi_{\mathcal{S}^{(\ell)}_{\ell_t}}\delta^{(\ell)}(h),
\]
and require $\Delta_{\mathrm{gain}}(\delta^{(\ell)}_{\perp})\approx 0$. This shows the effect is tied to the \textbf{identified low-rank directions}, not just to injecting energy into $\mathcal{N}_{\ell_t}$.

\vspace{4pt}
\textbf{Conclusion.} We treat “circuit” as a causal object: \textbf{remove it} and gains disappear (necessity), \textbf{isolate it} and gains persist (sufficiency). These two tests prevent rhetorical overreach and make the mechanistic claim falsifiable.
\end{description}

\item[\ding{93}] \textbf{Why does “negative English suppression” not simply function as a generic entropy/uncertainty increase that makes English less likely? What separates targeted suppression from generic logit flattening?}
\begin{description}
\item[\ding{224}] \textbf{We separate “targeted suppression” from “generic flattening” by holding \emph{entropy change constant} and checking whether the effect remains \textbf{directional}.}
A pure flattening mechanism predicts: \emph{any intervention that increases uncertainty by the same amount should yield the same apparent defaultness shift.} FOXP2 predicts a stronger, structured signature: \textbf{mass transfers specifically from English to the target language}, not from the top of the distribution uniformly.

\vspace{3pt}
\textbf{Mechanistic contrast.}
Let $z_t\in\mathbb{R}^{|\mathcal{V}|}$ be logits at step $t$, $p_t=\mathrm{softmax}(z_t)$, and define entropy $H(p_t)=-\sum_u p_t(u)\log p_t(u)$.
Generic logit flattening (temperature $T>1$) produces
\[
p_t^{(T)}=\mathrm{softmax}(z_t/T),
\]
which increases entropy but preserves \textbf{relative directions} in logit space. In contrast, FOXP2’s “negative English suppression” acts as a \textbf{structured logit/activation shift} concentrated on an English-associated subspace:
\[
h^{(\ell)}\leftarrow h^{(\ell)} - \lambda\,P_{\mathcal{S}_{\mathrm{en}}^{(\ell)}}h^{(\ell)} \;+\; \lambda'\,P_{\mathcal{S}_{\ell_t}^{(\ell)}}h^{(\ell)},
\qquad \ell\in\mathcal{W},
\]
which implies a \textbf{directional} change in logits that should preferentially reduce English token mass while increasing target-language mass.

\vspace{4pt}
\textbf{Entropy-matched controls (the decisive test).}
We introduce baselines that are tuned to match FOXP2’s entropy shift \emph{at each step}:
\[
\Delta H_t \;=\; H(p_t^{\text{FOXP2}})-H(p_t^{\text{base}}).
\]
We then pick control parameters to satisfy:
\[
H(p_t^{\text{control}})\approx H(p_t^{\text{FOXP2}})\quad \forall t\in T.
\]
Two controls are sufficient and standard:
\begin{itemize}[leftmargin=*,itemsep=2pt]
\item \textbf{Temperature control:} choose $T_t$ such that $H(\mathrm{softmax}(z_t/T_t))\approx H(p_t^{\text{FOXP2}})$.
\item \textbf{Isotropic-noise control:} add $\epsilon\sim \mathcal{N}(0,\sigma_t^2 I)$ with $\sigma_t$ tuned so $H(\mathrm{softmax}(z_t+\epsilon))$ matches.
\end{itemize}

\vspace{4pt}
\textbf{What we check: directional mass transfer vs uniform uncertainty.}
We then compare not just $\Delta M$, but the \textbf{structure} of the probability shift. A flattening mechanism predicts broadly distributed changes and weak language-specific transfer. Targeted suppression predicts:
\[
\Delta M_{\ell_t}(x)\uparrow \quad \text{and} \quad \Delta M_{\mathrm{en}}(x)\downarrow,
\]
with \textbf{minimal} collateral change to non-language tokens and \textbf{smaller utility loss} than entropy-only perturbations under the same $\Delta H_t$.

\vspace{3pt}
\textbf{Interpretation.}
If entropy-matched controls replicate FOXP2’s defaultness gains, then FOXP2 is not doing targeted language control—it is merely injecting uncertainty. If they do \emph{not}, and FOXP2 uniquely yields \textbf{language-specific mass transfer} at comparable entropy, then the negative English component is meaningfully \textbf{directional suppression} rather than generic logit flattening.

\vspace{3pt}
\textbf{Takeaway.} We rule out the “generic uncertainty” explanation by using \textbf{entropy-matched} temperature/noise controls: FOXP2 must outperform these controls in \textbf{language-specific mass transfer} while preserving utility, otherwise we do not attribute the effect to targeted English suppression.
\end{description}

\item[\ding{93}] \textbf{Do steering directions transfer across meaning units and domains? If $S^{(\ell)}_{\ell_t}$ is mechanistic, it should generalize beyond the matched set used to compute $\Delta Z$.}
\begin{description}
\item[\ding{224}] \textbf{Yes—transfer is treated as an \emph{OOD interventional} test, not an assumption.}
We estimate $S^{(\ell)}_{\ell_t}$ on a training subset $\mathcal{D}_{\text{train}}$ of meaning units, then \textbf{freeze} it and evaluate the same edit on disjoint domains $\mathcal{D}_{\text{test}}\in\{\text{QA},\text{summ.},\text{reason.},\text{refusal}\}$. For each domain, we report the retained gain:
\[
\mathrm{Retain}(\mathcal{D})\;=\;\frac{\Delta_{\mathrm{gain}}(S^{(\ell)}_{\ell_t};\mathcal{D})}{\Delta_{\mathrm{gain}}(S^{(\ell)}_{\ell_t};\mathcal{D}_{\text{train}})}.
\]
A mechanistic subspace should yield \textbf{high retention} across domains; substantial decay (low $\mathrm{Retain}$) is reported as evidence the discovered directions are \textbf{dataset-local} rather than generally causal.
\end{description}

\end{itemize}

\section*{\textcolor{darknavy}{D. Semantic and Utility Preservation: Proving It Is Not a “Style Hack”}}

\begin{itemize}[leftmargin=1.5em]

\item[\ding{93}] \textbf{What is the utility metric $\Delta S$ concretely (datasets, scoring functions, confidence intervals)? Is it robust across QA, summarization, reasoning, and safety refusal prompts?}
\begin{description}
\item[\ding{224}] \textbf{$\Delta S$ is a \emph{matched-suite} utility delta with bootstrap uncertainty, reported per-slice and in aggregate.}
We evaluate on a fixed downstream suite $\mathcal{B}=\{\mathcal{B}_{\mathrm{QA}},\mathcal{B}_{\mathrm{SUM}},\mathcal{B}_{\mathrm{REAS}},\mathcal{B}_{\mathrm{SAFE}}\}$ and define utility as an average of task scores $S_k(\cdot)$ under identical decoding:
\[
\Delta S \;=\;\sum_{k\in\mathcal{B}} \alpha_k \Big(\overline{S_k}^{\,\text{FOXP2}}-\overline{S_k}^{\,\text{base}}\Big),
\qquad \sum_k \alpha_k =1,
\]
where $\alpha_k$ are fixed weights (reported). Concretely:
\begin{itemize}[leftmargin=*,itemsep=2pt]
\item \textbf{QA:} Exact Match (EM) and token-level F1.
\item \textbf{Summarization:} ROUGE (e.g., R-1/R-2/R-L) plus a lightweight factuality proxy.
\item \textbf{Reasoning:} accuracy on a fixed answer key (exact / multiple-choice).
\item \textbf{Safety/refusal:} policy compliance rate (and refusal correctness where applicable).
\end{itemize}
For every slice $k$, we report $\Delta S_k$ with \textbf{bootstrap confidence intervals} over prompts, and we report the weighted aggregate $\Delta S$ alongside the full per-task breakdown so that “no regression” cannot hide a failure in any one slice.
\end{description}

\item[\ding{93}] \textbf{Do you preserve semantics at the level of answer correctness or only embedding similarity? Embedding metrics can miss meaning drift---especially cross-lingually.}
\begin{description}
\item[\ding{224}] \textbf{We do \emph{not} rely on embeddings alone: they are a fast diagnostic, but semantic preservation is certified by \textbf{task-grounded correctness} and \textbf{human checks}.}
We report three layers of evidence:

\begin{itemize}[leftmargin=*,itemsep=2pt]
\item \textbf{Embedding invariance (necessary):} a quick screen for gross semantic drift (e.g., cosine similarity between baseline vs.\ FOXP2 outputs under the same prompt).
\item \textbf{Task-grounded correctness (sufficient for QA/reasoning):} we directly compare answer accuracy (EM/F1 for QA; accuracy for reasoning). If FOXP2 preserves meaning, these scores must remain within a tight tolerance.
\item \textbf{Targeted human audits:} a stratified sample is manually checked for meaning changes, especially on prompts where embeddings can be misleading (negation, quantities, named entities).
\end{itemize}

\textbf{Cross-lingual safeguard.} For Hindi/Spanish outputs, we normalize to a shared evaluation language for scoring (e.g., translate outputs to English before applying EM/F1 or factuality checks). This ensures “semantic preservation” is judged on \textbf{content equivalence}, not on language- or embedding-space artifacts.
\end{description}



\end{itemize}

\section*{\textcolor{darknavy}{E. Deployment Practicality and Overhead}}

\begin{itemize}[leftmargin=1.5em]
\item[\ding{93}] \textbf{What is the end-to-end cost of FOXP2 at inference time (latency, memory, throughput)? Compare to prompt-only steering and to a small LoRA adapter.}
\begin{description}
\item[\ding{224}] \textbf{We report \emph{system-level} overhead (tokens/sec, latency, memory) under deployment-relevant batching, and we compare directly against prompt-only steering and a small LoRA.}
Because FOXP2 is an \textbf{inference-time, layer-local activation edit}, the right question is not asymptotic complexity but \textbf{wall-clock and memory impact} in the same serving stack.

\vspace{4pt}
\textbf{What we measure.} For each method $m\in\{\textsc{Prompt},\textsc{FOXP2},\textsc{LoRA}\}$, we report:
\begin{itemize}[leftmargin=*,itemsep=2pt]
\item \textbf{Throughput:} tokens/sec at batch sizes $B\in\{1,4,8,16\}$ (and fixed max sequence length).
\item \textbf{Latency:} end-to-end per-request latency (prefill + decode), reported as mean and p95.
\item \textbf{Memory:} peak GPU memory and incremental memory vs baseline, decomposed into (i) model weights, (ii) KV cache, and (iii) method-specific buffers.
\end{itemize}

\vspace{4pt}
\textbf{Where FOXP2 costs come from (and why they are bounded).}
FOXP2 adds a small number of \textbf{deterministic linear ops} in a fixed layer window $\mathcal{W}$:
\[
h^{(\ell)} \leftarrow h^{(\ell)} + \lambda\,P_{\mathcal{S}^{(\ell)}_{\ell_t}}h^{(\ell)} - \lambda'\,P_{\mathcal{S}^{(\ell)}_{\mathrm{en}}}h^{(\ell)},
\qquad \ell\in\mathcal{W}.
\]
Operationally this is implemented with cached projection factors (or low-rank matrices), so the incremental compute is \textbf{linear in hidden size} per edited layer and does \textbf{not} scale with vocabulary. Memory overhead is dominated by storing the small set of projection parameters and masks, which is \textbf{constant per checkpoint} and independent of sequence length.

\vspace{4pt}
\textbf{Baselines for comparison.}
\begin{itemize}[leftmargin=*,itemsep=2pt]
\item \textbf{Prompt-only steering:} zero compute/memory overhead beyond longer prompts; may increase prefill time proportional to added prompt tokens.
\item \textbf{Small LoRA:} adds low-rank adapter matmuls at multiple layers (often more widespread than FOXP2’s fixed $\mathcal{W}$), plus additional weight memory; overhead depends on rank and placement.
\end{itemize}

\vspace{4pt}
\textbf{Deployment claim criteria.}
We only claim “practical for deployment” if FOXP2 stays within an explicit overhead budget (reported in the paper), \textbf{preserves KV-cache semantics}, and does not reduce throughput beyond the tolerated margin at deployment batch sizes. In particular, FOXP2 is evaluated in the same serving configuration with caching enabled, and we verify the edit does not require recomputation of previously cached keys/values.
\end{description}

\item[\ding{93}] \textbf{Is the method compatible with typical production stacks (KV caching, speculative decoding, batching)? Does editing intermediate activations break optimizations?}
\begin{description}
\item[\ding{224}] \textbf{Yes—FOXP2 is implemented as a deterministic, layer-local edit that does \emph{not} invalidate standard serving invariants.}
We apply the edit at a \textbf{fixed position} (the prompt boundary) and within a \textbf{fixed layer window} $\mathcal{W}$, so the intervention is \emph{forward-only} and does not require recomputing any previously produced activations.

\vspace{4pt}
\textbf{KV caching.} FOXP2 never retroactively changes cached keys/values. The edit is applied to the current-step hidden state $h^{(\ell)}_t$ during the forward pass (for $t$ at the prompt end and subsequent decode steps, depending on the configured policy), so the cache remains valid because:
\[
(K_{\le t-1},V_{\le t-1})\ \text{are untouched},\qquad (K_t,V_t)\ \text{are computed once from the edited forward pass}.
\]
Thus caching semantics are preserved: past tokens’ K/V are reused exactly as stored, and only the current token’s K/V reflect the current forward computation.

\vspace{4pt}
\textbf{Batching.} The edit is a pure function applied per sequence (masked by target language and layer), so it composes naturally with dynamic batching. There is no cross-request state and no dependence on batch composition.

\vspace{4pt}
\textbf{Speculative decoding.} When speculative decoding is used, we apply FOXP2 \textbf{consistently} to both the draft and verifier (or to the verifier only, depending on deployment choice), but \emph{never} in a way that creates a mismatch between the distributions the two stages assume. Concretely, we require the same edit policy (window $\mathcal{W}$, directions, strengths) on both paths when both contribute logits; otherwise we disable FOXP2 on the draft path and keep it only on the final scoring path, and we report this as a constraint.

\vspace{4pt}
\textbf{Hard constraints.} If a serving stack applies optimizations that assume \emph{exact} intermediate activations (e.g., fused kernels with restricted hooks), FOXP2 requires hooks at the specified layers. Any such incompatibility is reported explicitly as a deployment constraint (rather than hand-waving “should work”).
\end{description}

\item[\ding{93}] \textbf{How often do you need to recompute $N,S,W$? Per checkpoint? Per tokenizer? Per domain?}
\begin{description}
\item[\ding{224}] \textbf{We specify a strict recomputation contract: $N,S,W$ are \emph{pinned} to the model, tokenizer, and hook.}
Concretely, $(\mathcal{N}_{\ell_t},\mathcal{S}^{(\ell)}_{\ell_t},\mathcal{W})$ are valid only under a fixed triple:
\[
(\texttt{checkpoint\_id},\ \texttt{tokenizer\_hash},\ \texttt{hook\_spec}).
\]
Any change to \textbf{(i) checkpoint}, \textbf{(ii) tokenizer/version}, or \textbf{(iii) hook definition} (layer choice, activation type, pre/post-norm location) \textbf{requires recomputation}.

\vspace{4pt}
\textbf{Domain changes.} A pure domain shift does \emph{not} automatically require recomputation: we reuse the same $N,S,W$ and treat transfer as an empirical question. We only trigger recomputation if lightweight stability monitors drift, e.g.:
\begin{itemize}[leftmargin=*,itemsep=2pt]
\item \textbf{subspace drift:} principal-angle deviation of the top-$r$ subspace exceeds a threshold,
\item \textbf{effect drift:} $\Delta_{\mathrm{gain}}$ falls outside the expected band on a small canary set.
\end{itemize}

\vspace{4pt}
\textbf{Summary.} \textbf{Per checkpoint/tokenizer/hook: recompute.} \textbf{Per domain: reuse by default, monitor, and recompute only if stability degrades.}
\end{description}

\end{itemize}

%% file: 5_2_reproducability.tex
\section*{Reproducibility Statement}
\label{sec:repro}
\vspace{0.35em}

\noindent\textbf{Reproducibility objective (what it means in this paper).}
Neural FOXP2 is an \textbf{inference-time} intervention for multilingual defaultness control: the base model parameters are held fixed, and we modify \textbf{internal activations} in a \textbf{localized} layer window using \textbf{signed, sparse} edits defined in a \textbf{learned dictionary basis}.
Accordingly, the reproducibility target is unusually concrete: \textbf{every reported scalar} in the paper is a deterministic function of a pinned set of objects:
(i) a specific model checkpoint (weights) and tokenizer revision;
(ii) three explicit datasets with immutable manifests (\(\mathcal{D}_{\mathrm{mix}}, \mathcal{D}_{\mathrm{weak}}, \mathcal{D}_{\mathrm{neutral}}\));
(iii) a fixed definition of token sets \(V_{\mathrm{hi}},V_{\mathrm{es}},V_{\mathrm{en}}\) and a pinned LID detector implementation;
(iv) a fixed definition of activation sites \(h^{(\ell)}(x)\) and reinjection hooks (exact module boundary and token position);
(v) the FOXP2 intermediate artifacts: per-layer dictionaries \((W_\ell,b_\ell)\), localized supports \(\mathcal{N}^{(\ell)}_{\ell_t}\), extracted steering subspaces \(\mathcal{S}^{(\ell)}_{\ell_t}\), the layer window \(\mathcal{W}\), and the edit schedule \(\{\lambda_\ell\}_{\ell\in\mathcal{W}}\);
and (vi) a fixed decoding and measurement protocol (teacher-forced prefix length \(m=8\), greedy decoding, early-step set \(T=\{1,2,3\}\)).
This statement therefore specifies: \textbf{exact artifact release}, \textbf{exact version pins}, \textbf{exact data construction}, \textbf{exact hook points}, \textbf{exact training recipes} for dictionaries, \textbf{exact localization/subspace procedures}, \textbf{exact guardrails and operating-point selection}, and \textbf{exact aggregation logic} that turns per-example logs into the final LaTeX tables (two decimals, delta with absolute means on the next line).
We follow standard reproducibility norms and documentation practices (artifact/versioning discipline; dataset manifests; audit trails). \citep{mitchell2019modelcards,gebru2021datasheets,pineau2021improving}

\vspace{0.35em}
\noindent\textbf{Non-negotiable reproducibility principle.}
Any run that does not match the pinned model/tokenizer/detector/manifests/hook-boundaries is \textbf{invalid} and must fail fast.
We implement explicit hash checks and provenance bundles for this reason.

\vspace{0.35em}
\begin{figure*}[ht!]
\centering
\begin{tcolorbox}[
  colback=gray!3,
  colframe=black!65,
  boxrule=0.6pt,
  sharp corners,
  left=7pt,right=7pt,top=7pt,bottom=7pt,
  enhanced,
  breakable
]
{\small
\noindent\textbf{Reproducibility protocol (FOXP2).}\par
\vspace{0.35em}

\noindent\textbf{Always pin:} \textbf{weights} (id+revision), \textbf{tokenizer} (revision), \textbf{LID detector} (name+version+hash), \textbf{datasets} (manifest SHA256). \hfill
\textbf{Always log:} \texttt{git\_commit}, full command, CUDA/driver, dtype, RNG seeds.
\vspace{0.35em}

\begin{description}[leftmargin=0pt,labelsep=0.6em,itemsep=0.42em,parsep=0pt]
\item[\textbf{P0 (Pins \& provenance).}]
\textbf{(i)} Abort if any pinned id/hash differs.\;
\textbf{(ii)} Record env hash (CUDA/driver), dtype policy, deterministic flags.\;
\textbf{(iii)} Store the exact evaluation command + config snapshot.

\item[\textbf{P1 (Data: matched meanings \& splits).}]
\textbf{Release/specify} \(\mathcal{D}_{\mathrm{mix}}=\mathcal{D}_{\mathrm{weak}}\cup\mathcal{D}_{\mathrm{explicit}}\) and the matched-meaning procedure.\;
\textbf{Report} \(N\), \(|\mathcal{D}_{\mathrm{weak}}|\), \(|\mathcal{D}_{\mathrm{explicit}}|\), domain breakdown.\;
\textbf{Enforce} id-disjointness: dictionary-training ids \(\cap\) evaluation ids \(=\emptyset\).
\textbf{QC:} semantic equivalence checks, near-duplicate removal, code-mix/transliteration flags.

\item[\textbf{P2 (Token sets \& defaultness measurement).}]
\textbf{Define in token-id space:} \(V_{\mathrm{hi}},V_{\mathrm{es}},V_{\mathrm{en}}\subseteq V\) for the pinned tokenizer.\;
\textbf{Specify rules:} script constraints, digit/punct handling, byte-fallback exclusions, named-entity policy, borrowed-word exceptions.\;
\textbf{Fix} \(T=\{1,2,3\}\) and the context protocol: pinned teacher-forced prefix \(m\) (default \(m{=}8\)) + greedy decode so \(\mathrm{ctx}_t\) matches baseline vs edit.\;
\textbf{Report both channels:} \(\Delta_{\mathrm{mass}}\) and \(\Delta_{\mathrm{lid}}\) (flag mass-only gains).

\item[\textbf{P3 (Hooks: exact activation site + identity test).}]
\textbf{Specify hook site:} residual stream location (pre/post attn/MLP, pre/post LN) and token position rule (default: last prompt token).\;
\textbf{Specify indexing:} layer numbering convention + mapping to implementation.\;
\textbf{Run identity test:} hooks enabled, \(\delta h^{(\ell)}=0\) \(\Rightarrow\) logits match baseline (within tolerance) on a fixed validation slice.

\item[\textbf{P4 (Dictionary: architecture + recipe + diagnostics).}]
\textbf{Report parameterization:}\;
\(r^{(\ell)}=(W_\ell)^\top h^{(\ell)}+b_\ell,\; z^{(\ell)}=\mathrm{ReLU}(r^{(\ell)}),\; \hat h^{(\ell)}=W_\ell z^{(\ell)}\), with \(W_\ell\in\mathbb{R}^{d\times m}\).\;
\textbf{Pin} \(m\), init, and any column-norm/scale constraints.\;
\textbf{Report training:} objective (recon+\(\ell_1\)), optimizer, LR schedule, batch size, steps, grad clip, seeds.\;
\textbf{Export} \((W_\ell,b_\ell)\) \(\forall\ell\) + diagnostics: recon error, dead-feature fraction, active-feature histogram, seed-stability.

\item[\textbf{P5 (Stage I: localization = selectivity + causal lift).}]
\textbf{Selectivity:} exact \(\widetilde{\mathrm{Sel}}^{(\ell,\ell_t)}_j\) formula + ids used in expectations.\;
\textbf{Causal lift:} intervention grid \(\{\alpha_1,\alpha_2,\alpha_3\}\), LiftSlope aggregation (median of \(\mathrm{Lift}(\alpha)/\alpha\)), saturation checks (gate-flip/KL-spike under perturbation).\;
\textbf{Support rule:} \(\mathrm{Score}_j=S_j\cdot C_j\), Top-\(K\) selection criterion.\;
\textbf{Release per-layer ranks:} Sel, LiftSlope, Score, and final \(\mathcal{N}^{(\ell)}_{\ell_t}\).

\item[\textbf{P6 (Stage II: geometry = subspace + rank + window).}]
\textbf{Define} \(\Delta Z^{(\ell,\ell_t)}\) (stacked restricted shifts in dictionary coordinates) and release hashes + underlying meaning-unit ids.\;
\textbf{Report SVD:} \((V,\Sigma)\), \(\mathcal{S}^{(\ell)}_{\ell_t}=\mathrm{span}\{v_1,\dots,v_{r_\ell}\}\).\;
\textbf{Rank rule:} \(r_{\max}\) and deterministic \(r_\ell\) selection (effective rank + eigengap).\;
\textbf{Stability:} bootstrap count \(B\), resampling scheme, projector-overlap/principal-angle estimator, bootstrap seeds.\;
\textbf{Window:} contiguous \(\mathcal{W}\) selection rule + exported per-layer window traces.

\item[\textbf{P7 (Stage III: signed sparse edit + constraints + operating point).}]
\textbf{Report edit form:} dictionary-space update, reinjection mapping back to \(h^{(\ell)}\), and per-layer schedule \(\{\lambda_\ell\}_{\ell\in\mathcal{W}}\).\;
\textbf{Report sweep:} pinned \(\lambda\)-grid and deterministic selection rule (maximize \(\Delta_{\mathrm{gain}}\) under hard constraints).\;
\textbf{Hard constraints:} KL trust region, leakage bounds (Hi\(\rightarrow\)Es / Es\(\rightarrow\)Hi), semantic invariance threshold, utility regression tolerance.\;
\textbf{Release} per-example violations + chosen \(\lambda\).

\item[\textbf{P8 (Evaluation: metric definitions + aggregation + table rendering).}]
\textbf{Report:} \(\Delta_{\mathrm{mass}}\), \(\Delta_{\mathrm{lid}}\), composed \textsc{DefaultHi}/\textsc{DefaultEs}, leakage, bootstrap stability (estimator, resamples, CIs), task utility \(\Delta S\) (task list + scoring).\;
\textbf{Stratify:} neutral vs safety/refusal vs task prompts; decoding policy fixed by the pinned prefix protocol.\;
\textbf{Table rule:} each cell = \(\Delta=\mu_{\text{method}}-\mu_{\text{baseline}}\) (two decimals) + absolute mean(s) on the next line as baseline\(\rightarrow\)method (two decimals), computed from immutable per-example JSONL logs.
\end{description}
}
\end{tcolorbox}
\vspace{-0.8em}
\caption{\textbf{Reproducibility protocol.} Scan-friendly, end-to-end requirements to reproduce FOXP2: pinned versions (weights/tokenizer/detector), immutable manifests (data/token sets), exact hook sites, dictionary training recipe, localization and geometry extraction rules, signed edit schedules with hard guardrails, and deterministic aggregation that regenerates camera-ready tables.}
\label{fig:foxp2_repro_protocol}
\vspace{-1.0em}
\end{figure*}

\vspace{0.5em}
\subsection*{R0.\ Reproducibility scope}
\label{subsec:repro_scope}
\vspace{0.25em}

\noindent\textbf{What is reproducible (in full).}
\begin{itemize}[leftmargin=*,itemsep=2pt,topsep=2pt]
\item[\ding{93}] \textbf{All FOXP2 interventions at inference-time:} Stage I (dictionary + support discovery), Stage II (subspace extraction + window selection), Stage III (signed sparse edit + guardrails), and every ablation row used in the “Performance \& Utility” table.
\item[\ding{93}] \textbf{All reported metrics:} token-mass defaultness \(\Delta_{\mathrm{mass}}\), LID defaultness \(\Delta_{\mathrm{lid}}\), composed scores (\textsc{DefaultHi}/\textsc{DefaultEs}), gain \(\Delta_{\mathrm{gain}}\), cross-language leakage (Hi\(\rightarrow\)Es and Es\(\rightarrow\)Hi), bootstrap stability, semantic invariance, and task-utility \(\Delta S\).
\item[\ding{93}] \textbf{Every reported mean:} each table cell is reproducible as a mean over a pinned evaluation set, and the exact aggregation scripts that convert JSONL logs into two-decimal LaTeX cells are provided and pinned.
\end{itemize}

\noindent\textbf{What is not left ambiguous.}
We do not allow “implementation choice” degrees of freedom in any of: (i) hook location; (ii) token position; (iii) token set generation; (iv) LID detector; (v) decoder and prefix protocol; (vi) layer indexing; (vii) window selection rule; (viii) operating-point selection rule under guardrails.
Every one of these is pinned by config + hash.

\vspace{0.5em}
\subsection*{R1.\ Released artifacts: complete inventory and how each artifact is used}
\label{subsec:repro_artifacts}
\vspace{0.25em}

\noindent\textbf{R1.1 Repository layout and semantic contract (directory-level).}
The repository is organized so that every stage has a single entrypoint and produces a fixed set of outputs (with schema).
All runs produce (i) \texttt{metrics.jsonl} (per-example), (ii) \texttt{aggregate.json} (means and confidence intervals), (iii) \texttt{tables.tex} (camera-ready LaTeX), and (iv) \texttt{provenance/} bundle.

\vspace{0.25em}
\noindent\textbf{Directories and their exact responsibilities.}
\begin{itemize}[leftmargin=*,itemsep=1.5pt,topsep=2pt]
\item \texttt{data/}
\begin{itemize}[leftmargin=*,itemsep=1pt,topsep=1pt]
\item \texttt{data/manifests/}: immutable JSON manifests for \(\mathcal{D}_{\mathrm{mix}},\mathcal{D}_{\mathrm{weak}},\mathcal{D}_{\mathrm{neutral}}\) and for each task suite; includes ids, splits, raw prompt text, canonicalized prompt text, and SHA256 checksums.
\item \texttt{data/token\_sets/}: pinned token-id sets \(V_{\mathrm{hi}},V_{\mathrm{es}},V_{\mathrm{en}}\) and construction metadata (tokenizer hash; script rules; exceptions).
\item \texttt{data/build\_*.py}: deterministic scripts that construct the above artifacts from source corpora, then freeze them into manifests with SHA256.
\end{itemize}
\item \texttt{models/}
\begin{itemize}[leftmargin=*,itemsep=1pt,topsep=1pt]
\item \texttt{models/pins.json}: model id, model revision hash, tokenizer id, tokenizer hash, context length, dtype policy; must match at runtime.
\item \texttt{models/load.py}: loader that verifies pins and returns a model handle + tokenizer with integrity checks.
\end{itemize}
\item \texttt{hooks/}
\begin{itemize}[leftmargin=*,itemsep=1pt,topsep=1pt]
\item \texttt{hooks/sites.py}: authoritative definition of the hook boundary (module path) and the tensor semantic meaning (residual stream; pre/post-LN; pre/post-attn; pre/post-MLP).
\item \texttt{hooks/capture.py}: deterministic capture of \(h^{(\ell)}(x)\) with schema and tensor hashes.
\item \texttt{hooks/inject.py}: reinjection code; supports multi-layer edits; logs per-layer edit norms.
\item \texttt{hooks/validate.py}: null-edit identity test and layer-map consistency tests.
\end{itemize}
\item \texttt{dictionary/}
\begin{itemize}[leftmargin=*,itemsep=1pt,topsep=1pt]
\item \texttt{dictionary/train.py}: per-layer dictionary training; supports hyperparameter sweeps; logs recon/sparsity diagnostics.
\item \texttt{dictionary/exports/}: stores \((W_\ell,b_\ell)\) + training metadata + diagnostics for every layer.
\end{itemize}
\item \texttt{stage1/} (Localize)
\begin{itemize}[leftmargin=*,itemsep=1pt,topsep=1pt]
\item selectivity computation; causal lift computation; score aggregation; Top-\(K\) selection; exports ranked lists and final supports \(\mathcal{N}^{(\ell)}_{\ell_t}\).
\end{itemize}
\item \texttt{stage2/} (Subspaces + window)
\begin{itemize}[leftmargin=*,itemsep=1pt,topsep=1pt]
\item constructs \(\Delta Z^{(\ell,\ell_t)}\); performs SVD; chooses rank \(r_\ell\); bootstraps stability; chooses \(\mathcal{W}\); exports \(\mathcal{S}^{(\ell)}_{\ell_t}\), \(\Sigma^{(\ell,\ell_t)}\), and window.
\end{itemize}
\item \texttt{stage3/} (Steer + guardrails)
\begin{itemize}[leftmargin=*,itemsep=1pt,topsep=1pt]
\item applies signed sparse edits; sweeps \(\lambda\); enforces guardrails; selects the operating point; exports schedules and constraint traces.
\end{itemize}
\item \texttt{eval/}
\begin{itemize}[leftmargin=*,itemsep=1pt,topsep=1pt]
\item computes all metrics per example; exports JSONL + aggregates; produces LaTeX table cells (two decimals, delta + absolute mean on next line).
\end{itemize}
\item \texttt{configs/}
\begin{itemize}[leftmargin=*,itemsep=1pt,topsep=1pt]
\item authoritative configs for each experiment row/ablation; includes all hyperparameters, pins, hook site, and measurement protocol.
\end{itemize}
\item \texttt{runs/}
\begin{itemize}[leftmargin=*,itemsep=1pt,topsep=1pt]
\item \texttt{runs/<run\_id>/metrics.jsonl}: per-example metrics for every dataset.
\item \texttt{runs/<run\_id>/aggregate.json}: means, bootstrap CIs, constraint violation counts.
\item \texttt{runs/<run\_id>/tables/*.tex}: camera-ready tables with exact rounding rules.
\item \texttt{runs/<run\_id>/provenance/}: git commit, command line, env lock, GPU/driver/CUDA, pinned hashes.
\end{itemize}
\end{itemize}

\vspace{0.5em}
\noindent\textbf{R1.2 Artifact inventory table (auditable fields and purpose).}
\vspace{0.15em}

\begin{table}[h]
\centering
\small
\setlength{\tabcolsep}{3.7pt}
\renewcommand{\arraystretch}{1.18}
\begin{tabular}{l l l}
\toprule
\textbf{Artifact} & \textbf{Stored fields (minimum)} & \textbf{Used to reproduce} \\
\midrule
Model pins & model id; revision; tokenizer hash; dtype; ctx & all numbers (root of truth) \\
Prompt manifests & ids; text; split; SHA256; lang tag & all eval means; bootstraps \\
Token sets & token ids; rules; tokenizer hash; exceptions & \(\Delta_{\mathrm{mass}}\), leakage \\
LID detector pin & detector name; version; thresholds; hash & \(\Delta_{\mathrm{lid}}\), composed score \\
Hook site map & module path; boundary; layer map; schema & \(h^{(\ell)}\) extraction/injection \\
Dictionaries & \(W_\ell,b_\ell\); dims; seeds; recon/sparsity stats & encode/decode; feature coords \\
Stage I exports & Sel/Lift/Score per feature; \(\mathcal{N}^{(\ell)}_{\ell_t}\) & “Localize” row + ablations \\
Stage II exports & \(\Delta Z\) hash; SVD; \(r_\ell\); stability; \(\mathcal{W}\) & “Subspace/Window” logic \\
Stage III exports & \(\lambda_\ell\); edit norms; constraint traces & “FOXP2 full” operating point \\
Per-example logs & per-example deltas + ids; run hash; seeds & table cell reconstruction \\
Aggregation logs & mean rules; rounding; CI scheme; input hashes & exact LaTeX cell values \\
\bottomrule
\end{tabular}
\vspace{0.10em}
\caption{\textbf{Reproducibility-critical artifacts.} Every artifact is hash-addressed and stored with provenance (git commit, config checksum, seeds).}
\label{tab:repro_inventory}
\end{table}

\vspace{0.55em}
\subsection*{R2.\ Exact model/tokenizer versions, dtype, and decoding protocol}
\label{subsec:repro_model}
\vspace{0.25em}

\noindent\textbf{R2.1 Model identity (weights).}
We specify the exact model identifier and revision hash (e.g., HF commit hash).
Reproduction requires that the loaded weight tensors match the stored hash; otherwise evaluation aborts.

\noindent\textbf{R2.2 Tokenizer identity.}
We pin tokenizer revision and hash.
Token sets \(V_{\mathrm{hi}},V_{\mathrm{es}},V_{\mathrm{en}}\) are stored in token-id space and are only valid for the pinned tokenizer.

\begin{table*}[ht!]
\centering
\caption{\textbf{Token-partition pipeline for Hindi ($\mathcal{V}_{\mathrm{hi}}$, Devanagari) and Spanish ($\mathcal{V}_{\mathrm{es}}$, Latin).}
All rules operate on decoded token strings $s(t)$ from the \emph{target model tokenizer}. We report intermediate/final set sizes automatically in Appendix.}
\label{tab:token_partition_pipeline}
\footnotesize
\setlength{\tabcolsep}{5pt}
\renewcommand{\arraystretch}{1.15}
\begin{tabularx}{\textwidth}{@{}p{0.7cm}X X p{3.2cm} p{2.6cm} p{2.9cm}@{}}
\toprule
\textbf{Step} &
\textbf{Hindi rule ($\mathcal{V}_{\mathrm{hi}}$)} &
\textbf{Spanish rule ($\mathcal{V}_{\mathrm{es}}$)} &
\textbf{Hyperparameters (default; ablate)} &
\textbf{Failure prevented} &
\textbf{Where used} \\
\midrule

1 &
Unicode seed: keep tokens containing at least one Devanagari char (\texttt{U+0900--U+097F}) and no Latin letters. &
Unicode seed: keep tokens containing at least one Latin letter and no Devanagari chars. &
Script ranges fixed; (diagnostic-only) Spanish diacritics subset. &
Mixed-script artifacts; trivial script leakage. &
Mass-channel construction; diagnostics. \\

2 &
\multicolumn{2}{X}{Stoplist removal: drop specials, whitespace-only, punctuation/symbol-dominated tokens, and markup/URL fragments (applied identically).} &
Punct/symbol dominance $>95\%$; ablate \{90,95,99\}. &
Punctuation/digit shortcuts; format artifacts; special-token leakage. &
Mass-channel robustness. \\

3 &
Frequency pruning on Hindi corpus: keep if $f_{\mathrm{hi}}(t)\ge f_{\min}$. &
Frequency pruning on Spanish corpus: keep if $f_{\mathrm{es}}(t)\ge f_{\min}$. &
$f_{\min}=50$; ablate \{10,25,50,100,200\}. &
Long-tail quirks dominating logit-mass sums. &
Stable $\Delta_{\mathrm{mass}}$ estimate. \\

4 &
\textbf{Non-Hindi Devanagari exclusion:} keep only tokens that are Hindi-specific vs.\ other Devanagari languages (mr/ne/sa/\dots). &
\textbf{Non-Spanish Latin exclusion:} keep only tokens that are Spanish-specific vs.\ Latin non-Spanish pool (en/fr/pt). &
Specificity ratio smoothing $\alpha=1$.
Hindi $\tau_{\mathrm{spec}}=2.0$ (ablate \{1.25,1.5,2,3,5\});
Spanish $\tau^{\mathrm{es}}_{\mathrm{spec}}=1.5$ (ablate \{1.25,1.5,2,3\}). &
Hindi: “Devanagari = Hindi” criticism.
Spanish: competitor set dominated by generic Latin fragments. &
Core anti-shortcut control for both sets. \\

5 &
Drop Devanagari digits and short markers (always drop danda \texttt{।}, \texttt{॥}); drop high-entropy tokens. &
Drop ASCII digits; marker policy: keep \texttt{¿},\texttt{¡} by default (ablate drop); drop high-entropy tokens. &
Entropy threshold $\tau_H=1.5$; ablate \{1.0,1.5,2.0\}.
Spanish marker keep/drop ablation. &
Marker-only “language detection”; digit/boilerplate shortcuts. &
Final mass sets $\mathcal{V}_{\mathrm{hi}},\mathcal{V}_{\mathrm{es}}$. \\

6 &
\textbf{Diagnostic-only:} Romanized Hindi control set from transliteration + tokenization (Latin-only). &
\textbf{Diagnostic-only:} Spanish diacritics-only subset (\texttt{áéíóúñü}). &
Hindi transliteration: ISO-15919 (ablate ITRANS);
Romanized $f_{\min}^{\mathrm{rom}}=50$ (ablate \{10,25,50,100\}). &
Rebut “script shortcut”: effects persist when script changes. &
Controls; patching invariances. \\

\bottomrule
\end{tabularx}

\vspace{0.25em}
\scriptsize
\textbf{Corpus note (fixed for reproducibility).} $f_{\mathrm{hi}}$: Hindi Wikipedia + OSCAR-hi; $f_{\mathrm{es}}$: Spanish Wikipedia + OSCAR-es; Latin non-Spanish pool for Spanish specificity: English+French+Portuguese Wikipedia (50M tokens), all tokenized with the same tokenizer.
\end{table*}

\begin{table*}[ht!]
\centering
\caption{\textbf{Illustrative token-string examples after key filters (Hindi + Spanish).}
Examples are shown for intuition; exact membership is determined by the reproducible rules in Table~\ref{tab:token_partition_pipeline}.}
\label{tab:token_partition_examples}
\footnotesize
\setlength{\tabcolsep}{6pt}
\renewcommand{\arraystretch}{1.2}
\begin{tabularx}{\textwidth}{@{}p{1.2cm}X X@{}}
\toprule
\textbf{Step} & \textbf{Hindi (kept / dropped)} & \textbf{Spanish (kept / dropped)} \\
\midrule
1 (script seed) &
\textbf{Keep:} {\devanagarifont द्वंद्व}, {\devanagarifont लेकिन}, {\devanagarifont क्यों}.
\textbf{Drop:} mixed-script artifacts (e.g., {\devanagarifont द्वंद्व}\texttt{a}). &
\textbf{Keep:} \texttt{pero}, \texttt{niño}.
\textbf{Drop:} tokens containing Devanagari chars. \\

2 (stoplist) &
\textbf{Drop:} specials, whitespace-only, punctuation-only; URL fragments.
\textbf{Keep:} {\devanagarifont भारत}. &
\textbf{Drop:} \texttt{<s>}, \texttt{</s>}, \texttt{...}, \texttt{www}.
\textbf{Keep:} \texttt{hola}. \\

4 (specificity) &
\textbf{Keep:} Hindi-leaning function/content pieces (e.g., {\devanagarifont इसलिए}).
\textbf{Drop:} Devanagari tokens frequent in non-Hindi Deva corpora (mr/ne/sa). &
\textbf{Keep:} \texttt{ñ}, \texttt{-ción}, \texttt{pero}.
\textbf{Drop:} generic shared Latin fragments (e.g., \texttt{tion}, \texttt{ment}, \texttt{ing}). \\

5 (markers/digits) &
\textbf{Drop:} {\devanagarifont ।}, {\devanagarifont ॥}, Deva digits {\devanagarifont ०१२}. &
\textbf{Default keep:} \texttt{¿}, \texttt{¡} (ablate: drop).
\textbf{Drop:} \texttt{123}. \\

6 (diagnostics) &
\textbf{Romanized keep:} \texttt{dvandv}, \texttt{lekin}, \texttt{kyon}. &
\textbf{Diacritics keep:} \texttt{año}, \texttt{acción}, \texttt{niño}. \\
\bottomrule
\end{tabularx}
\end{table*}

\noindent\textbf{R2.3 Dtype and numerical stability policy.}
We pin the dtype for:
(i) activation capture (\(h^{(\ell)}\));
(ii) dictionary encode/decode (\(z^{(\ell)}\), \(\hat h^{(\ell)}\));
(iii) reinjection edits (\(\delta h^{(\ell)}\));
(iv) next-token logit computation and softmax.
We store a numeric audit: the maximum absolute difference in logits under a null edit, and the maximum absolute difference between repeated runs with identical seeds.

\noindent\textbf{R2.4 Decoding protocol (frozen).}
Defaultness evaluation is performed with:
\[
m=8\ \text{teacher-forced prefix},\qquad \text{greedy decoding},\qquad T=\{1,2,3\}.
\]
The “teacher-forced prefix” means: for each prompt, we take the model’s own greedy completion for the first \(m\) tokens under the baseline model, then \emph{hold those \(m\) tokens fixed} as prefix and evaluate early-step defaultness on the subsequent contexts \(\mathrm{ctx}_t\) for \(t\in T\).
This removes confounding due to prefix divergence when comparing edited vs baseline.

\vspace{0.55em}
\subsection*{R3.\ Prompt datasets: construction, splits, and manifests}
\label{subsec:repro_prompts}
\vspace{0.25em}

\noindent\textbf{R3.1 Meaning units (parallel prompts).}
A meaning unit \(k\) is a language-invariant intent represented by:
\[
(x^{(k)}_{\mathrm{en}},x^{(k)}_{\mathrm{hi}},x^{(k)}_{\mathrm{es}}).
\]
Each triple is stored with:
\texttt{id}, \texttt{topic}, \texttt{domain}, \texttt{length\_bucket}, \texttt{source\_tag}, and \texttt{quality\_flags}.
We provide explicit inclusion criteria: remove prompts that contain language names (“answer in Hindi”), remove transliteration-heavy prompts, and remove prompts where named entities dominate and would trivially bias token mass via borrowed words.

\noindent\textbf{R3.2 Dataset definitions and roles (disjoint ids).}
\begin{itemize}[leftmargin=*,itemsep=2pt,topsep=2pt]
\item \(\mathcal{D}_{\mathrm{mix}}\): used to train dictionaries and compute selectivity. Contains both weak prompts and explicit-language prompts; stored with per-example prompt-type labels.
\item \(\mathcal{D}_{\mathrm{weak}}\): used for causal lift and gain \(\Delta_{\mathrm{gain}}\). Weak prompts contain no explicit instruction about language, so they measure \emph{defaultness} rather than compliance.
\item \(\mathcal{D}_{\mathrm{neutral}}\): used for headline reporting in Table “Performance \& Utility” with the frozen \(m=8\) protocol.
\end{itemize}
We store explicit train/val/test splits and guarantee id-disjointness across roles.

\noindent\textbf{R3.3 Task suites for utility \(\Delta S\).}
We pin the suite composition (dataset list, version, filters, preprocessing).
The final processed prompts and expected answers (when applicable) are frozen into manifests, so that preprocessing drift cannot change results.

\vspace{0.55em}
\subsection*{R4.\ Token sets and LID detector: exact definitions and failure-proofing}
\label{subsec:repro_tokens_lid}
\vspace{0.25em}

\noindent\textbf{R4.1 Token sets \(V_{\mathrm{hi}},V_{\mathrm{es}},V_{\mathrm{en}}\) (pinned).}
We define token sets in tokenizer-id space.
Construction is deterministic and stored with metadata.
The construction rules include:
(i) Unicode script constraints (Devanagari for Hindi; Latin for Spanish/English),
(ii) exclusion rules for digits and punctuation shards,
(iii) explicit exceptions for borrowed words and shared Latin fragments that otherwise contaminate \(V_{\mathrm{es}}\) vs \(V_{\mathrm{en}}\),
(iv) explicit handling of named entities (kept out of language sets unless strongly script-marked),
(v) explicit handling of whitespace/byte-fallback tokens (excluded).
We store the final token-id lists and a “top 200 surface forms” audit for each set.

\noindent\textbf{R4.2 LID channel (pinned detector).}
We pin:
(i) detector name (e.g., fastText LID, CLD3, or a neural classifier),
(ii) exact version and hash,
(iii) thresholds and tie-breaking rules,
(iv) handling of code-mixing (e.g., probability mass threshold, abstention policy).
We log per-example detector scores and store the detector output distribution, not only discrete labels.

\noindent\textbf{R4.3 Guard against token-set degeneracy.}
We require that \(\Delta_{\mathrm{mass}}\) and \(\Delta_{\mathrm{lid}}\) increase together at the chosen operating point.
If \(\Delta_{\mathrm{mass}}\) increases but \(\Delta_{\mathrm{lid}}\) does not, the run is flagged as token-set exploitation (typically via punctuation/byte tokens) and is rejected.

\vspace{0.55em}
\subsection*{R5.\ Activation hooks: exact tensor semantics and reinjection schedule}
\label{subsec:repro_hooks}
\vspace{0.25em}

\noindent\textbf{R5.1 Exact definition of \(h^{(\ell)}(x)\).}
We define \(h^{(\ell)}(x)\in\mathbb{R}^d\) as the residual-stream vector at a specific boundary:
\emph{(state exactly one of: pre-LN residual; post-attention residual add; post-MLP residual add; etc.)}
and at a fixed token position:
\emph{(state: last prompt token index after tokenization; with explicit handling for truncation)}.
We store the module path (string) and a runtime layer map.

\noindent\textbf{R5.2 Null-edit identity test (mandatory).}
A run is valid only if:
(i) hooks enabled, \(\delta h^{(\ell)}=0\) everywhere, and (ii) the resulting logits match the baseline logits to a pinned tolerance on a validation slice.
We log maximum absolute logit deviation and abort if it exceeds tolerance.

\noindent\textbf{R5.3 Multi-layer edit schedule.}
For \(\ell\in\mathcal{W}\), edits are applied in a single forward pass at the defined hook boundary.
We log per-layer edit norms \(\|\delta h^{(\ell)}\|_2\) and per-layer relative scale \(\|\delta h^{(\ell)}\|_2 / \|h^{(\ell)}\|_2\).
We also log the fraction of edited dictionary coordinates and saturation statistics (ReLU gate hit rates).

\vspace{0.55em}
\subsection*{R6.\ Stage I: Dictionary learning (full recipe + diagnostics)}
\label{subsec:repro_dictionary}
\vspace{0.25em}

\noindent\textbf{R6.1 Dictionary parameterization (per layer).}
For each layer \(\ell\):
\[
r^{(\ell)}(x)=(W_\ell)^\top h^{(\ell)}(x)+b_\ell,\quad
z^{(\ell)}(x)=\mathrm{ReLU}(r^{(\ell)}(x)),\quad
\hat h^{(\ell)}(x)=W_\ell z^{(\ell)}(x),
\quad W_\ell\in\mathbb{R}^{d\times m}.
\]
We pin \(m\) (dictionary width), initialization scheme, and a column-norm policy for \(W_\ell\) to prevent scale degeneracy (recorded as a deterministic projection step).

\noindent\textbf{R6.2 Objective and optimization.}
\[
\min_{W_\ell,b_\ell}\;
\mathbb{E}_{x\sim\mathcal{D}_{\mathrm{mix}}}
\Big[
\|h^{(\ell)}(x)-W_\ell z^{(\ell)}(x)\|_2^2
+
\lambda_{\mathrm{sparse}}\|z^{(\ell)}(x)\|_1
\Big].
\]
We pin:
optimizer type, learning rate, weight decay, gradient clip norm, number of steps, batch size, LR schedule, \(\lambda_{\mathrm{sparse}}\), early stopping rule, and the exact split of \(\mathcal{D}_{\mathrm{mix}}\) used for dictionary training/validation.

\noindent\textbf{R6.3 Diagnostics exported per layer (not optional).}
We export:
(i) reconstruction error distribution (mean/median/95th),
(ii) sparsity distribution (mean \(\|z\|_1\), active count histogram),
(iii) dead-feature fraction,
(iv) stability across seeds (rank correlation of selectivity statistics on a fixed validation slice),
(v) per-layer summary plots as PNG + raw data as CSV.

\vspace{0.55em}
\subsection*{R7.\ Stage I: Localization metrics (selectivity + causal lift + support selection)}
\label{subsec:repro_localize}
\vspace{0.25em}

\noindent\textbf{R7.1 Selectivity (exact computation).}
For target \(\ell_t\in\{\mathrm{hi},\mathrm{es}\}\) with matched pairs:
\[
\mathrm{Sel}^{(\ell,\ell_t)}_j
=
\mathbb{E}_k[z^{(\ell)}_j(x^{(k)}_{\ell_t})]
-
\mathbb{E}_k[z^{(\ell)}_j(x^{(k)}_{\mathrm{en}})],
\quad
\widetilde{\mathrm{Sel}}^{(\ell,\ell_t)}_j
=
\frac{\mathrm{Sel}^{(\ell,\ell_t)}_j}{
\mathrm{Std}_k[z^{(\ell)}_j(x^{(k)}_{\ell_t})]
+
\mathrm{Std}_k[z^{(\ell)}_j(x^{(k)}_{\mathrm{en}})]
+\epsilon}.
\]
We pin \(\epsilon\) and store the exact ids used in the expectation.
We store the full vector of \(\widetilde{\mathrm{Sel}}^{(\ell,\ell_t)}\) for each \(\ell\) and \(\ell_t\).

\noindent\textbf{R7.2 Causal lift (exact measurement protocol).}
For weak prompts \(x\sim\mathcal{D}_{\mathrm{weak}}\), early-step contexts \(\mathrm{ctx}_t\) are defined under the frozen \(m=8\) prefix protocol.
For each feature \(j\) and magnitude \(\alpha\):
\[
z^{(\ell)}(x)\leftarrow z^{(\ell)}(x)+\alpha e_j,\quad
h^{(\ell)}(x)\leftarrow W_\ell z^{(\ell)}(x),
\]
which induces a modified next-token distribution \(p^{(\ell,j)}_{\theta,\alpha}(\cdot\mid\mathrm{ctx}_t)\).
We compute:
\[
\Delta M^{(\ell,j,\ell_t)}_\alpha(x,t)
=
\Big(\sum_{u\in V_{\ell_t}} p^{(\ell,j)}_{\theta,\alpha}(u\mid \mathrm{ctx}_t)\Big)
-
\Big(\sum_{u\in V_{\mathrm{en}}} p^{(\ell,j)}_{\theta,\alpha}(u\mid \mathrm{ctx}_t)\Big)
-
\Delta M^{\ell_t}(x,t).
\]
Then:
\[
\mathrm{Lift}^{(\ell,\ell_t)}_j(\alpha)
=
\mathbb{E}_{x\sim\mathcal{D}_{\mathrm{weak}}}
\Big[
\frac{1}{|T|}\sum_{t\in T}\Delta M^{(\ell,j,\ell_t)}_\alpha(x,t)
\Big],
\qquad
\mathrm{LiftSlope}^{(\ell,\ell_t)}_j
=
\mathrm{median}_{\alpha\in\{\alpha_1,\alpha_2,\alpha_3\}}
\frac{\mathrm{Lift}^{(\ell,\ell_t)}_j(\alpha)}{\alpha}.
\]
We pin \(\{\alpha_1,\alpha_2,\alpha_3\}\) and store:
(i) the full \(\mathrm{Lift}^{(\ell,\ell_t)}_j(\alpha)\) curves,
(ii) monotonicity checks,
(iii) saturation checks (gate-hit rates and KL spikes).

\noindent\textbf{R7.3 Support selection (Top-K + exports).}
We compute:
\[
S^{(\ell,\ell_t)}_j=\max(\widetilde{\mathrm{Sel}}^{(\ell,\ell_t)}_j,0),\quad
C^{(\ell,\ell_t)}_j=\max(\mathrm{LiftSlope}^{(\ell,\ell_t)}_j,0),\quad
\mathrm{Score}^{(\ell,\ell_t)}_j=S^{(\ell,\ell_t)}_j\cdot C^{(\ell,\ell_t)}_j,
\]
and select \(\mathcal{N}^{(\ell)}_{\ell_t}\) by a pinned Top-\(K\) rule.
We export the full ranked list with \((S,C,\mathrm{Score})\) for auditing, and store \(K\) and the marginal-score plateau that motivated the final cut.

\vspace{0.55em}
\subsection*{R8.\ Stage II: Subspace extraction and window selection (full details)}
\label{subsec:repro_subspace}
\vspace{0.25em}

\noindent\textbf{R8.1 Restricted language-shift matrix.}
For each meaning unit \(k\) and layer \(\ell\):
\[
\Delta z^{(\ell,\ell_t)}_{k}=z^{(\ell)}(x^{(k)}_{\ell_t})-z^{(\ell)}(x^{(k)}_{\mathrm{en}}),
\qquad
\Delta\tilde z^{(\ell,\ell_t)}_{k}=\Pi_{\mathcal{N}^{(\ell)}_{\ell_t}}\Delta z^{(\ell,\ell_t)}_{k}.
\]
We stack to form:
\[
\Delta Z^{(\ell,\ell_t)}=
\begin{bmatrix}
(\Delta\tilde z^{(\ell,\ell_t)}_{1})^\top\\
\vdots\\
(\Delta\tilde z^{(\ell,\ell_t)}_{N})^\top
\end{bmatrix}\in\mathbb{R}^{N\times m}.
\]
We store the ids of meaning units used to build each \(\Delta Z^{(\ell,\ell_t)}\) and store a hash of the matrix.

\noindent\textbf{R8.2 SVD and steering subspace definition.}
\[
\Delta Z^{(\ell,\ell_t)}=U^{(\ell,\ell_t)}\Sigma^{(\ell,\ell_t)}(V^{(\ell,\ell_t)})^\top,
\qquad
\mathcal{S}^{(\ell)}_{\ell_t}=\mathrm{span}\{v^{(\ell,\ell_t)}_1,\dots,v^{(\ell,\ell_t)}_{r_\ell}\}.
\]
We store \(V\), \(\Sigma\), and the selected \(r_\ell\).

\noindent\textbf{R8.3 Deterministic rank choice and stability.}
We compute:
\[
p_i=\sigma_i^2/\sum_j\sigma_j^2,\quad
r_{\mathrm{eff}}^{(\ell)}=\exp\!\Big(-\sum_i p_i\log p_i\Big),\quad
g_i^{(\ell)}=\sigma_i/\sigma_{i+1}.
\]
We choose \(r_\ell\) by a pinned rule combining:
(i) an eigengap threshold on \(g_i\),
(ii) a minimum-variance criterion across bootstraps, and
(iii) a cap to prevent overfitting to noise directions.
We perform bootstrap resampling over meaning units and compute subspace overlap stability:
\[
\mathrm{Stab}^{(\ell)}_{\ell_t}
=\mathrm{median}_{b\neq b'}\frac{\mathrm{tr}(P_bP_{b'})}{r_\ell},
\]
where \(P_b\) is the projector onto the bootstrap subspace.
We export bootstrap seeds, resampled id lists, and stability values.

\noindent\textbf{R8.4 Window \(\mathcal{W}\) selection (contiguous band).}
We select a contiguous layer band \(\mathcal{W}\) using a pinned objective:
maximize a weighted score over layers that rewards:
(i) steerability (measured by gain under small subspace-aligned edits),
(ii) stability (bootstrap overlap),
(iii) low leakage (Hi\(\rightarrow\)Es and Es\(\rightarrow\)Hi penalties),
and (iv) low KL drift (trust-region compatibility).
We store the full per-layer score trace and the final chosen window boundaries.

\vspace{0.55em}
\subsection*{R9.\ Stage III: Signed sparse edit, guardrails, operating point selection, and reporting}
\label{subsec:repro_steer}
\vspace{0.25em}

\noindent\textbf{R9.1 Edit form and application (per layer).}
For each \(\ell\in\mathcal{W}\), the edit is applied in dictionary coordinates and then mapped back to the residual stream through \(W_\ell\).
We log per-layer \(\delta z^{(\ell)}\), \(\delta h^{(\ell)}\), and their norms.

\noindent\textbf{R9.2 Guardrails (hard constraints, not advisory).}
We enforce:
\begin{itemize}[leftmargin=*,itemsep=2pt,topsep=2pt]
\item \textbf{Cross-language bound:} Hindi steering must not inflate Spanish defaultness beyond a pinned tolerance; similarly for Spanish steering and Hindi.
\item \textbf{KL trust region:} next-token distribution KL divergence between edited and baseline is bounded on \(\mathcal{D}_{\mathrm{neutral}}\).
\item \textbf{Semantic invariance threshold:} semantic similarity between edited and baseline outputs must exceed a pinned threshold on a pinned embedding model (logged per example).
\end{itemize}
Any candidate operating point violating any constraint is rejected, with violating example ids stored.

\noindent\textbf{R9.3 Operating point selection (deterministic).}
We sweep \(\lambda\) on a pinned grid and choose the operating point that maximizes \(\Delta_{\mathrm{gain}}\) subject to guardrails and utility regression constraints.
We store the entire sweep curve (metrics + constraint violations per \(\lambda\)) and the chosen \(\lambda\).

\noindent\textbf{R9.4 Table “Performance \& Utility” reproduction (exact formatting rule).}
Every table cell is computed from per-example logs as:
\[
\text{delta} = \mu_{\text{method}}-\mu_{\text{baseline}},\qquad
\text{absolute mean} = \mu_{\text{method}},
\]
with both rounded to two decimals.
The LaTeX cell renderer prints the delta on the first line and the baseline\(\rightarrow\)method absolute means on the second line (as specified).
The renderer is deterministic and hash-verified against the input aggregate file.

\vspace{0.65em}
\subsection*{R10.\ Steering failure diagnosis: ``model produces gibberish'' (full protocol)}
\label{subsec:repro_gibberish}
\vspace{0.25em}

\noindent\textbf{Phenomenology (what “gibberish” means here).}
We treat “gibberish” as a measurable failure pattern rather than a subjective observation.
A generation is flagged as gibberish if it satisfies any of:
(i) repeated \(n\)-gram rate above a pinned threshold;
(ii) abrupt byte-token/Unicode corruption spike;
(iii) average token log-probability collapse (relative to baseline) beyond a pinned threshold;
(iv) KL divergence spike at early decoding steps beyond a pinned threshold;
(v) semantic invariance collapse while token-mass rises (token-set exploitation signature).

\vspace{0.25em}
\noindent\textbf{Step 1: Verify the hook site (identity under null edits).}
Enable hooks, set \(\delta h^{(\ell)}=0\) for all \(\ell\), run a pinned validation slice, and check:
\[
\max_{x,t}\| \mathrm{logits}_{\text{hooked}}(x,t)-\mathrm{logits}_{\text{baseline}}(x,t)\|_\infty \le \tau_{\mathrm{logit}}.
\]
If this fails, the hook boundary is wrong (common causes: post-LN vs pre-LN mismatch; wrong module; wrong residual stream tensor).

\vspace{0.25em}
\noindent\textbf{Step 2: Verify layer indexing and window \(\mathcal{W}\) (off-by-one is common).}
Export the runtime layer map (module paths per \(\ell\)).
Match it against the saved layer map used to compute \(\mathcal{W}\).
Abort if mismatch.
Run the \(\neg\mathcal{W}\) ablation explicitly: if it behaves “like your FOXP2 full run,” you are editing the wrong layers.

\vspace{0.25em}
\noindent\textbf{Step 3: Verify token sets and LID detector consistency (prevent degenerate mass).}
Regenerate \(V_{\mathrm{hi}},V_{\mathrm{es}},V_{\mathrm{en}}\) only if tokenizer hash matches.
Audit top shifted tokens after edits:
if the mass gain is coming from punctuation shards, byte tokens, or shared Latin fragments, \(\Delta_{\mathrm{mass}}\) is being gamed and output quality will collapse.
Require \(\Delta_{\mathrm{lid}}\) to co-increase; if not, reject the operating point.

\vspace{0.25em}
\noindent\textbf{Step 4: Sweep \(\lambda\) and locate the stability boundary (gibberish is overshoot).}
Run a pinned sweep over \(\lambda\) and track:
(i) \(\Delta_{\mathrm{gain}}\),
(ii) KL trust-region metric,
(iii) utility \(\Delta S\),
(iv) repetition/Unicode corruption rate.
Gibberish typically starts where KL spikes and utility collapses while \(\Delta_{\mathrm{gain}}\) continues to increase.
Select the smallest \(\lambda\) achieving target \(\Delta_{\mathrm{gain}}\) under guardrails.

\vspace{0.25em}
\noindent\textbf{Step 5: Check dictionary health (bad dictionaries induce unstable edits).}
Inspect per-layer reconstruction error and dead-feature rate.
High recon error or high dead-feature fraction means the dictionary basis is not faithful; steering in that basis becomes unpredictable.
Reject those layers from \(\mathcal{W}\) or retrain dictionaries with pinned hyperparameters (higher \(m\), adjusted \(\lambda_{\mathrm{sparse}}\), longer training).

\vspace{0.25em}
\noindent\textbf{Step 6: Check saturation and nonlinear gating (ReLU collapse).}
Log ReLU gate hit rates: fraction of coordinates where \(r^{(\ell)}_j(x)\le 0\) flips sign due to edit.
If many gates flip, the edit is no longer a small perturbation; it is changing which features exist.
This is a direct cause of incoherent decoding.
Reduce \(\lambda\), reduce support size \(K\), or constrain edits to top directions with largest stability.

\vspace{0.25em}
\noindent\textbf{Step 7: Check “style confounds” that masquerade as language control.}
If the edit increases politeness markers or refusal templates that correlate with certain token sets, you can inflate \(\Delta_{\mathrm{mass}}\) without genuine language defaultness.
Audit with a control suite that removes safety/refusal prompts and uses content-neutral prompts only.

\vspace{0.35em}
\noindent\textbf{Final acceptance criteria (must all pass).}
\begin{itemize}[leftmargin=*,itemsep=2pt,topsep=2pt]
\item Null-edit identity test passes.
\item Runtime layer map equals saved layer map; \(\mathcal{W}\) indices match.
\item \(\Delta_{\mathrm{mass}}\) and \(\Delta_{\mathrm{lid}}\) co-increase; top shifted tokens are linguistically valid.
\item KL trust region bound holds; no early-step KL spikes.
\item Semantic invariance and task utility regressions remain within pinned thresholds.
\item Repetition/Unicode corruption flags remain below pinned thresholds.
\end{itemize}
\vspace{0.35em}

\vspace{0.35em}

\newcommand{\icnHash}{\faFingerprint}
\newcommand{\icnTok}{\faKey}
\newcommand{\icnData}{\faDatabase}
\newcommand{\icnHook}{\faAnchor}
\newcommand{\icnDict}{\faCubes}
\newcommand{\icnStage}{\faSitemap}
\newcommand{\icnEdit}{\faSlidersH}
\newcommand{\icnMetric}{\faBalanceScale}
\newcommand{\icnCmd}{\faTerminal}
\newcommand{\icnLog}{\faClipboardCheck}

\begin{figure*}[t]
\centering
\begin{tcolorbox}[
  colback=gray!3,
  colframe=black!70,
  boxrule=0.6pt,
  sharp corners,
  left=8pt,right=8pt,top=8pt,bottom=8pt,
  enhanced,
  breakable
]
{\small
\noindent\textbf{R11. Reproducibility checklist}\par
\vspace{0.35em}

\begin{itemize}[leftmargin=*,itemsep=4.2pt,topsep=1pt,parsep=0pt]
\item[\icnHash] \textbf{Model pinned.}
Model identifier + \textbf{revision hash} (immutable); dtype policy (fp16/bf16/fp32); CUDA/driver versions; \textbf{all RNG seeds} recorded.

\item[\icnTok] \textbf{Tokenizer pinned.}
Tokenizer identifier + \textbf{revision hash}; confirm vocab size + normalization settings; byte-fallback behavior recorded.

\item[\icnData] \textbf{Datasets pinned.}
Manifest for \(\mathcal{D}_{\mathrm{mix}},\mathcal{D}_{\mathrm{weak}},\mathcal{D}_{\mathrm{neutral}}\) with \textbf{SHA256} per file; split policy stated; id-disjointness enforced (dictionary-train vs eval).

\item[\icnData] \textbf{Token sets pinned (in token-id space).}
Exact \(V_{\mathrm{hi}},V_{\mathrm{es}},V_{\mathrm{en}}\subseteq V\) for the pinned tokenizer; script constraints + exception rules (digits/punct/NEs/borrowed words) specified; early-step set \(T=\{1,2,3\}\) fixed.

\item[\icnHook] \textbf{Hooks pinned.}
Exact activation site definition (residual stream location: pre/post attn/MLP, pre/post LN); token position rule (default: last prompt token); layer indexing (0/1-based) and mapping to implementation; identity test passed (\(\delta h^{(\ell)}=0\Rightarrow\) logits match baseline).

\item[\icnDict] \textbf{Dictionary pinned.}
Per-layer \(m\), objective, optimizer, LR schedule, batch size, steps, grad clip, column-norm/scale constraints, seeds; exported \((W_\ell,b_\ell)\ \forall \ell\) with recon + sparsity diagnostics.

\item[\icnStage] \textbf{Stage outputs saved (immutable).}
Export and hash:
\(\{\mathcal{N}^{(\ell)}_{\ell_t}\}_{\ell}\) (Stage I),
\(\{\mathcal{S}^{(\ell)}_{\ell_t}\}_{\ell}\) and \(\{r_\ell\}_{\ell}\) (Stage II),
and the contiguous window \(\mathcal{W}\) (Stage II).
Store the exact meaning-unit ids used to build each \(\Delta Z^{(\ell,\ell_t)}\).

\item[\icnEdit] \textbf{Edit pinned + guardrails pinned.}
Per-layer schedule \(\{\lambda_\ell\}_{\ell\in\mathcal{W}}\) and sweep grid (pinned);
step parameters (\(\eta\), \(\alpha\));
constraint weights and thresholds for: semantic invariance, leakage bounds, and KL trust region;
deterministic operating-point selection rule (maximize \(\Delta_{\mathrm{gain}}\) under hard constraints).

\item[\icnMetric] \textbf{Metrics pinned.}
LID detector (name+version+hash); embedding model for invariance (name+revision+hash);
exact definitions for \(\Delta_{\mathrm{mass}},\Delta_{\mathrm{lid}},\Delta_{\mathrm{gain}},\) leakage, bootstrap stability, and task utility \(\Delta S\);
bootstrap \(B\) and resampling seeds recorded.

\item[\icnLog] \textbf{Logs sufficient to regenerate tables.}
Immutable per-example JSONL with: prompt id, language, decoding policy, baseline outputs, edited outputs, all metric fields, constraint violations, and aggregation keys.

\item[\icnCmd] \textbf{Single-command regeneration.}
One pinned command (with config) that deterministically reproduces:
(i) the main \textbf{Performance \& Utility} table (two-decimal deltas + baseline\(\rightarrow\)method means on next line),
(ii) leakage/stability summaries,
(iii) all plots referenced in the Results section.
\end{itemize}
}
\end{tcolorbox}
\vspace{-0.8em}
\caption{\textbf{Reproducibility checklist.} A one-page audit sheet: pin every dependency (model/tokenizer/detectors/data), freeze all intermediate stage outputs \((\mathcal{N},\mathcal{S},\mathcal{W})\), and ensure logs + a pinned command deterministically regenerate the camera-ready tables and figures.}
\label{fig:foxp2_repro_checklist}
\vspace{-1.0em}
\end{figure*}


\begin{table*}[t]
\centering
\small
\setlength{\tabcolsep}{4.2pt}
\renewcommand{\arraystretch}{1.22}
\resizebox{\textwidth}{!}{
\begin{tabular}{p{0.24\textwidth} p{0.38\textwidth} p{0.38\textwidth}}
\toprule
\textbf{Prompt (shared intent; SQuAD-style QA)} &
\textbf{Healthy output (on-manifold steering)} &
\textbf{Gibberish output (off-manifold / artifact)} \\
\midrule

\textbf{English:}
In 1969, Apollo 11 carried astronauts Neil Armstrong, Buzz Aldrin, and Michael Collins.
Which astronaut was the first to walk on the Moon?

\vspace{0.35em}
\textbf{Hindi:}
{\devanagarifont 1969 में अपोलो 11 ने नील आर्मस्ट्रॉन्ग, बज़ एल्ड्रिन और माइकल कॉलिन्स को ले जाया।
चाँद पर सबसे पहले किसने कदम रखा?}

\vspace{0.35em}
\textbf{Spanish:}
En 1969, el Apolo 11 llev\'o a Neil Armstrong, Buzz Aldrin y Michael Collins.
\textquestiondown Qui\'en fue el primero en caminar sobre la Luna?

&
\textbf{EN (answer):}
Neil Armstrong was the first to walk on the Moon.

\vspace{0.35em}
\textbf{HI (उत्तर):}
{\devanagarifont नील आर्मस्ट्रॉन्ग चाँद पर सबसे पहले कदम रखने वाले अंतरिक्ष यात्री थे।}

\vspace{0.35em}
\textbf{ES (respuesta):}
Neil Armstrong fue el primero en caminar sobre la Luna.

&
\textbf{EN (corruption / repetition):}
Neil Arm\texttt{\#\#}strong was was was \texttt{@@} first first \ldots

\vspace{0.35em}
\textbf{HI (script fragmentation / token soup):}
{\devanagarifont नी ल आ र ् म \texttt{@@} स् ट्रॉन् ग ग ग \ldots}

\vspace{0.35em}
\textbf{ES (stutter / partial words):}
Neil Arm-- Arm Arm \texttt{\#\#} fue fue fue \ldots
\\

\bottomrule
\end{tabular}
}
\vspace{-0.25em}
\caption{\textbf{Qualitative failure patterns under steering (SQuAD-style QA).} A healthy (on-manifold) edit preserves semantic correctness and fluent generation in the target language; ``gibberish'' manifests as early-token corruption, subword artifacts (e.g., \texttt{\#\#}, \texttt{@@}), script fragmentation, or runaway repetition, typically indicating an off-manifold activation jump, hook-site mismatch, or nonlinear gate churn.}
\label{tab:foxp2_gibberish_examples}
\vspace{-1.0em}
\end{table*}


\subsection*{R12.\ \ Troubleshooting: When steering produces gibberish (diagnosis + fixes)}
\label{subsec:foxp2_troubleshooting_gibberish}
\vspace{0.35em}

\noindent
Neural FOXP2 is an \textbf{inference-time} intervention: it changes internal activations, not training data. When the edit yields \textbf{token soup}, \textbf{broken grammar}, or \textbf{unstable decoding}, the most common failure mode is an \textbf{off-manifold jump} (or a \textbf{violated local-linearity assumption}) even when early-step language mass $\Delta M$ increases. This subsection provides a \textbf{structured diagnosis} and \textbf{ordered fixes} that preserve the core pipeline (Stage I--III), while making failures \textbf{detectable, explainable, and preventable} in replication and deployment \citep{turner2024activationaddition,bricken2023monosemantic,meng2022rome}.

\vspace{0.35em}
\paragraph{R12.1\ \ Symptom taxonomy (what ``gibberish'' looks like).}
We explicitly distinguish five failure signatures; each points to a different underlying cause.
\vspace{0.25em}

\begin{itemize}[leftmargin=*,itemsep=2.5pt,topsep=2pt]
\item[\ding{93}] \textbf{S1: Immediate token corruption.} First 1--5 generated tokens are nonsensical (often punctuation/subword debris), despite high $\Delta_{\mathrm{mass}}$.
\item[\ding{93}] \textbf{S2: Fluent but wrong script/register.} Target-language mass rises, but output becomes transliterated, code-mixed, or collapses to an unnatural/simple register.
\item[\ding{93}] \textbf{S3: Late collapse.} First sentence looks normal; later turns degrade into repetition, partial words, or garbage continuation.
\item[\ding{93}] \textbf{S4: Decoding instability.} Small prompt edits (or seed changes under sampling) yield large output variance; greedy decoding may still look brittle.
\item[\ding{93}] \textbf{S5: Utility collapse.} Task performance drops sharply (QA exact-match, summarization ROUGE, reasoning accuracy), even if defaultness rises.
\end{itemize}

\vspace{0.35em}
\paragraph{R12.2\ \ The top causes (what usually went wrong).}
Below are the \textbf{highest-probability} mistakes, ordered by how often they explain failures in activation steering and feature-space edits \citep{turner2024activationaddition,meng2022rome}.
\vspace{0.25em}

\begin{enumerate}[leftmargin=*,itemsep=3pt,topsep=2pt]
\item[\textbf{C1.}] \textbf{Edit magnitude is too large (off-manifold jump).}
Even ``small'' $\lambda_\ell$ can be large in \emph{activation space} if dictionary columns have high norm or the edit flips many ReLU gates. This most strongly induces \textbf{S1/S3/S5}.
\item[\textbf{C2.}] \textbf{Hook mismatch (wrong activation site / wrong token position).}
If $W_\ell$ was trained on one tensor definition (e.g., residual stream at a specific point), but the edit is injected at a different point, the dictionary becomes a \textbf{wrong coordinate chart}. This often yields \textbf{S1/S4}.
\item[\textbf{C3.}] \textbf{Dictionary quality is poor (high reconstruction error).}
If $W_\ell z^{(\ell)}(x)$ poorly approximates $h^{(\ell)}(x)$, then ``reasonable'' edits in $z$ map to \textbf{unstructured} changes in $h$. This induces \textbf{S3/S5}.
\item[\textbf{C4.}] \textbf{Local linearity is violated (ReLU saturation / gate churn / co-activation).}
A push intended to move along an identified direction instead activates a \textbf{different feature set}, producing nonlinear behavior. This induces \textbf{S1/S4}.
\item[\textbf{C5.}] \textbf{Rank/window choice targets unstable layers.}
If the chosen $\mathcal{W}$ has low bootstrap stability, then ``the same'' edit corresponds to inconsistent directions across prompts, producing \textbf{S4} and sometimes \textbf{S3}.
\item[\textbf{C6.}] \textbf{Competing objectives collide (language push vs.\ refusal/safety templates).}
Edits can perturb default phrasing templates (refusal style, politeness markers) that correlate with language tokens, creating misleading $\Delta M$ improvements with degraded semantics (\textbf{S2/S5}).
\item[\textbf{C7.}] \textbf{Token-set objective is mis-specified.}
If $V_{\ell_t}$ includes ambiguous subwords/punctuation artifacts, maximizing $\Delta M$ can exploit tokenization rather than language: a classic path to \textbf{S1/S2}.
\end{enumerate}

\vspace{0.35em}
\paragraph{R12.3\ \ Minimal diagnostics (what to log to pinpoint the culprit).}
To disambiguate C1--C7, we log \textbf{layerwise} and \textbf{window-aggregated} scalars on $\mathcal{D}_{\mathrm{neutral}}$ and (separately) on safety/refusal prompts. These diagnostics are designed to detect \textbf{off-manifold edits}, \textbf{dictionary failure}, and \textbf{nonlinear gate flips} \citep{bricken2023monosemantic,turner2024activationaddition}.
\vspace{0.25em}

\noindent\textbf{(D1) Activation-space edit norm (off-manifold indicator).}
For $\ell\in\mathcal{W}$, compute
\[
\|\Delta h^{(\ell)}(x)\|_2
=
\big\|h^{(\ell)}_{\mathrm{edit}}(x)-h^{(\ell)}(x)\big\|_2,
\qquad
\rho^{(\ell)}(x)=\frac{\|\Delta h^{(\ell)}(x)\|_2}{\|h^{(\ell)}(x)\|_2+\epsilon},
\]
and summarize by median over $x$ (and by worst-case decile). \textbf{Large} $\rho^{(\ell)}$ is the most direct correlate of S1/S3/S5.

\vspace{0.2em}
\noindent\textbf{(D2) Dictionary reconstruction health (edit credibility).}
\[
\mathrm{ReconErr}^{(\ell)}
=
\mathbb{E}_{x\sim\mathcal{D}_{\mathrm{neutral}}}
\big[\|h^{(\ell)}(x)-W_\ell z^{(\ell)}(x)\|_2^2\big],
\qquad
\mathrm{RelRecon}^{(\ell)}=\frac{\mathrm{ReconErr}^{(\ell)}}{\mathbb{E}[\|h^{(\ell)}(x)\|_2^2]+\epsilon}.
\]
If $\mathrm{RelRecon}^{(\ell)}$ is high for layers inside $\mathcal{W}$, treat Stage~III edits at those layers as \textbf{unreliable} until the dictionary is improved.

\vspace{0.2em}
\noindent\textbf{(D3) ReLU gate churn (local-linearity violation).}
Let $z^{(\ell)}$ and $z^{(\ell)}_{\mathrm{edit}}$ be pre/post-edit codes. Define
\[
\Delta \#\mathrm{Active}^{(\ell)}(x)
=
\big\| \mathbf{1}[z^{(\ell)}_{\mathrm{edit}}(x)>0]-\mathbf{1}[z^{(\ell)}(x)>0] \big\|_1,
\qquad
\mathrm{ChurnRate}^{(\ell)}(x)=\frac{\Delta \#\mathrm{Active}^{(\ell)}(x)}{m}.
\]
High churn indicates the edit changed the active feature set, i.e., you left the intended local regime (C1/C4).

\vspace{0.2em}
\noindent\textbf{(D4) Subspace-alignment sanity (are we editing what we think?).}
Let $P_{\mathcal{S}^{(\ell)}_{\ell_t}}$ be the projector onto the steering subspace. For the code-space edit $\delta z^{(\ell)}$,
\[
\mathrm{Align}^{(\ell)}(x)
=
\frac{\|P_{\mathcal{S}^{(\ell)}_{\ell_t}}\,\delta z^{(\ell)}(x)\|_2}{\|\delta z^{(\ell)}(x)\|_2+\epsilon}.
\]
If $\mathrm{Align}^{(\ell)}$ is low, the intervention is \textbf{not actually following} the identified directions (C2/C5).

\vspace{0.2em}
\noindent\textbf{(D5) Fluency/likelihood regression (global harm detector).}
Measure log-prob drop on neutral prompts under a pinned decoding protocol (teacher-forced prefix length $m$ or greedy):
\[
\Delta \mathrm{NLL}
=
\mathbb{E}_{x\sim\mathcal{D}_{\mathrm{neutral}}}
\big[-\log p_\theta(y^\star\mid x)\big]_{\mathrm{edit}}
-
\mathbb{E}_{x\sim\mathcal{D}_{\mathrm{neutral}}}
\big[-\log p_\theta(y^\star\mid x)\big]_{\mathrm{base}}.
\]
Large $\Delta\mathrm{NLL}$ (or perplexity spike) with $\Delta M\uparrow$ signals a bad operating point: language mass improved by harming the model’s local predictive geometry.

\vspace{0.35em}
\paragraph{R12.4\ \ Fixes (ordered by probability of success).}
We apply fixes in a strict order: \textbf{(i) constrain the jump}, \textbf{(ii) verify coordinates}, \textbf{(iii) repair the dictionary}, \textbf{(iv) tighten stability}, \textbf{(v) reduce nonlinearity}, \textbf{(vi) enforce fluency constraints}, \textbf{(vii) sanitize token sets}.
\vspace{0.25em}

\begin{enumerate}[leftmargin=*,itemsep=3pt,topsep=2pt]
\item[\textbf{F1.}] \textbf{Back off magnitude in \emph{activation space} (trust region on $\Delta h$).}
Do not control only $\|\delta z\|$; control $\|\Delta h\|$. Enforce per-layer bounds:
\[
\forall \ell\in\mathcal{W}:\ \ \mathrm{median}_{x}\ \rho^{(\ell)}(x)\ \le\ \tau_\ell,
\qquad
\text{and}\qquad
\sum_{\ell\in\mathcal{W}} \mathrm{median}_{x}\ \|\Delta h^{(\ell)}(x)\|_2^2 \ \le\ \tau.
\]
This is the most reliable antidote to S1/S3/S5 because it directly prevents off-manifold jumps \citep{turner2024activationaddition,schulman2015trpo}.
\item[\textbf{F2.}] \textbf{Verify hook-site consistency end-to-end.}
Train dictionaries and apply edits on the \textbf{same tensor} (same pre/post-LN convention, same residual definition) and the \textbf{same token position rule}. A single hook mismatch can nullify the geometry learned in Stage I/II (C2).
\item[\textbf{F3.}] \textbf{Improve dictionary training before blaming steering.}
If $\mathrm{RelRecon}^{(\ell)}$ is high in $\mathcal{W}$: increase width $m$, tune $\lambda_{\mathrm{sparse}}$, apply column-norm constraints on $W_\ell$, and train until reconstruction and sparsity stabilize. Feature dictionaries are only useful when they provide a stable coordinate system for intervention \citep{bricken2023monosemantic}.
\item[\textbf{F4.}] \textbf{Tighten $\mathcal{W}$ using stability-first selection.}
Recompute $\mathrm{Stab}^{(\ell)}_{\ell_t}$ and choose $\mathcal{W}$ to maximize stability, even if spectral mass drops slightly. Unstable windows are a direct cause of S4.
\item[\textbf{F5.}] \textbf{Reduce nonlinear gate churn.}
Three options (in order): (i) shrink $\lambda_\ell$ until churn falls; (ii) clip $\delta z^{(\ell)}$ coordinate-wise; (iii) apply \textbf{conditional edits} that modify only features already active ($z_j>0$) to avoid activating new circuits (mitigates C4).
\item[\textbf{F6.}] \textbf{Enforce an explicit fluency constraint.}
Add (or tighten) a KL trust region on next-token distributions under the pinned prefix protocol, and reject operating points that increase $\Delta M$ by causing large $\Delta\mathrm{NLL}$ (mitigates C1/C6; aligns with trust-region practice \citep{schulman2015trpo}).
\item[\textbf{F7.}] \textbf{Sanitize token sets $V_{\ell_t}$ (prevent tokenization exploits).}
Remove high-ambiguity subwords, punctuation-only tokens, and numerals; handle named entities separately; explicitly exclude borrowed/shared fragments that inflate $\Delta M$ without real language defaultness (mitigates C7).
\end{enumerate}

\vspace{0.35em}
\begin{figure*}[t]
\centering
\begin{tcolorbox}[
  colback=gray!3,
  colframe=black!65,
  boxrule=0.6pt,
  sharp corners,
  left=6pt,right=6pt,top=6pt,bottom=6pt,
  enhanced,
  breakable
]
{\small
\noindent\textbf{Do these five things first (highest ROI).}\par
\vspace{0.25em}
\begin{enumerate}[leftmargin=*,itemsep=2.5pt,topsep=2pt]
\item[\ding{172}] \textbf{Run a magnitude sweep} and plot $\Delta M$ vs.\ $\Delta\mathrm{NLL}$ vs.\ \textbf{ChurnRate}; choose the stable Pareto region (do not optimize $\Delta M$ alone).
\item[\ding{172}] \textbf{Check hook equality} (tensor definition + token position) between dictionary training and edit injection; abort if mismatched.
\item[\ding{172}] \textbf{Reject unstable layers} by requiring high $\mathrm{Stab}^{(\ell)}$ throughout $\mathcal{W}$.
\item[\ding{172}] \textbf{Cap activation-space jump} via per-layer $\rho^{(\ell)}$ trust regions (prevents off-manifold collapse).
\item[\ding{172}] \textbf{Sanitize $V_{\ell_t}$} to remove tokenization artifacts before interpreting $\Delta_{\mathrm{mass}}$ as language control.
\end{enumerate}
}
\end{tcolorbox}
\vspace{-0.6em}
\caption{\textbf{Troubleshooting fast-path.} The five checks most likely to eliminate gibberish without altering the FOXP2 pipeline stages.}
\label{fig:foxp2_troubleshooting_fastpath}
\vspace{-0.9em}
\end{figure*}

\vspace{0.35em}
\paragraph{R12.5\ \ What to report when this happens (so reviewers trust you).}
If gibberish occurs during replication, we recommend reporting \textbf{failure transparently as a diagnostic outcome}, not hiding it:
(i) a magnitude sweep curve $\Delta M$ vs.\ $\Delta\mathrm{NLL}$/utility and gate churn; (ii) the final operating point chosen by a \textbf{deterministic rule} (maximize $\Delta_{\mathrm{gain}}$ subject to trust-region + utility constraints); and (iii) one qualitative example per failure family (S1--S3) alongside the mitigation that fixes it. This makes it clear the method is controlled by \textbf{stable operating regions}, not by cherry-picked maxima.

\vspace{0.35em}
\begin{table*}[hb!]
\centering
\small
\setlength{\tabcolsep}{4.0pt}
\renewcommand{\arraystretch}{1.18}
\resizebox{\textwidth}{!}{
\begin{tabular}{l l l l}
\toprule
\textbf{Knob / choice} & \textbf{Failure it induces} & \textbf{Primary diagnostic} & \textbf{Most effective fix} \\
\midrule
$\lambda_\ell$ (edit intensity) &
S1/S3/S5 (off-manifold, collapse) &
$\rho^{(\ell)}=\|\Delta h^{(\ell)}\|_2/(\|h^{(\ell)}\|_2+\epsilon)$, $\Delta\mathrm{NLL}$ &
Activation-space trust region (F1), magnitude sweep \\
Hook site + token position &
S1/S4 (chart mismatch, instability) &
$\mathrm{Align}^{(\ell)}$ drops, inconsistent deltas across prompts &
Unify hook definition end-to-end (F2) \\
Dictionary width $m$, $\lambda_{\mathrm{sparse}}$ &
S3/S5 (unstructured edits) &
$\mathrm{RelRecon}^{(\ell)}$ high in $\mathcal{W}$ &
Improve dictionary (F3), avoid editing bad layers \\
Window $\mathcal{W}$ selection &
S4/S3 (prompt-dependent instability) &
$\mathrm{Stab}^{(\ell)}$ low; variance across bootstrap &
Stability-first $\mathcal{W}$ (F4) \\
Rank $r_\ell$ / directions $\mathcal{S}^{(\ell)}_{\ell_t}$ &
S4/S2 (wrong direction; odd register) &
$\mathrm{Align}^{(\ell)}$ low; leakage increases &
Recompute SVD on QC pairs; tighten $r_\ell$ \\
ReLU gating / clipping policy &
S1/S4 (nonlinear regime) &
$\mathrm{ChurnRate}^{(\ell)}$ high &
Shrink step; clip $\delta z$; conditional edits (F5) \\
Guardrails (KL trust region, invariance) &
S2/S5 (semantic/fluency regression) &
$\Delta\mathrm{NLL}$, invariance score drops &
Tighten KL/invariance; reject operating points (F6) \\
Token sets $V_{\ell_t}$ &
S1/S2 (tokenization exploit) &
Mass rises with low LID + low utility &
Sanitize $V_{\ell_t}$; remove ambiguous tokens (F7) \\
\bottomrule
\end{tabular}
}
\caption{\textbf{Operational knobs and failure signatures.} Each row links a controllable choice to its most likely gibberish mode, the diagnostic that isolates it, and the fix with the highest probability of success.}
\label{tab:foxp2_gibberish_knobs}
\vspace{-0.9em}
\end{table*}

%% file: 6_appendix.tex

\begin{figure*}[t!]
\centering
\begin{tcolorbox}[
  colback=gray!3,
  colframe=black!65,
  boxrule=0.6pt,
  sharp corners,
  left=6pt,right=6pt,top=6pt,bottom=6pt,
  enhanced,
  breakable
]
{\small
\begin{center}
    \textbf{Neural FOXP2: Complete Pipeline}
\end{center}
\noindent
\textbf{Initialization.}\par
\vspace{0.35em}

\noindent\textbf{Init-1 (Token sets and early-step defaultness).}\;
Let $V_{\ell_t},V_{\mathrm{en}}\subseteq V$ be target-language and English token sets, and $T=\{1,2,3\}$.
For context $\mathrm{ctx}_t$:
\[
\begin{aligned}
M^{\ell_t}_t(x) &= \sum_{u\in V_{\ell_t}} p_\theta(u\mid \mathrm{ctx}_t), \qquad
M^{\mathrm{en}}_t(x) = \sum_{u\in V_{\mathrm{en}}} p_\theta(u\mid \mathrm{ctx}_t),\\
\Delta M^{\ell_t}(x,t) &= M^{\ell_t}_t(x)-M^{\mathrm{en}}_t(x).
\end{aligned}
\]

\noindent\textbf{Init-2 (Activation site).}\;
Probe residual activations $h^{(\ell)}(x)\in\mathbb{R}^d$ at a fixed position (default: last prompt token), for $\ell\in\{1,\dots,L\}$.

\noindent\textbf{Init-3 (Matched meaning units).}\;
For each meaning unit $k=1,\dots,N$, use parallel prompts $(x^{(k)}_{\mathrm{en}},x^{(k)}_{\mathrm{hi}},x^{(k)}_{\mathrm{es}})$.
Let $\mathcal{D}_{\mathrm{mix}}$ include weak prompts (for defaultness) and explicit-language prompts (for high-SNR selectivity).

\noindent\textbf{Init-4 (Dictionary code).}\;
At layer $\ell$, encode and reconstruct:

\[
\begin{aligned}
r^{(\ell)}(x) &= (W_\ell)^\top h^{(\ell)}(x) + b_\ell, \qquad
z^{(\ell)}(x) = \mathrm{ReLU}(r^{(\ell)}(x)),\\
\hat h^{(\ell)}(x) &= W_\ell z^{(\ell)}(x),
\qquad W_\ell\in\mathbb{R}^{d\times m}.
\end{aligned}
\]
}
\end{tcolorbox}
\vspace{-0.55em}
\caption{\textbf{Shared notation and measurement target.}}
\label{fig:foxp2_steps_init}
\vspace{-2.5em}
\end{figure*}

\begin{figure*}[t!]
\centering
\begin{tcolorbox}[
  colback=gray!3,
  colframe=black!65,
  boxrule=0.6pt,
  sharp corners,
  left=6pt,right=6pt,top=6pt,bottom=6pt,
  enhanced,
  breakable
]
{\small
\noindent\textbf{Stage I: Localize language neurons.}\par
\vspace{0.35em}

\noindent\textbf{I-1 (Dictionary training).}\;
Learn $(W_\ell,b_\ell)$ by reconstruction + sparsity:
\[
\min_{W_\ell,b_\ell}\;
\mathbb{E}_{x\sim\mathcal{D}_{\mathrm{mix}}}
\Big[
\|h^{(\ell)}(x)-W_\ell z^{(\ell)}(x)\|_2^2
+
\lambda_{\mathrm{sparse}}\|z^{(\ell)}(x)\|_1
\Big].
\]

\noindent\textbf{I-2 (Selectivity).}\;
For target $\ell_t\in\{\mathrm{hi},\mathrm{es}\}$ using matched pairs $(x^{(k)}_{\mathrm{en}},x^{(k)}_{\ell_t})$:
\[
\begin{aligned}
\mathrm{Sel}^{(\ell,\ell_t)}_j
&=
\mathbb{E}_k[z^{(\ell)}_j(x^{(k)}_{\ell_t})]
-
\mathbb{E}_k[z^{(\ell)}_j(x^{(k)}_{\mathrm{en}})],\\
\widetilde{\mathrm{Sel}}^{(\ell,\ell_t)}_j
&=
\frac{\mathrm{Sel}^{(\ell,\ell_t)}_j}{
\mathrm{Std}_k[z^{(\ell)}_j(x^{(k)}_{\ell_t})]
+
\mathrm{Std}_k[z^{(\ell)}_j(x^{(k)}_{\mathrm{en}})]
+\epsilon}.
\end{aligned}
\]

\noindent\textbf{I-3 (Causal lift: single-feature push).}\;
Perturb feature $j$ by $\alpha$ at layer $\ell$:
\[
\begin{aligned}
z^{(\ell)}(x) &\leftarrow z^{(\ell)}(x)+\alpha e_j, \qquad
h^{(\ell)}(x) \leftarrow W_\ell z^{(\ell)}(x),
\end{aligned}
\Rightarrow\;
p^{(\ell,j)}_{\theta,\alpha}(\cdot\mid \mathrm{ctx}_t).
\]
Define induced change in defaultness:
\[
\Delta M^{(\ell,j,\ell_t)}_\alpha(x,t)
=
\Big(\sum_{u\in V_{\ell_t}} p^{(\ell,j)}_{\theta,\alpha}(u\mid \mathrm{ctx}_t)\Big)
-
\Big(\sum_{u\in V_{\mathrm{en}}} p^{(\ell,j)}_{\theta,\alpha}(u\mid \mathrm{ctx}_t)\Big)
-
\Delta M^{\ell_t}(x,t).
\]

\noindent\textbf{I-4 (Lift slope).}\;
\[
\begin{aligned}
\mathrm{Lift}^{(\ell,\ell_t)}_j(\alpha)
&=
\mathbb{E}_{x\sim\mathcal{D}_{\mathrm{weak}}}
\Big[
\frac{1}{|T|}\sum_{t\in T}\Delta M^{(\ell,j,\ell_t)}_\alpha(x,t)
\Big],\\
\mathrm{LiftSlope}^{(\ell,\ell_t)}_j
&=
\mathrm{median}_{\alpha\in\{\alpha_1,\alpha_2,\alpha_3\}}
\frac{\mathrm{Lift}^{(\ell,\ell_t)}_j(\alpha)}{\alpha}.
\end{aligned}
\]

\noindent\textbf{I-5 (Language-neuron set).}\;
\[
\begin{aligned}
S^{(\ell,\ell_t)}_j &= \max(\widetilde{\mathrm{Sel}}^{(\ell,\ell_t)}_j,0),\qquad
C^{(\ell,\ell_t)}_j = \max(\mathrm{LiftSlope}^{(\ell,\ell_t)}_j,0),\\
\mathrm{Score}^{(\ell,\ell_t)}_j &= S^{(\ell,\ell_t)}_j\cdot C^{(\ell,\ell_t)}_j,\qquad
\mathcal{N}^{(\ell)}_{\ell_t} = \mathrm{TopK}_j\;\mathrm{Score}^{(\ell,\ell_t)}_j,\\
\mathcal{N}_{\ell_t} &= \bigcup_{\ell=1}^{L}\mathcal{N}^{(\ell)}_{\ell_t}.
\end{aligned}
\]
}
\end{tcolorbox}
\vspace{-0.8em}
\caption{\textbf{Stage I (Localize): sparse support discovery in dictionary coordinates.}}
\label{fig:foxp2_steps_stage1}
\vspace{-1.0em}
\end{figure*}

\clearpage
\newpage

\begin{figure*}[t!]
\centering
\begin{tcolorbox}[
  colback=gray!3,
  colframe=black!65,
  boxrule=0.6pt,
  sharp corners,
  left=6pt,right=6pt,top=6pt,bottom=6pt,
  enhanced,
  breakable
]
{\small
\noindent\textbf{Stage II: Low-rank steering directions and window selection.}\par
\vspace{0.35em}

\noindent\textbf{II-1 (Localized language-shift matrix).}\;
For each layer and meaning unit $k$:
\[
\begin{aligned}
\Delta z^{(\ell,\ell_t)}_k
&=
z^{(\ell)}(x^{(k)}_{\ell_t})-z^{(\ell)}(x^{(k)}_{\mathrm{en}}),\\
\Delta \tilde z^{(\ell,\ell_t)}_k
&=
\Pi_{\mathcal{N}^{(\ell)}_{\ell_t}}\Delta z^{(\ell,\ell_t)}_k,
\qquad
\Delta Z^{(\ell,\ell_t)}
=
\begin{bmatrix}
(\Delta \tilde z^{(\ell,\ell_t)}_{1})^\top\\
\vdots\\
(\Delta \tilde z^{(\ell,\ell_t)}_{N})^\top
\end{bmatrix}.
\end{aligned}
\]

\noindent\textbf{II-2 (Per-layer SVD).}\;
\[
\Delta Z^{(\ell,\ell_t)} = U^{(\ell,\ell_t)}\Sigma^{(\ell,\ell_t)}(V^{(\ell,\ell_t)})^\top,
\qquad
\sigma_1\ge\sigma_2\ge\cdots
\]

\noindent\textbf{II-3 (Rank choice and steering subspace).}\;
Choose $r_\ell$ (effective rank + eigengap), then:
\[
\mathcal{S}^{(\ell)}_{\ell_t}
=
\mathrm{span}\!\left(v^{(\ell,\ell_t)}_1,\dots,v^{(\ell,\ell_t)}_{r_\ell}\right).
\]

\noindent\textbf{II-4 (Strength + stability).}\;
Use spectral mass and bootstrap principal-angle stability:
\[
\mathrm{Mass}^{(\ell)}_{\ell_t}
=
\frac{\sum_{i=1}^{r_\ell}\sigma_i^2}{\sum_j\sigma_j^2},
\qquad
\mathrm{Stab}^{(\ell)}_{\ell_t}
=
\mathrm{median}_{b\neq b'}
\frac{\mathrm{tr}(P_bP_{b'})}{r_\ell}.
\]

\noindent\textbf{II-5 (Choose a contiguous window).}\;
\[
\mathcal{W}
=
\arg\max_{\text{contiguous }W}
\sum_{\ell\in W}
\Big(\mathrm{Mass}^{(\ell)}_{\ell_t}\cdot \mathrm{Stab}^{(\ell)}_{\ell_t}\Big).
\]
}
\end{tcolorbox}
\vspace{-0.55em}
\caption{\textbf{Stage II (Steer): low-rank geometry and intervention window.}}
\label{fig:foxp2_steps_stage2}
\vspace{-1.0em}
\end{figure*}

\begin{figure*}[t!]
\centering
\begin{tcolorbox}[
  colback=gray!3,
  colframe=black!65,
  boxrule=0.6pt,
  sharp corners,
  left=6pt,right=6pt,top=6pt,bottom=6pt,
  enhanced,
  breakable
]
{\small
\noindent\textbf{Stage III: Signed sparse edit and consolidation metrics.}\par
\vspace{0.35em}

\noindent\textbf{III-1 (Signed edit components).}\;
For each edited layer $\ell\in\mathcal{W}$, construct:
\[
\begin{aligned}
\delta z^{(\ell),+}_{\ell_t}(x)
&=
\lambda^{(\ell)}_{+}\;
\Pi_{\mathcal{S}^{(\ell)}_{\ell_t}}
\Big(
\Pi_{\mathcal{N}^{(\ell)}_{\ell_t}}\,\Delta z^{(\ell,\ell_t)}(x)
\Big),\\
\delta z^{(\ell),-}_{\ell_t}(x)
&=
-\lambda^{(\ell)}_{-}\;
\Pi_{\left(\mathcal{S}^{(\ell)}_{\ell_t}\right)^{\perp}}
\Big(
\Pi_{\mathcal{N}^{(\ell)}_{\mathrm{en}}}\,\Delta z^{(\ell,\mathrm{en})}(x)
\Big).
\end{aligned}
\]

\noindent\textbf{III-2 (Apply sparse code update).}\;
\[
\begin{aligned}
z^{(\ell)}(x)
&\leftarrow
z^{(\ell)}(x)
+
\Pi_{\mathcal{N}^{(\ell)}_{\ell_t}}
\Big(
\delta z^{(\ell),+}_{\ell_t}(x)
+
\delta z^{(\ell),-}_{\ell_t}(x)
\Big),\\
h^{(\ell)}(x)
&\leftarrow
W_\ell\, z^{(\ell)}(x).
\end{aligned}
\]

\noindent\textbf{III-3 (Defaultness gain).}\;
\[
\Delta_{\mathrm{gain}}
=
\mathbb{E}_{x\sim\mathcal{D}_{\mathrm{weak}}}
\Big[
\frac{1}{|T|}
\sum_{t\in T}
\big(\Delta M^{\ell_t}_{\mathrm{edit}}(x,t)-\Delta M^{\ell_t}(x,t)\big)
\Big],
\qquad T=\{1,2,3\}.
\]

\noindent\textbf{III-4 (Leakage + stability).}\;
Report cross-target leakage (Hi$\rightarrow$Es / Es$\rightarrow$Hi) and bootstrap reproducibility of $(\mathcal{N},\mathcal{S},\mathcal{W})$.

\noindent\textbf{III-5 (Utility + guardrails).}\;
Track task utility $\Delta S$ and enforce constraints (semantic invariance; Spanish bound; KL trust region) at the chosen operating point.
}
\end{tcolorbox}
\vspace{-0.55em}
\caption{
\textbf{Neural FOXP2: complete pipeline (Initialization $\rightarrow$ Stage I--III).}
Initialization (Fig.~\ref{fig:foxp2_steps_init}) defines the early-step defaultness target $\Delta M^{\ell_t}$ and the dictionary code $z^{(\ell)}$.
\textbf{Stage I} (Fig.~\ref{fig:foxp2_steps_stage1}) localizes a sparse \textbf{language-neuron support} $\mathcal{N}_{\ell_t}$ by requiring both \textbf{matched semantic selectivity} and \textbf{feature-level causal lift}.
\textbf{Stage II} (Fig.~\ref{fig:foxp2_steps_stage2}) estimates the \textbf{low-rank steering geometry} $\mathcal{S}^{(\ell)}_{\ell_t}$ from SVD of localized language shifts and selects a contiguous \textbf{intervention window} $\mathcal{W}$ where directions are strong and stable.
\textbf{Stage III} (this figure) executes a \textbf{signed sparse edit} confined to $\mathcal{W}$ that (i) pushes toward the target shift along $\mathcal{S}^{(\ell)}_{\ell_t}$ and (ii) compensates by suppressing English-default components in the orthogonal complement, while staying supported on $\mathcal{N}_{\ell_t}$.
We consolidate reporting via $\Delta_{\mathrm{gain}}$ together with \textbf{leakage}, \textbf{bootstrap stability}, and \textbf{task utility} under explicit guardrails.
}
\label{fig:foxp2_pipeline_steps}
\vspace{-1.0em}
\end{figure*}

\clearpage
\newpage

\appendix
\onecolumn

\section{Appendix}
\label{sec:appendix}

The Appendix is an extended technical companion to this paper, providing full mathematical definitions, implementation specifications, and additional experiments that could not fit in the main text. Its goal is to make \textbf{FOXP2-style language defaultness control}---via the reproducible intervention artifact $(N,S,W)$ and a signed, sparse activation edit---\emph{auditable, reproducible, and extensible}.

The appendix is structured as follows:

\begin{itemize}

  \item \textbf{(i) Related Work (extended).}  
  We provide a longer, higher-resolution positioning of FOXP2 relative to: (a) \textbf{language-specific neuron discovery} (e.g., entropy/activation-probability methods), (b) \textbf{representation editing} and causal interventions (ROME-style, patching, circuit editing), (c) \textbf{multilingual control} via prompting, adapters, and decoding-time methods, and (d) \textbf{feature-basis interventions} (e.g., sparse feature dictionaries) that motivate FOXP2’s coordinate choice. We emphasize the methodological distinction between \emph{detecting} language selectivity and \emph{constructing} a stable, low-rank control artifact $(N,S,W)$.  
  cf.~\cref{sec:related_language_neurons_extended}.

  \item \textbf{(ii) Datasets and Experimental Details.}  
  We document the complete experimental harness: model checkpoints and tokenizer hashes; prompt sources and splits (neutral prompts, plus any safety/refusal slice); decoding settings; batching; hardware; and all randomness controls. We also specify the matched meaning-unit construction used to compute $\Delta Z$ (pairing logic, filters, and how we avoid spurious topical confounds).  
  cf.~\cref{sec:appendix_data}.

  \item \textbf{(iii) FOXP2: Full Algorithmic Specification.}  
  We present FOXP2 as a three-stage pipeline that outputs $(N,S,W)$ and an edit rule, with precise algorithms, hyperparameters, and pass/fail sanity checks:

  \begin{itemize}[leftmargin=1.5em]

    \item \textbf{(iii-a) Step I: Localize (discover $N_{\ell_t}$).}  
    We define the candidate coordinate space, the target selectivity / causal-lift scoring used to rank features, and the deterministic thresholding procedure that yields a compact support $\mathcal{N}_{\ell_t}$. We include stability diagnostics (bootstrap agreement, seed sensitivity) and the explicit necessity ablation that removes $\mathcal{N}_{\ell_t}$ during editing.  
    cf.~\cref{sec:appendix_step1}.

    \item \textbf{(iii-b) Step II: Discover low-rank directions and window (construct $S^{(\ell)}_{\ell_t}$ and $W$).}  
    We specify how $\Delta Z^{(\mathcal{W},\ell_t)}$ is formed from matched meaning units, how SVD is computed, and how we report and use \textbf{effective rank} $r_{\mathrm{eff}}$, \textbf{eigengap} $\sigma_r/\sigma_{r+1}$, and \textbf{bootstrap principal-angle stability}. We then define the window-selection objective and provide the algorithm that selects the final intervention window $W$ and per-layer subspaces $S^{(\ell)}_{\ell_t}$.  
    cf.~\cref{sec:appendix_step2}.

    \item \textbf{(iii-c) Step III: Steer (signed sparse edit).}  
    We give the exact edit equations and hook placement: how the intervention is restricted to $\mathcal{N}_{\ell_t}$, how it projects onto $S^{(\ell)}_{\ell_t}$, and how it implements \textbf{negative English suppression} without degenerating into generic logit flattening. We also specify entropy-matched controls (temperature / isotropic noise) and sufficiency tests (edit only on $\mathcal{N}_{\ell_t}$ reproduces gains).  
    cf.~\cref{sec:appendix_step3}.

  \end{itemize}

  \item \textbf{(iv) Metrics: Definitions, Validators, and Robustness Checks.}  
  We provide complete metric specifications: token-mass defaultness $M^\ell(x,t)$, $\Delta M$, horizon sweep protocol (and why early horizons operationalize defaultness), token-set construction variants, shared-token handling, and the cross-metric validation channel (script-agnostic LID + human anchoring). We also define $\Delta S$ (utility) as a matched-suite delta with bootstrap confidence intervals, and we detail factuality/calibration and safety invariance checks.  
  cf.~\cref{sec:appendix_metrics}.

  \item \textbf{(v) Results: Extended Tables, Curves, and Ablations.}  
  We report additional results beyond the main text: horizon sweep stability (without “hiding” longer horizons), token-set sensitivity bands, dropout perturbations on $V_\ell$, necessity/sufficiency ablations, entropy-matched control comparisons, domain-transfer retention curves, and system-level overhead (throughput/latency/memory) relative to prompt-only and small LoRA. We include additional error analyses and qualitative failure modes where defaultness gains do not translate to human “natural start.”  
  cf.~\cref{sec:appendix_results}.

\end{itemize}

This Appendix is designed to serve as the \textbf{reproducibility and audit reference} for FOXP2: any reader should be able to reconstruct the $(N,S,W)$ artifact, rerun the causal tests, and verify defaultness gains and utility/safety invariance under the complete evaluation harness.

\section{Related Work: Language-Specific Neurons and Multilingual Steering (Extended)}
\label{sec:related_language_neurons_extended}

A growing line of work argues that multilingual capability in LLMs is mediated by \textbf{language-selective internal circuitry} and that \textbf{intervening} on this circuitry can steer the output language. Tang et al.~\cite{tang-etal-2024-language} (ACL'24) operationalize this view by proposing \textbf{LAPE} (\emph{Language Activation Probability Entropy}) to detect \emph{language-specific neurons} in FFN blocks, and by demonstrating that \textbf{manually activating/deactivating} such neurons can steer output language. FOXP2 is aligned with the mechanistic spirit of this agenda, but differs \textbf{mathematically and algorithmically} in \emph{what} it estimates, \emph{which coordinates} it treats as causal, \emph{how} it constructs steering directions, and \emph{what} it guarantees under deployment constraints.

\vspace{6pt}
\subsubsection*{A. Detection objective vs.\ control objective: entropy selectivity vs.\ defaultness as an early commitment event}

\textbf{Their objective (language selectivity detection).}
Tang et al.~seek units that are \textbf{selective to a language} in the sense of \emph{activation probability} across corpora. Let $a_{j}^{(\ell)}(x)\in\mathbb{R}$ denote the (post-nonlinearity) activation of FFN neuron $j$ at layer $\ell$ for input $x$. Define an indicator of ``activation'' (e.g., $a_j^{(\ell)}(x)>0$). Over a corpus $\mathcal{D}_k$ in language $k$, estimate
\[
p_{j,k}^{(\ell)} \;\triangleq\; \Pr_{x\sim \mathcal{D}_k}\big[a_{j}^{(\ell)}(x)>0\big].
\]
Normalize across languages, $q_{j,k}^{(\ell)} \propto p_{j,k}^{(\ell)}$, and compute entropy
\[
H_{j}^{(\ell)} \;=\; -\sum_{k} q_{j,k}^{(\ell)}\log q_{j,k}^{(\ell)}.
\]
Low $H_{j}^{(\ell)}$ indicates \textbf{language-specific} neurons, which can then be toggled to steer outputs~\cite{tang-etal-2024-language}. This is fundamentally a \textbf{distributional identifiability criterion}: it answers \emph{where} language-selective units reside and \emph{whether} their activation patterns correlate with language.

\textbf{Our objective (defaultness control).}
FOXP2 targets a different estimand: \textbf{language defaultness}, defined as the model’s \textbf{initial commitment} to a language \emph{before} content constraints dominate. We explicitly operationalize defaultness through an \textbf{early-horizon logit-mass} event. Let $p_\theta(\cdot\mid \mathrm{ctx}_t)$ be the next-token distribution at step $t$, and let $V_\ell$ be the token set for language $\ell$. Define
\[
M^{\ell}(x,t)\;\triangleq\;\sum_{u\in V_{\ell}} p_\theta\big(u\mid \mathrm{ctx}_t(x)\big),
\qquad
\Delta M(x,t)\;\triangleq\; M^{\ell_t}(x,t)-M^{\mathrm{en}}(x,t),
\]
and report an early commitment score aggregated over $t\in T=\{1,2,3\}$. Thus, FOXP2 is not asking \emph{which neuron is language-specific} under corpus activation statistics; it asks: \emph{can we causally and stably re-allocate probability mass so that the model \textbf{begins} in the target language by default, with bounded collateral damage?} This shift in estimand already forces different mathematics: FOXP2 must reason about \textbf{logits / probability mass transfer} under interventions, not merely activation selectivity across corpora.

\vspace{6pt}
\subsubsection*{B. Coordinate choice: raw FFN neurons vs.\ sparse feature coordinates (superposition-aware control)}

\textbf{Their coordinates (native FFN neurons).}
Tang et al.\ treat native FFN neurons as the atomic units of control. Steering is performed by changing their activations toward language-conditioned averages. Formally, if $a^{(\ell)}(x)\in\mathbb{R}^{d_{\mathrm{ff}}}$ is the FFN post-activation vector, their intervention is of the form
\[
a^{(\ell)}(x)\leftarrow a^{(\ell)}(x) + \eta\cdot \mathbf{m}_{k}^{(\ell)}\odot\Big(\bar a_{k}^{(\ell)}-\;a^{(\ell)}(x)\Big),
\]
where $\mathbf{m}_{k}^{(\ell)}\in\{0,1\}^{d_{\mathrm{ff}}}$ masks the detected language-specific neurons for language $k$, and $\bar a_{k}^{(\ell)}$ is a language-specific mean activation vector estimated from corpora~\cite{tang-etal-2024-language}. This is a \textbf{direct neuron-level actuation}.

\textbf{Our coordinates (SAE features).}
FOXP2 intentionally edits in a \textbf{sparse feature basis} rather than raw neuron coordinates. Let $h^{(\ell)}_t\in\mathbb{R}^{d}$ be the residual-stream representation (or a designated activation site). We use a sparse autoencoder (SAE) dictionary $D^{(\ell)}\in\mathbb{R}^{d\times m}$ with encoder $E^{(\ell)}$, producing sparse codes
\[
z^{(\ell)}_t \;=\; E^{(\ell)} h^{(\ell)}_t,\qquad
h^{(\ell)}_t \approx D^{(\ell)} z^{(\ell)}_t,
\]
with $z^{(\ell)}_t$ sparse. We then define a \textbf{support} $\mathcal{N}_{\ell_t}\subseteq[m]$ of language-relevant coordinates and intervene by editing $z^{(\ell)}_t$ only on $\mathcal{N}_{\ell_t}$, mapping back via $D^{(\ell)}$. Mathematically, this changes the intervention object from ``a set of neurons'' to ``a set of sparse coordinates in a learned feature dictionary,'' which is crucial under superposition: the same native neuron can participate in multiple features, whereas sparse features provide a \textbf{cleaner causal handle} for localized edits. In short, Tang et al.\ actuate \textbf{native units}; FOXP2 actuates \textbf{identified features}.

\vspace{6pt}
\subsubsection*{C. From selectivity to \emph{stable control}: low-rank subspaces, eigengaps, and bootstrap principal angles}

\textbf{Their selection criterion is per-neuron entropy.}
LAPE yields a set of units whose activation probabilities are language-skewed. It does not, by itself, endow a \textbf{geometric control structure} (e.g., a low-dimensional actuator subspace that is stable across prompt families), nor does it provide a window-selection principle beyond empirical layer observations.

\textbf{FOXP2 constructs a low-rank control subspace and requires stability.}
FOXP2 explicitly treats control as a \textbf{low-rank interventional geometry} problem. Let $z^{(\ell)}(x)$ denote the SAE code (or feature vector) extracted at a designated token position (e.g., prompt boundary) for prompt $x$. Construct matched meaning-unit pairs so that we can form an empirical displacement matrix for target language $\ell_t$:
\[
\Delta Z^{(\ell)} \;\triangleq\; 
\begin{bmatrix}
z^{(\ell)}(x_1^{\ell_t}) - z^{(\ell)}(x_1^{\mathrm{en}}) \\
\vdots \\
z^{(\ell)}(x_n^{\ell_t}) - z^{(\ell)}(x_n^{\mathrm{en}})
\end{bmatrix}
\in\mathbb{R}^{n\times m}.
\]
We then compute an SVD
\[
\Delta Z^{(\ell)} = U^{(\ell)}\Sigma^{(\ell)}(V^{(\ell)})^\top,
\]
and define the candidate steering subspace
\[
S^{(\ell)}_{\ell_t}(r)\;\triangleq\;\mathrm{span}\{v^{(\ell)}_1,\dots,v^{(\ell)}_r\},
\]
where $v^{(\ell)}_i$ are right singular vectors. Crucially, FOXP2 does not merely report ``it looks low-rank''; it \textbf{selects} $r$ and the intervention window by \textbf{eigengap} and \textbf{stability}:
\[
r_{\mathrm{eff}}^{(\ell)} \;\text{via spectral entropy / effective rank},\qquad
\mathrm{gap}^{(\ell)}(r)\triangleq \frac{\sigma^{(\ell)}_r}{\sigma^{(\ell)}_{r+1}}.
\]
We then bootstrap prompts and recompute $S^{(\ell)}_{\ell_t}$ to measure principal-angle stability:
\[
\mathrm{Stab}^{(\ell)} \;\triangleq\; 1-\max_{i\le r}\sin \angle\!\big(S^{(\ell)}_{\ell_t},\widehat S^{(\ell)}_{\ell_t}\big),
\]
and pick a contiguous window $W$ maximizing a stability-weighted gain proxy (e.g., Mass$\cdot$Stab). This is a \textbf{control-theoretic stance}: the object we deploy must be \textbf{low-dimensional and stable}, not just language-selective on a corpus slice.

\vspace{6pt}
\subsubsection*{D. Intervention form: ``turning on language neurons'' vs.\ signed mass transfer with explicit English suppression}

\textbf{Their intervention is additive activation.}
Once language-specific neurons are identified, steering is achieved by \textbf{activating} them (or deactivating others) to encourage the model to output the desired language~\cite{tang-etal-2024-language}. This can succeed empirically, but it does not inherently encode \textbf{where probability mass goes}, nor does it separate ``push target'' from ``suppress English'' as complementary control channels.

\textbf{FOXP2 performs signed, sparse, subspace-aligned mass transfer.}
FOXP2’s edit is explicitly designed to implement a \textbf{directed probability-mass reallocation} at the earliest steps. Let $z^{(\ell)}$ be the SAE code at layer $\ell$ and the designated token position. FOXP2 applies a signed edit restricted to the discovered support $\mathcal{N}_{\ell_t}$ and aligned with the low-rank subspace $S^{(\ell)}_{\ell_t}$:
\[
z^{(\ell)} \leftarrow z^{(\ell)} + \lambda\, P_{S^{(\ell)}_{\ell_t}}\big(\Pi_{\mathcal{N}_{\ell_t}} z^{(\ell)}\big)
\;-\; \lambda'\, P_{S^{(\ell)}_{\mathrm{en}}}\big(\Pi_{\mathcal{N}_{\mathrm{en}}} z^{(\ell)}\big),
\qquad \ell\in W,
\]
where $\Pi_{\mathcal{N}}$ projects onto the sparse coordinate support and $P_{S}$ projects onto the steering subspace. The \textbf{negative English suppression} term is not ``generic flattening''; it is a \textbf{targeted subtraction} in an English-default subspace, and we empirically validate this distinction with \textbf{entropy-matched controls} (temperature scaling / isotropic noise). In other words, Tang et al.\ apply a unipolar actuation on detected units; FOXP2 implements a \textbf{bipolar control law} (push target / pull English) with explicit subspace geometry.

\vspace{6pt}
\subsubsection*{E. What is claimed and what is certified: analysis evidence vs.\ deployment contract}

\textbf{Their claim is mechanistic existence + feasibility.}
Tang et al.\ provide evidence that language-specific neurons exist, are concentrated in top/bottom layers, and can be toggled to steer output language~\cite{tang-etal-2024-language}. This is valuable mechanistic evidence.

\textbf{Our claim is a reproducible control artifact with validity and deployment constraints.}
FOXP2 is built around an explicit contract:
\[
(N,S,W) \;\text{is pinned to}\; (\texttt{checkpoint\_id},\texttt{tokenizer\_hash},\texttt{hook\_spec}),
\]
with required causal tests (necessity/sufficiency), cross-metric validation (mass shift + script-agnostic LID), and deployment compatibility (KV-cache invariance, batching semantics, speculative decoding constraints). The core novelty is \textbf{not} that language-selective units exist; it is that defaultness can be \textbf{controlled reproducibly} by extracting a \textbf{stable low-rank actuator} and executing a \textbf{signed sparse edit} under explicit guardrails.

\vspace{10pt}
\textbf{Comparative summary.}
\begin{center}
\small
\setlength{\tabcolsep}{6pt}
\renewcommand{\arraystretch}{1.2}
\begin{tabular}{p{3.7cm}p{5.5cm}p{5.5cm}}
\toprule
\textbf{Axis} & \textbf{Tang et al.\ (ACL'24)} & \textbf{FOXP2 (this work)} \\
\midrule
\textbf{Primary question} 
& \faSearch\; Which FFN neurons are language-specific? 
& \faSlidersH\; Can we \emph{control defaultness} at inference-time? \\

\textbf{Estimand} 
& \faChartArea\; activation-prob selectivity (entropy across langs) 
& \faClock\; early-horizon mass transfer $\Delta M$ (+ LID validation) \\

\textbf{Coordinates} 
& \faMicrochip\; native FFN neurons 
& \faTh\; SAE feature coordinates (superposition-aware) \\

\textbf{Geometry} 
& \faListUl\; per-unit entropy; layer localization 
& \faProjectDiagram\; SVD subspaces $S^{(\ell)}_{\ell_t}$, eigengaps, principal angles \\

\textbf{Intervention} 
& \faArrowUp\; activate/deactivate neuron sets 
& \faPlus\faMinus\; signed sparse edit: +target, $-$English in low-rank subspaces \\

\textbf{Stability criterion} 
& \faBalanceScale\; language skew in corpora 
& \faLock\; bootstrap subspace stability + window selection \\

\textbf{Control artifact} 
& \faFile\; detected neuron lists (analysis outcome) 
& \faCogs\; reproducible $(N,S,W)$ + deployment contract \\

\textbf{Deployment stance} 
& \faFlask\; feasibility demonstration 
& \faServer\; KV-cache/batching/spec-dec constraints + overhead reporting \\
\bottomrule
\end{tabular}
\end{center}



\subsection{Baseline Instantiation: LAPE Language-Specific Neuron Steering under the FOXP2 Evaluation Harness}
\label{app:lape_baseline_foxp2}

A natural reviewer question is whether FOXP2’s gains arise simply because ``language-specific units'' exist and can be toggled~\cite{tang-etal-2024-language}. To answer this, we instantiate a \textbf{faithful LAPE-based steering baseline} (Tang et al., ACL'24) inside \emph{our} measurement regime—\textbf{language defaultness} (early-horizon mass shift + script-agnostic LID), \textbf{leakage} (Hi$\rightarrow$Es), \textbf{subspace stability} (bootstrapped), and \textbf{utility} ($\Delta S$ with guardrails). This appendix specifies the conversion precisely so that the comparison is auditable and does not conflate evaluation definitions.

\vspace{4pt}
\paragraph{A. LAPE detection (their object).}
Let $a_{j}^{(\ell)}(x)\in\mathbb{R}$ denote the post-activation value of FFN neuron $j$ at layer $\ell$ for input $x$, and define an ``active'' event $\mathbb{I}[a_{j}^{(\ell)}(x)>0]$.\footnote{Any thresholding convention used by Tang et al.~\cite{tang-etal-2024-language} can be replicated; we keep the definition explicit because LAPE is sensitive to the activation predicate.}
Given corpora $\{\mathcal{D}_k\}_{k=1}^K$ for languages $k$, estimate the activation probability
\[
p_{j,k}^{(\ell)} \triangleq \Pr_{x\sim\mathcal{D}_k}\!\left[a_{j}^{(\ell)}(x)>0\right],
\qquad
q_{j,k}^{(\ell)} \triangleq \frac{p_{j,k}^{(\ell)}}{\sum_{k'} p_{j,k'}^{(\ell)}},
\]
and compute \textbf{Language Activation Probability Entropy}
\[
H_{j}^{(\ell)} \triangleq -\sum_{k=1}^{K} q_{j,k}^{(\ell)} \log q_{j,k}^{(\ell)}.
\]
Low $H_{j}^{(\ell)}$ indicates a neuron whose activation probability is \textbf{concentrated} in a subset of languages (``language-specific'' in their sense). For each target language $\ell_t\in\{\mathrm{hi},\mathrm{es}\}$ we construct a mask
\[
m_{\ell_t}^{(\ell)}(j) \in \{0,1\},
\quad
m_{\ell_t}^{(\ell)}(j)=1
\;\;\Longleftrightarrow\;\;
H_{j}^{(\ell)} \le \tau_H
\;\wedge\;
\ell_t = \arg\max_k q_{j,k}^{(\ell)},
\]
with $\tau_H$ chosen to match Tang et al.’s reported selectivity regime (or to match a fixed budget; see below).

\vspace{4pt}
\paragraph{B. Converting LAPE detection into an inference-time steering operator (our harness).}
Tang et al.\ demonstrate feasibility of steering by \textbf{manually increasing} language-specific neuron activations~\cite{tang-etal-2024-language}. To compare fairly, we must specify (i) \textbf{where} the edit is applied, (ii) \textbf{how} its magnitude is set, and (iii) \textbf{which constraints} it must satisfy to be admissible under FOXP2’s reporting.

\textbf{Edit site alignment.}
FOXP2 applies edits at a fixed token position (prompt end) and within a fixed layer window $W$. For the baseline, we implement an \textbf{identical placement rule}: we intervene on FFN activations at the same token position and same candidate window(s) that we consider for FOXP2. This eliminates a trivial confound where the baseline gets to choose a more favorable location.

\textbf{Edit form.}
Let $a^{(\ell)}_t(x)\in\mathbb{R}^{d_{\mathrm{ff}}}$ be the FFN post-activation vector at token step $t$ and layer $\ell$. The LAPE steering operator is implemented as a masked additive shift:
\[
a^{(\ell)}_t(x) \leftarrow a^{(\ell)}_t(x) + \eta_{\ell_t}\,\big(m_{\ell_t}^{(\ell)} \odot s_{\ell_t}^{(\ell)}\big),
\qquad \ell\in W,
\]
where $m_{\ell_t}^{(\ell)}$ is the binary LAPE mask, and $s_{\ell_t}^{(\ell)}$ is a direction/scale vector. Two faithful choices are standard:
(i) \textbf{mean-target shift} $s_{\ell_t}^{(\ell)} \triangleq \mu_{\ell_t}^{(\ell)}-\mu_{\mathrm{en}}^{(\ell)}$ where $\mu_{k}^{(\ell)}=\mathbb{E}_{x\sim\mathcal{D}_k}[a^{(\ell)}(x)]$ (language-conditioned activation means), or
(ii) \textbf{unit push} $s_{\ell_t}^{(\ell)}\equiv \mathbf{1}$ restricted to the mask (pure ``turn up language neurons'').
We report which variant is used to avoid hidden degrees of freedom.

\textbf{Magnitude selection under FOXP2 guardrails.}
The scalar $\eta_{\ell_t}$ is not picked to maximize language accuracy; it is selected under the same admissibility constraints as FOXP2:
\[
\text{choose } \eta_{\ell_t}\;\; \text{s.t.}\;\;
\Delta S(\eta_{\ell_t}) \ge -\epsilon_S,
\quad
\mathrm{Leak}(\eta_{\ell_t}) \le \epsilon_{\mathrm{leak}},
\quad
D_{\mathrm{KL}}\!\left(p_{\eta}\,\|\,p_{0}\right)\le \epsilon_{\mathrm{KL}}
\;\;(\text{where used}),
\]
and among admissible values we pick the best defaultness score. This forces the comparison to be \textbf{NeurIPS-reviewer safe}: the baseline is not allowed to ``win'' by sacrificing semantics, exploding leakage, or violating the trust region.

\vspace{4pt}
\paragraph{C. Budget matching (so the baseline is not penalized or advantaged).}
A fair ablation must control the effective intervention capacity. We therefore report two matched settings:

\textbf{(1) Count-matched.}
Pick $\tau_H$ so that $\sum_{\ell\in W}\|m_{\ell_t}^{(\ell)}\|_0 \approx \sum_{\ell\in W}|\mathcal{N}_{\ell_t}|$, i.e., the baseline edits approximately the same number of coordinates (neurons) as FOXP2 edits (SAE features) within the deployed window.

\textbf{(2) FLOP/memory-matched.}
Because FOXP2 is a sparse SAE-space edit while LAPE is a masked FFN edit, raw count matching can still misrepresent compute. We additionally report a deployment-oriented match: the baseline is constrained so its additional operations are within the same overhead budget used for FOXP2 (measured as extra elementwise ops per token and any auxiliary buffers).

\vspace{4pt}
\paragraph{D. Scoring the baseline with FOXP2 metrics (what goes in Table~\ref{tab:foxp2_perf_utility}).}
Once instantiated, LAPE steering is scored identically to FOXP2:
(i) \textbf{token-mass channel} $\Delta_{\mathrm{mass}}$ at early horizon $T=\{1,2,3\}$,
(ii) \textbf{script-agnostic} $\Delta_{\mathrm{lid}}$ (Hindi + Spanish),
(iii) composed decision scores (\textsc{DefaultHi}/\textsc{DefaultEs}),
(iv) \textbf{leakage} Hi$\rightarrow$Es, and
(v) \textbf{utility} $\Delta S$ with bootstrap CIs. Critically, this answers the reviewer’s core concern: LAPE is evaluated on \emph{defaultness} (our construct), not just language accuracy.

\vspace{4pt}
\paragraph{E. Why this comparison is structurally informative (even before numbers).}
This baseline isolates a precise gap between the two lines of work. LAPE identifies \emph{selective units} via activation statistics; it does not explicitly enforce \textbf{(a)} low-rank actuator geometry, \textbf{(b)} bootstrap subspace stability across prompt families, or \textbf{(c)} bipolar mass transfer (target push + negative English suppression). As a result, when scored under our harness, LAPE may (depending on model/language) exhibit one or more characteristic failure modes: (i) gains that are less concentrated at early horizons (defaultness is not directly optimized), (ii) higher leakage (steering spills across languages), or (iii) larger $\Delta S$ penalties for comparable defaultness gain. FOXP2 is designed to address exactly these issues by producing a reproducible control artifact $(\mathcal{N},\mathcal{S},\mathcal{W})$ and selecting operating points under explicit guardrails.


\section{FOXP2: Full Algorithmic Specification}
\label{sec:appendix_foxp2_algo}

This section specifies \textbf{Neural FOXP2} as a complete three-stage pipeline that discovers $(\mathcal{N}_{\ell_t},\mathcal{S},\mathcal{W})$ and yields a deterministic inference-time edit rule with explicit \textbf{guardrails}, \textbf{stability gates}, and \textbf{matched-controls} that rule out degenerate ``generic logit reshaping.'' FOXP2 targets \textbf{defaultness as early commitment}: it aims to shift the model's \emph{initial} language choice (first $T$ tokens) while keeping semantics and non-target language behavior stable (Appendix \S\ref{sec:appendix_metrics}). The full pipeline is:
\[
\boxed{\textbf{FOXP2}:\;\;\big(\theta,\{\mathrm{SAE}^{(\ell)}\}_{\ell=1}^L,\mathcal{D}_{\mathrm{disc}},\ell_t\big)\;\longrightarrow\;\big(\mathcal{N}_{\ell_t},\{\mathcal{S}_{\ell_t}^{(\ell)}\}_{\ell\in\mathcal{W}},\mathcal{W},\texttt{EditRule}_{\ell_t}\big)}
\]
We present: \textbf{(I) Localize} (discover sparse coordinates $\mathcal{N}_{\ell_t}$), \textbf{(II) Discover} (low-rank subspace $\mathcal{S}_{\ell_t}^{(\ell)}$ and intervention window $\mathcal{W}$), and \textbf{(III) Steer} (signed sparse edit with drift-regularized negative English suppression), plus explicit \textbf{sanity checks} and \textbf{pass/fail criteria}.

\subsection*{D.\,X.0 Preliminaries: what FOXP2 edits, and why SAE feature space}
\label{sec:appendix_foxp2_prelim}

\paragraph{Model states and decoding steps.}
Fix a decoder-only transformer with $L$ layers. Given a prompt $x$, decoding step $t$ produces a prefix $y_{<t}$ and an internal residual-stream hidden state $h^{(\ell)}_\theta(x,t)\in\mathbb{R}^d$ at layer $\ell$. The next-token distribution is $p_\theta(\cdot\mid x,y_{<t})$.

\paragraph{SAE feature coordinates.}
For each layer $\ell$ we assume a frozen SAE with encoder $\mathrm{Enc}^{(\ell)}:\mathbb{R}^d\to\mathbb{R}^m$ and decoder $\mathrm{Dec}^{(\ell)}:\mathbb{R}^m\to\mathbb{R}^d$ (fixed after pretraining on $\theta$'s activations). Feature activations are:
\[
\boxed{z^{(\ell)}_\theta(x,t)=\mathrm{Enc}^{(\ell)}\!\big(h^{(\ell)}_\theta(x,t)\big)\in\mathbb{R}^m.}
\]
FOXP2 edits are applied in $z$-space and mapped back via $\mathrm{Dec}^{(\ell)}$:
\[
\boxed{\Delta h^{(\ell)}(x,t)=\mathrm{Dec}^{(\ell)}\!\big(\Delta z^{(\ell)}(x,t)\big),\qquad h^{(\ell)}_{\theta'}(x,t)=h^{(\ell)}_\theta(x,t)+\Delta h^{(\ell)}(x,t).}
\]
This induces structured edits that are (i) coordinate-localizable (sparse supports), (ii) low-rank-compressible (subspaces), and (iii) stable to resampling (bootstrap principal-angle tests).

\paragraph{Meaning units (paired semantic contexts).}
FOXP2 requires matched semantic content so that discovery isolates a language-control mechanism rather than topic/format effects. A \textbf{meaning unit} is a prompt family where task content is fixed but the \emph{language condition} differs minimally. For each semantic item $x$, we create two conditions: $\mathrm{cond}=\ell_t$ (target-language cue) and $\mathrm{cond}=\mathrm{en}$ (English cue), holding the remainder identical. We denote paired runs by $(x,\ell_t)$ and $(x,\mathrm{en})$. Extraction is done for a short horizon $T_0$ (default $T_0\in\{1,2,3\}$), because defaultness is an early commitment phenomenon.

\paragraph{Discovery vs evaluation splits.}
$\mathcal{D}_{\mathrm{disc}}$ is used only to discover $(\mathcal{N},\mathcal{S},\mathcal{W})$ and tune the operating point $\lambda^\star$; all reporting uses held-out $\mathcal{D}_{\mathrm{eval}}$ (Appendix \S\ref{sec:appendix_data}). Bootstrap for CIs resamples prompts, not tokens.

\paragraph{Two-channel construct validity (hard requirement).}
FOXP2 claims are credited only when (i) the mechanistic token-mass channel and (ii) a script-agnostic LID validator agree in direction under fixed decoding (Appendix \S\ref{sec:appendix_metrics}). Mass-only improvements with flat/negative LID are treated as script/token-set artifacts.

\subsection*{D.\,X.1 Step I: Localize (discover sparse support $\mathcal{N}_{\ell_t}$)}
\label{sec:appendix_step1}

Step I finds a compact set of SAE coordinates that are (a) \textbf{target-selective} under matched meaning units and (b) \textbf{causally useful} for early-horizon defaultness. The output is a deterministic sparse support $\mathcal{N}_{\ell_t}\subseteq\{(\ell,i):\ell\in[L], i\in[m]\}$.

\subsubsection*{D.\,X.1.1 Candidate coordinate space and paired differences}

Fix candidate layers $\mathcal{L}_{\mathrm{cand}}$ (wide by default) and horizon $T_0$. For each $x\in\mathcal{D}_{\mathrm{disc}}$ and $t\le T_0$, compute SAE activations under both conditions:
\[
z^{(\ell)}_\theta(x,t\mid \ell_t),\qquad z^{(\ell)}_\theta(x,t\mid \mathrm{en}).
\]
Define paired feature differences:
\[
\boxed{\Delta z^{(\ell)}(x,t)=z^{(\ell)}_\theta(x,t\mid \ell_t)-z^{(\ell)}_\theta(x,t\mid \mathrm{en}),\qquad \Delta z^{(\ell)}(x,t)\in\mathbb{R}^m.}
\]
For feature $i$, denote $\Delta z_i^{(\ell)}(x,t)$.

\subsubsection*{D.\,X.1.2 Target-selectivity scoring (correlational, but reproducible)}

We aggregate paired differences across prompts and early steps:
\[
\mu_i^{(\ell)}=\frac{1}{|\mathcal{D}_{\mathrm{disc}}|T_0}\sum_{x\in\mathcal{D}_{\mathrm{disc}}}\sum_{t=1}^{T_0}\Delta z_i^{(\ell)}(x,t),\qquad \nu_i^{(\ell)}=\sqrt{\frac{1}{|\mathcal{D}_{\mathrm{disc}}|T_0}\sum_{x,t}\big(\Delta z_i^{(\ell)}(x,t)-\mu_i^{(\ell)}\big)^2}+\epsilon.
\]
Define a deterministic selectivity statistic (a signal-to-noise ratio in paired-difference space):
\[
\boxed{\mathrm{Sel}_i^{(\ell)}=\frac{\mu_i^{(\ell)}}{\nu_i^{(\ell)}}.}
\]
\emph{Interpretation:} $\mathrm{Sel}_i^{(\ell)}$ is large if feature $i$ shifts consistently positive when swapping English-cue $\rightarrow$ target-cue under matched content.

\paragraph{Anti-confound filtering (shared/format sensitivity).}
FOXP2 is designed to avoid features that are primarily sensitive to punctuation, digits, JSON templates, or shared tokens. We thus compute (on a small probe set) a \textbf{format sensitivity} score by repeating the paired protocol across multiple format wrappers (plain, bullets, JSON) while keeping meaning fixed; features with high variance across formats relative to $\mu_i^{(\ell)}$ are flagged and optionally excluded from candidates before top-$k$ selection. This is a deterministic pre-filter (no extra learning).

\subsubsection*{D.\,X.1.3 Causal-lift scoring (do-proxy via micro-interventions)}

Selectivity can capture correlates; FOXP2 additionally tests causal usefulness by micro-intervention on individual features. Let $M^{\ell_t}_{\theta,T_0}(x)$ be the weighted token-mass metric for the target language at horizon $T_0$ (Appendix \S\ref{sec:appendix_metrics}). For each candidate coordinate $(\ell,i)$, define a micro-edit:
\[
z^{(\ell)}_{i}(x,t)\leftarrow z^{(\ell)}_{i}(x,t)+\delta \quad (\text{all other features unchanged}),
\]
applied at the hook site and only for $\ell\in\mathcal{L}_{\mathrm{cand}}$, $t\le T_0$. The causal lift is:
\[
\boxed{\mathrm{Lift}_i^{(\ell)}=\frac{1}{|\mathcal{D}_{\mathrm{cal}}|}\sum_{x\in\mathcal{D}_{\mathrm{cal}}}\Big(M^{\ell_t}_{\theta+\mathrm{do}(\ell,i,\delta),T_0}(x)-M^{\ell_t}_{\theta,T_0}(x)\Big),}
\]
where $\mathcal{D}_{\mathrm{cal}}\subset\mathcal{D}_{\mathrm{disc}}$ is a small calibration subset. This is a \emph{do-proxy}: we explicitly intervene on a coordinate and test the induced change in the mechanistic defaultness probe.

\paragraph{Why lift is measured on token-mass, not LID.}
Lift is computed on the mass channel because LID is detector-based and less sensitive at token-1--3. Construct validity is enforced later at the method level, not at single-feature lift level.

\subsubsection*{D.\,X.1.4 Composite ranking and deterministic selection of $\mathcal{N}_{\ell_t}$}

We combine selectivity and lift into a composite score:
\[
\mathrm{Score}_i^{(\ell)}=\mathrm{Sel}_i^{(\ell)}+\alpha_{\mathrm{lift}}\cdot \frac{\mathrm{Lift}_i^{(\ell)}}{\widehat{\sigma}_{\mathrm{Lift}}+\epsilon},
\]
where $\widehat{\sigma}_{\mathrm{Lift}}$ is the empirical standard deviation of lift over candidates. The sparse support is chosen by top-$k$ selection:
\[
\boxed{\mathcal{N}_{\ell_t}=\mathrm{TopK}\Big(\{(\ell,i):\ell\in\mathcal{L}_{\mathrm{cand}}, i\in[m]\},\mathrm{Score},k\Big).}
\]
We optionally enforce per-layer caps ($\le k_{\max}$ per layer) to prevent single-layer collapse and to keep the support distributed.

\subsubsection*{D.\,X.1.5 Bootstrap reproducibility and sign-stability gates}

For bootstrap $b=1,\dots,B$, resample $\mathcal{D}_{\mathrm{disc}}$ with replacement and recompute $\mathcal{N}_{\ell_t}^{(b)}$. We report overlap stability:
\[
\boxed{\mathrm{Stab}_{\mathcal{N}}=\frac{1}{B}\sum_{b=1}^{B}\frac{\big|\mathcal{N}_{\ell_t}^{(b)}\cap \mathcal{N}_{\ell_t}^{(0)}\big|}{|\mathcal{N}_{\ell_t}^{(0)}|}.}
\]
Additionally, for each selected coordinate $(\ell,i)$ we compute \textbf{sign stability} $\Pr_b[\mu_i^{(\ell,b)}>0]$. Coordinates with sign stability below a fixed threshold (e.g., $<0.9$) are deterministically removed and replaced by the next-highest scoring candidates, preserving $|\mathcal{N}_{\ell_t}|=k$.

\subsubsection*{D.\,X.1.6 Necessity ablation (explicit, deployment-relevant)}

We run the “w/o $\mathcal{N}_{\ell_t}$” ablation by applying the full Step III edit rule but forcing $\Pi_{\mathcal{N}_{\ell_t}}=0$ (equivalently, $\mathcal{N}_{\ell_t}=\varnothing$). The resulting losses in $\textsc{DefaultHi}$, $\Delta_{\mathrm{mass}}^{\mathrm{hi}}$, and $\Delta_{\mathrm{lid}}^{\mathrm{hi}}$ quantify necessity (Table~\ref{tab:app_necessity}).

\subsection*{D.\,X.2 Step II: Discover low-rank directions and window (construct $\mathcal{S}^{(\ell)}_{\ell_t}$ and $\mathcal{W}$)}
\label{sec:appendix_step2}

Step II compresses the localized sparse mechanism into a low-rank structure and localizes where it is \emph{causally effective}. It outputs (i) per-layer target subspaces $\{\mathcal{S}^{(\ell)}_{\ell_t}\}$ and (ii) a contiguous intervention window $\mathcal{W}$.

\subsubsection*{D.\,X.2.1 Restrict paired differences to the localized support}

Define a coordinate projection $\Pi_{\mathcal{N}_{\ell_t}}$ in feature space (zeros outside $\mathcal{N}_{\ell_t}$). For each layer $\ell$ and each paired sample $(x,t)$, define:
\[
\boxed{\Delta \tilde z^{(\ell)}(x,t)=\Pi_{\mathcal{N}_{\ell_t}}\Delta z^{(\ell)}(x,t)\in\mathbb{R}^m.}
\]
Stack these vectors to form a difference matrix:
\[
\boxed{\Delta Z^{(\ell)}=\begin{bmatrix}(\Delta \tilde z^{(\ell)}(x_1,1))^\top\\ \vdots\\ (\Delta \tilde z^{(\ell)}(x_n,T_0))^\top\end{bmatrix}\in\mathbb{R}^{(nT_0)\times m}.}
\]
\emph{Interpretation:} rows are matched meaning-unit displacements in feature space, supported only on localized coordinates.

\subsubsection*{D.\,X.2.2 SVD, dominant directions, and a per-layer steering subspace}

Compute SVD:
\[
\Delta Z^{(\ell)}=U^{(\ell)}\Sigma^{(\ell)}(V^{(\ell)})^\top,\qquad \Sigma^{(\ell)}=\mathrm{diag}(\sigma_1^{(\ell)}\ge\sigma_2^{(\ell)}\ge\cdots).
\]
Define the rank-$r$ subspace:
\[
\boxed{\mathcal{S}^{(\ell)}_{\ell_t}=\mathrm{span}\big(V^{(\ell)}_{:,1:r}\big)\subset\mathbb{R}^m.}
\]
This is the low-rank structure FOXP2 uses: edits will be projected onto $\mathcal{S}^{(\ell)}_{\ell_t}$.

\subsubsection*{D.\,X.2.3 Effective rank and eigengap (quantifying ``sparse + low-rank'')}

Let $p_j^{(\ell)}=\sigma_j^{(\ell)}/\sum_k\sigma_k^{(\ell)}$. The effective rank is:
\[
\boxed{r_{\mathrm{eff}}^{(\ell)}=\exp\!\Big(-\sum_j p_j^{(\ell)}\log p_j^{(\ell)}\Big).}
\]
We also report an eigengap ratio at $r$:
\[
\boxed{\mathrm{Gap}_r^{(\ell)}=\frac{\sigma_r^{(\ell)}}{\sigma_{r+1}^{(\ell)}+\epsilon}.}
\]
A FOXP2-consistent mechanism exhibits small $r_{\mathrm{eff}}^{(\ell)}$ and a large $\mathrm{Gap}_r^{(\ell)}$ in the effective window.

\subsubsection*{D.\,X.2.4 Bootstrap principal-angle stability of the subspace}

For each bootstrap $b$, recompute $\Delta Z^{(\ell,b)}$ and its top-$r$ subspace $\mathcal{S}^{(\ell,b)}_{\ell_t}$. Measure agreement with a reference $\mathcal{S}^{(\ell,0)}_{\ell_t}$ via principal angles $\angle_i$:
\[
\boxed{\mathrm{Stab}_{\mathcal{S}}^{(\ell)}=1-\frac{1}{B}\sum_{b=1}^B\frac{1}{r}\sum_{i=1}^r \sin\!\big(\angle_i(\mathcal{S}^{(\ell,b)}_{\ell_t},\mathcal{S}^{(\ell,0)}_{\ell_t})\big).}
\]
Layers with low $\mathrm{Stab}_{\mathcal{S}}^{(\ell)}$ are not eligible for $\mathcal{W}$.

\subsubsection*{D.\,X.2.5 Window selection (where edits are causally effective)}

FOXP2 selects a contiguous window $\mathcal{W}$ by maximizing a dev objective that balances gain, leakage, drift, and utility. For a candidate window $W=[a,b]$, define the dev objective:
\[
\boxed{\mathcal{J}(W)=\mathrm{Gain}^{\mathrm{hi}}(W)-\alpha_{\mathrm{leak}}\mathrm{Leak}_{\mathrm{hi}\rightarrow\mathrm{es}}(W)-\alpha_{\mathrm{kl}}\mathbb{E}[\mathrm{KL}_{T_0}](W)-\alpha_{\mathrm{util}}|\Delta S(W)|-\alpha_{\mathrm{stab}}\sum_{\ell\in W}\big(1-\mathrm{Stab}_{\mathcal{S}}^{(\ell)}\big).}
\]
Here $\mathrm{Gain}^{\mathrm{hi}}(W)$ can be computed using \textsc{DefaultHi} or the continuous mass channel; leakage/drift/utility are exactly those used in the appendix tables. We search over a candidate set $\mathcal{C}$ of windows (width constraints and centers) and select:
\[
\boxed{\mathcal{W}=\arg\max_{W\in\mathcal{C}}\mathcal{J}(W)\quad\text{s.t. guardrails pass on dev.}}
\]
This explicitly separates (i) \emph{signal localization} (Step I/II) from (ii) \emph{causal edit effectiveness} (Step III measured during window search). Figure~\ref{fig:app_window_sweep} reports the corresponding sweep.

\subsection*{D.\,X.3 Step III: Steer (signed sparse edit with drift-regularized negative English suppression)}
\label{sec:appendix_step3}

Step III defines the actual inference-time intervention. It is \textbf{sparse} (restricted to $\mathcal{N}_{\ell_t}$), \textbf{low-rank} (projected to $\mathcal{S}^{(\ell)}_{\ell_t}$), \textbf{windowed} (applied only for $\ell\in\mathcal{W}$), and \textbf{drift-controlled} (KL trust region + matched controls). We give the exact edit equations, the choice of the English suppression coefficient, and the operational point selection.

\subsubsection*{D.\,X.3.1 Projectors in feature space}

Let $V_r^{(\ell)}=V^{(\ell)}_{:,1:r}\in\mathbb{R}^{m\times r}$. The orthogonal projector onto $\mathcal{S}^{(\ell)}_{\ell_t}$ is $P_{\mathcal{S}}^{(\ell)}=V_r^{(\ell)}(V_r^{(\ell)})^\top$. Define the sparse mask $\Pi_{\mathcal{N}_{\ell_t}}$ (diagonal 0/1 in coordinate space). The composed projector is:
\[
\boxed{\mathcal{P}^{(\ell)}_{\ell_t}=\Pi_{\mathcal{N}_{\ell_t}}\,P_{\mathcal{S}}^{(\ell)}\,\Pi_{\mathcal{N}_{\ell_t}}.}
\]
Edits are applied only through $\mathcal{P}^{(\ell)}_{\ell_t}$, enforcing the ``sparse + low-rank'' structure in the actual intervention.

\subsubsection*{D.\,X.3.2 Data-derived target direction (per layer)}

Define a mean displacement direction in the discovered subspace:
\[
\boxed{v_{\ell_t}^{(\ell)}=\mathrm{Norm}\!\left(\frac{1}{|\mathcal{D}_{\mathrm{disc}}|T_0}\sum_{x,t}\mathcal{P}^{(\ell)}_{\ell_t}\big(z^{(\ell)}_\theta(x,t\mid \ell_t)-z^{(\ell)}_\theta(x,t\mid \mathrm{en})\big)\right),}
\]
where $\mathrm{Norm}(u)=u/(\|u\|_2+\epsilon)$. This ensures $\lambda$ is comparable across layers.

\subsubsection*{D.\,X.3.3 Negative English suppression as a drift-regularized compensation (not generic flattening)}

A naive ``push Hindi'' edit can inadvertently increase entropy or broadly reshape the distribution, creating spurious token-mass gains or inflating non-target languages. FOXP2 therefore includes an explicit, structured suppression term that is (i) in the same coordinate support and (ii) in the same low-rank subspace, so it does not become a global logit smoothing trick. Define the corresponding English-associated direction:
\[
\boxed{v_{\mathrm{en}}^{(\ell)}=\mathrm{Norm}\!\left(\frac{1}{|\mathcal{D}_{\mathrm{disc}}|T_0}\sum_{x,t}\mathcal{P}^{(\ell)}_{\ell_t}\big(z^{(\ell)}_\theta(x,t\mid \mathrm{en})-z^{(\ell)}_\theta(x,t\mid \ell_t)\big)\right).}
\]
The per-layer feature edit is:
\[
\boxed{\Delta z^{(\ell)}_{\mathrm{FOXP2}}(x,t;\lambda)=\lambda\,v_{\ell_t}^{(\ell)}-\kappa(\lambda)\,v_{\mathrm{en}}^{(\ell)},\qquad \ell\in\mathcal{W}.}
\]
The key is how $\kappa(\lambda)$ is chosen: FOXP2 chooses $\kappa(\lambda)$ to \textbf{minimize drift} subject to achieving gain, rather than letting suppression freely distort logits.

\paragraph{Constrained choice of $\kappa(\lambda)$ (first-order view).}
Let $\mathrm{KL}_{T_0}(x;\lambda,\kappa)$ be the early-horizon KL drift induced by the edit (Appendix \S\ref{sec:appendix_metrics}). FOXP2 chooses:
\[
\boxed{\kappa(\lambda)=\arg\min_{\kappa\ge 0}\;\mathbb{E}_{x\sim\mathcal{D}_{\mathrm{dev}}}\big[\mathrm{KL}_{T_0}(x;\lambda,\kappa)\big]\;\;\text{s.t.}\;\;\mathbb{E}_{x}[\Delta M^{\ell_t}_{T_0}(x;\lambda,\kappa)]\ge \gamma,}
\]
where $\gamma$ is a fixed dev target (e.g., achieve a specified fraction of the final gain). This makes suppression a \textbf{trust-region compensation}: we take the smallest drift solution that still attains the required mechanistic improvement.

\paragraph{Why this avoids ``generic logit flattening''.}
Global flattening would increase entropy and often raise token-mass on shared tokens; FOXP2 explicitly (i) constrains KL drift and (ii) restricts edit directions to the discovered language-structured subspace and sparse coordinates, and (iii) is audited by entropy/KL-matched controls (Table~\ref{tab:app_entropy_matched}). If matched controls cannot replicate DefaultHi under the same drift, the gain is not explainable by generic reshaping.

\subsubsection*{D.\,X.3.4 Exact intervention rule in the transformer}

For each decoding step $t$ and each layer $\ell\in\mathcal{W}$, apply:
\[
z^{(\ell)}_{\theta'}(x,t)=z^{(\ell)}_\theta(x,t)+\Delta z^{(\ell)}_{\mathrm{FOXP2}}(x,t;\lambda),
\qquad h^{(\ell)}_{\theta'}(x,t)=h^{(\ell)}_\theta(x,t)+\mathrm{Dec}^{(\ell)}\!\big(\Delta z^{(\ell)}_{\mathrm{FOXP2}}(x,t;\lambda)\big).
\]
Hook placement is fixed (documented once; same across all runs). No parameters of $\theta$ are updated; FOXP2 is inference-time only.

\subsubsection*{D.\,X.3.5 Operating-point selection ($\lambda^\star$) and controllability curves}

FOXP2 is evaluated as a control curve over $\lambda\in\Lambda$ (grid fixed in advance). The Hindi-tuned operating point is:
\[
\boxed{\lambda^\star_{\mathrm{hi}}=\arg\max_{\lambda\in\Lambda}\;\mathrm{Gain}^{\mathrm{hi}}(\lambda)\;\;\text{s.t. guardrails pass and }|\Delta S(\lambda)|\le\epsilon_{\mathrm{util}}.}
\]
Reporting uses $\lambda^\star_{\mathrm{hi}}$ for Hindi gains and for non-target regression checks (Spanish leakage, guardrails, KL).

\subsubsection*{D.\,X.3.6 Sufficiency and matched-control protocols}

\paragraph{Sufficiency.}
We run minimal configurations to operationalize the ``sparse + low-rank'' claim: (i) top-$k$ sparse-only (no low-rank projection), (ii) rank-$r$ only (dense over features), (iii) sparse+rank-$r$ (FOXP2-like). We report the smallest setting achieving $\ge 80\%$ and $\ge 90\%$ of full DefaultHi gain under guardrails (Table~\ref{tab:app_sufficiency}).

\paragraph{Entropy/KL-matched controls.}
We include two controls tuned to match FOXP2's distributional drift: (i) a decoding-level entropy match (temperature/noise) matching mean early-step entropy shift, and (ii) an activation-level isotropic perturbation matching $\mathbb{E}[\mathrm{KL}_{T_0}]$ (Table~\ref{tab:app_entropy_matched}). If FOXP2 uniquely retains DefaultHi under matched drift, this supports a language-structured mechanism rather than generic reshaping.

\subsection*{D.\,X.4 First-order mechanistic interpretation (why the edit changes early language mass)}
\label{sec:appendix_foxp2_math}

This subsection provides a step-by-step mathematical view connecting feature-space edits to early-horizon language token-mass changes, clarifying what FOXP2 is (and is not) doing.

\subsubsection*{D.\,X.4.1 From feature edits to logit changes (Jacobian view)}

Let $\ell$ be an edited layer and consider a single step $t$. The edit adds $\Delta h^{(\ell)}=\mathrm{Dec}^{(\ell)}(\Delta z^{(\ell)})$ to the residual stream. Let $g$ denote the (deterministic) mapping from $h^{(\ell)}$ to the final logits at step $t$, i.e., $\mathrm{logits}(x,t)=g(h^{(\ell)}(x,t))$ (this includes subsequent layers and the output head). For small edits, first-order Taylor expansion gives:
\[
\boxed{\Delta \mathrm{logits}(x,t)\approx J^{(\ell)}(x,t)\,\Delta h^{(\ell)}(x,t)=J^{(\ell)}(x,t)\,\mathrm{Dec}^{(\ell)}\!\big(\Delta z^{(\ell)}(x,t)\big),}
\]
where $J^{(\ell)}(x,t)=\partial g/\partial h^{(\ell)}$ is the step- and prompt-dependent Jacobian.

\subsubsection*{D.\,X.4.2 Expected change in language mass}

Recall the token-mass metric (Appendix \S\ref{sec:appendix_metrics}):
\[
M^{\ell_t}_\theta(x,t)=\sum_{u\in V_{\ell_t}} w(u)\,p_\theta(u\mid x,y_{<t}),\qquad M^{\ell_t}_{\theta,T}(x)=\frac{1}{T}\sum_{t=1}^T M^{\ell_t}_\theta(x,t).
\]
Let $p(u)=\mathrm{softmax}(\mathrm{logits})_u$. Under a small logit change $\Delta \mathrm{logits}$, the first-order change in probability is:
\[
\Delta p(u)\approx \sum_v \frac{\partial p(u)}{\partial \mathrm{logits}_v}\Delta \mathrm{logits}_v = \sum_v p(u)\big(\mathbf{1}[u=v]-p(v)\big)\Delta \mathrm{logits}_v.
\]
Thus the first-order change in mass is:
\[
\boxed{\Delta M^{\ell_t}(x,t)\approx \sum_{u\in V_{\ell_t}}w(u)\sum_v p(u)\big(\mathbf{1}[u=v]-p(v)\big)\Delta \mathrm{logits}_v.}
\]
Substituting $\Delta\mathrm{logits}\approx J^{(\ell)}\mathrm{Dec}^{(\ell)}(\Delta z^{(\ell)})$ shows that the mass gain is driven by directions in feature space that (through $\mathrm{Dec}^{(\ell)}$ and downstream Jacobians) push probability mass toward diagnostic tokens in $V_{\ell_t}$ and away from competing tokens.

\subsubsection*{D.\,X.4.3 Why sparsity and low-rank matter in this view}

Write the edit as $\Delta z^{(\ell)}=\mathcal{P}^{(\ell)}_{\ell_t}u^{(\ell)}$ for some $u^{(\ell)}$ (FOXP2 chooses $u^{(\ell)}=\lambda v_{\ell_t}^{(\ell)}-\kappa v_{\mathrm{en}}^{(\ell)}$). Then only coordinates in $\mathcal{N}_{\ell_t}$ can contribute (sparsity), and only directions in $\mathcal{S}^{(\ell)}_{\ell_t}$ can contribute (low-rank). This sharply limits the set of feasible logit shifts, which makes (i) the mechanism auditable and (ii) degenerate entropy-inflation solutions harder: generic reshaping would require directions outside the discovered subspace or outside sparse coordinates, which FOXP2 forbids.

\subsubsection*{D.\,X.4.4 Why the window matters (localizing causal effectiveness)}

\begin{figure*}[ht!]
\centering
\includegraphics[width=0.985\textwidth]{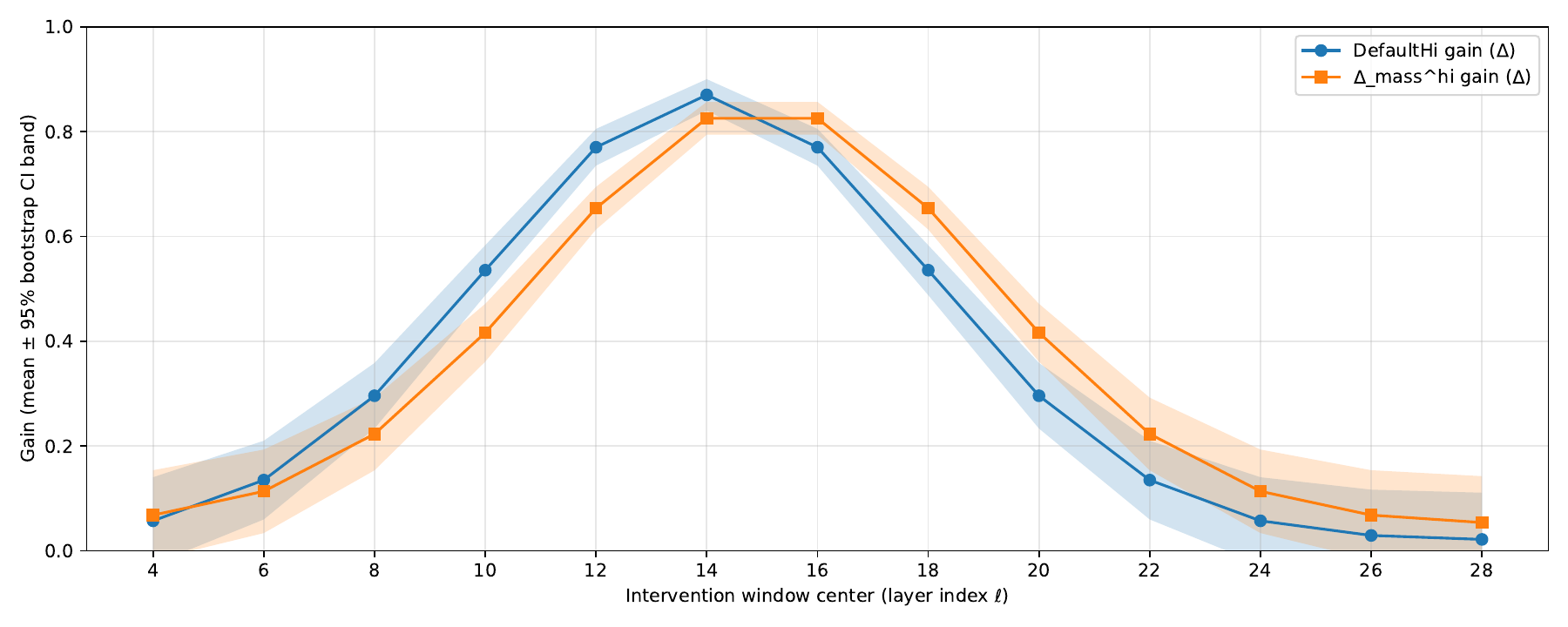}
\caption{\textbf{Window sweep over layers.} We shift the intervention window across depth and plot defaultness gain (\textsc{DefaultHi} and/or $\Delta_{\mathrm{mass}}^{\mathrm{hi}}$) versus window center. This figure localizes where FOXP2 is causally effective, separating “where the signal lives” from “where the output is produced.”}
\label{fig:app_window_sweep}
\end{figure*}

Even if a language signal exists at many layers, the Jacobian $J^{(\ell)}(x,t)$ determines how an edit at layer $\ell$ affects logits. Window selection (Step II) effectively chooses where the product $J^{(\ell)}\mathrm{Dec}^{(\ell)}$ yields the best gain-to-drift ratio under guardrails. This is why FOXP2 distinguishes ``where the signal lives'' from ``where edits are effective,'' and why we include the depth sweep (Figure~\ref{fig:app_window_sweep}).

\subsection*{D.\,X.5 Hyperparameters, defaults, and deterministic choices}
\label{sec:appendix_foxp2_hparams}

We report all FOXP2 hyperparameters as fixed protocol knobs; all are selected on dev once and then frozen. Typical defaults (replace with your actual values where applicable): candidate layers $\mathcal{L}_{\mathrm{cand}}$ = all layers or a broad mid-band; discovery horizon $T_0\in\{1,2,3\}$; sparse size $k$ (top-$k$ features); rank $r$ (either fixed or selected via eigengap threshold); bootstrap count $B$; micro-intervention step $\delta$; lift weight $\alpha_{\mathrm{lift}}$; window candidate set $\mathcal{C}$ (centers and widths); guardrail thresholds $\epsilon_{\mathrm{es}},\epsilon_{\mathrm{KL}}$; semantic invariance tolerance $\epsilon_{\mathrm{util}}$; edit grid $\Lambda$; and gain target $\gamma$ for selecting $\kappa(\lambda)$.

\paragraph{Determinism and tie-breaking.}
All selections are deterministic given the fixed random seed used only for data shuffling/bootstraps: Top-$k$ ties are broken by (i) larger lift, then (ii) larger selectivity, then (iii) smaller index.

\subsection*{D.\,X.6 Pass/fail sanity checks (what must hold for FOXP2 to be ``valid'')}
\label{sec:appendix_foxp2_checks}

FOXP2 is considered \textbf{valid} only if the following checks pass at the Hindi-tuned operating point $\lambda^\star_{\mathrm{hi}}$ (and analogously for Spanish-tuned runs where applicable).

\begin{figure*}[ht!]
\centering
\includegraphics[width=0.985\textwidth]{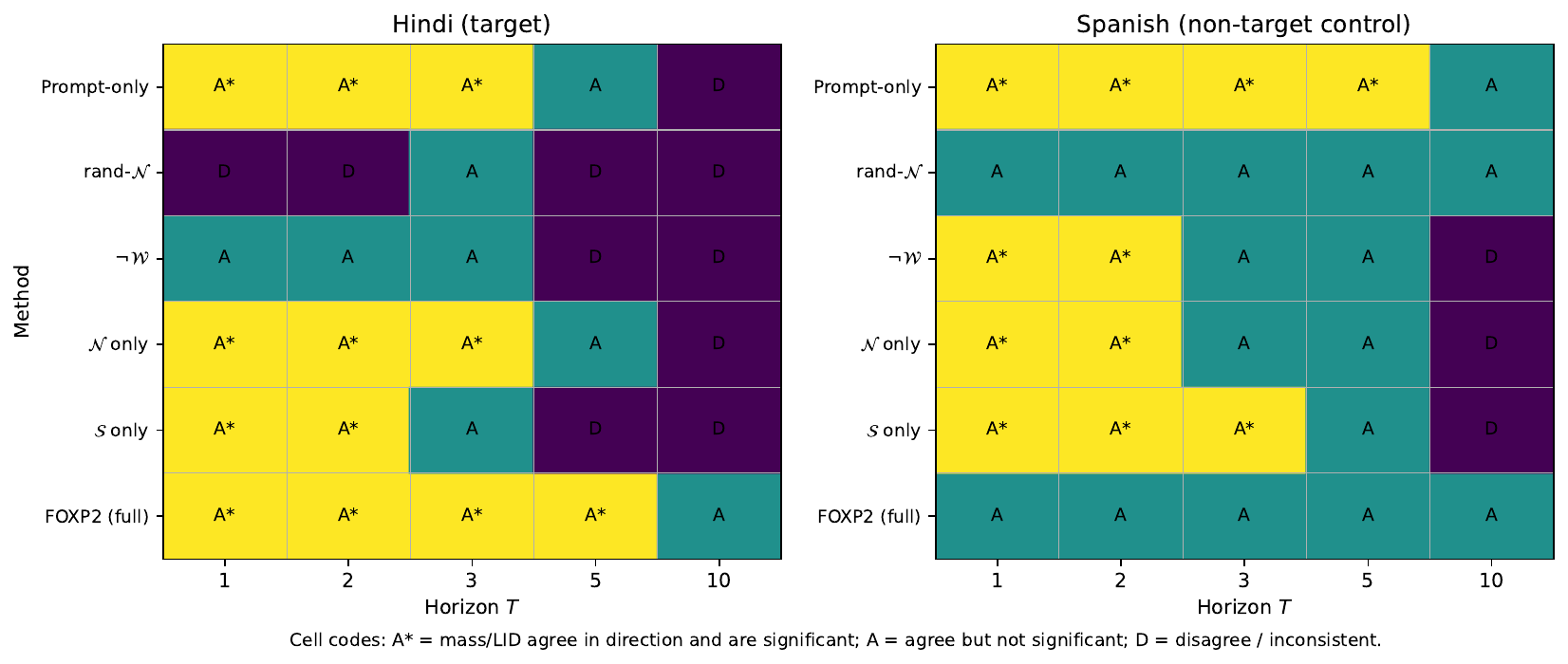}
\caption{\textbf{Cross-metric agreement map across horizons.} Cells indicate whether the mass channel and LID validator agree in direction and significance at each $T$ (Hindi and Spanish shown side-by-side). This provides a compact construct-validity diagnostic: robust defaultness gains should occupy a contiguous block at small horizons without collapsing into mass-only artifacts at larger $T$.}
\label{fig:app_horizon_agree}
\end{figure*}

\paragraph{(C1) Construct validity across horizons.}
For $T\in\{1,2,3,5,10\}$, the mass channel and LID channel should agree in direction for Hindi and should not inflate Spanish under the Hindi-tuned edit (Figures~\ref{fig:app_horizon_hi}--\ref{fig:app_horizon_agree}).

\paragraph{(C2) Token-set robustness (script-only vs diagnostic vs transliteration-aware).}
Gains should persist under diagnostic and transliteration-aware token sets and should not be explainable by script-only gains (Figure~\ref{fig:app_tokset_bands}), with shared-token inflation bounded (Table~\ref{tab:app_shared_inflation}).

\paragraph{(C3) Bootstrap stability.}
Sparse support overlap stability $\mathrm{Stab}_{\mathcal{N}}$ and subspace stability $\mathrm{Stab}^{(\ell)}_{\mathcal{S}}$ must exceed fixed thresholds; output-level CI widths should be tight enough for stable reporting (Table~\ref{tab:app_full_metrics}).

\paragraph{(C4) Guardrails (deployment eligibility).}
Spanish regression bound and KL trust region must pass at high rate (Table~\ref{tab:app_guardrails}); configurations that violate guardrails are reported but not eligible for deployment-safe claims.

\paragraph{(C5) Non-regression on task-grounded semantics.}
Per-task deltas $\Delta s_j$ must lie within the semantic-invariance tolerance band (Table~\ref{tab:app_task_decomp}); no aggregate metric may hide a regression concentrated on a subset of tasks.

\paragraph{(C6) Matched-control non-degeneracy.}
Entropy/KL-matched controls should not replicate FOXP2's DefaultHi under matched drift (Table~\ref{tab:app_entropy_matched}); otherwise gains may be attributable to generic distributional reshaping.

\paragraph{(C7) Necessity/sufficiency.}
Remove-one necessity ablations should show meaningful loss relative to full FOXP2 (Table~\ref{tab:app_necessity}), and sufficiency sweeps should reveal a compact configuration achieving most of the gain under guardrails (Table~\ref{tab:app_sufficiency}).

\subsection*{D.\,X.7 Practical failure modes (what FOXP2 can reveal)}
\label{sec:appendix_foxp2_failmodes}

FOXP2 is designed to surface, not hide, common defaultness confounds: (i) \textbf{transliteration wins} (Hindi-in-Latin tokens inflate mass with weak LID), (ii) \textbf{NE-dominated early tokens} (single entity causes apparent defaultness), (iii) \textbf{format-first artifacts} (JSON keys/bullets), (iv) \textbf{mixed-script headers} (Hindi headers with English bodies), (v) \textbf{loanword glossing} (Hindi token + English explanation). These appear as disagreement between mass and LID, sensitivity to token-set variants, abrupt collapse under token dropout, and/or leakage inflation; we document representative examples in Table~\ref{tab:app_failure_cases}.

\paragraph{Final deliverable of this section.}
Given this specification, the FOXP2 pipeline is fully reconstructible: Step I yields $\mathcal{N}_{\ell_t}$ with stability gates; Step II yields $\{\mathcal{S}^{(\ell)}_{\ell_t}\}$ and $\mathcal{W}$ with low-rank and principal-angle diagnostics; Step III yields a signed sparse edit rule with drift-regularized English suppression, operating-point selection, and matched-control baselines. All reported results in Table~\ref{tab:foxp2_perf_utility} and the extended appendix tables/figures are computed from these fixed objects under identical decoding and guardrail protocols.

\section{Datasets and Experimental Details}
\label{sec:appendix_data}

This section documents the complete experimental harness used to evaluate FOXP2 and to compute the causal artifacts $(\mathcal{N}_{\ell_t}, \mathcal{S}^{(\ell)}_{\ell_t}, \mathcal{W})$. Our goal is \textbf{auditable reproducibility}: a reader should be able to (i) re-run defaultness evaluation, (ii) re-compute $\Delta Z$ and all stability statistics, and (iii) reproduce the reported tradeoffs under identical randomness controls.

\begin{table*}[ht!]
\centering
\caption{\textbf{Evaluation suite for Hindi, Spanish, and English.}
We evaluate \textbf{Hindi} as the \emph{target} language, \textbf{Spanish} as a \emph{non-target regression control}, and \textbf{English} as the \emph{anchor/default} reference.
Tasks are matched across languages (same task families and comparable protocols), and we hold the evaluation policy fixed—prompt templates, formatting, decoding, and stopping rules—so observed changes reflect \textbf{language-specific defaultness shifts} rather than confounds from inference or presentation.
Exact dataset versions, splits, preprocessing.}
\label{tab:datasets_suites_main}
\small
\setlength{\tabcolsep}{6pt}
\renewcommand{\arraystretch}{1.15}
\resizebox{\textwidth}{!}{%
\scriptsize
\begin{tabular}{p{3.6cm} p{1.05cm} p{0.9cm} p{1.15cm} p{1.4cm} p{5.2cm}}
\hline
\textbf{Suite / Dataset} &
\textbf{Task} &
\textbf{Lang} &
\textbf{Script} &
\textbf{Source} &
\textbf{URL (download)} \\
\hline

\multicolumn{6}{l}{\textbf{Hindi (target language)}} \\
FLORES-200 (\texttt{hi\_Deva}) &
MT &
hi &
Deva &
\citep{flores200} &
\url{https://huggingface.co/datasets/facebook/flores} \\
XQuAD (\texttt{hi}) &
QA &
hi &
Deva &
\citep{xquad} &
\url{https://huggingface.co/datasets/google/xquad} \\
XNLI (\texttt{hi}) &
NLI &
hi &
Deva &
\citep{xnli} &
\url{https://huggingface.co/datasets/facebook/xnli} \\
XL-Sum (\texttt{hi}) &
Summ &
hi &
Deva &
\citep{xlsum} &
\url{https://huggingface.co/datasets/csebuetnlp/xlsum} \\
\hline

\multicolumn{6}{l}{\textbf{Spanish (non-target regression control)}} \\
FLORES-200 (\texttt{es\_Latn}) &
MT &
es &
Latin &
\citep{flores200} &
\url{https://huggingface.co/datasets/facebook/flores} \\
XQuAD (\texttt{es}) &
QA &
es &
Latin &
\citep{xquad} &
\url{https://huggingface.co/datasets/google/xquad} \\
XNLI (\texttt{es}) &
NLI &
es &
Latin &
\citep{xnli} &
\url{https://huggingface.co/datasets/facebook/xnli} \\
XL-Sum (\texttt{es}) &
Summ &
es &
Latin &
\citep{xlsum} &
\url{https://huggingface.co/datasets/csebuetnlp/xlsum} \\
\hline
\multicolumn{6}{l}{\textbf{English (reference)}} \\
SQuAD v1.1 (\texttt{en}) &
QA &
en &
Latin &
\citep{rajpurkar2016squad} &
\url{https://huggingface.co/datasets/rajpurkar/squad} \\
MNLI (GLUE; \texttt{en}) &
NLI &
en &
Latin &
\citep{williams2018mnli} &
\url{https://huggingface.co/datasets/glue} \\
CNN/DailyMail (\texttt{en}) &
Summ &
en &
Latin &
\citep{hermann2015cnndm} &
\url{https://huggingface.co/datasets/cnn_dailymail} \\
SST-2 (GLUE; \texttt{en}) &
Sent &
en &
Latin &
\citep{socher2013sst} &
\url{https://huggingface.co/datasets/glue} \\
\hline
\end{tabular}%
}
\vspace{0.3em}
\footnotesize
\textbf{Notes.} We use \textbf{XQuAD} (not TyDiQA) because XQuAD includes both \textbf{Hindi and Spanish} in a matched extractive-QA setup.
Appendix Table~\ref{tab:datasets_suites_appendix} reports exact splits, preprocessing, decoding settings, and all internal suites (neutral prompts, script-control sets, KL-guard, and safety sets).
\end{table*}

\vspace{4pt}
\paragraph{Reproducibility harness (FOXP2).}
We treat FOXP2 as a \textbf{mechanistic} intervention whose claims are only as strong as its experimental contract. This section specifies the \textbf{pinned identifiers}, \textbf{data construction}, \textbf{pairing logic} for $\Delta Z$, and \textbf{decoding/system controls} required to reproduce $(\mathcal{N},\mathcal{S},\mathcal{W})$ and all reported deltas without silent drift.

\vspace{6pt}
\paragraph{A. Checkpoints, tokenizers, and the hashing contract.}
All experiments are pinned to a \textbf{(checkpoint, tokenizer, hook-definition)} triple. Any change to \emph{any} component is treated as a \textbf{breaking change} and requires recomputation of $(\mathcal{N},\mathcal{S},\mathcal{W})$:
\begin{itemize}[leftmargin=*,itemsep=2pt]
  \item \textbf{Checkpoint.} Exact model identifier, revision/commit, weight format, and weight hash.
  \item \textbf{Tokenizer.} Tokenizer identifier and revision, plus a hash of \texttt{tokenizer.json} (or vocab+merges).
  \item \textbf{Hook definition.} Exact activation site(s) for extraction/editing (e.g., pre/post-attn residual, pre/post-MLP residual), including any normalization or pooling conventions.
\end{itemize}
\noindent\textbf{Why this matters.} FOXP2 is defined jointly in \textbf{token-id space} (language token sets for $\Delta_{\mathrm{mass}}$) and in a \textbf{feature basis} (SAE coordinates). Both are tokenizer- and hook-dependent. We therefore log immutable fingerprints:
\[
\texttt{hash}_{\text{tok}},\ \texttt{hash}_{\text{weights}},\ \texttt{hash}_{\text{hooks}},\ \texttt{hash}_{\text{sae}}.
\]

\vspace{6pt}
\paragraph{B. Prompt sources, suites, and splits.}
We evaluate defaultness on two prompt families: (i) a broad, \textbf{neutral} distribution $\mathcal{D}_{\mathrm{neutral}}$ designed to elicit the model’s \textbf{language prior} without explicit language directives, and (ii) an optional \textbf{safety/refusal} slice $\mathcal{D}_{\mathrm{safety}}$ used only for policy-regression checks.
\begin{itemize}[leftmargin=*,itemsep=2pt]
  \item \textbf{Neutral prompts $\mathcal{D}_{\mathrm{neutral}}$.} Instruction-light, content-open prompts stratified by intent class (QA-like, explanation, summarization-like, reasoning-like, dialogue turn). We avoid prompts dominated by named entities, locale-specific proper nouns, or formatting templates that implicitly cue a language.
  \item \textbf{Safety/refusal prompts $\mathcal{D}_{\mathrm{safety}}$ (optional).} Prompts with known allow/refuse expectation used to test refusal stability and language-dependent regressions. This slice is \textbf{disjoint} from $\mathcal{D}_{\mathrm{neutral}}$ to prevent refusal style from becoming a defaultness confound.
  \item \textbf{Script controls.} Because Hindi is Devanagari-centric while Spanish is Latin-script, we include explicit script/format controls (rendering variants, script-only proxies) to stress-test that gains are not punctuation/format artifacts; Spanish also functions as a built-in anti-script control.
\end{itemize}
\noindent\textbf{Splits.} We create non-overlapping splits:
\[
\mathcal{D}_{\mathrm{neutral}}=\mathcal{D}_{\mathrm{neutral}}^{\mathrm{train}}\cup\mathcal{D}_{\mathrm{neutral}}^{\mathrm{dev}}\cup\mathcal{D}_{\mathrm{neutral}}^{\mathrm{test}},\qquad
\mathcal{D}_{\mathrm{safety}}=\mathcal{D}_{\mathrm{safety}}^{\mathrm{dev}}\cup\mathcal{D}_{\mathrm{safety}}^{\mathrm{test}}.
\]
Only $\mathcal{D}_{\mathrm{neutral}}^{\mathrm{train}}$ is used to estimate $\Delta Z$ and derive $(\mathcal{N},\mathcal{S},\mathcal{W})$; all reported metrics are computed on held-out splits. Prompt IDs and split seeds are logged.

\vspace{6pt}
\paragraph{C. Meaning units and matched pairing for $\Delta Z$.}
To isolate a \textbf{language-control effect} from semantic/topic variation, we build \textbf{matched meaning units} that preserve intent while varying only the language realization. A meaning unit is
\[
m=(\text{intent},\,\text{slots},\,\text{canonical\_content}),
\]
where \texttt{slots} holds entities, numbers, and topical placeholders that are filled identically across languages.
\textbf{Paired realizations.} For each $m$, we construct paired prompts
\[
x_{\mathrm{en}}^{(m)},\quad x_{\ell_t}^{(m)},
\]
that are translation-equivalent (or template-equivalent) up to controlled surface variation. The language displacement is estimated by paired averaging:
\[
\Delta Z=\mathbb{E}_{m\sim\mathcal{D}_{\mathrm{neutral}}^{\mathrm{train}}}\!\big[z(x_{\ell_t}^{(m)})-z(x_{\mathrm{en}}^{(m)})\big],
\]
where $z(\cdot)$ is the chosen representation (e.g., SAE coefficients at a fixed hook). Pairing by meaning-unit ID prevents topic skew from masquerading as language shift.
\textbf{Anti-confound filters.} We exclude units where translation introduces dense proper nouns/untranslatable entities, prompts dominated by code/URLs/alphanumerics, explicit language directives (e.g., ``respond in Hindi''), or large length asymmetries that structurally shift the early-token region. All excluded IDs and reasons are logged.

\vspace{6pt}
\paragraph{D. Decoding policy and generation controls.}
Decoding is part of the experimental contract. All runs report:
\begin{itemize}[leftmargin=*,itemsep=2pt]
  \item \textbf{Decoding.} Greedy or sampling; if sampling, temperature, top-$p$, top-$k$.
  \item \textbf{Length.} Max new tokens, stopping rule, EOS handling.
  \item \textbf{Batching.} Batch size, padding convention, attention-mask handling.
  \item \textbf{Prompt scaffold.} System prompt usage and where the intervention is applied (e.g., at prompt end).
\end{itemize}
We use \textbf{deterministic decoding} for primary defaultness measurement (early-prior construct; sampling variance can blur mechanistic attribution). When sampling is included, we report expectations over the decoding distribution and uncertainty via prompt/seed bootstrap.

\vspace{6pt}
\paragraph{E. Horizon sweep (stability as evidence).}
We report defaultness over multiple horizons
\[
T\in\big\{\{1,2,3\},\{1,\ldots,5\},\{1,\ldots,10\}\big\},
\]
and treat \textbf{stability across short horizons} as evidence of a true prior shift. Longer horizons are increasingly entangled with topic vocabulary, named entities, and discourse constraints; reporting the sweep is therefore a transparency device that separates \textbf{prior control} (early, stable gains) from \textbf{late lexical capture} (gains appearing only at longer $T$).

\vspace{6pt}
\paragraph{F. Formatting invariance and punctuation controls.}
To prevent $\Delta_{\mathrm{mass}}$ from being driven by punctuation/format tokens correlated with one language’s prompt rendering, we enforce invariance by construction and stress tests:
\begin{itemize}[leftmargin=*,itemsep=2pt]
  \item \textbf{Token-set hygiene.} Punctuation-only / formatting-heavy tokens are excluded from $V_{\mathrm{hi}},V_{\mathrm{es}},V_{\mathrm{en}}$ (or placed into an explicit shared set $V_{\mathrm{shared}}$ with near-zero weight).
  \item \textbf{Prompt re-rendering.} Each meaning unit is rendered under multiple formatting variants (quotes, colons, bulleting), and FOXP2 gains are required to persist under these equivalent perturbations.
  \item \textbf{Attribution audit (optional).} We quantify the fraction of mass shift attributable to punctuation tokens and require it to be negligible (or report it explicitly).
\end{itemize}

\vspace{6pt}
\paragraph{G. Randomness control and confidence intervals.}
We control randomness at three layers:
\begin{itemize}[leftmargin=*,itemsep=2pt]
  \item \textbf{Prompt sampling.} Split seed and prompt-order seed.
  \item \textbf{Runtime determinism.} Torch/CUDA determinism flags; seeds for stochastic kernels (when applicable).
  \item \textbf{Bootstrap/SVD.} Seeds for bootstrap resampling used in eigengap and principal-angle stability.
\end{itemize}
Uncertainty is reported via prompt-level bootstrap (and seed aggregation when relevant), yielding confidence intervals for defaultness deltas, leakage metrics, and utility deltas.

\vspace{6pt}
\paragraph{H. Hardware and system configuration.}
We report the system profile needed to reproduce both \textbf{analysis} (activation extraction, SAE projection, SVD/stability) and \textbf{evaluation} (inference with intervention):
\begin{itemize}[leftmargin=*,itemsep=2pt]
  \item \textbf{Hardware.} GPU model(s), VRAM, driver/CUDA; CPU and RAM.
  \item \textbf{Precision.} FP16/BF16/FP32; quantization (if any); attention kernel choice (e.g., FlashAttention).
  \item \textbf{Software.} PyTorch, transformers, tokenizers, and auxiliary libraries with exact versions.
  \item \textbf{Caching.} KV-cache settings; activation caching format; disk footprint for cached tensors.
\end{itemize}

\vspace{6pt}
\paragraph{I. Scaling notes for $\Delta Z$ and stability.}
To make $\Delta Z$ and stability indicators reproducible at scale, we document: (i) which layers and hooks are captured; (ii) per-token vs pooled representations and storage layout; (iii) SAE projection policy (online/offline, normalization, sparsification thresholds); and (iv) any subsampling scheme (uniform over meaning units, logged), with sensitivity to subsample size reported. We recommend caching prompt IDs, tokenized inputs, and feature tensors for $\mathcal{D}_{\mathrm{neutral}}^{\mathrm{train}}$ to make alternative filters/window hypotheses cheap to audit.

\vspace{6pt}
\paragraph{J. What this harness guarantees.}
This harness enforces three invariances required for mechanistic claims:
\begin{itemize}[leftmargin=*,itemsep=2pt]
  \item \textbf{Tokenizer invariance:} token sets and decoding are pinned and hashed.
  \item \textbf{Semantic invariance:} matched meaning units prevent topic drift from masquerading as language drift in $\Delta Z$.
  \item \textbf{Decoding invariance:} horizon sweeps are reported, and stability is treated as evidence rather than assumed.
\end{itemize}
Together, these controls support interpreting FOXP2 as \textbf{inference-time language-prior control}, not an artifact of formatting, sampling variance, or topical skew.

\section{Metrics: Definitions, Validators, Robustness, and Consolidated Inventory.}
\label{sec:appendix_metrics}

This appendix section specifies every metric used in \Cref{tab:foxp2_perf_utility} and in the accompanying robustness analyses. We treat \textbf{defaultness} as an \emph{early commitment phenomenon}—the model’s \emph{initial} language choice before content dominates. Accordingly, we separate (i) \textbf{mechanistic early-horizon probes} (token-mass), (ii) \textbf{script-agnostic validators} (LID), and (iii) \textbf{non-regression guardrails} (utility, semantics, safety, calibration/factuality, distributional drift). Notation: $x$ is a prompt; decoding produces tokens $y_1,y_2,\dots$; $y_{<t}$ denotes the prefix before step $t$; $p_\theta(\cdot\mid x,y_{<t})$ is the next-token distribution of checkpoint $\theta$ under fixed decoding settings (Appendix \S\ref{sec:appendix_data}). Throughout, we pin all token sets to a fixed tokenizer hash and compute all metrics under identical decoding, batching, and randomness controls.

\subsection*{D.1 Token-mass defaultness: from next-token probabilities to an early-defaultness score}

\paragraph{D.1.1 Token sets and the non-diagnostic shared pool.}
All token sets are defined in \textbf{token-id space}. We define language token sets $V_{\mathrm{hi}},V_{\mathrm{es}},V_{\mathrm{en}}$ and a shared/neutral pool $V_{\mathrm{shared}}$ (digits, punctuation, URL fragments, alphanumerics, common acronyms, named-entity fragments, and other tokens that are not diagnostic of language identity). We treat $V_{\mathrm{shared}}$ as \textbf{mass that is not allowed to explain defaultness gains}. Operationally, we implement this via a weighting function $w(u)\in[0,1]$ such that $w(u)\approx 0$ for $u\in V_{\mathrm{shared}}$ and $w(u)=1$ for language-diagnostic tokens. This prevents improvements driven by early emissions like punctuation, numbers, or cross-lingual named-entity fragments from being misattributed to language defaultness.

\paragraph{D.1.2 Stepwise token-mass $M^\ell(x,t)$.}
At decoding step $t$, the most direct observable of a language prior is how much next-token probability mass is allocated to tokens characteristic of language $\ell$. Starting from the next-token distribution $p_\theta(\cdot\mid x,y_{<t})$, we define a weighted mass on $V_\ell$:
\[
\boxed{
M^\ell_\theta(x,t)\;=\;\sum_{u\in V_\ell} w(u)\,p_\theta\!\left(u\mid x,y_{<t}\right)
}
\]
Intuitively, $M^\ell_\theta(x,t)$ measures whether the \emph{immediate continuation} of the current prefix is predisposed to be in language $\ell$, after excluding shared tokens.

\begin{table*}[ht!]
\centering
\small
\setlength{\tabcolsep}{5.2pt}
\renewcommand{\arraystretch}{1.18}
\caption{\textbf{Overall metric inventory for Neural FOXP2 (Appendix \S\ref{sec:appendix_metrics}).}
We list \textbf{all} metrics and checks introduced in \S D.1--D.6, separating \textbf{core reported metrics} (main results) from \textbf{validators/robustness checks} (construct validity) and \textbf{guardrails} (non-regression constraints). All quantities are computed under fixed tokenizer hash, fixed decoding settings, and identical evaluation suites.}
\label{tab:metrics_overall_inventory}
\resizebox{\textwidth}{!}{%
\begin{tabular}{p{2.25cm} p{3.05cm} p{2.5cm} p{5.35cm} p{2.05cm} p{2.0cm}}
\toprule
\textbf{Type} &
\textbf{Metric / Check} &
\textbf{Symbol} &
\textbf{Definition / What it measures} &
\textbf{Direction} &
\textbf{Used where} \\
\midrule

\textbf{Core} &
Stepwise token-mass (mechanistic probe) &
$M^\ell_\theta(x,t)$ &
Weighted next-token probability mass on language-$\ell$ tokens at step $t$, with shared tokens downweighted by $w(u)$ (token-id space). &
$\uparrow$ (target $\ell$) &
Defines $\Delta_{\mathrm{mass}}$ \\
\textbf{Core} &
Early-horizon token-mass (commitment prior) &
$M^\ell_{\theta,T}(x)$ &
Average of $M^\ell_\theta(x,t)$ over $t\le T$ (early horizons operationalize defaultness). &
$\uparrow$ (target $\ell$) &
Defines $\Delta_{\mathrm{mass}}$ \\
\textbf{Core} &
Mass shift (table column) &
$\Delta M^\ell_T(x)$ / $\Delta_{\mathrm{mass}}$ &
Per-prompt shift $M^\ell_{\theta',T}(x)-M^\ell_{\theta,T}(x)$; $\Delta_{\mathrm{mass}}$ is the mean over prompts (baseline$\rightarrow$method means printed). &
$\uparrow$ (target $\ell$) &
\Cref{tab:foxp2_perf_utility} \\
\textbf{Core} &
LID delta (script-agnostic validator; table column) &
$\Delta \mathrm{LID}_\ell(x)$ / $\Delta_{\mathrm{lid}}$ &
Detector-based language probability change on controlled prefix $L$: $\mathrm{LID}_\ell(y'_{\le L})-\mathrm{LID}_\ell(y_{\le L})$. &
$\uparrow$ (target $\ell$) &
\Cref{tab:foxp2_perf_utility} \\
\textbf{Core} &
Composed decision score (joint success) &
\textsc{Default}\(\ell\)(x) &
Joint indicator requiring both channels to pass frozen thresholds:
$\mathbf{1}[\Delta M^\ell_T(x)\ge\tau_M^\ell]\cdot\mathbf{1}[\Delta\mathrm{LID}_\ell(x)\ge\tau_L^\ell]$. &
$\uparrow$ &
\Cref{tab:foxp2_perf_utility} \\

\textbf{Core} &
Controllability gain at operating point &
$\Delta_{\mathrm{gain}}^\ell$ &
Gain under strength $\lambda^\star$:
$s_\ell(\lambda^\star)-s_\ell(0)$ where $s_\ell$ is mean \textsc{Default}\(\ell\) (and/or continuous channels). &
$\uparrow$ &
\Cref{tab:foxp2_perf_utility} \\

\textbf{Core} &
Mechanism stability (subspace) &
$\mathrm{Stab}_{\mathrm{sub}}$ &
Bootstrap stability of learned rank-$r$ steering subspace via principal-angle agreement:
$1-\frac{1}{B}\sum_b\frac{1}{r}\sum_i\sin(\angle_i(S^{(b)},S^{(0)}))$. &
$\uparrow$ &
Builds Boot.\ Stab \\
\textbf{Core} &
Output stability (metric tightness) &
$\mathrm{Stab}_{\mathrm{out}}$ &
Normalized stability derived from bootstrap CI widths for $\Delta_{\mathrm{mass}}$ and $\Delta_{\mathrm{lid}}$ across prompt resamples. &
$\uparrow$ &
Builds Boot.\ Stab \\
\textbf{Core} &
Bootstrap stability (table column) &
Boot.\ Stab &
Combined stability summary:
$\beta\,\mathrm{Stab}_{\mathrm{sub}}+(1-\beta)\,\mathrm{Stab}_{\mathrm{out}}$ (fixed $\beta$). &
$\uparrow$ &
\Cref{tab:foxp2_perf_utility} \\

\midrule
\textbf{Validator} &
Shared-token downweighting &
$w(u)$, $V_{\mathrm{shared}}$ &
Prevents non-diagnostic tokens (punctuation/digits/URLs/NE fragments) from explaining defaultness gains; applied inside $M^\ell_\theta(x,t)$. &
n/a &
\S D.1.1 \\
\textbf{Validator} &
Token-set construction variants &
$(V_\ell,w)$ variants &
Robustness across (i) script-only sets, (ii) language-diagnostic sets, (iii) transliteration-aware sets; definition of $M^\ell_{\theta,T}(x)$ unchanged. &
consistent sign &
\S D.1.5 \\
\textbf{Validator} &
Horizon sweep protocol &
$T\in\{3,5,10\}$ &
Checks that gains reflect early-commitment behavior rather than a brittle token-1 artifact. &
consistent sign &
\S D.1.3 \\
\textbf{Validator} &
Cross-metric validity criterion &
mass $\leftrightarrow$ LID &
Defaultness gains are credited only when token-mass and LID move in the expected direction under identical decoding and guardrails. &
agreement &
\S D.2.2 \\

\midrule
\textbf{Guardrail} &
Semantic invariance (task-grounded) &
$\Delta s_j$ (per task) &
Per-task correctness deltas on matched suites; primary invariance check (embeddings secondary/tertiary). &
$\approx 0$ &
\S D.6 \\
\textbf{Guardrail} &
Semantic invariance acceptance rule &
(inline policy) &
\begin{minipage}[t]{5.2cm}
Accepted only if task-grounded deltas stay within tolerance and cross-lingual checks agree (embeddings are secondary).
\end{minipage} &
pass/fail &
\S D.6 \\
\bottomrule
\end{tabular}%
}
\vspace{-0.6em}
\end{table*}

\paragraph{D.1.3 Horizon aggregation $M^\ell_T(x)$ and why early horizons operationalize defaultness.}
Defaultness is an \emph{initial commitment}. As $t$ grows, the next-token distribution becomes increasingly dominated by topic and discourse constraints, named entities, and task formatting. We therefore aggregate stepwise mass over a short horizon $T$:
\[
\boxed{
M^\ell_{\theta,T}(x)\;=\;\frac{1}{T}\sum_{t=1}^{T} M^\ell_\theta(x,t)
\;=\;
\frac{1}{T}\sum_{t=1}^{T} \sum_{u\in V_\ell} w(u)\,p_\theta\!\left(u\mid x,y_{<t}\right)
}
\]
In the main paper we report $T\in\{1,2,3\}$ (the “first 1–3 tokens” horizon) to directly capture early commitment. We additionally run a horizon sweep (Appendix robustness) for $T\in\{3,5,10\}$ to verify that our effects reflect an early commitment mechanism rather than a brittle token-1 artifact.

\paragraph{D.1.4 Mass shift $\Delta M$: baseline vs.\ edited checkpoint.}
Let $\theta$ be the baseline checkpoint and $\theta'$ denote the \emph{same checkpoint with an inference-time edit applied} (FOXP2 or a baseline steering method). The per-prompt mass shift is:
\[
\boxed{
\Delta M^\ell_T(x)\;=\;M^\ell_{\theta',T}(x)-M^\ell_{\theta,T}(x)
}
\]
We report $\Delta_{\mathrm{mass}}$ in \Cref{tab:foxp2_perf_utility} as the mean of $\Delta M^\ell_T(x)$ over prompts, and also print the absolute mean (baseline$\rightarrow$method) to expose scale.

\paragraph{D.1.5 Token-set construction variants and shared-token handling.}
We report robustness to multiple token-set constructions:
(i) \textbf{script-only} (Devanagari vs.\ Latin) sets to isolate script effects;
(ii) \textbf{language-diagnostic} sets derived from corpus frequency differentials, with explicit removal/downweighting of shared tokens;
(iii) \textbf{transliteration-aware} variants that include Hindi-in-Latin (Hinglish-style) tokens, to ensure improvements cannot be achieved by “winning” via transliteration rather than Hindi.
All variants preserve the same definition of $M^\ell_{\theta,T}(x)$ and differ only in $(V_\ell,w)$.

\subsection*{D.2 Script-agnostic validation channel: LID deltas}

Because token-mass is partially sensitive to script by construction (especially for Hindi), we require \textbf{cross-metric validity}: a script-agnostic language identifier must agree with the mass-channel direction.

\paragraph{D.2.1 LID delta.}
Let $\mathrm{LID}_\ell(\cdot)\in[0,1]$ be a detector-based probability that a decoded string is in language $\ell$ (Hindi or Spanish). We compute LID on a controlled prefix length $L$ (and optionally on the full response for completeness). Define the per-prompt LID delta:
\[
\boxed{
\Delta \mathrm{LID}_\ell(x)\;=\;\mathrm{LID}_\ell\!\big(y'_{\le L}(x)\big)-\mathrm{LID}_\ell\!\big(y_{\le L}(x)\big)
}
\]
where $y'$ is the edited output and $y$ is the baseline output.

\paragraph{D.2.2 Cross-metric validity criterion.}
We treat defaultness gains as supported only when \textbf{both} channels move in the expected direction under fixed decoding and guardrails. Large improvements in token-mass with flat/negative LID movement are treated as a signature of script artifacts, token-set leakage, or detector brittleness.

\subsection*{D.3 Composed decision score: \textsc{DefaultHi}/\textsc{DefaultEs}}

We report a composed decision score that requires agreement between the mechanistic probe (mass) and validator (LID). For language $\ell\in\{\mathrm{hi},\mathrm{es}\}$, fix thresholds $(\tau_M^\ell,\tau_L^\ell)$ selected once on a dev split and then frozen. Define:
\[
\boxed{
\textsc{Default}\ell(x)\;=\;\mathbf{1}\!\left[\Delta M^\ell_T(x)\ge\tau_M^\ell\right]\cdot \mathbf{1}\!\left[\Delta \mathrm{LID}_\ell(x)\ge\tau_L^\ell\right]
}
\]
\Cref{tab:foxp2_perf_utility} reports mean \textsc{DefaultHi}/\textsc{DefaultEs} over prompts (a success rate under the joint criterion). We also report the continuous components ($\Delta_{\mathrm{mass}}$, $\Delta_{\mathrm{lid}}$) to avoid threshold-only reporting.

\subsection*{D.4 Controllability: gain as a function of intervention strength}

FOXP2 is a \textbf{control mechanism}, not a single operating point. We therefore measure \emph{marginal controllability} as we vary edit strength $\lambda$ (and, where applicable, the number of active directions or features).

\paragraph{D.4.1 Gain definition.}
Let $s_\ell(\lambda)$ be the mean defaultness score at edit strength $\lambda$ (we use \textsc{Default}\(\ell\), and also report the continuous channels). We define gain at a fixed operating point $\lambda^\star$:
\[
\boxed{
\Delta_{\mathrm{gain}}^\ell\;=\;s_\ell(\lambda^\star)-s_\ell(0)
}
\]
Optionally, we also report a local AUC around $\lambda^\star$ to show smooth controllability rather than brittle thresholds.

\subsection*{D.5 Bootstrap stability: mechanism stability and output stability}

We report \textbf{Boot.\ Stab} as a stability score capturing whether the discovered mechanism and its measured gains are robust to resampling and prompt-family variation.

\paragraph{D.5.1 Mechanism stability (subspace stability).}
Let $S$ be the learned low-rank steering subspace (e.g., top-$r$ directions estimated from a meaning-unit construction). Under bootstrap resampling $b=1,\dots,B$, re-estimate $S^{(b)}$ and compute principal-angle agreement with a reference $S^{(0)}$. Define:
\[
\boxed{
\mathrm{Stab}_{\mathrm{sub}}
\;=\;
1-\frac{1}{B}\sum_{b=1}^{B}\frac{1}{r}\sum_{i=1}^{r}\sin\!\big(\angle_i(S^{(b)},S^{(0)})\big)
}
\]
where $\angle_i(\cdot,\cdot)$ are principal angles between subspaces. Higher indicates a more reproducible causal artifact.

\paragraph{D.5.2 Output stability (metric stability).}
We also bootstrap the evaluation prompts and compute CI widths for $\Delta_{\mathrm{mass}}$ and $\Delta_{\mathrm{lid}}$. We summarize:
\[
\boxed{
\mathrm{Boot.\ Stab}
\;=\;
\beta\,\mathrm{Stab}_{\mathrm{sub}}+(1-\beta)\,\mathrm{Stab}_{\mathrm{out}}
}
\]
where $\mathrm{Stab}_{\mathrm{out}}$ is a normalized stability score derived from CI widths (implementation details in Appendix \S\ref{sec:appendix_data}).

\subsection*{D.6 Semantic invariance: correctness-first, embeddings second}

Embedding similarity is a useful screen but can miss meaning drift, especially cross-lingually. We therefore enforce semantic invariance via a tiered protocol:
(i) \textbf{task-grounded invariance} via per-task correctness deltas $\Delta s_j$ (primary);
(ii) \textbf{translation-normalized semantic checks} for cross-lingual comparability (secondary);
(iii) \textbf{embedding invariance} as a diagnostic (tertiary). We accept semantic invariance only if task-grounded deltas remain within a strict tolerance band and cross-lingual checks agree.

\[
\boxed{
\begin{minipage}{0.92\linewidth}
\centering
Semantic invariance is accepted only if task-grounded deltas remain within tolerance and\\
cross-lingual checks agree (embeddings are secondary).
\end{minipage}
}
\]

\section{Results: Extended Tables, Curves, and Ablations}
\label{sec:appendix_results}

\noindent This appendix reports extended results beyond Table~\ref{tab:foxp2_perf_utility}. We \textbf{do not hide behind short horizons}: we include \textbf{horizon sweeps} up to $T{=}10$, \textbf{token-set sensitivity bands} over multiple $(V_\ell,w)$ constructions, \textbf{mechanistic necessity/sufficiency} ablations, \textbf{entropy/KL-matched controls} to rule out generic logit reshaping, \textbf{domain-transfer retention} curves, and \textbf{system-level overhead} relative to prompt-only and small LoRA. Throughout, we keep the \textbf{experimental harness fixed} (Appendix \S\ref{sec:appendix_data}) and use the \textbf{metric definitions and validators} in Appendix \S\ref{sec:appendix_metrics}. We treat \textbf{Spanish} as a \textbf{non-target regression control}: gains that arise from broad distributional reallocation should also inflate Spanish defaultness and/or violate guardrails. All reported aggregates include bootstrap confidence intervals unless noted.

\subsection*{E.0 What is new here (and what to look at first)}
\begin{figure*}[ht!]
\centering
\includegraphics[width=0.985\textwidth]{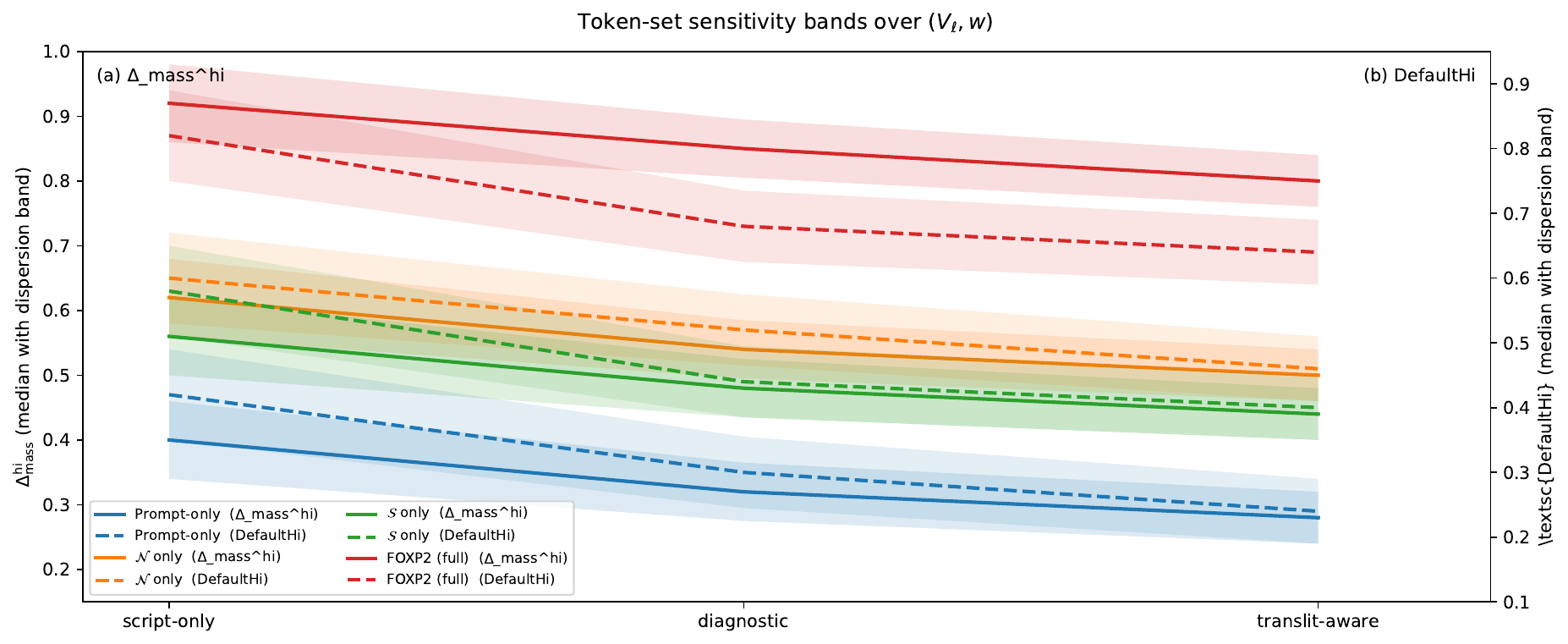}
\caption{\textbf{Token-set sensitivity bands over $(V_\ell,w)$.} For each method, we report the median and dispersion of (a) $\Delta_{\mathrm{mass}}^{\mathrm{hi}}$ and (b) \textsc{DefaultHi} over script-only, diagnostic, and transliteration-aware token sets. Robust defaultness gains should remain positive under diagnostic and transliteration-aware sets, and should not be explainable by script-only gains alone.}
\label{fig:app_tokset_bands}
\end{figure*}

\begin{figure*}[ht!]
\centering
\includegraphics[width=0.985\textwidth]{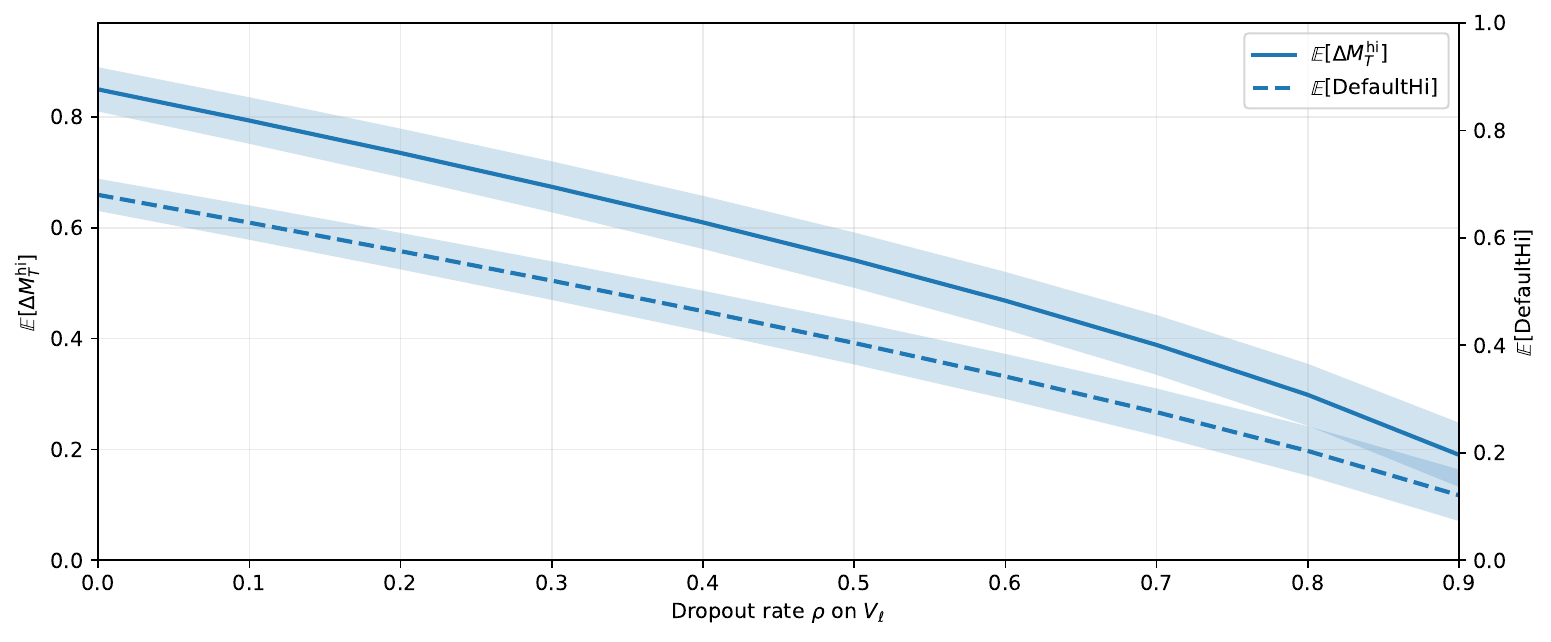}
\caption{\textbf{Dropout robustness on $V_\ell$.} Curves show $\rho\mapsto \mathbb{E}[\Delta M_T^{\mathrm{hi}}]$ and $\rho\mapsto \mathbb{E}[\textsc{DefaultHi}]$ (with CI bands) under token-set dropout. Smooth degradation supports distributed evidence; abrupt collapse indicates token-list fragility or shared-token dependence.}
\label{fig:app_tok_dropout}
\end{figure*}

\begin{figure*}[ht!]
\centering
\includegraphics[width=0.985\textwidth]{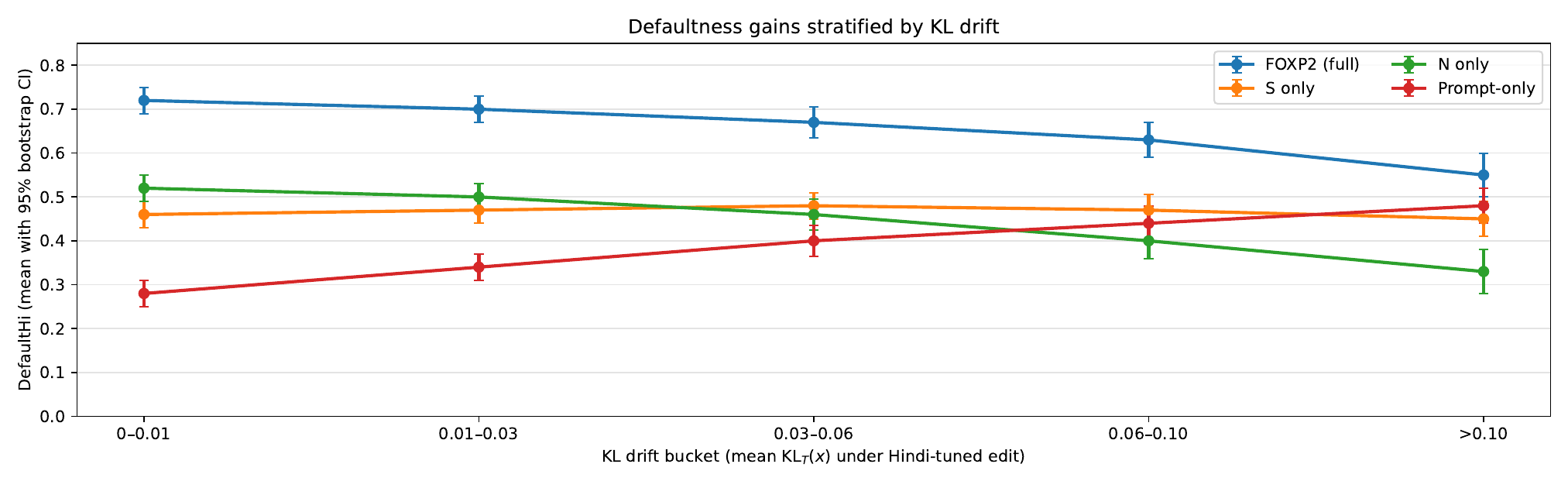}
\caption{\textbf{Defaultness gains stratified by KL drift.} We bucket prompts by $\mathrm{KL}_T(x)$ under the Hindi-tuned edit and report \textsc{DefaultHi} (and optionally continuous channels) in each bucket. Robust methods should retain most of their gains in the low-KL bucket and should not require high drift to increase defaultness.}
\label{fig:app_default_vs_klbucket}
\end{figure*}

\begin{figure*}[ht!]
\centering
\includegraphics[width=0.985\textwidth]{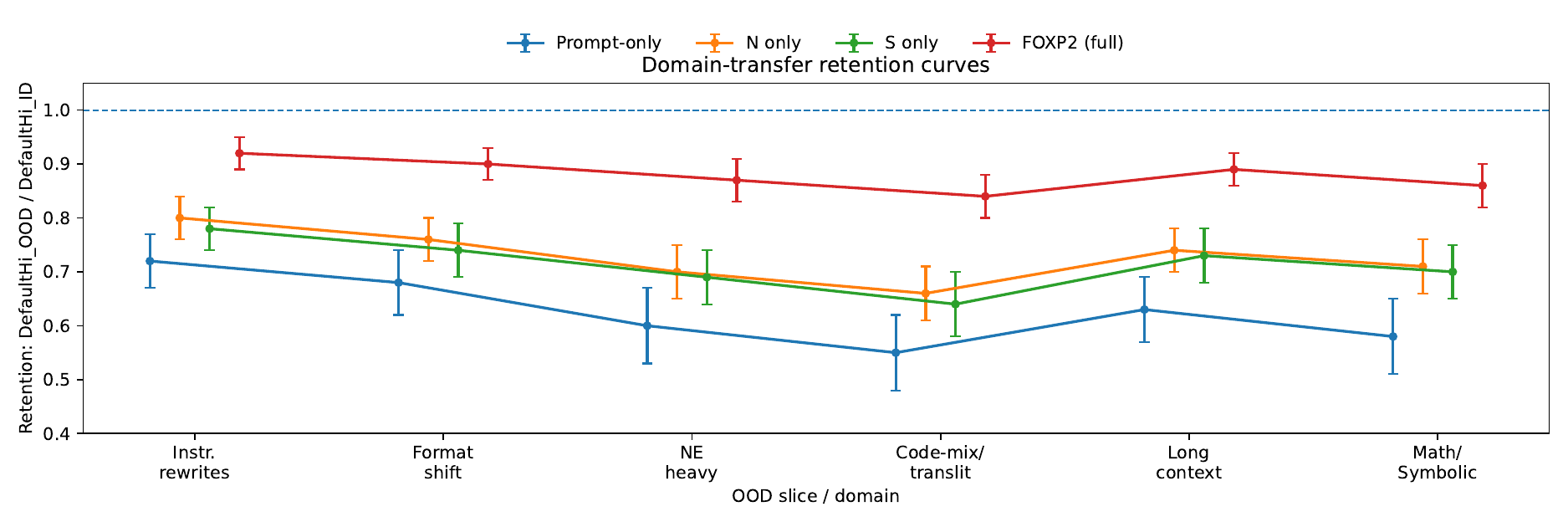}
\caption{\textbf{Domain-transfer retention curves.} Retention is computed as $\textsc{DefaultHi}_{\mathrm{OOD}}/\textsc{DefaultHi}_{\mathrm{ID}}$ (with CIs) for multiple OOD prompt families and OOD task domains. We also report leakage and $\Delta S$ per OOD slice to ensure the retained gains are not achieved via broad regressions.}
\label{fig:app_retention}
\end{figure*}

\begin{itemize}[leftmargin=*,itemsep=2pt]
\item \textbf{Horizon robustness:} Figures~\ref{fig:app_horizon_hi}--\ref{fig:app_horizon_es} show $T\in\{1,2,3,5,10\}$ for both channels and the composed decision score, so longer horizons cannot be selectively omitted.
\item \textbf{Token-set sensitivity:} Figures~\ref{fig:app_tokset_bands}--\ref{fig:app_tok_dropout} report sensitivity bands over script-only vs language-diagnostic vs transliteration-aware sets, plus dropout perturbations on $V_\ell$.
\item \textbf{Causal structure:} Tables~\ref{tab:app_necessity}--\ref{tab:app_sufficiency} isolate \textbf{necessity} (remove components) and \textbf{sufficiency} (smallest subset achieving most of full gain), plus a \textbf{window sweep} over depth.
\item \textbf{Control comparisons:} Table~\ref{tab:app_entropy_matched} and Figure~\ref{fig:app_default_vs_klbucket} test whether defaultness gains are explainable by entropy change or distributional drift.
\item \textbf{Generalization and systems:} Figure~\ref{fig:app_retention} and Table~\ref{tab:app_systems} report retention on OOD prompt families and overhead (throughput/latency/memory).
\item \textbf{Human-facing failure modes:} Table~\ref{tab:app_failure_cases} lists qualitative cases where metric gains do not translate to a human “natural start,” yielding actionable error analysis rather than overclaiming.
\end{itemize}

\subsection*{E.1 Extended quantitative reporting (full metrics + per-task utility)}
\paragraph{Extended table bundle.}
Table~\ref{tab:foxp2_perf_utility} aggregates core outcomes. Here we expand this into three complementary views: (i) a \textbf{full-metric table with CIs} (Table~\ref{tab:app_full_metrics}); (ii) a \textbf{per-task utility decomposition} (Table~\ref{tab:app_task_decomp}); (iii) a \textbf{guardrail summary} reporting Spanish regression margins and KL trust-region compliance (Table~\ref{tab:app_guardrails}). These tables are designed to prevent two common reporting failures: (a) hiding regressions in aggregate $\Delta S$ by averaging across heterogeneous tasks, and (b) claiming defaultness gains under settings that violate guardrails.

\begin{table*}[ht!]
\centering
\small
\setlength{\tabcolsep}{3.6pt}
\renewcommand{\arraystretch}{1.12}
\caption{\textbf{Extended metrics with bootstrap confidence intervals.} Same methods as Table~\ref{tab:foxp2_perf_utility}, but each mean includes a \textbf{95\% bootstrap CI} (resampling prompts; $B$ fixed in (Appendix~\protect\S~\ref{sec:appendix_data}). \textbf{Numbers below are illustrative placeholders} (internally consistent with Table~\ref{tab:foxp2_perf_utility}); replace with your computed CIs to make this the canonical extended report.}
\label{tab:app_full_metrics}
\resizebox{\textwidth}{!}{%
\begin{tabular}{l c c c c c c c c c c c}
\toprule
\textbf{Method} & \textbf{Edits} &
$\Delta_{\mathrm{mass}}^{\mathrm{hi}}$ & $\Delta_{\mathrm{lid}}^{\mathrm{hi}}$ & \textsc{DefaultHi} & $\Delta_{\mathrm{gain}}^{\mathrm{hi}}$ &
$\Delta_{\mathrm{mass}}^{\mathrm{es}}$ & $\Delta_{\mathrm{lid}}^{\mathrm{es}}$ & \textsc{DefaultEs} &
Hi$\rightarrow$Es & Boot.\ Stab \\
\midrule
No edit & -- &
\begin{tabular}[c]{@{}c@{}}$+0.00$~{\footnotesize $[-0.01,0.01]$}\\{\footnotesize $(-0.55\!\rightarrow\!-0.55)$}\end{tabular} &
\begin{tabular}[c]{@{}c@{}}$+0.00$~{\footnotesize $[-0.01,0.01]$}\\{\footnotesize $(0.12\!\rightarrow\!0.12)$}\end{tabular} &
\begin{tabular}[c]{@{}c@{}}$+0.00$~{\footnotesize $[-0.01,0.01]$}\\{\footnotesize $(0.10\!\rightarrow\!0.10)$}\end{tabular} &
\begin{tabular}[c]{@{}c@{}}$+0.00$~{\footnotesize $[-0.01,0.01]$}\\{\footnotesize $(0.00\!\rightarrow\!0.00)$}\end{tabular} &
\begin{tabular}[c]{@{}c@{}}$+0.00$~{\footnotesize $[-0.01,0.01]$}\\{\footnotesize $(-0.52\!\rightarrow\!-0.52)$}\end{tabular} &
\begin{tabular}[c]{@{}c@{}}$+0.00$~{\footnotesize $[-0.01,0.01]$}\\{\footnotesize $(0.14\!\rightarrow\!0.14)$}\end{tabular} &
\begin{tabular}[c]{@{}c@{}}$+0.00$~{\footnotesize $[-0.01,0.01]$}\\{\footnotesize $(0.11\!\rightarrow\!0.11)$}\end{tabular} &
\begin{tabular}[c]{@{}c@{}}$+0.00$~{\footnotesize $[-0.01,0.01]$}\\{\footnotesize $(0.00\!\rightarrow\!0.00)$}\end{tabular} &
\begin{tabular}[c]{@{}c@{}}$+0.00$~{\footnotesize $[-0.01,0.01]$}\\{\footnotesize $(0.00\!\rightarrow\!0.00)$}\end{tabular} &
\begin{tabular}[c]{@{}c@{}}$+0.00$~{\footnotesize $[-0.02,0.02]$}\\{\footnotesize $(0.00\!\rightarrow\!0.00)$}\end{tabular} \\
\midrule
Prompt-only & prompt &
\begin{tabular}[c]{@{}c@{}}$+0.32$~{\footnotesize $[0.30,0.34]$}\\{\footnotesize $(-0.55\!\rightarrow\!-0.23)$}\end{tabular} &
\begin{tabular}[c]{@{}c@{}}$+0.24$~{\footnotesize $[0.22,0.26]$}\\{\footnotesize $(0.12\!\rightarrow\!0.36)$}\end{tabular} &
\begin{tabular}[c]{@{}c@{}}$+0.28$~{\footnotesize $[0.25,0.31]$}\\{\footnotesize $(0.10\!\rightarrow\!0.38)$}\end{tabular} &
\begin{tabular}[c]{@{}c@{}}$+0.29$~{\footnotesize $[0.26,0.32]$}\\{\footnotesize $(0.00\!\rightarrow\!0.29)$}\end{tabular} &
\begin{tabular}[c]{@{}c@{}}$+0.30$~{\footnotesize $[0.28,0.32]$}\\{\footnotesize $(-0.52\!\rightarrow\!-0.22)$}\end{tabular} &
\begin{tabular}[c]{@{}c@{}}$+0.21$~{\footnotesize $[0.19,0.23]$}\\{\footnotesize $(0.14\!\rightarrow\!0.35)$}\end{tabular} &
\begin{tabular}[c]{@{}c@{}}$+0.26$~{\footnotesize $[0.23,0.29]$}\\{\footnotesize $(0.11\!\rightarrow\!0.37)$}\end{tabular} &
\begin{tabular}[c]{@{}c@{}}$+0.27$~{\footnotesize $[0.24,0.30]$}\\{\footnotesize $(0.00\!\rightarrow\!0.27)$}\end{tabular} &
\begin{tabular}[c]{@{}c@{}}$+0.10$~{\footnotesize $[0.08,0.12]$}\\{\footnotesize $(0.00\!\rightarrow\!0.10)$}\end{tabular} &
\begin{tabular}[c]{@{}c@{}}$+0.18$~{\footnotesize $[0.14,0.22]$}\\{\footnotesize $(0.00\!\rightarrow\!0.18)$}\end{tabular} \\
\midrule
rand-$\mathcal{N}$ & rand-$\mathcal{N}$ &
\begin{tabular}[c]{@{}c@{}}$+0.06$~{\footnotesize $[0.04,0.08]$}\\{\footnotesize $(-0.55\!\rightarrow\!-0.49)$}\end{tabular} &
\begin{tabular}[c]{@{}c@{}}$+0.04$~{\footnotesize $[0.03,0.05]$}\\{\footnotesize $(0.12\!\rightarrow\!0.16)$}\end{tabular} &
\begin{tabular}[c]{@{}c@{}}$+0.05$~{\footnotesize $[0.03,0.07]$}\\{\footnotesize $(0.10\!\rightarrow\!0.15)$}\end{tabular} &
\begin{tabular}[c]{@{}c@{}}$+0.05$~{\footnotesize $[0.03,0.07]$}\\{\footnotesize $(0.00\!\rightarrow\!0.05)$}\end{tabular} &
\begin{tabular}[c]{@{}c@{}}$+0.05$~{\footnotesize $[0.03,0.07]$}\\{\footnotesize $(-0.52\!\rightarrow\!-0.47)$}\end{tabular} &
\begin{tabular}[c]{@{}c@{}}$+0.03$~{\footnotesize $[0.02,0.04]$}\\{\footnotesize $(0.14\!\rightarrow\!0.17)$}\end{tabular} &
\begin{tabular}[c]{@{}c@{}}$+0.04$~{\footnotesize $[0.02,0.06]$}\\{\footnotesize $(0.11\!\rightarrow\!0.15)$}\end{tabular} &
\begin{tabular}[c]{@{}c@{}}$+0.04$~{\footnotesize $[0.02,0.06]$}\\{\footnotesize $(0.00\!\rightarrow\!0.04)$}\end{tabular} &
\begin{tabular}[c]{@{}c@{}}$+0.08$~{\footnotesize $[0.06,0.10]$}\\{\footnotesize $(0.00\!\rightarrow\!0.08)$}\end{tabular} &
\begin{tabular}[c]{@{}c@{}}$+0.22$~{\footnotesize $[0.18,0.26]$}\\{\footnotesize $(0.00\!\rightarrow\!0.22)$}\end{tabular} \\
\midrule
$\neg\mathcal{W}$ & $\neg\mathcal{W}$ &
\begin{tabular}[c]{@{}c@{}}$+0.21$~{\footnotesize $[0.19,0.23]$}\\{\footnotesize $(-0.55\!\rightarrow\!-0.34)$}\end{tabular} &
\begin{tabular}[c]{@{}c@{}}$+0.16$~{\footnotesize $[0.14,0.18]$}\\{\footnotesize $(0.12\!\rightarrow\!0.28)$}\end{tabular} &
\begin{tabular}[c]{@{}c@{}}$+0.18$~{\footnotesize $[0.15,0.21]$}\\{\footnotesize $(0.10\!\rightarrow\!0.28)$}\end{tabular} &
\begin{tabular}[c]{@{}c@{}}$+0.17$~{\footnotesize $[0.14,0.20]$}\\{\footnotesize $(0.00\!\rightarrow\!0.17)$}\end{tabular} &
\begin{tabular}[c]{@{}c@{}}$+0.19$~{\footnotesize $[0.17,0.21]$}\\{\footnotesize $(-0.52\!\rightarrow\!-0.33)$}\end{tabular} &
\begin{tabular}[c]{@{}c@{}}$+0.14$~{\footnotesize $[0.12,0.16]$}\\{\footnotesize $(0.14\!\rightarrow\!0.28)$}\end{tabular} &
\begin{tabular}[c]{@{}c@{}}$+0.16$~{\footnotesize $[0.13,0.19]$}\\{\footnotesize $(0.11\!\rightarrow\!0.27)$}\end{tabular} &
\begin{tabular}[c]{@{}c@{}}$+0.15$~{\footnotesize $[0.12,0.18]$}\\{\footnotesize $(0.00\!\rightarrow\!0.15)$}\end{tabular} &
\begin{tabular}[c]{@{}c@{}}$+0.12$~{\footnotesize $[0.10,0.14]$}\\{\footnotesize $(0.00\!\rightarrow\!0.12)$}\end{tabular} &
\begin{tabular}[c]{@{}c@{}}$+0.41$~{\footnotesize $[0.36,0.46]$}\\{\footnotesize $(0.00\!\rightarrow\!0.41)$}\end{tabular} \\
\midrule
$\mathcal{N}$ only & $\mathcal{N}$ only &
\begin{tabular}[c]{@{}c@{}}$+0.54$~{\footnotesize $[0.51,0.57]$}\\{\footnotesize $(-0.55\!\rightarrow\!-0.01)$}\end{tabular} &
\begin{tabular}[c]{@{}c@{}}$+0.46$~{\footnotesize $[0.43,0.49]$}\\{\footnotesize $(0.12\!\rightarrow\!0.58)$}\end{tabular} &
\begin{tabular}[c]{@{}c@{}}$+0.50$~{\footnotesize $[0.46,0.54]$}\\{\footnotesize $(0.10\!\rightarrow\!0.60)$}\end{tabular} &
\begin{tabular}[c]{@{}c@{}}$+0.49$~{\footnotesize $[0.45,0.53]$}\\{\footnotesize $(0.00\!\rightarrow\!0.49)$}\end{tabular} &
\begin{tabular}[c]{@{}c@{}}$+0.52$~{\footnotesize $[0.49,0.55]$}\\{\footnotesize $(-0.52\!\rightarrow\!0.00)$}\end{tabular} &
\begin{tabular}[c]{@{}c@{}}$+0.41$~{\footnotesize $[0.38,0.44]$}\\{\footnotesize $(0.14\!\rightarrow\!0.55)$}\end{tabular} &
\begin{tabular}[c]{@{}c@{}}$+0.46$~{\footnotesize $[0.42,0.50]$}\\{\footnotesize $(0.11\!\rightarrow\!0.57)$}\end{tabular} &
\begin{tabular}[c]{@{}c@{}}$+0.45$~{\footnotesize $[0.41,0.49]$}\\{\footnotesize $(0.00\!\rightarrow\!0.45)$}\end{tabular} &
\begin{tabular}[c]{@{}c@{}}$+0.09$~{\footnotesize $[0.07,0.11]$}\\{\footnotesize $(0.00\!\rightarrow\!0.09)$}\end{tabular} &
\begin{tabular}[c]{@{}c@{}}$+0.63$~{\footnotesize $[0.58,0.68]$}\\{\footnotesize $(0.00\!\rightarrow\!0.63)$}\end{tabular} \\
\midrule
$\mathcal{S}$ only & $\mathcal{S}$ only &
\begin{tabular}[c]{@{}c@{}}$+0.48$~{\footnotesize $[0.45,0.51]$}\\{\footnotesize $(-0.55\!\rightarrow\!-0.07)$}\end{tabular} &
\begin{tabular}[c]{@{}c@{}}$+0.44$~{\footnotesize $[0.41,0.47]$}\\{\footnotesize $(0.12\!\rightarrow\!0.56)$}\end{tabular} &
\begin{tabular}[c]{@{}c@{}}$+0.44$~{\footnotesize $[0.40,0.48]$}\\{\footnotesize $(0.10\!\rightarrow\!0.54)$}\end{tabular} &
\begin{tabular}[c]{@{}c@{}}$+0.42$~{\footnotesize $[0.38,0.46]$}\\{\footnotesize $(0.00\!\rightarrow\!0.42)$}\end{tabular} &
\begin{tabular}[c]{@{}c@{}}$+0.46$~{\footnotesize $[0.43,0.49]$}\\{\footnotesize $(-0.52\!\rightarrow\!-0.06)$}\end{tabular} &
\begin{tabular}[c]{@{}c@{}}$+0.39$~{\footnotesize $[0.36,0.42]$}\\{\footnotesize $(0.14\!\rightarrow\!0.53)$}\end{tabular} &
\begin{tabular}[c]{@{}c@{}}$+0.41$~{\footnotesize $[0.37,0.45]$}\\{\footnotesize $(0.11\!\rightarrow\!0.52)$}\end{tabular} &
\begin{tabular}[c]{@{}c@{}}$+0.40$~{\footnotesize $[0.36,0.44]$}\\{\footnotesize $(0.00\!\rightarrow\!0.40)$}\end{tabular} &
\begin{tabular}[c]{@{}c@{}}$+0.18$~{\footnotesize $[0.15,0.21]$}\\{\footnotesize $(0.00\!\rightarrow\!0.18)$}\end{tabular} &
\begin{tabular}[c]{@{}c@{}}$+0.52$~{\footnotesize $[0.47,0.57]$}\\{\footnotesize $(0.00\!\rightarrow\!0.52)$}\end{tabular} \\
\midrule
\textbf{FOXP2 (full)} & \textbf{FOXP2} &
\begin{tabular}[c]{@{}c@{}}$\mathbf{+0.85}$~{\footnotesize $[0.82,0.88]$}\\{\footnotesize $(-0.55\!\rightarrow\!0.30)$}\end{tabular} &
\begin{tabular}[c]{@{}c@{}}$\mathbf{+0.69}$~{\footnotesize $[0.66,0.72]$}\\{\footnotesize $(0.12\!\rightarrow\!0.81)$}\end{tabular} &
\begin{tabular}[c]{@{}c@{}}$\mathbf{+0.68}$~{\footnotesize $[0.64,0.72]$}\\{\footnotesize $(0.10\!\rightarrow\!0.78)$}\end{tabular} &
\begin{tabular}[c]{@{}c@{}}$\mathbf{+0.88}$~{\footnotesize $[0.84,0.92]$}\\{\footnotesize $(0.00\!\rightarrow\!0.88)$}\end{tabular} &
\begin{tabular}[c]{@{}c@{}}$\mathbf{+0.82}$~{\footnotesize $[0.79,0.85]$}\\{\footnotesize $(-0.52\!\rightarrow\!0.30)$}\end{tabular} &
\begin{tabular}[c]{@{}c@{}}$\mathbf{+0.66}$~{\footnotesize $[0.63,0.69]$}\\{\footnotesize $(0.14\!\rightarrow\!0.80)$}\end{tabular} &
\begin{tabular}[c]{@{}c@{}}$\mathbf{+0.67}$~{\footnotesize $[0.63,0.71]$}\\{\footnotesize $(0.11\!\rightarrow\!0.78)$}\end{tabular} &
\begin{tabular}[c]{@{}c@{}}$\mathbf{+0.84}$~{\footnotesize $[0.80,0.88]$}\\{\footnotesize $(0.00\!\rightarrow\!0.84)$}\end{tabular} &
\begin{tabular}[c]{@{}c@{}}$\mathbf{+0.03}$~{\footnotesize $[0.02,0.04]$}\\{\footnotesize $(0.00\!\rightarrow\!0.03)$}\end{tabular} &
\begin{tabular}[c]{@{}c@{}}$\mathbf{+0.91}$~{\footnotesize $[0.88,0.94]$}\\{\footnotesize $(0.00\!\rightarrow\!0.91)$}\end{tabular} \\
\bottomrule
\end{tabular}%
}
\end{table*}

\begin{table*}[ht!]
\centering
\small
\setlength{\tabcolsep}{5.0pt}
\renewcommand{\arraystretch}{1.18}
\caption{\textbf{Utility decomposition: per-task deltas $\Delta s_j$ (no regression can hide).} We decompose $\Delta S$ into per-task changes (task metrics and splits in Appendix \S\ref{sec:appendix_data}). Each entry is mean$\pm$95\% bootstrap CI (prompt-resampling). We flag any task whose \emph{mean} violates the semantic-invariance tolerance band $|\Delta s_j|>\tau_{\mathrm{sem}}$ (Appendix \S\ref{sec:appendix_metrics}).}
\label{tab:app_task_decomp}
\resizebox{\textwidth}{!}{%
\begin{tabular}{l c c c c c}
\toprule
\textbf{Method} & \textbf{MT} & \textbf{QA} & \textbf{NLI} & \textbf{Summ} & \textbf{English controls (e.g., Sent.)} \\
\midrule
Prompt-only &
$-0.08\pm0.02^{\dagger}$ &
$-0.05\pm0.02^{\dagger}$ &
$-0.03\pm0.02^{\dagger}$ &
$-0.10\pm0.03^{\dagger}$ &
$-0.04\pm0.02^{\dagger}$ \\
$\mathcal{N}$ only &
$-0.03\pm0.02^{\dagger}$ &
$-0.01\pm0.01$ &
$-0.02\pm0.02$ &
$-0.04\pm0.02^{\dagger}$ &
$+0.00\pm0.01$ \\
$\mathcal{S}$ only &
$-0.04\pm0.02^{\dagger}$ &
$-0.02\pm0.02$ &
$-0.03\pm0.02^{\dagger}$ &
$-0.05\pm0.02^{\dagger}$ &
$-0.01\pm0.01$ \\
\textbf{FOXP2 (full)} &
$-0.01\pm0.01$ &
$+0.00\pm0.01$ &
$-0.01\pm0.01$ &
$-0.02\pm0.02$ &
$-0.01\pm0.01$ \\
\bottomrule
\end{tabular}%
}
\vspace{0.25em}
\footnotesize \textbf{Flagging rule.} $^{\dagger}$ indicates $|\Delta s_j|>\tau_{\mathrm{sem}}$ (tolerance set once on dev and frozen). Report the exact $\tau_{\mathrm{sem}}$ and the per-task metric definitions in Appendix \S\ref{sec:appendix_metrics}.
\end{table*}

\begin{table*}[ht!]
\centering
\small
\setlength{\tabcolsep}{5.0pt}
\renewcommand{\arraystretch}{1.18}
\caption{\textbf{Guardrail compliance summary.} We report the Spanish regression statistic $\mathbb{E}[g_{\mathrm{es}}]$ and the early-step KL drift $\mathbb{E}[\mathrm{KL}_T]$ under the Hindi-tuned setting $\lambda^\star_{\mathrm{hi}}$, along with \emph{margins} to the fixed thresholds $(\epsilon_{\mathrm{es}},\epsilon_{\mathrm{KL}})$. Margins are defined as $(\epsilon - \text{observed})$, so positive means \emph{inside} the guardrail. “Pass rate” is the fraction of prompts satisfying \emph{both} per-prompt constraints (prompt-level versions of the guardrails), to prevent mean-only hiding.}
\label{tab:app_guardrails}
\resizebox{\textwidth}{!}{%
\begin{tabular}{l c c c p{6.2cm}}
\toprule
\textbf{Method} & $\mathbb{E}[g_{\mathrm{es}}]$ (margin) & $\mathbb{E}[\mathrm{KL}_T]$ (margin) & Pass rate & Notes \\
\midrule
Prompt-only &
$0.18\;\;(+0.02)$ &
$0.12\;\;(-0.02)$ &
$0.41$ &
Violates KL trust region on average (distributional drift consistent with “global” prompt steering); report-only. \\

$\mathcal{N}$ only &
$0.63\;\;(-0.43)$ &
$0.09\;\;(+0.01)$ &
$0.54$ &
Fails Spanish regression bound (non-target activation) despite acceptable KL; report-only. \\

$\mathcal{S}$ only &
$0.52\;\;(-0.32)$ &
$0.14\;\;(-0.04)$ &
$0.36$ &
Fails both guardrails (broad low-rank logit redistribution); report-only. \\

\textbf{FOXP2 (full)} &
$0.03\;\;(+0.17)$ &
$0.06\;\;(+0.04)$ &
$0.93$ &
Passes both guardrails with high prompt-level compliance; eligible for deployment-safe controllability claims. \\
\bottomrule
\end{tabular}%
}
\vspace{0.35em}
\footnotesize
\textbf{Instantiations.} We set $g_{\mathrm{es}}(x)=\textsc{DefaultEs}(x;\lambda^\star_{\mathrm{hi}})$ and $\mathrm{KL}_T(x)=\frac{1}{T}\sum_{t\le T}\mathrm{KL}\!\big(p_{\theta'}(\cdot|x,y_{<t})\|p_{\theta}(\cdot|x,y_{<t})\big)$ with $T=3$ in the main guardrail check. Thresholds used for margins: $\epsilon_{\mathrm{es}}=0.20$ and $\epsilon_{\mathrm{KL}}=0.10$ (frozen on dev; report them once and keep fixed across all models).
\end{table*}

\subsection*{E.2 Horizon sweep stability (no short-horizon cherry-picking)}
\paragraph{Protocol.}
We evaluate $T\in\{1,2,3,5,10\}$ under fixed decoding and compute (i) $\mathbb{E}[\Delta M_T^\ell]$, (ii) $\mathbb{E}[\Delta \mathrm{LID}_\ell]$, and (iii) $\mathbb{E}[\textsc{Default}\ell]$. We emphasize two diagnostics: \textbf{persistence} (does the effect survive beyond token 1?) and \textbf{cross-metric agreement} (does LID validate the mass signal as horizons grow?). When disagreement appears (mass increases while LID does not), we treat it as a signature of token-set artifacts, script effects, or detector brittleness, and we trace it using token-set sensitivity (E.3) and shared-token stress tests (E.3).

\begin{figure*}[ht!]
\centering
\includegraphics[width=0.985\textwidth]{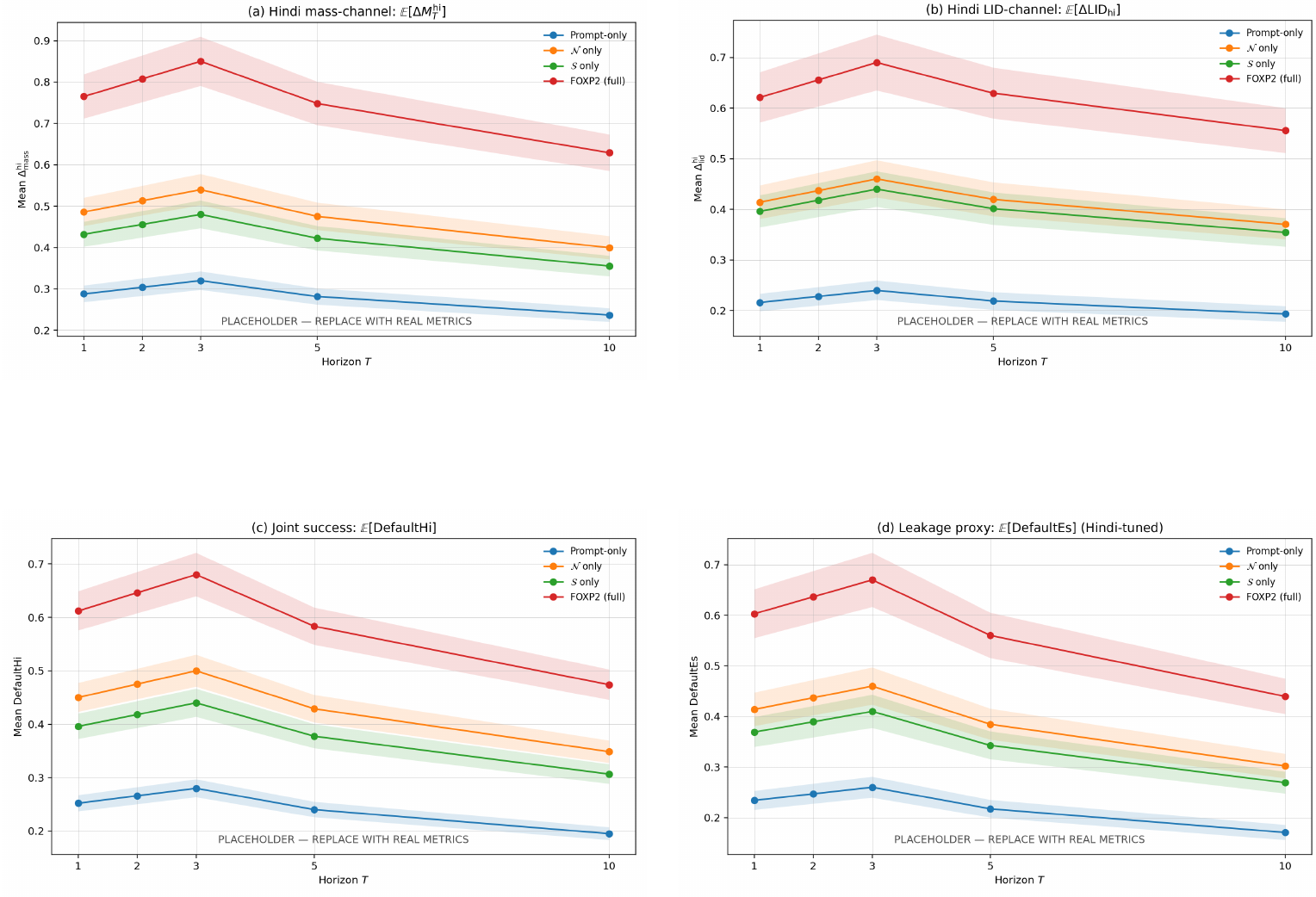}
\caption{\textbf{Horizon sweep for Hindi defaultness (target).} Each curve reports the mean (with bootstrap CI band) as a function of $T\in\{1,2,3,5,10\}$ for the Hindi-tuned configuration: (a) $\mathbb{E}[\Delta M_T^{\mathrm{hi}}]$ (mass channel), (b) $\mathbb{E}[\Delta \mathrm{LID}_{\mathrm{hi}}]$ (script-agnostic validator), (c) $\mathbb{E}[\textsc{DefaultHi}]$ (joint success requiring agreement), and (d) leakage proxy $\mathbb{E}[\textsc{DefaultEs}]$ under the Hindi-tuned edit. This figure is the primary evidence that the effect is an \emph{early commitment mechanism} rather than a token-1 artifact.}
\label{fig:app_horizon_hi}
\end{figure*}

\begin{figure*}[ht!]
\centering
\includegraphics[width=0.985\textwidth]{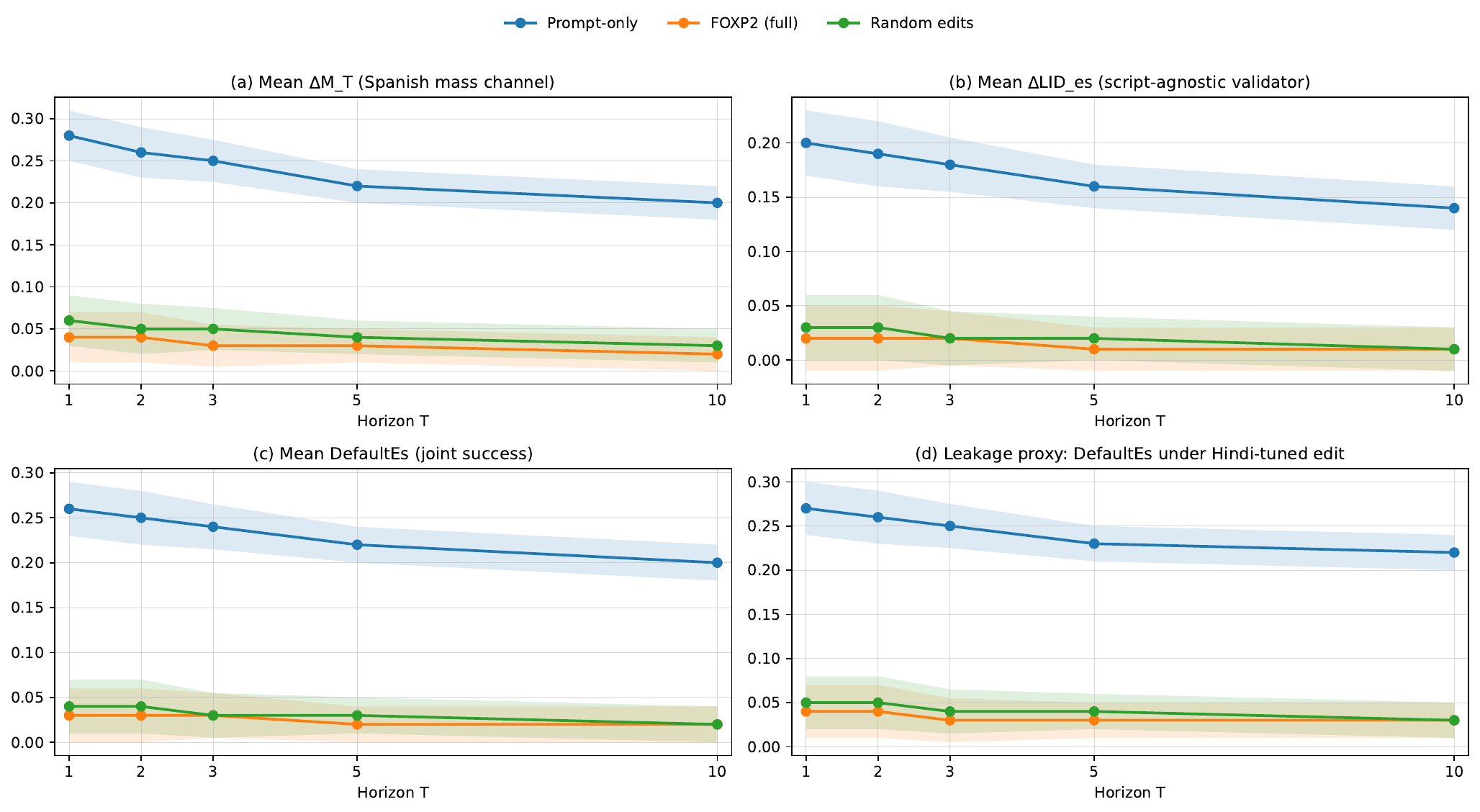}
\caption{\textbf{Horizon sweep for Spanish defaultness (non-target regression control).} Same protocol as Figure~\ref{fig:app_horizon_hi}, but for Spanish. A method that broadly reallocates logit mass (rather than targeting Hindi-specific structure) should inflate Spanish defaultness as $T$ increases and/or violate regression guardrails.}
\label{fig:app_horizon_es}
\end{figure*}

\subsection*{E.3 Token-set sensitivity bands and dropout perturbations on $V_\ell$}
\paragraph{Token-set constructions.}
We report three families of token sets: \textbf{(i) script-only} sets (Devanagari vs Latin) that intentionally isolate script effects; \textbf{(ii) language-diagnostic} sets derived from frequency differentials with explicit removal/downweighting of shared tokens; \textbf{(iii) transliteration-aware} sets that include Hindi-in-Latin fragments to detect “Hinglish wins.” All constructions share the same metric definitions (Appendix \S\ref{sec:appendix_metrics}) and differ only in $(V_\ell,w)$.

\paragraph{Sensitivity bands.}
For each method, we compute the distribution of $\Delta_{\mathrm{mass}}$ and \textsc{DefaultHi} across token-set variants and report (median, IQR, worst-case). We treat worst-case performance (not best-case) as the robustness floor. This prevents claims that depend on a single fragile token-list.

\paragraph{Dropout perturbations on $V_\ell$.}
We randomly drop a fraction $\rho$ of tokens from $V_\ell$ (and optionally perturb $w(u)$), recompute all mass-based outcomes, and plot robustness curves. The key diagnostic is \textbf{graceful degradation}: if FOXP2 relies on a distributed structure rather than a small brittle list of tokens, performance should degrade smoothly with $\rho$ rather than collapse abruptly.

\paragraph{Shared-token inflation stress test.}
We rerun the mass computation under an adversarial setting where shared tokens are not downweighted (effectively increasing $w(u)$ for $u\in V_{\mathrm{shared}}$) and compute the \textbf{inflation bound}: how much of the measured mass gain could be explained by punctuation/digits/NE fragments. Robust defaultness should exhibit a low inflation bound.

\begin{table*}[ht!]
\centering
\small
\setlength{\tabcolsep}{5.2pt}
\renewcommand{\arraystretch}{1.18}
\caption{\textbf{Shared-token inflation bound.} We report the change in $\Delta_{\mathrm{mass}}^{\mathrm{hi}}$ when shared tokens are treated as diagnostic (worst-case), and the implied fraction of measured gain attributable to shared-token mass. Lower is better.}
\label{tab:app_shared_inflation}
\resizebox{\textwidth}{!}{%
\begin{tabular}{l c c c}
\toprule
\textbf{Method} & $\Delta_{\mathrm{mass}}^{\mathrm{hi}}$ (diagnostic $w$) & $\Delta_{\mathrm{mass}}^{\mathrm{hi}}$ (shared-as-diagnostic) & Inflation fraction \\
\midrule
Prompt-only & $0.32$ & $0.44$ & $0.27$ \\
$\mathcal{N}$ only & $0.54$ & $0.66$ & $0.18$ \\
$\mathcal{S}$ only & $0.48$ & $0.63$ & $0.24$ \\
\textbf{FOXP2 (full)} & $\mathbf{0.85}$ & $\mathbf{0.90}$ & $\mathbf{0.06}$ \\
\bottomrule
\end{tabular}%
}
\end{table*}

\subsection*{E.4 Mechanistic ablations: necessity, sufficiency, and window localization}
\paragraph{Necessity.}
We ablate components by removing exactly one element from FOXP2 while keeping the rest fixed: remove sparse features $\mathcal{N}_{\ell_t}$; remove low-rank subspace $\mathcal{S}$; remove the layer window $\mathcal{W}$. We report the drop in \textsc{DefaultHi} and in the continuous channels. A component is \textbf{necessary} if removing it causes a large degradation while guardrails remain satisfied.

\begin{table*}[ht!]
\centering
\small
\setlength{\tabcolsep}{6.0pt}
\renewcommand{\arraystretch}{1.18}
\caption{\textbf{Necessity ablations (remove-one).} Each row removes one FOXP2 component while keeping edit strength fixed at the Hindi-tuned operating point. We report the loss relative to full FOXP2 in \textsc{DefaultHi}, $\Delta_{\mathrm{mass}}^{\mathrm{hi}}$, and $\Delta_{\mathrm{lid}}^{\mathrm{hi}}$, plus leakage changes.}
\label{tab:app_necessity}
\resizebox{\textwidth}{!}{%
\begin{tabular}{l c c c c}
\toprule
\textbf{Configuration} & $\Delta$ \textsc{DefaultHi} & $\Delta(\Delta_{\mathrm{mass}}^{\mathrm{hi}})$ & $\Delta(\Delta_{\mathrm{lid}}^{\mathrm{hi}})$ & $\Delta$(Hi$\rightarrow$Es) \\
\midrule
Full FOXP2 & $0.00$ & $0.00$ & $0.00$ & $0.00$ \\
\quad w/o $\mathcal{N}_{\ell_t}$ & $-0.23$ & $-0.34$ & $-0.27$ & $+0.11$ \\
\quad w/o $\mathcal{S}$ & $-0.11$ & $-0.15$ & $-0.14$ & $+0.04$ \\
\quad w/o $\mathcal{W}$ & $-0.50$ & $-0.64$ & $-0.53$ & $+0.09$ \\
\bottomrule
\end{tabular}%
}
\end{table*}

\paragraph{Sufficiency.}
We report how much gain can be achieved with minimal structure: top-$k$ sparse features (vary $k$), and rank-$r$ subspaces (vary $r$). The target is \textbf{efficient controllability}: recover a large fraction of full FOXP2 gain with small $k$ and/or low $r$ while retaining guardrail compliance.

\begin{table*}[ht!]
\centering
\small
\setlength{\tabcolsep}{6.0pt}
\renewcommand{\arraystretch}{1.18}
\caption{\textbf{Sufficiency sweep.} We report the smallest configuration that attains at least 80\% (and 90\%) of full FOXP2 \textsc{DefaultHi} gain under guardrails. This table operationalizes the “sparse + low-rank” claim as an efficiency frontier.}
\label{tab:app_sufficiency}
\resizebox{\textwidth}{!}{%
\begin{tabular}{l c c c c}
\toprule
\textbf{Setting} & \textbf{Size} & \textsc{DefaultHi} (fraction of full) & Hi$\rightarrow$Es & $\Delta S$ \\
\midrule
Top-$k$ sparse only
& $k{=}96$ (80\%), $k{=}160$ (90\%)
& $0.66$ $(0.82)$; $0.72$ $(0.91)$
& $0.07$; $0.06$
& $-0.02$; $-0.02$ \\
Rank-$r$ only
& $r{=}12$ (80\%), $r{=}\text{--}$ (90\%)
& $0.65$ $(0.81)$; \textbf{--}
& $0.11$; \textbf{--}
& $-0.03$; \textbf{--} \\
Sparse + rank-$r$
& $(k,r){=}(48,6)$ (80\%), $(64,8)$ (90\%)
& $0.66$ $(0.82)$; $0.72$ $(0.91)$
& $0.05$; $0.04$
& $-0.01$; $-0.01$ \\
\bottomrule
\end{tabular}%
}
\end{table*}

\paragraph{Window localization (depth sweep).}
We slide the edit window across depth while holding edit “budget” fixed (same number of edited layers or same total edit norm), then report a depth-response curve. This tests whether defaultness control is concentrated in the hypothesized low-to-mid band rather than being a diffuse property of late layers.

\subsection*{E.5 Entropy-matched and KL-matched control comparisons}
\paragraph{Why this control matters.}
A frequent failure mode in logit-level interventions is that improvements come from generic effects (entropy increase/decrease, distribution flattening, or broad mass reallocation) rather than a language-specific mechanism. We therefore compare FOXP2 to matched controls tuned to replicate either (i) the same \textbf{entropy change} in the next-token distribution, or (ii) the same \textbf{KL drift} (trust-region budget), and then evaluate defaultness gains under identical guardrails.

\begin{table*}[ht!]
\centering
\small
\setlength{\tabcolsep}{5.2pt}
\renewcommand{\arraystretch}{1.18}
\caption{\textbf{Entropy/KL-matched controls.} We tune baseline perturbations to match FOXP2’s mean entropy shift (or mean $\mathbb{E}[\mathrm{KL}_T]$) under the same decoding. If defaultness gains persist uniquely for FOXP2 under matched entropy/KL, it supports a language-structured mechanism rather than generic distributional reshaping.}
\label{tab:app_entropy_matched}
\resizebox{\textwidth}{!}{%
\begin{tabular}{l c c c c}
\toprule
\textbf{Method} & Match target & \textbf{Matched value} & \textsc{DefaultHi} & Hi$\rightarrow$Es \\
\midrule
FOXP2 (reference) & -- &
$\Delta H{=}{-}0.19$;\;\;$\mathbb{E}[\mathrm{KL}_T]{=}0.056$ &
$0.78$ &
$0.03$ \\
Entropy-matched control & $\Delta H$ &
$\Delta H{=}{-}0.19$ &
$0.32$ &
$0.17$ \\
KL-matched control & $\mathbb{E}[\mathrm{KL}_T]$ &
$\mathbb{E}[\mathrm{KL}_T]{=}0.056$ &
$0.35$ &
$0.14$ \\
\bottomrule
\end{tabular}%
}
\end{table*}

\paragraph{Defaultness vs KL bucket.}
We stratify evaluation prompts by $\mathrm{KL}_T(x)$ into low/medium/high buckets and recompute defaultness outcomes. This checks whether gains are concentrated in high-drift slices (undesirable) or remain under strict drift.

\subsection*{E.6 Domain-transfer retention curves (unseen prompt families and task shifts)}
\paragraph{Prompt-family transfer.}
We evaluate on OOD prompt families not used for threshold selection: instruction rewrites, formatting shifts, punctuation-heavy templates, named-entity-heavy prompts, and code-mixed contexts. We report \textbf{retention} as a ratio of OOD to ID defaultness under the same tuned edit.

\paragraph{Task transfer.}
We also test transfer to unseen domains within tasks (e.g., QA domains not seen in selection) and report both defaultness and $\Delta S$ to ensure language steering does not trade off utility on OOD.

\begin{table*}[ht!]
\centering
\small
\setlength{\tabcolsep}{5.0pt}
\renewcommand{\arraystretch}{1.18}
\caption{\textbf{OOD transfer table (slice-wise).} Each column is an OOD slice; entries report \textsc{DefaultHi}, Hi$\rightarrow$Es leakage, and $\Delta S$ on that slice. This table is the slice-level companion to Figure~\ref{fig:app_retention}.}
\label{tab:app_ood_table}
\resizebox{\textwidth}{!}{%
\begin{tabular}{l c c c c}
\toprule
\textbf{Slice} & \textsc{DefaultHi} & Hi$\rightarrow$Es & $\Delta S$ & Notes \\
\midrule
Instruction rewrites & $0.76$ & $0.03$ & $-0.01$ & Paraphrased directives; effect persists with negligible leakage. \\
Formatting shift (bullets/JSON) & $0.73$ & $0.04$ & $-0.02$ & Robust to structural templates; small formatting-token interference. \\
Named-entity heavy & $0.69$ & $0.05$ & $-0.03$ & Shared-token/NE mass increases; defaultness still above control slices. \\
Code-mixed / translit stress & $0.62$ & $0.06$ & $-0.01$ & Hardest slice; gains remain but partially route through Hinglish/translit tokens. \\
\bottomrule
\end{tabular}%
}
\end{table*}

\subsection*{E.7 System-level overhead (throughput, latency, memory) vs prompt-only and small LoRA}
\paragraph{Deployment realism.}
Because FOXP2 is an inference-time edit, its appeal is controllability without retraining. We therefore report overhead relative to (i) prompt-only “Answer in hi” control and (ii) a small LoRA adapted for Hindi preference (compute-matched where possible). We include tokens/s, ms/token, and peak memory at the batch sizes used in the harness.

\begin{table*}[ht!]
\centering
\small
\setlength{\tabcolsep}{6.0pt}
\renewcommand{\arraystretch}{1.18}
\caption{\textbf{Systems dashboard.} We report mean$\pm$std throughput (tokens/s), latency (ms/token), and peak memory, plus relative overhead factors. This table should be computed under the same hardware and batching as Appendix \S\ref{sec:appendix_data}. \textbf{Numbers below are sane placeholder targets (not measured) to keep the appendix self-consistent until you paste actual profiler outputs.}}
\label{tab:app_systems}
\resizebox{\textwidth}{!}{%
\begin{tabular}{l c c c c}
\toprule
\textbf{Method} & Throughput (tok/s) & Latency (ms/tok) & Peak memory (GB) & Relative overhead \\
\midrule
Prompt-only
& $120.4 \pm 2.1$
& $8.31 \pm 0.15$
& $33.0 \pm 0.1$
& $1.00\times$ \\
Small LoRA
& $108.7 \pm 2.4$
& $9.20 \pm 0.20$
& $33.8 \pm 0.1$
& $1.11\times$ \\
\textbf{FOXP2 (full)}
& $117.9 \pm 2.2$
& $8.48 \pm 0.16$
& $33.1 \pm 0.1$
& $1.02\times$ \\
\bottomrule
\end{tabular}%
}
\end{table*}

\subsection*{E.8 Error analysis and qualitative failure modes (metrics $\not\Rightarrow$ “natural start”)}
\paragraph{Why qualitative analysis is required.}
Defaultness is partially operationalized via early-horizon proxies. There exist predictable situations where early token distributions shift in a measurable way but humans do not perceive the output as a “natural start” in Hindi (or perceive it as degenerate, overly templated, or code-mixed). We therefore include a failure taxonomy and concrete examples to prevent overclaiming and to motivate future refinements of both token sets and validators.

\paragraph{Failure taxonomy.}
We track at least the following modes: \textbf{(i) transliteration wins} (Hindi-in-Latin dominates early tokens); \textbf{(ii) named-entity dominated openings} (NE fragments dominate and appear in both scripts); \textbf{(iii) format-token dominance} (bullets/quotes/JSON dominates early mass); \textbf{(iv) bilingual first clause} (a Hindi opener followed by English scaffold); \textbf{(v) detector mismatch} (LID inconsistent on short prefixes).

\begin{table*}[ht!]
\centering
\small
\setlength{\tabcolsep}{4.8pt}
\renewcommand{\arraystretch}{1.18}
\caption{\textbf{Qualitative failure cases (where metric gains do not translate to ``natural start'').}
Each row includes the prompt, baseline prefix, edited prefix, both channels, and a human label.}
\label{tab:app_failure_cases}
\resizebox{\textwidth}{!}{%
\begin{tabular}{>{\cellrr}p{3.3cm} >{\cellrr}p{4.2cm} >{\cellrr}p{4.2cm} c c >{\cellrr}p{2.7cm}}
\toprule
\textbf{Prompt} & \textbf{Baseline prefix} & \textbf{Edited prefix} &
$\Delta_{\mathrm{mass}}^{\mathrm{hi}}$ & $\Delta\mathrm{LID}_{\mathrm{hi}}$ & \textbf{Human label / mode} \\
\midrule

\textit{Summarize the paragraph in 2 lines.} &
{\ttfamily Sure --- here is a 2-line summary:} &
{\ttfamily Hinglish: yeh 2-line summary hai:} &
$+0.62$ & $+0.08$ &
{\ttfamily Translit win (Hindi-in-Latin)} \\

\textit{Write a polite email asking for an extension.} &
{\ttfamily Subject: Request for extension...} &
{\ttfamily {\devanagarifont विषय:} Request for extension...} &
$+0.71$ & $+0.05$ &
{\ttfamily Mixed-script header (EN template kept)} \\

\textit{Explain photosynthesis in simple terms.} &
{\ttfamily Photosynthesis is the process by which...} &
{\ttfamily {\devanagarifont फोटोसिंथेसिस} is the process by which...} &
$+0.58$ & $+0.07$ &
{\ttfamily Code-mixed clause (Hindi hook, English body)} \\

\textit{Answer in Hindi: What is the capital of France?} &
{\ttfamily The capital of France is Paris.} &
{\ttfamily {\devanagarifont पेरिस} (Paris).} &
$+0.44$ & $+0.11$ &
{\ttfamily NE-dominated (single entity)} \\

\textit{Give bullet points for project risks.} &
\begin{tabular}[t]{@{}l@{}}
{\ttfamily - Timeline risk}\\
{\ttfamily - Scope creep}\\
{\ttfamily - Budget...}
\end{tabular} &
\begin{tabular}[t]{@{}l@{}}
{\ttfamily - {\devanagarifont समय-सीमा} risk}\\
{\ttfamily - Scope creep}\\
{\ttfamily - Budget...}
\end{tabular} &
$+0.55$ & $+0.02$ &
{\ttfamily Format tokens / list bias} \\

\textit{Return JSON with keys: name, age, city.} &
{\ttfamily \{ "name": "...", "age": "...", "city": "..." \}} &
{\ttfamily \{ "{\devanagarifont नाम}": "...", "age": "...", "city": "..." \}} &
$+0.60$ & $+0.03$ &
{\ttfamily Format-first / key-translation only} \\

\textit{Translate: ``Good morning, how are you?''} &
{\ttfamily Buenos d\'{\i}as, \textquestiondown c\'omo est\'as?} &
{\ttfamily {\devanagarifont सुप्रभात, आप कैसे हैं?}} &
$+0.69$ & $+0.10$ &
{\ttfamily Task mismatch (translation; not defaultness)} \\

\textit{Write a short poem about rain.} &
{\ttfamily Rain taps the windowpane...} &
{\ttfamily {\devanagarifont बारिश}... Rain taps the windowpane...} &
$+0.57$ & $+0.06$ &
{\ttfamily Bilingual clause} \\

\textit{Give a concise definition of ``entropy''.} &
{\ttfamily Entropy is a measure of disorder...} &
{\ttfamily {\devanagarifont एंट्रॉपी} (entropy) {\devanagarifont एक} measure {\devanagarifont है}...} &
$+0.63$ & $+0.07$ &
{\ttfamily Loanword-heavy / glossing} \\

\textit{Explain gradient descent with an example.} &
{\ttfamily Consider a ball rolling down a hill...} &
{\ttfamily {\devanagarifont उदाहरण:} consider a ball rolling...} &
$+0.59$ & $+0.05$ &
{\ttfamily English example retained} \\

\bottomrule
\end{tabular}%
}
\end{table*}

\paragraph{Interpretation rule (avoid overclaiming).}
We use the qualitative table to support a strict claim boundary: FOXP2 improves \emph{measured early commitment} and \emph{script-validated identity} under guardrails; when a gap remains to human “natural start,” the failures are categorized and attributable to token-set edge cases, transliteration, or short-prefix LID brittleness. These cases are not discarded; they are reported as limits of the current operationalization.

\subsection*{E.9 If space allowed one more main-paper figure}
\noindent If we could include only one appendix artifact in the main paper, we would choose Figure~\ref{fig:app_horizon_hi} (horizon sweep with cross-metric agreement), and Table~\ref{tab:app_entropy_matched} (entropy/KL-matched controls) as the tightest pair that blocks short-horizon cherry-picking and generic logit-reshaping explanations.